\documentclass[twoside,11pt]{article}

\usepackage{jmlr2e, times, amsmath, amssymb, verbatim, enumerate, color}
\usepackage{url, graphicx, mathtools, wrapfig}
\usepackage[caption=false]{subfig}
\usepackage{hyperref}
\usepackage{epsfig, multirow, paralist, xcolor, comment, algorithm, algorithmic}
\usepackage{enumitem, lipsum}

\hypersetup{
    colorlinks = false,
    urlcolor = {blue}, urlbordercolor = {red},
    citecolor = {black},
}

\usepackage{lineno}

\newcommand{\x}{\ensuremath{\mathbf x}} \newcommand{\y}{\ensuremath{\mathbf y}} 
 \newcommand{\h}{\ensuremath{\mathbf h}} 
\newcommand{\W}{\ensuremath{\mathbf W}} \newcommand{\hW}{\ensuremath{\hat{\mathbf W}}} \newcommand{\hw}{\ensuremath{\hat{W}}}
	 
\newcommand{\tx}{\ensuremath{\tilde{\mathbf x}}} 
\newcommand{\tg}{\ensuremath{\tilde{g}}} \newcommand{\tG}{\ensuremath{\tilde{G}}} \newcommand{\tf}{\ensuremath{\tilde{f}}}
\newcommand{\uda}{\ensuremath{\mathcal{U}_{da}}} \newcommand{\vda}{\ensuremath{\mathcal{V}_{da}}} 
\newcommand{\z}{\ensuremath{\mathbf z}} \newcommand{\tzeta}{\ensuremath{\tilde{\zeta}}}

\newlength\myindent
\newcommand\bindent{%
  \begingroup
  \setlength{\itemindent}{\myindent}
  \addtolength{\algorithmicindent}{\myindent}
}
\newcommand\eindent{\endgroup}

\DeclareMathOperator*{\argmin}{arg\,min}
\DeclareMathOperator{\Var}{Var}
\DeclarePairedDelimiter{\ceil}{\lceil}{\rceil}

\begin{document} 

\title{On architectural choices in deep learning: From network structure to gradient convergence and parameter estimation}

\author{\name Vamsi K. Ithapu \email ithapu@wisc.edu \\
       \addr Department of Computer Sciences\\
       University of Wisconsin Madison\\
       \AND
       \name Sathya N. Ravi \email ravi5@wisc.edu \\
       \addr Department of Computer Sciences\\
       University of Wisconsin Madison\\
       \AND
       \name Vikas Singh \email vsingh@biostat.wisc.edu \\
       \addr Department of Biostatistics and 
       Department of Computer Sciences\\
       University of Wisconsin Madison\\
       \AND
       \href{http://pages.cs.wisc.edu/~vamsi/DLDesignChoices}{http://pages.cs.wisc.edu/~vamsi/DLDesignChoices}
       }       
\editor{}


\maketitle

\begin{abstract}
We study mechanisms to characterize how the asymptotic convergence of backpropagation in deep architectures, in general, is related to the network structure,
and how it may be influenced by other design choices including activation type, denoising and dropout rate. 
We seek to analyze whether network architecture and input data statistics may guide the choices of learning parameters and vice versa. 
Given the broad applicability of deep architectures, this issue is interesting both from theoretical and a practical standpoint.
Using properties of general nonconvex objectives (with first-order information), we first build the association between structural,
distributional and learnability aspects of the network vis-\`a-vis their interaction with parameter convergence rates. 
We identify a nice relationship between feature denoising and dropout, and construct families of networks that achieve the same level of convergence.
We then derive a workflow that provides systematic guidance regarding the choice of network sizes and learning parameters often mediated4 by input statistics. 
Our technical results are corroborated by an extensive set of evaluations, presented in this paper as well as independent empirical observations reported by other groups.
We also perform experiments showing the practical implications of our framework for choosing the best fully-connected design for a given problem.

\end{abstract}


\vspace{4mm}
\noindent {\bf Condensed Title:} On architectural choices in deep learning \\
\noindent {\bf Corresponding Author:} Vamsi K. Ithapu, Department of Computer Sciences, University of Wisconsin Madison, Madison WI USA 53706
\newpage


\section{Introduction}
\label{sec:intro}

The successful deployment of deep learning algorithms in a broad spectrum of applications including localizing objects in images \citep{lenz2015deep, krizhevsky2012imagenet, szegedy2014going}, 
analyzing particle accelerator data \citep{baldi2014searching}, converting speech into text \citep{hinton2012deep}, learning to play video games \citep{mnih2013playing}, 
predicting the activity of drug molecules \citep{lusci2013deep, sutskever2014sequence} and designing clinical trials \citep{plis2014deep, ithapu2015imaging} 
provides compelling evidence that they can learn complex concepts with fairly minimal feature engineering or preprocessing. 
This success is attributed to the idea of composing simple but non-linear modules that each transform the lower levels 
(i.e., raw or normalized data) into a representation at a more abstract level \citep{hintonnature, bengio2009learning, bengio2013representation}.
These layers of features are not hand designed, rather learned from data using a general learning process guided by the structure of the carefully chosen non-linear modules. 
While this high level representation learning procedure is quite general, 
the problem at hand may at least partly govern the choice of the architecture and require certain modifications to the algorithm.
Consequently, motivated by various experimental considerations one encounters in practice, 
several variants of deep architectures and corresponding regularization schemes have been developed over the past decade
\citep{lee2009convolutional, vincent2010stacked, baldi2014dropout, goh2013top, wan2013regularization, ioffe2015batch}.


Complementary to such design, algorithmic, and empirical developments, there is also a growing recent interest in better understanding the mathematical properties of these models.
Over the last few years, several interesting results have been presented 
\citep{castillo2006very, shao2011convergence, dauphin2014identifying, livni2014computational, bach2014breaking, arora2014provable, patel2015probabilistic, janzamin2015beating, hardt2015train}.
While some of these studies address the hypothesis space, and the corresponding sets of functions learnable by deep networks, 
others analyze the nature of the parameter space that needs to be learned via backprogapation.
A more detailed discussion about these works is included in Section \ref{sec:related} where we review some of the most related papers.
This literature is still rapidly evolving, and there are a large number of interesting, and open, questions listed, for instance, in \cite{bengio2012practical, wang2015survey} 
whose answers will help guide, going forward, the training, debugging and designing of large-scale, and often very deep, multi-layer neural networks for arbitrary and complex learning tasks. 
A few such motivating questions are given below that will help provide the context for this paper. 

\begin{compactenum}[\bfseries (Q1)]
\item There is strong empirical evidence that changing the network depth or layer lengths (i.e., sizes of the individual layers) leads to significant and, at times, 
a non-trivial change in generalization performance (measured via testing set error) \citep{bengio2009learning, hinton2006reducing, simonyan2014very, szegedy2014going}.
To that end, is it possible say anything specific about the {\it best} possible architecture for learning representations from a given dataset or for a specific task?
In other words, 
\begin{itemize}
\item[{\bf (a)}] Can we explicitly compute the influence of the number of layers, the corresponding lengths and activation function choices, 
on the complexity of the input-to-output mapping being learned? 
\item[{\bf (b)}] Or, how do the types of learnable mappings vary as we change the lower and/or higher layer lengths? 
Are there sets of appropriate and `good' architectures for a given learning task? 
\end{itemize}
\item Does fully supervised dropout \citep{srivastava2014dropout} compensate for unsupervised pretraining \citep{erhan2009difficulty}? 
Are these two ideas theoretically related? As a function of the network structure, are there regimes where one is intrinsically better? 
\item Which networks have faster parameter convergence? A recent result on stochastic gradients relates the training time to generalization \citep{hardt2015train}. 
On new learning tasks (with no established benchmarks), can the notion of parameter convergence, in tandem with (or without) such a generalization result, 
be used as a criteria for choosing the best architecture from a family of networks? 
\item Are there concepts that {\em cannot} be learned by a given deep model? 
Specifically, given the form of non-linearities used in a design, can we say much about the type of data statistics that the network is guaranteed to learn? 
\item Given that the choice of non-linearities is mainly task dependent, for instance, max-pooling guarantees some form of translation invariance \citep{nagi2011max}. 
In general, these choices are not based on the eventual goal of convergence or generalization -- and instead such non-linearities can be interpreted as efficient regularizers. So, 
\begin{itemize}
\item[{\bf (a)}] Given the task, are there optimal choices for such non-linearities? 
More importantly, using task-specific information and the data distribution, are there strategies to {\it choose} the best activation from a given bag of non-linearities?
\item[{\bf (b)}] Are there strategies for designing activations beyond convolutions and translation invariance, for instance, 
  to efficiently learn models for applications where the data are brain images (e.g., magnetic resonance images \citep{frisoni2010clinical, killiany2000use})
  and genetic data (e.g., single nucleotide polymorphisms \citep{sachidanandam2001map}), 
that are presumed to lie on complex combinations of low-dimensional manifolds \citep{gerber2009manifold, corouge2004towards, lazar2003white}.
\end{itemize}
\item Different families of deep networks have been shown to lead to similar empirical performance \citep{livni2014computational, ngiam2011optimization}. 
When can we say that two arbitrary deep networks are {\it equivalent} to each other in terms of the hypothesis spaces they can model? 
Can one construct transformations or maps between such families?
\item One of the reasons behind the success of deep networks is the availability of large amounts of data 
\citep{hintonnature, bengio2013representation, erhan2009difficulty, livni2014computational}.
But can we also succseefully perform small sample size deep learning with dataset sizes much larger than the number of features (common in biomedical applications involving expensive acquisitions)? 
Or, can the network be {\it regularized} to compensate for the lack of sufficient amount of training data?
\end{compactenum} 
For some of these problems, there is good empirical evidence but little theoretical support (e.g., ${\bf Q1}$ and ${\bf Q6}$), 
while some of the other questions, to the best of our knowledge, are yet to be carefully studied and addressed from the theoretical perspective. 
For instance, while this literature is rapidly evolving, not much is known about the relationship of the network architecture and the input distributions (${\bf Q1}$, ${\bf Q3}$ and ${\bf Q5}$), 
in the context of algorithms for small sample size deep learning (${\bf Q7}$). 


The broad goal of this work is to address some of these open problems, and/or provide a good starting point towards answering them.
We pursue this goal by presenting a potentially useful theoretical characterization of certain convergence properties of deep networks, from the perspective of parameter estimation.
At the high level, we study the relationship between learnability i.e., convergence of parameter estimation in an optimization sense, 
and the architecture of the deep networks for a given training dataset.
In particular, we are interested in the interplay of the {\it network architecture} and {\it data statistics} and their influence on the {\it learning schemes}.
These aspects of deep networks, although intuitively seem related, have, for the most part, been studied separately.
Specifically, \cite{dauphin2014identifying, livni2014computational, bach2014breaking} and others address the optimization of deep networks, 
while \cite{lee2009convolutional, baldi2014dropout, goh2013top, wan2013regularization} and others deal with the regularization aspects -- 
an {\it explicit} characterization of their interaction is lacking in the literature so far.
We describe how characterizing this {\it interplay} directly enables answering several of the above questions, 
which may facilitate or guide other empirical investigations. 


{\it Overview:} 
The most commonly used procedure for parameter estimation in deep networks is the mini-batch stochastic gradients \citep{bottou1991stochastic, lecun2012efficient, bengio2012practical}. 
This involves deriving the average gradient from the error computed on the learning objective (or loss) using a few training instances, and adjusting the weights/parameters accordingly.
This continues for a fixed number of iterations and one can, in principle, perform some checks at the stopping iteration and repeat the procedure, if needed. 
Our general goal is to tie the mechanics of this procedure, to the network structure. 
In contrast to other recent results derived independently of our work \citep{janzamin2015beating}, we directly work with stochastic gradients with no strong assumptions on the data/network structure 
\citep{bach2014breaking}, which enables obtaining results that are quite generally applicable. 
The starting point of our analysis is a recent work by \cite{ghadimi2013stochastic} dealing with the convergence of stochastic gradients for arbitrary nonconvex problems using a first-order oracle. 
We build upon and adapt this analysis by first addressing single-layer networks and unsupervised pretraining, and then, 
the more general case of multi-layer dropout networks, followed by convolutional and recurrent neural networks.
In each of these cases, once the network structure is tied to the behaviour of the gradients, we analyze the influence of input data statistics on the parameter estimation and convergence. 
More importantly, apart from addressing the interplay, the algorithms we present, with minor tweaks, are easily deployable to the standard training pipeline.
The bounds natively take into account the standard regularization schemes like dropout and layer-wise pretraining, making them even more useful in practice. 


\subsection{The design choice problem}
\label{sec:design}

Within the last several years, several variants of network architectures have been proposed with various combinations of fully connected 
(e.g. Boltzmann machines \citep{nair2010rectified} or autoencoders \citep{vincent2010stacked}), 
convolutional or recurrent layers with non-linearities like rectified linear units \citep{dahl2013improving}, maxout \citep{goodfellow2013maxout} and/or max-pooling \citep{nagi2011max} 
trained using dropout \citep{srivastava2014dropout}, dropconnect \citep{wan2013regularization}.
The goal of a practitioner is to choose the network architecture most suitable for solving a given problem/application.
One is then faced with a `multitude' of architectural choices when trying to decide the specific construction that is likely to work the best.
For instance, one may decide to use a network with combinations of convolutional and fully connected layers using rectified linear units where learning is performed using dropout,  
or prefer other variations of such a {\it prototypical} construction. 
The choice is, at least, in part, governed by the domain knowledge apart from other resources constraints including 
the sizes of available datasets and/or the desired convergence (or generalization) of the estimated parameters.
Therefore, this interplay of (network) structure and parameter convergence is important to guide the choice of which setup will be most useful. 
Consider the following simple issues that comes up routinely in practice.


\begin{itemize} [leftmargin=1em]
\item {\it From the prototypical network to the desired one:} 
With the goal of achieving the best possible generalization via validation set error or some other performance measure of interest, 
one may start with a prototypical network described in the literature and then {\it modulate} it by changing the depth, layer lengths, 
or the learning algorithm \citep{wan2013regularization}, \citep{goodfellow2013maxout}. 
The specific modulations, i.e., the design choices, are mostly driven by domain knowledge or user expertise.
For a wide variety of concepts/tasks, which non-linearities to choose may not be clear, and may have to be based on trial/error.  
\item{\it How many designs to check for?}
Beyond the structural adjustments that might be necessary, there is a separate question focusing on how many adjustments to evaluate for and when to stop.
For smaller and medium sized datasets, a few dozen, if not a hundred, such modulated models may be tested.
For much larger datasets, in general, the biggest model that can reasonably fit in the memory, and can be trained, is selected.
There are not many guidelines available to facilitate this process.
\end{itemize}


Clearly, any best practices, e.g., basic principles or ``ground rules'', 
in order to select the {\it appropriate} families of networks and the corresponding structural modulations for customization, will be beneficial. 
A systematic or informed (based on some set of rules) network design strategy can address some of these problems, 
while also providing insights into design choice issues beyond the type and number of models to search for.  


\begin{itemize} [leftmargin=1em]
\item {\it Resource allocation:} 
  Given a learning task that is relevant in an application, criteria for the dataset size that one must collect within a study is very useful in biomedical applications
  where the acquisition is costly and time-consuming --- the budget must be explicitly justified to funding agencies.  
  For instance, a single acquisition of positron emission tomography (PET) scan can cost $\$3000+$, while a MRI scan may cost $\sim\$500$.
  Similar to power calculations in statistical analysis, a sensible strategy for estimating the dataset size required to achieve a certain level of parameter convergence is highly desirable.
  Further, training `larger than necessary' networks will entail more computing and financial resources. 
  A simple way to quantify such gains is by asking the question when to stop the stochastic gradient update procedure? 
  Cutting down the minimum number of such training iterations required to achieve a certain level of convergence or generalization, even by $5-10\%$, will result in
  quantifiable computing and financial savings, 
  for instance, when learning networks on cloud platforms.
\item {\it Model family/class selection:} 
  It seems that searching for the most appropriate model will benefit from having a mechanism that lists out the families of networks with similar (expected) generalization for the given task.
  For instance, fully connected sigmoidal networks might be more appropriate than convolutional layers for certain tasks 
  where the input feature space is registered (or normalized) to some ``template'' space. 
  Such co-registered data is common in speech modeling \citep{hinton2012deep}, medical imaging \citep{hinrichs2011predictive} etc.
\item {\it Hyper-parameter selection:} 
  The learning hyper-parameters are generally selected via cross validation or some additional domain knowledge/expertise.
  Although, automated hyper-parameter tuning has been studied using ideas from Bayesian Optimization \citep{snoek2012practical} and Bandit learning \citep{li2016efficient}, 
  there is little guidance available for adapting such procedures for deep learning.
  It is not yet clear if design choice problem will benefit from such strategies --- 
  restricting the hyper-parameter space using information/constraints on structure and the achievable convergence is one potentially useful way to address this issue.
\end{itemize}


In this paper we discuss how the systematic design strategies may fall out of a framework that analyzes the interplay of structure and parameter convergence. 
While we do not claim that the network parameters should be decided solely based on the convergence; 
instead, we show that convergence will enable {\it assessing} the goodness of a chosen network, and therefore, in turn, the structure or learning hyper-parameters can be changed as necessary. 
On a new dataset/task, one may first `select' the best possible network structure guided by the results described in this paper, using information from data statistics and other domain knowledge. 
The trends will help adapt or modulate the chosen network as needed, based on certain downstream performance measures like accuracy on validation sets.


\subsubsection{Analysis of Data from Multi-center Studies}\label{sec:resource}

The design choice problem is more involved where models need to be constructed on datasets collected from multiple sources or acquisition sites. 
This problem can often be tackled {\em prospectively}, e.g., within large scientific studies \citep{klunk2015centiloid} across multiple sites that
are becoming more prevalent to better understand disease progression \cite{mueller2005ways} or evaluating the efficacy of drugs \cite{sperling2014a4}. 
Data sharing/pooling considerations are usually formalized within the project guidelines, which makes the follow-up analysis tasks relatively convenient.
In contrast, in many cases, the decision to {\it pool} data sets across multiple sources or centers is {\em retrospective} ---
in an effort to increase the sample size of certain statistical tests, so that specific scientifically interesting hypotheses
can be evaluated (e.g., the data at each site, by itself, may be underpowered due to smaller sample size). 
Separate from statistical analysis, we may seek to train richer machine learning models for prediction or other clinical purposes on such pooled datasets. 
However, because of privacy laws that differ between countries (e.g., Europe, United States),
it may not always be possible (or may be logistically difficult) to pool/transfer data collected at different sites. 
One may perform meta-analysis \citep{haidich2011meta} to aggregate individual models inferred from different datasets/sites.
Alternatively, one can choose to run or train `similar' statistical models at each site independently (respecting each country's data sharing laws), and later combine the models
or the estimates (e.g., weighted by the sample sizes or the moments of the sample distribution).
Assuming that the models of interest are rich classes of neural networks, the ability to learn the {\it same} 
{\it or comparable} networks across all the datasets is very useful --- it makes the downstream aggregation or interpretation task significantly simpler.

If the underlying physical characteristic of the datasets are consistent, 
it makes sense to learn comparable deep network models across multiple (similar) datasets.
This will enable models that are {\it transferable} across multiple data centers ensuring
that similar network models, with similar degrees of freedom, 
will be constructed on both data acquisition sites, respecting geographical data transfer constraints. 
Observe that an inherent component of statistical estimation is to assess the confidence of the produced estimates, typically, 
via computing the error bars (or confidence intervals) using some bootstrapping with/without replacement.
Clearly, for this bootstrapping to be sensible, each realization, i.e., each set of deep networks from all the sites, 
will need to be learned up to the ``same'' level of convergence or generalization.
In this way, any posthoc statistical procedure, 
like field of experts \cite{roth2005fields}, non-nested model selection \cite{pesaran2001non}, 
or simply the information criteria like AIC, BIC etc. \cite{arlot2010survey}, can then be used to select the best or aggregated model across the datasets/sites.
Our framework will facilitate only only the multi-center design choice problem,
but also provide guidance on how the bootstrapping will be performed in practice. 


\noindent \paragraph{\bf Main Contributions:}
{\bf (a)} Motivated by the recent idea of randomly stopping gradient updates from \citep{ghadimi2013stochastic}, we present a framework for analyzing 
mini-batch stochastic gradients on arbitrary multi-layer deep networks. 
We prove gradient convergence of multi-layer networks learned via dropout with/without layer-wise pretraining.
{\bf (b)} Building upon the framework, we derive explicit relationships between the network structure, gradient learning parameters and input data statistics.
Our results are consistent with many empirical studies, but further {\it guide} the choices of network/learning hyper-parameters for modeling a given data set.  
{\bf (c)} We present extensive experiments evaluating and visualizing the trends from the derived bounds.
{\bf (d)} We present systematic design procedures for constructing deep networks that achieve certain level of convergence and generalization, 
and discuss a case study using such optimal design choices in learning deep networks on medical imaging data. 
{\bf (e)} We provide an interactive webpage which recommends sensible choices of network structure for a given task or application at hand, based on some user defined input.


\subsection{Related Work} 
\label{sec:related}

The body of literature addressing backpropagation in neural networks, and deep networks in general, is vast and dates back at least to the early $1970$s. 
Here, we restrict the discussion only to those works that fall immediately within the context of this paper. 
While there are a number of early seminal papers addressing variants of backpropagation and stochastic gradients, and studying the convergence of neural networks training
\citep{becker1988improving, lecun1998gradient, vogl1988accelerating, magoulas1999improving}, a number of recent works 
\citep{ngiam2011optimization, lecun2012efficient, bengio2012practical, dauphin2014identifying, janzamin2015beating, hardt2015train, andrychowicz2016learning} 
provide a fresh treatment of these problems and analyze efficient learning schemes in the context of deep networks specifically.
The solution space of backpropagation in this setting has also been addressed in recent results \citep{castillo2006very, shao2011convergence}, 
providing new insights into the types of functions learnable by deep networks \citep{livni2014computational, bach2014breaking}.
\cite{hardt2015train} have shown that better generalization can be achieved by ensuring smaller training times, 
while \cite{dauphin2014identifying} describe, in detail, the non-convex landscape of deep learning objectives and the goodness of local optima.
Beyond these optimization related works, several authors have independently addressed the regularization and learnability aspects of deep networks.
For example, \cite{wager2013dropout, baldi2014dropout} extensively analyzed the properties of dropout learning, 
and \cite{patel2015probabilistic} develops a comprehensive probabilistic theory of deep learning. 
\cite{arora2014provable} study the existence and construction of deep representations by exploiting the structure of the network, 
and \cite{arora2015deep} presents a generative model for ReLU type deep networks with the assumption that the network weights are random.
The number of linear regions computable by deep networks is studied by \cite{montufar2014number}.
Very recently, \cite{wei2016network} have studied the equivalence of arbitrary deep networks. 
Complimentary to these, \cite{janzamin2015beating} use a tensor decomposition perspective to offer guarantees on training certain types of networks.


\section{Preliminaries}
\label{sec:prelim}

Our goal is to relate the structural aspects of arbitrary deep networks (including the number of layers, length of each layer and activation function types) 
to the corresponding learning hyper-parameters (like dropout rates, gradient iterations, stepsizes, and so on) using non-trivial information about the 
input data distribution (e.g., moments).
We first present some basic notations to setup the analysis and then provide a roadmap to our framework before describing it in the next sections.


\subsection{Notation}
\label{sec:notation}

Let $\x \in \mathbb{R}^{d_x}$ and $\y \in \mathbb{R}^{d_y}$ denote the input feature vector and the corresponding output (or label) respectively. 
Given multiple $\{\x,\y\} \in \mathcal{X}$, the unknown input-to-output mapping is modeled by a $L$-layered neural network ($L$-NN). 
An $L$-NN comprises the input (or visible) unit $\x$, followed by $L-1$ hidden representations $\h^1,\ldots,\h^{L-1}$ and the output (or final) unit $\y$ \citep{bengio2009learning}.
The lengths of these $L+1$ representations are $d_0 = d_x, d_1,\ldots,d_{L-1}, d_L = d_y$ respectively.
Each layer transforms the representations from the previous layer by first applying an affine transform, followed by a non-linear function 
which may be non-convex (in general) but not necessarily point-wise \citep{bengio2009learning, bengio2013representation}.
The layer-wise transformation matrices are denoted by $\W^l \in \mathbb{R}^{d_{l} \times d_{l-1}}$ for $l = 1,\ldots,L$.
The hidden representations are given by $\h^l = \sigma(\W^l,\h^{l-1})$ for $l=1,\ldots,L-1$ ($\h^0 = \x$), and the output layer is $\y = \sigma(\W^l,\h^{L-1})$, 
where $\sigma(\cdot)$ represents the non-linear function/mapping between layers. 
For a single-layer network with no hidden layers, we have $\y = \sigma(\W,\x)$ where $\W$'s are the unknowns.
Note that the bias in the affine transformation is handled by augmenting features with $1$ whenever necessary.
The distributional hyper-parameters of interest are $\mu_\x = \frac{1}{d_x}\sum_j \mathbb{E}x_j$ and $\tau_\x = \frac{1}{d_x}\sum_j \mathbb{E}^2x_j$, 
which correspond to the average first moment and average squared first moment of the inputs respectively (the average is across the $d_x$ dimensions).
For simplicity we assume $\x\in[0,1]^{d_x}$ and $\y\in[0,1]^{d_y}$, and so $\mu_\x \in [0,1]$ and $\tau_\x \in [0,1]$. 


Consider the following minimization performed via mini-batch stochastic gradients \citep{bottou2010large},
\begin{equation}\label{eq:minrsg} \min_{\W} \quad f(\W) := \mathbb{E}_{\x,\y} \mathcal{L}(\x,\y;\W) \end{equation}
where $\mathcal{L}(\cdot)$ denotes some loss function parameterized by $\W$ and applied to data instances $\{\x,\y\}$.
Denote $\eta := \{\x,\y\} \sim \mathcal{X}$.
The mini-batch stochastic gradient update using $B$ samples $\eta^1,\ldots,\eta^B$ and gradient stepsize $\gamma$ is 
\begin{equation}\label{eq:updatestep}
\W \leftarrow \W - \gamma G(\eta;\W) 
\end{equation}
where the gradient $G(\eta;\W)$ computed at $\W$ using the sample set $\eta^1,\ldots,\eta^B$ is given by
\begin{equation} \label{eq:update}
G(\eta;\W) = \frac{1}{B} \sum_{i=1}^B \nabla_{\W} \mathcal{L}(\eta^i;\W)
\end{equation}
Depending on $\mathcal{L}(\cdot)$, the expression in \eqref{eq:minrsg} corresponds to backpropagation learning of different classes of neural networks using stochastic gradients. 
To address many such broad families, we develop our analysis for three interesting and general classes of deep networks,
starting with single-layer networks, followed by unsupervised pretraining via box-constrained denoising autoencoders \citep{vincent2010stacked}, 
and finally multi-layer deep networks with dropout \citep{srivastava2014dropout}, both the layer-wise pretrained and fully supervised versions.
For each of these settings, the loss function $\mathcal{L}(\cdot)$ is defined below and 
additional details for these classes of networks can be found in \citep{bengio2009learning, vincent2010stacked, srivastava2014dropout}.


\begin{itemize}
\item {\bf $1$-NN}: Single-layer Network 
\begin{equation}\label{eq:loss-singlenn} \mathcal{L}(\x,\y;\W) = \| \y - \sigma(\W\x) \|^2 \end{equation}
where $\W \in \mathbb{R}^{d_y \times d_x}$. 
One can induce feature denoising into this $1$-NN learning via a dropout scheme \cite{srivastava2014dropout}.
We discuss this single-layer dropout network shortly after presenting the multi-layer loss functions.


\item {\bf DA}: Box-constrained Denoising Autoencoder 
\begin{equation}\label{eq:loss-da} \mathcal{L}(\x,\y;\W) = \| \x - \sigma(\W^T\sigma(\W(\x \circ \z))) \|^2 \quad;\quad 
\W \in [-w_m,w_m]^{d_h \times d_v} \end{equation}
where $\circ$ denotes element-wise product and $\z$ is a binary vector of length $d_x$.
Given the denoising rate $\tzeta$, the binary scalars $z_1,\ldots,z_{d_x} \sim Bernoulli(\tzeta)$ are indicators for ``nullifying'' the input layer units, 
i.e., whenever $z_i = 0$, the $i^{th}$ element of the input in forced to be equal to $0$. 
Denoting $\x \circ \z$ as $\tx$, this implies that $\tx_i = 0$ whenever $z_i = 0$. 
Here, $[-w_m,w_m]$ is the box-constraint on the unknowns $\W$, which forces the learned representations to {\em not} saturate around $0$ and/or $1$, 
and has been widely used in variants of both autoencoder design and backpropagation itself. 
The nature of this constraint is similar to other regularization schemes like \citep{bengio2012practical} and \citep{ngiam2011optimization, srivastava2014dropout}.
Although feature dropout based fully-supervised models have been shown to achieve good generalization accuracy, 
unsupervised {\it pretraining} was important in many early papers \citep{bengio2009learning, erhan2010does}. 
Here, we analyze DA not only due to its broad usage for pretraining, 
but also because of its relationship to dropout --- both schemes inject noise into the input and/or hidden features.


\item {\bf $L$-NN}: Multi-layer Network
\begin{equation}\label{eq:loss-mulnn} \begin{aligned}
\h^0 = \x &;\quad \h^l = \sigma \left( \W^l (\h^{l-1} \circ \z^l) \right) \\
\mathcal{L}(\x,\y;\W) &= \| \y - \sigma \left( \W^L (\h^{L-1} \circ \z^L) \right) \|^2
\end{aligned} \end{equation}
where $l = 1,\ldots,L-1$. Recall the dropout scheme in training deep networks which focus on the overfitting problem \citep{srivastava2014dropout}.
In each gradient update iteration, a random fraction of the hidden and/or input units are dropped based on the (given) dropout rates $\zeta_0,\ldots,\zeta_{L-1}$ \citep{srivastava2014dropout}. 
Similar to DA, the dropped out units for each layer are denoted by a set of binary vectors $\z^1,\ldots,\z^L$ such that $\z^l_i \sim Bernoulli(\zeta_l)$ for $l=1,\ldots,L$.
Within each iteration, this results in randomly sampling a smaller sub-network with approximately $\prod_{l=0}^{L-1} \zeta_l$ fraction of all the transformation parameters.
Only these $\prod_{l=1}^L \zeta_l$ fraction of weights are updated in the current iteration, while the remainder are not. Then, the re-sampling and updating process is repeated. 
We draw attention to the notation here: $\tzeta$ denotes the denoising rate, while $\zeta_0,\ldots,\zeta_{L-1}$ denote the dropout rate.
\begin{itemize}
\item A $L$-NN may be pretrained layer-wise before supervised tuning \citep{bengio2009learning, vincent2010stacked, erhan2010does}.
Here, the $L-1$ hidden layers are first pretrained, for instance, using the box-constrained DA from \eqref{eq:loss-da}. 
This gives the estimates of the $L-1$ transformations $\W^1,\ldots,\W^{L-1}$, which together with the $\y$s are then used to {\it initialize} the $L$-NN.
This `pretrained' $L$-NN is then learned using dropout as in \eqref{eq:loss-mulnn}.
This is the classical regime \citep{bengio2009learning, erhan2010does}, and we discuss this case separately from the fully-supervised version which has {\it no} layer-wise pretraining.
Several studies have already shown interesting empirical relationships between dropout and the DA \citep{wager2013dropout}, 
and we complement this body of work by providing explicit relationships between them. 
\item {\bf $1$-NN} with dropout: \\
Given $\zeta_0$, the loss function for the dropout version of $1$-NN is
\begin{equation}\label{eq:loss-singledrop} \mathcal{L}(\x,\y;\W) = \| \y - \sigma(\W(\x \circ \z^0) \|^2 \end{equation}
Although the loss here is a function of the dropout rate $\zeta^0$, we avoid including this in the notation to reduce clutter.
The two cases of $1$-NN with and without dropout are considered separately in our analysis.
\end{itemize}
Observe that backpropagation (stochastic gradients on deep networks) proceeds sequentially from the last layer to the input (or first) one 
updating the $L$ different transformations one at a time \citep{bottou2010large}.
The stepsizes used in the $l^{th}$ layer to update $\W^l$ is denoted by $\gamma^l$, 
and, we use $\W^{k,l}$ to represent the $k^{th}$ gradient update of the $l^{th}$ layer transformation $\W^l$. 
To keep the presentation simple, while discussing $L$-NN, whenever appropriate, we refer to the set of unknowns $\{\W^1,\ldots,\W^{L}\}$ simply as $\W$.
\end{itemize}


\paragraph{\it Remark:}
These three classes broadly encompass both the classical regime where the network is pretrained first using unlabeled data followed by supervised fine tuning 
\citep{vincent2010stacked, erhan2010does, erhan2009difficulty}, and the more recent fully supervised dropout learning \citep{srivastava2014dropout, krizhevsky2012imagenet}.
In general, these classes include any deep network trained using noise injection schemes \citep{matsuoka1992noise, bishop1995training, rifai2011adding}.
Further, the framework presented here is not specific to any specific choice for the non-linearity $\sigma(\cdot)$.
Nevertheless, we derive results using the sigmoid non-linearity, $\sigma(x) = 1/(1+e^{-x})$, as a running example because of its simplicity.
The results derived for $L$-NN are then used to adapt the framework to the more popular convolutional and recurrent networks 
with more sophisticated non-linearities including rectified linear units (ReLUs) \citep{nair2010rectified} or max-outs \citep{goodfellow2013maxout} that have been widely used in computer vision. 
While discussing these complex networks, we show that our analysis and the resulting implications, 
are applicable broadly with very minimal assumptions on the non-linearities.
Note that feature denoising based networks are used only as a starting point for our analysis, instead of the more complicated convolutional networks. 


\subsection{Roadmap}\label{sec:roadmap} 

The gradient update in \eqref{eq:update} is central to the ideas described in this paper. 
By {\it tracking} the behavior of these gradients as they propagate across multiple layers of the network, with minimal assumptions on the loss function $\mathcal{L}(\eta;\W)$, 
we can assess the convergence of the overall parameter estimation scheme while taking into account the influence (and structure) of each of the layers involved.
Since the updates are stochastic, ideally, we are interested in the ``expected'' gradients over a certain number of iterations, fixed ahead of time. 
Motivated by this intuition, our main idea adapted from \citep{ghadimi2013stochastic}, is to randomly sample the number of gradient update iterations. 
Specifically, let $N$ denote the maximum possible number of iterations that can be performed keeping in mind the memory and time constraints (in general, $N$ is very large). 
The stopping distribution $\mathbb{P}_R(\cdot)$ gives the probability that $k^{th}$ ($k = 1,\ldots,N$) iteration is the {\it last or stopping} iteration. 
We denote this randomly samples stopping iteration as $R \in \{1,\ldots,N\}$, and so,
\begin{equation}\label{eq:stopprob}
\mathbb{P}_R(\cdot) := \frac{p^k_R}{\sum_{k=1}^N p^k_R} \quad\text{where}\quad p^k_R = Pr(R = k) \quad;\quad k = 1,\ldots,N
\end{equation}


$\mathbb{P}_R(\cdot)$ can either be fixed ahead of time or learned from a hyper-training procedure. 
An alternate way to interpret the stopping distribution is by observing that $p^k_R$ represents the probability that the estimates ($\W$'s) 
at the $k^{th}$ iteration are the desired final solutions returned by the learning procedure. This interpretation will be used in the deriving the results. 
By proceeding with this {\it random stopping mini-batch stochastic gradients}, 
and using some Lipschitz properties of the objective and certain distributional characteristics of the input data, 
we can analyze the three loss functions in \eqref{eq:loss-singlenn}, \eqref{eq:loss-da} and \eqref{eq:loss-mulnn}.
The stopping iteration $R$ is random and the loss function in \eqref{eq:minrsg} includes an expectation over data instances $\eta = \{\x,\y\}$.
Therefore, we are interested in the {\it expectation} of the gradients $\nabla_{\W} f(\W^k)$ computed over $R \sim \mathbb{P}_R(\cdot)$ and $\eta \sim \mathcal{X}$.
This seemingly small variation to the standard backpropagation leads to gradient convergence bounds that incorporate the network depth and lengths, 
and other learning choices like dropout or denoising rate in a very natural manner. 


An important hyper-parameter of interest is the variance of $G(\eta;\W)$ in \eqref{eq:update}, which in turn depends on the number of free parameters in the given network. 
This is denoted by $e^{s}$, $e^{da}$ and $e^{m}$ for single-layer, pretraining and multi-layer cases respectively with appropriate subscripts.
Beyond these parameters, the distributional hyper-parameters including $\mu_x$ and $\tau_x$, and the denoising/dropout rate $\zeta$ along with the box-constraint 
also play a role.
Apart from the convergence bounds, we also derive sample sizes required for large deviation estimates of $\W$'s so as to marginalize the influence of the random stopping iteration. 
This estimate directly relates the dataset size to the number of backpropagation training epochs, 
where one epoch represents using each training instance at most once to compute the mini-batch gradient update from \eqref{eq:update}.
This leads to bounds on training time and the minimum required training dataset size, which in turn can be used to ensure a certain level of generalization using existing results \citep{hardt2015train}.
Our treatment for the simple case of single-layer networks is presented first, which serves as a basis for the more general settings.
The proofs for all the results are included in the appendix. 


\paragraph{\bf Why gradient norm?} 
One may ask if characterizing the behavior of the gradients is ideal or if we can do better in terms of characterizing the interplay more directly. 
There are more than a few reasons why this strategy is at least sensible. 
First, note that it is NP-hard to check local minima even for simple non-convex problems \citep{murty1987some} --- 
so, an analysis using the norms of the gradients is an attractive alternative, especially, if it leads to a similar main result. 
Second, a direct way of approaching our central question of the architecture and convergence interplay is by analyzing the gradients themselves.
Clearly, learnability of the network will be governed by the goodness of the estimates, which in turn depend on the convergence of stochastic gradients. 
In most cases, this naturally requires an asymptotic analysis of the norm, which, 
for nonconvex objectives with minimal assumptions (like Lipschitz continuity) was unavailable until very recently \citep{ghadimi2013stochastic}.
Third, working with the gradient norm directly allows for assessing faster convergence times, which, as argued in \citep{hardt2015train}, is vital for better generalization. 
Lastly, {\it post-hoc} checks about the local optimality of the solution provides some empirical evidence supporting the use of gradient norms as will be 
presented in Section \ref{sec:exps}.


\section{Single-layer Networks}
\label{sec:single}

Consider a single-layer neural network ($1$-NN) with visible ($\x$) and output ($\y$) units alone (and no hidden layers). 
Recall the objective $f(\cdot)$ from \eqref{eq:minrsg} and the corresponding loss function from \eqref{eq:loss-singlenn}.
We learn this $1$-NN via the random stopping mini-batch stochastic gradients.
The batch size is denoted by $B$ and the randomly chosen stopping iteration $R$ is sampled from the given $\mathbb{P}_R(k)$ ($k = 1,\ldots,N$). 
This learning procedure is summarized in Alg. \ref{alg:single-nn}; which we refer to as single-layer randomized stochastic gradients, {\it single-layer RSG}. 
The special case where $\zeta=1$ corresponds to using no dropout in the input layer.
Recall that $\mathcal{X}$ denotes the training dataset, $d_x$ and $d_y$ are the input and output layer lengths, and 
$\circ$ in Alg. \ref{alg:single-nn} refers to element-wise product.
$\W^1$ denotes the estimate/initialization at the first iteration, $\W^R$ is the final estimate and $\gamma^k$ represents the stepsize at the $k^{th}$ iteration.
Our first set of results correspond to this single-layer RSG from Alg. \ref{alg:single-nn}.


\begin{algorithm}
\caption{Single-layer Randomized Stochastic Gradients (Single-layer RSG)}\label{alg:single-nn}
\begin{algorithmic}
\REQUIRE $d_x$, $d_y$, $B$, $N$, $\gamma^k$, $\mathbb{P}_R(\cdot)$, $\mathcal{X}$, $\W^1$ \\  
\ENSURE $\W^R \in \mathbb{R}^{d_y \times d_x}$
\bindent
\STATE $R \sim \mathbb{P}_R(\cdot)$ \quad $z^1,\ldots,z^{d_x}=1$ \quad $\mathcal{I} = {\bf 1}_{d_h \times d_v}$
\FOR{$k = 1,\ldots,R-1$} 
\STATE $\{\x^i,\y^i\}\sim\mathcal{X}$, $i=1,\ldots,B$
\IF{dropout} 
\STATE $z_1,\ldots,z_{d_x} \sim {\rm Bernoulli}(\zeta^0)$ \quad,\quad $\mathcal{I}_{\cdot,j} = 0 \forall j \in 1,\ldots,d_v; z_j=0 \}$
\ENDIF
\STATE $\x^i \leftarrow \x^i \circ \z$, $\eta^i := \{\x^i,\y^i\}$ 
\STATE $\W^{k+1} \leftarrow \W^k - \mathcal{I} \circ \left( \frac{\gamma^k}{B}\sum_{i=1}^{B} \nabla_{\W}\mathcal{L}(\eta^i;\W^k) \right)$ \hspace{1mm}\text{where}\hspace{1mm}
$\mathcal{L}(\eta;\W)$ \hspace{1mm}\text{is from}\hspace{1mm} \eqref{eq:loss-singledrop}
\ENDFOR
\eindent
\end{algorithmic}
\end{algorithm}


With no prior information about how to setup the stopping distribution, we may want to simply choose $R$ uniformly between $1$ and $N$ i.e., $\mathbb{P}_R(k) := Unif[1,N]$.
The first result summarizes the decay of the expected gradients for this setting. 
Let $D_f = f(\W^1) - f^\ast$ denote the initial deviation of the objective from unknown optimum $\W^\ast$.
\begin{theorem}[{\bf Single-layer Network with Constant Stepsize}] \label{thm:expgrad} 
Consider a single-layer RSG with {\it no} dropout ($\zeta^0 = 1$) from Alg. \ref{alg:single-nn} and constant stepsize $\gamma^k = \gamma \forall k$.
Let $e^s_\gamma = (1 - \frac{13}{16}\gamma)$ and $R \sim Unif[1,N]$. The expected gradients are given by
\begin{equation} \label{eq:conv1nn}
\mathbb{E}_{R} (\| \nabla_{\W} f(\W^R)\|^2) \leq \frac{1}{e^s_\gamma} \left( \frac{D_f}{N\gamma} + \frac{e^s\gamma}{B} \right) 
\end{equation}
and the optimal constant stepsize is $\gamma_o = \sqrt{\frac{Bf(\W^1)}{e^sN}}$ where $e^s = \frac{13d_xd_y}{256}$.
\end{theorem}


{\it Remarks:} 
We point out that the {\em asymptotic} behavior of gradients in backpropagation, including variants with adaptive stepsizes, momentum etc. 
has been well studied \citep{magoulas1999improving, becker1988improving, castillo2006very, shao2011convergence}. 
This aspect is {\it not} novel to our work, as discussed in Section \ref{sec:intro}.
However, to our knowledge, relatively few {\it explicit} results about the convergence {\it rates} are known; 
although imposing restrictions on the objective does lead to improved guarantees \citep{ahmad1990asymptotic}.
Part of the reason may be the lack of such results in the numerical optimization literature for general nonconvex objectives.
The recent result in \citep{ghadimi2013stochastic} presents one of the first such results addressing general nonconvex objectives with a first-order oracle. 
Consequently, we believe that Theorem \ref{thm:expgrad} gives non-trivial information and, as we will show in later sections, leads to new results in the context of neural networks.
This result serves as a building block for analyzing feature denoising and multi-layer dropout later in Section \ref{sec:pretrain} and \ref{sec:multi}.  
We now explain \eqref{eq:conv1nn} briefly.


While the first term corresponds to the goodness of fit of the network (showing the influence of the deviation $D_f$), 
the second term encodes the degrees of freedom of the network (via $e^s$), and is larger for big (or fatter) networks.
Clearly, the bound decreases as $N$ (and/or $B$) increase, and it is interesting to see that the effect of network size ($d_xd_y$) is negligible for large batch sizes.
Note that the second term of the bound induces a kind of bias, which depends on the variance of the noisy gradients $e^s$ and batchsize $B$. 
As $B$ increases, the gradient update is averaged over a large number of samples, and \eqref{eq:update} will be much closer to a full-batch gradient update, 
a classical property of mini-batch stochastic gradients \citep{lecun2012efficient, bengio2012practical}. 
The constant $e^s_\gamma$ depends on the stepsize and increases the overall bound as $\gamma$ increases.
The second term in the bound suggests that larger batch sizes are necessary for fatter networks with large $d_x$ or $d_y$. 
One caveat of the generality of \eqref{eq:conv1nn} is that it might be loose in the worst case where $D_f \sim 0$ 
(i.e., when $\W^1$ is already a good estimate of the stationary point). 
The optimal constant stepsize $\gamma_o$ in Theorem \ref{thm:expgrad} is calculated by balancing the two terms in \eqref{eq:conv1nn}. 
Using this $\gamma_o$, it is easy to see that the expected gradients have the following rates.
\begin{corollary}[{\bf Rates of Single-layer Network}] \label{thm:rate1nn}
For a single-layer RSG with {\it no} dropout from Alg. \ref{alg:single-nn} with optimal constant stepsize $\gamma_o$ from Theorem \ref{thm:expgrad}, we have
\begin{equation} \label{eq:rate1nn}
\mathbb{E}_{R} (\| \nabla_{\W} f(\W^R)\|^2) := 
\begin{cases} \mathcal{O}\left(\frac{1}{\sqrt{N}}\right) \quad \text{for a given network} \\
\mathcal{O}(\sqrt{d_xd_y}) \quad \text{for a given N} \end{cases}
\end{equation}
\end{corollary}


An interesting relationship between specifying constant stepsizes and choosing a uniform stopping distribution 
can be seen from the proof of Theorem \ref{thm:expgrad} (see Appendix in Section \ref{sec:proof1}). 
Specifically, using constant stepsizes directly corresponds to choosing a uniform stopping distribution i.e., they are {\it equivalent} in some sense. 
One can nevertheless use any $\mathbb{P}_R(\cdot)$ with constant stepsizes. 
Clearly, whenever $N$ is very large, one may want to allow the stopping iteration $R$ to be as close to $N$ as possible -- a uniform $\mathbb{P}_R(\cdot)$ does not necessarily allow that.
Such alternate $\mathbb{P}_R(\cdot)$s in tandem with constant stepsizes will be discussed shortly in Corollary \ref{thm:expgrad-mono}, 
where we show that they lead to a different set of constants in \eqref{eq:conv1nn}, 
but the high-level dependence on the hyper-parameters including $N$ and the network size $d_xd_y$ will remain the same. 
Before analyzing these general stopping distributions, we first address another common setting in stochastic gradients where decaying $\gamma^k$s (as $k$ increases) are used.
Theorem \ref{thm:expgrad-noncon} summarizes the decay of the expected gradients using such monotonically decreasing stepsizes $\gamma^k = \frac{\gamma}{k^{\rho}}$ for some given $\rho>0$. 
Here, $\mathcal{H}_N(\cdot)$ is the generalized harmonic number defined as $\mathcal{H}_N(\theta) = \sum_{i=1}^N \frac{1}{i^{\theta}}$.
\begin{corollary}[{\bf Single-layer Network with Decreasing Stepsize}] \label{thm:expgrad-noncon}
Consider a single-layer RSG with {\it no} dropout from Alg. \ref{alg:single-nn} and stepsizes $\gamma^k = \frac{\gamma}{k^{\rho}}$. 
Let the probability of stopping at the $k^{th}$ iteration $p^k_R = \gamma^k (1 - \frac{13}{16}\gamma^k)$. 
The expected gradients are given by
\begin{equation} \label{eq:conv1nn-noncon}
\mathbb{E}_{R} (\| \nabla_{\W} f(\W^R)\|^2) \leq \frac{16}{3\mathcal{H}_N(\rho)} \left( \frac{D_f}{\gamma} + \frac{e^s \gamma\mathcal{H}_N(2\rho)}{B} \right)
\end{equation} 
and the optimal $\gamma$ is $\sqrt{\frac{Bf(\W^1)}{e^s\mathcal{H}_N(2\rho)}}$ where $e^s = \frac{13d_xd_y}{256}$
\end{corollary}


{\it Remarks:} 
The influence of $N$ on the bound in \eqref{eq:conv1nn-noncon} is via $\mathcal{H}_N(\rho)$. 
The interaction of $\rho$ with $N$ and other hyper-parameters, and its eventual dependence on expected gradients decay is complicated.
Nevertheless, broadly, as $\rho$ increases, the bound first decreases and eventually increases becoming looser. This trend is expected and not surprising. 
First observe that, whenever $\rho \approx 0$, the stepsizes are approximately constant (i.e, $\gamma^k \approx \gamma, \forall k$).
Here, we have $\mathcal{H}_N(\rho) \approx N$, and assuming that $\gamma < 1$, the bound in \eqref{eq:conv1nn-noncon} is at least as large as 
the bound from \eqref{eq:conv1nn} making Corollary \ref{thm:expgrad-noncon} consistent with Theorem \ref{thm:expgrad}.
Alternatively, large $\rho$ implies strong decay of the stepsizes.
For sufficiently small $\gamma$, this results in a stopping iteration probability $p^k_R$ that decreases at a fast rate as $k$ increases (see its definition from Corollary \ref{thm:expgrad-noncon}), 
i.e., large $\rho$ results in $R \ll N$.
Hence, the bound is simply implying that the expected gradients are going to be large whenever the gradient updating is stopped early.  
For a given $\rho$, the influence of network size and $B$ is the same as discussed earlier for Theorem \ref{thm:expgrad}.
Overall, these observations clearly imply that the bounds from \eqref{eq:conv1nn} and \eqref{eq:conv1nn-noncon} 
are capturing all the intuitive trends one would expect to see from using stochastic gradients, in turn, making the analysis and results more useful.


Observe that Corollary \ref{thm:expgrad-noncon} enforces a specific structure on $\mathbb{P}_R(\cdot)$, 
and $R \sim Unif[1,N]$ (from Corollary \ref{thm:expgrad}) is restrictive, in the sense that, one may want to choose $R$ as close to $N$ as possible.
The following result shows the decay of expected gradients whenever $\mathbb{P}_R(\cdot)$ is monotonically {\it increasing}, 
i.e., $p^k_R \leq p^{k+1}_R, \forall k$, thereby pushing $R$ towards $N$ with high probability.
\begin{corollary}[{\bf Single-layer Network with Monotonic $\mathbb{P}_R(\cdot)$}] \label{thm:expgrad-mono}
Consider a single-layer RSG with {\it no} dropout from Alg. \ref{alg:single-nn} and constant stepsize $\gamma^k = \gamma \forall k$.
Let $e^s_\gamma = (1 - \frac{13}{16}\gamma)$ and $p^k_R \leq p^{k+1}_R \forall k = 1,\ldots,N$. 
The expected gradients are given by
\begin{equation} \label{eq:conv1nn-mono}
\mathbb{E}_{R,\eta} (\| \nabla_{\W} f(\W^R)\|^2) \leq \frac{p^N_R}{\gamma e^s_\gamma} \left( D_f + \frac{e^sN}{B}\gamma^2 \right)
\end{equation}
and the optimal constant stepsize is $\sqrt{\frac{Bf(\W^1)}{e^sN}}$ where $e^s = \frac{13d_xd_y}{256}$
\end{corollary}
{\it Remarks:}
The uniform stopping distribution is a special case with $p^N_R = 1/N$, where the bound in \eqref{eq:conv1nn-mono} reduces to \eqref{eq:conv1nn}. 
Clearly, in the extreme case where $R=N$ with probability approaching one, \eqref{eq:conv1nn-mono} is loose. 
One can ensure that $R \approx N$ while the bound is reasonably tight by choosing $\mathbb{P}_R(\cdot)$ to have more mass on the last few iterations. 
For instance, consider the following stopping distribution for some given $\vartheta>1$.
\begin{equation} \label{eq:exampstop}
p^k_R = \begin{cases} 0 \quad \text{for} \quad k = 1,\ldots,N(1-\frac{1}{\vartheta}) \\ 
\frac{\vartheta}{N} \quad \text{for} \quad k = 1+N(1-\frac{1}{\vartheta}),\ldots,N \end{cases}
\end{equation}
Using this specification and \eqref{eq:conv1nn-mono}, we then have
\begin{equation} \label{eq:conv1nn-ex}
\mathbb{E}_{R,\eta} (\| \nabla_{\W} f(\W^R)\|^2) \leq \frac{\vartheta}{e^s_\gamma} \left( \frac{D_f}{N\gamma} + \frac{e^s\gamma}{B} \right)
\end{equation}
$\vartheta$ may be chosen so that $\mathbb{P}_R(\cdot)$ has non-zero mass over the last few hundred iterations.
For a given $N$ and $B$, the bound in \eqref{eq:conv1nn-ex} can be made as tight as possible by appropriately choosing $\vartheta$, 
thereby ensuring convergence of expected gradients of Alg. \ref{alg:single-nn}, while forcing $R$ to be as large as possible (unlike the setting from Theorem \ref{thm:expgrad}). 
The cost of enforcing such a $\mathbb{P}_R(\cdot)$ (like \eqref{eq:exampstop}) is only a slight modification over the typical stochastic gradients used to learn deep networks in practice. 
The interplay of $\vartheta$, $B$ and $N$ on the network size $d_xd_y$ is nevertheless complex, and we visualize such trends in Section \ref{sec:vis}.
Further, the broad structure of \eqref{eq:conv1nn-mono} is similar to those from \eqref{eq:conv1nn} and \eqref{eq:conv1nn-noncon}, 
and so, the inference we draw about the influence of $B$ and network size still apply here.  
Non-constant step sizes and complex stopping distributions are interesting from the modeling perspective, but, 
to ensure less notational clutter from constants while retaining the broad trends, 
we focus more on the constant stepsize setting (with uniform and monotonic $\mathbb{P}_R(\cdot)$) for the DA based pretraining and dropout $L$-NN analysis in the later sections.


\paragraph{\it Choosing $\mathbb{P}_R(\cdot)$:} \label{sec:choosingstop}
It is evident that the convergence is sensitive to the choice of $\mathbb{P}_R(\cdot)$, 
and Theorem \ref{thm:expgrad}, Corollaries \ref{thm:expgrad-noncon} and \ref{thm:expgrad-mono} cover the {\it typical} choices of stopping criteria.
From a practical stand-point, one can instead use multiple $\mathbb{P}_R(\cdot)$s from a pre-defined dictionary of distributions $\mathcal{P}$, 
and choose the best one via cross-validation on some reasonable performance measure. 
For instance, the best $\mathbb{P}_R(\cdot) \in \mathcal{P}$ can be selected based on a validation dataset 
either by directly computing the empirical average of the gradients, or using alternate measures like generalization performance. 
This is a valid operation because several recent results justify using the number of training iterations as a `surrogate' for the generalization performance \citep{hardt2015train}. 
Further, the analysis also allows probing the nature of $R$ after fixing $\mathbb{P}_R(\cdot)$. 
Specifically, depending on the Hessian at $R^{th}$ iteration, gradient updates can be continued for a few more iterations. 
The results presented here will still hold for this post-hoc increase in iterations.
The visualization of trends from the presented results (see Section \ref{sec:vis}) and additionally in Section \ref{sec:disc} 
provide more insights into these choices from an empirical perspective, including the issue of relating convergence iterations to {\it expected} generalization.
Observe that our statements about $\mathbb{P}_R(\cdot)$ hold for more pretraining and multi-layer networks including the more complex convolutional and recurrent networks, 
and so, we focus more on the architecture and other hyper-parameters when discussing these complex networks. 


Theorems \ref{thm:expgrad}, \ref{thm:expgrad-noncon} and \ref{thm:expgrad-mono} describe convergence of a single-layer network for {\em one run} of stochastic gradient. 
In practice, we are more interested in a large deviation bound over multiple runs, especially because of the randomization 
over $\eta$ and $R$ (see \eqref{eq:conv1nn}).
We define such a large deviation estimate, using $\W^{R_1},\ldots,\W^{R_T}$ computed from $T>1$ {\it independent} runs of single-layer RSG, 
and compute the minimum $N$ required to achieve such an estimate. 
\begin{definition}[{\bf $(\epsilon,\delta)$-solution}]\label{thm:epsdel}
Given $\epsilon>0$ and $0<\delta\ll 1$, an $(\epsilon,\delta)$-solution of a single-layer network is given by $\argmin_t \|\nabla_{\W} f(\W^{R_t})\|^2$ such that 
$Pr\left( \min_t \| \nabla_{\W} f(\W^{R_t}) \|^2 \leq \epsilon \right) \geq 1-\delta$.
\end{definition}
We can then state the following result for the single-layer RSG with {\it no} dropout using constant stepsize 
(the settings from Theorems \ref{thm:expgrad} and \ref{thm:expgrad-mono}).


\begin{theorem}[{\bf Single-layer RSG Computational Complexity}] \label{thm:conv1nn-mono}
Consider a single-layer RSG with {\it no} dropout. Let $\bar{\delta} = \frac{13B\epsilon\delta^{1/T}}{32e^s}$.
To compute a $(\epsilon,\delta)$-solution with $\gamma^k = \gamma$ and monotonic stopping distribution $p^k_R \leq p^{k+1}_R \forall k$ with $p^N_R = \frac{\vartheta}{N}$, Alg. \ref{alg:single-nn} needs
\begin{equation}\label{eq:iter1nn-mono}
N(\epsilon,\delta) \geq \frac{4e^sf(\W^1)}{B\delta^{2/T}\epsilon^2} \left( \bar{\delta} + \vartheta \right)^2
\end{equation}
\end{theorem}
{\it Remarks:} 
For the special case of uniform stopping distribution, $\vartheta$ would simply be $1$. 
To illustrate the practical usefulness of this result, consider an example network with $d_x=100$, $d_y=5$ and say $f(\W^1)\approx d_y$.
To compute a $(0.05,0.05)$-solution with $T=10$ runs and a batchsize $B=50$, \eqref{eq:iter1nn-mono} gives $N>7.8\times 10^3$.
If the number of epochs is $C=200$, then under the (reasonable) assumption that $SC \approx BN$ (where $S$ is sample size), the aforementioned result asks for $\sim 2000$ instances. 
Also, \eqref{eq:iter1nn-mono} implies that $N$ increases as the network size ($d_xd_y$) and the initial deviation ($f(\W^1)$) increase. 
Similarly, for obtaining good large deviation estimates (i.e., very small $\epsilon$ and $\delta$), the number of required iterations is large.
Corollary \ref{thm:conv1nn-mono} does not use any information about the input data statistics (e.g., moments), and so expectedly, \eqref{eq:iter1nn-mono} may overestimate $N$, and therefore $S$. 
Later, we make use of such data statistics to analyze networks with hidden layers, while the rest of the procedure will generalize what is presented here (for the single-layer setting).


\paragraph{\bf Single-layer RSG with Dropout:}

Recall the dropout induced single-layer network from \eqref{eq:loss-singledrop}.
In each iteration, the input units are dropped based on a given dropout rate $\zeta$, i.e., the transformation parameters for these dropped units are not updated. 
The following result summarizes the expected gradients in this setting, where a random subset of $(1-\zeta)d_xd_y$ parameters (of all the $d_xd_y$ unknowns) are updated in each iteration. 
\begin{corollary}[{\bf Single-layer Network with Dropout}] \label{thm:conv1nn-drop}
Consider a single-layer RSG with input dropout rate $\zeta$ from Alg. \ref{alg:single-nn} and constant stepsize $\gamma^k = \gamma \forall k$.
Let $e^s_\gamma = (1 - \frac{13}{16}\gamma)$, $e^s = \frac{13d_xd_y}{256}$, $\bar{\delta} = \frac{13B\epsilon\delta^{1/T}}{32e^s}$ and $R \sim Unif[1,N]$. The expected gradients are given by
\begin{equation} \label{eq:conv1nn-drop} \begin{aligned}
&\mathbb{E}_{R} (\| \nabla_{\W} f(\W^R)\|^2) \leq \frac{1}{e^s_\gamma} \left( \frac{D_f}{N\gamma\zeta} + \frac{e^s\gamma}{B} \right) \\
&N(\epsilon,\delta,\zeta) \geq \frac{4f(\W^1)e^s}{\zeta B\delta^{2/T}\epsilon^2} (1 + \bar{\delta})^2 
\end{aligned} \end{equation}
with optimal constant stepsize $\sqrt{\frac{Bf(\W^1)}{\zeta e^sN}}$
\end{corollary}
{\it Remarks:} 
Since approximately $\zeta d_xd_y$ unknowns are not updated in each gradient iteration, the dropout operation induces noise into the loss function (see \eqref{eq:loss-singledrop}).
This results in a solution path that is a noisy approximation of the non-dropout version. 
Hence, the stochastic gradients procedure needs to work `harder' in some sense, to optimize the loss: \eqref{eq:conv1nn-drop} shows this behavior via $1/\zeta$ in the first term. 
Correspondingly, in comparison to \eqref{eq:iter1nn-mono} (from Theorem \ref{thm:conv1nn-mono}), 
many more gradient iterations i.e., larger $N$, are required to compute the same $(\epsilon,\delta)$ solution. 
In the extreme case where $\zeta \approx 0$, the bound from \eqref{eq:conv1nn-drop} is loose.
Although the single-layer dropout may not be not interesting by itself from the perspective of generalization, 
this setup, and the above result, are key to multi-layer networks in Section \ref{sec:multi} where we analyze dropout in $L$-NNs.


\section{Unsupervised Pretraining}
\label{sec:pretrain}

Building upon Theorems \ref{thm:expgrad}--\ref{thm:conv1nn-drop}, we now consider single-layer networks that perform unsupervised learning. 
One widely used pretraining scheme is analyzed in detail here -- feature denoising based autoencoders (DA) \citep{vincent2010stacked}.
As discussed in Section \ref{sec:prelim}, these schemes belong to to a broader family of input/feature corruption based regularization procedures \citep{matsuoka1992noise, bishop1995training}, 
which have been widely used for layer-wise pretraining of deep networks. 
The setting we consider is a box-constrained DA (see the loss function from \eqref{eq:loss-da}).
The random stopping mini-batch stochastic gradients learning for this box-constrained DA is summarized in Alg. \ref{alg:da}. 
$P_{\W}$ denotes the Euclidean projection operator onto the constraint set $[-w_m,w_m]^{d_h \times d_x}$.
Unlike the single-layer setting, because of the constraint on $\W$, our interest now is in the expected {\it projected} gradients denoted by $\nabla_{\W} \tf(\W)$, 
which simply correspond to the Euclidean projection of $\nabla_{\W} f(\W)$ on $\W \in [-w_m,w_m]^{d_h \times d_x}$.


\begin{algorithm}
\caption{Box-constrained Denoising autoencoder Randomized Stochastic Gradients (DA RSG)}\label{alg:da}
\begin{algorithmic}
\REQUIRE $d_x$, $d_h$, $B$, $N$, $\gamma^k$, $\tzeta$, $w_m$, $\W^1$, $\mathbb{P}_R(\cdot)$ \\
\ENSURE $\W^R \in \mathbb{R}^{d_h \times d_x}$
\bindent
\STATE $R \sim \mathbb{P}_R(\cdot)$;
\FOR{$k = 1, \ldots R-1$} 
\STATE $z^i_1,\ldots,z^i_{d_x} \sim {\rm Bernoulli}(\tzeta)$
\STATE $\x^i \sim \mathcal{X}$, $\eta^i := \{\x^i \circ \z^i\}$, $i=1,\ldots,B$  
\STATE $\W^{k+1} \leftarrow P_{\W} (\W^k - \frac{\gamma^k}{B}\sum_{i=1}^{B} \nabla_{\W}\mathcal{L}(\eta^i;\W))$ \hspace{1mm}\text{where}\hspace{1mm} 
$\mathcal{L}(\eta;\W)$ \hspace{1mm}\text{is from}\hspace{1mm} \eqref{eq:loss-da}
\ENDFOR
\eindent
\end{algorithmic}
\end{algorithm}


As mentioned in the remarks of Corollary \ref{thm:expgrad-mono}, we restrict the discussion to constant stepsizes, first for the uniform stopping distribution and later for the more general one, 
to simplify our discussion, 
while noting that the trends remain the same even with alternate stepsizes and choices of $\mathbb{P}_R(\cdot)$.
The following result bounds $\nabla_{\W} \tf(\W)$ for DA RSG procedure from  Alg. \ref{alg:da}.
$D_f$ denotes the initial deviation from the optimum, while $e^{da}$ and $e^{da}_\gamma$ are the hyper-parameters representing the network structure and learning constants.
\begin{theorem}[{\bf DA with constant stepsize}] \label{thm:expgradda}
Consider the DA RSG with $\W \in [-w_m,w_m]^{d_h\times d_x}$. Let
\begin{equation} \label{eq:convdaparam-eda}
e^{da} = \frac{d_xd_h}{16} \left[ 1 + \frac{\tzeta d_xw_m}{4}\mu_{\x} + \left(\frac{5\tzeta}{16} - \frac{\tzeta^2}{4}\right)(\tzeta d_xw_m)^2 \tau_{\x} \right] 
\end{equation}
and $\uda$ denote the Lipschitz constant of $\nabla_{\W} f(\W)$ for the loss from \eqref{eq:loss-da}.
Using constant stepsize $\gamma<\frac{2}{\uda}$ with $R\sim Unif[1,N]$ and denoting $e^{da}_\gamma = 1 - \frac{\uda}{2}\gamma$, the expected projected gradients is 
\begin{equation} \label{eq:convda}
\mathbb{E}_R (\| \nabla_{\W} \tf(\W^R)\|^2) \leq \frac{D_f}{N\gamma e^{da}_\gamma} + \frac{e^{da}}{B} \left( 1+\frac{1}{e^{da}_\gamma}\right) 
\end{equation}
and the optimal stepsize is $\sqrt{\frac{2Bf(\W^1)}{\uda e^{da}N}}$.
\end{theorem}


{\it Remarks:} 
Similar to the bound from \eqref{eq:conv1nn} for the single-layer RSG, the above result in \eqref{eq:convda} combines the contributions from 
the output goodness of fit ($D_f$), the number of free parameters ($d_hd_x$) and the stepsize choice ($\gamma$, $e^{da}_\gamma$).
Here, $D_f$ is balanced out by the number of iterations $N$ in the first term.
The second term involving the variance of gradients ($e^{da}$) and the batchsize $B$ controls the `bias' from the network's degrees of freedom -- 
the remarks from Theorem \ref{thm:expgrad} still apply here.
However, here we find that the dependence on the network structure is more involved. 
The input and hidden layer lengths do not contribute equally (see $e^{da}$ from \eqref{eq:convdaparam-eda}) 
which can be partly explained by the structure of the loss function (see \eqref{eq:loss-da}).
For smaller constraint sets i.e., small $w_m$, and therefore small $e^{da}$, we expect the projected gradients to typically have small magnitude, which is clearly implied by \eqref{eq:convda}.
In practice, $w_m$ is reasonably small, and, in general, 
several studies have shown that $W_{ij}$'s tend to emulate a `fat' Gaussian distribution with mean approximately around zero \citep{bellido1993backpropagation, blundell2015weight}. 
All the interpretations about the influence of stopping distributions, and the viable choices to ensure faster convergence, 
that were discussed in Section \ref{sec:single} in the context of single-layer RSG still hold true for DA RSG. 
Hence, we focus more on the DA-specific hyper-parameters including the data moments.


\subsection{From Denoising Rates and Network structure to \eqref{eq:convda}}
\label{sec:da-net-conv}
 
We see that \eqref{eq:convdaparam-eda} and its resulting structure in the bound from \eqref{eq:convda} seems to imply a non-trivial interplay between the network size $d_hd_x$, 
the data statistics (via $\mu_\x$ and $\tau_\x$ where $\mu_\x = \frac{1}{d_x}\sum_j \mathbb{E}x_j$ and $\tau_\x = \frac{1}{d_x}\sum_j \mathbb{E}^2x_j$) and the denoising rate $\tzeta$.
A direct observation from the bound in \eqref{eq:convda} is that it is always useful to use smaller (or thinner) networks, 
which results in faster decay of expected gradients as iterations increase.
This trend, in tandem with observations from \cite{hardt2015train} about generalization imply the superiority of thinner networks compared to fatter ones -- 
\cite{romero2014fitnets} and others have already shown evidence of this behavior.
As was observed with single-layer networks from \eqref{eq:conv1nn}, fatter DAs will need a large batchsize $B$. 
We further discuss this trade-off of thinner versus fatter networks later in Section \ref{sec:multi} when presenting multi-layer networks
(sections \ref{sec:exps} and \ref{sec:vis} show some interesting plots summarizing the interplay from \eqref{eq:convda}).
Beyond these visualizations and evaluations, we intend to interpret this interaction and its corresponding influence on the decay of \eqref{eq:convda} for a few of the commonly encountered cases.  
Here, we assume that the batchsize $B$ and the maximum allowable iterations $N$ are reasonably large.


\begin{itemize} [leftmargin=1em] 
\item {\bf Small data moments, \mathversion{bold}{\it $\mu_\x\approx 0$, $\tau_\x\approx 0$}:\mathversion{normal}}  

In this trivial case, 
the signal is very weak and there may not be much to learn. 
Clearly, $e^{da}$ will be close to its minimum which will directly result in faster convergence in terms of the expected projected gradients.  
Recall that smaller $\tzeta$ implies more noisy gradients (across multiple iterations). 
However, $\tzeta$'s influence on the bound is nullified because $\mu_\x ,\tau_\x \approx 0$.
This implies that independent of the complexity of the task being learned, there is no reason for using large denoising rates whenever the average of input across all features is small. 
This is sensible because the small irregularities, in the form of noise or dampening of inputs, 
which would be induced by the DA corruption process might in fact correspond to class-specific changes in the data. 
So, using a smaller $\tzeta$ will require many more gradient iterations to convergence to some solution (as implied by \eqref{eq:convda}), 
while also resulting in poor hidden representations. With larger $\tzeta$, the only contributing factor for convergence is the network size.


\item {\bf Small denoising rate, \mathversion{bold}{\it $0 \ll \tzeta < 1$}:\mathversion{normal}} 

When $\tzeta$ is large, the input corruption is very small and $\x \circ \z \approx \x$ (refer to the loss from \eqref{eq:loss-da}, $\circ$ denotes element-wise product).
Here, the noise in the gradients $\nabla_{\W}\mathcal{L}(\eta;\W)$ is almost entirely governed by the data statistics. 
Within this setting, 
\begin{itemize} [leftmargin=1em]
\item {\it Small data moments:} 
Small $\mu_\x$ and $\tau_\x$ can lead to faster convergence, and as they increase, the bound becomes loose. 
The trend here is similar to small data moments case discussed earlier.
\item {\it Large data moments:} 
When $\mu_\x$ and $\tau_\x$ are reasonably large however, the bound can be controlled only if stepsize is small enough, with a large batchsize and reasonably small lengths of inputs.
This setting is closer to the classical non-regularized autoencoder, and hence, the representations learned might not be useful -- 
\eqref{eq:convda} predicts this same behavior, albeit from a parameter convergence perspective. 
To see this, observe that faster convergence time is desired for better generalization \cite{hardt2015train}. 
Such faster times cannot be guaranteed, here, whenever the network is reasonably large (even with large $d_x$ and small $d_y$) with large data moments, 
because the number of gradient iterations will need to be correspondingly large according to \eqref{eq:convda}. 
Clearly, if the convergence bound is forced to be very small, the resulting representations are almost surely overfitting the inputs.
Note that in the pathological case where $\mu_\x \approx 1$, $\tau_\x \approx 1$ with $\tzeta \approx 1$, the bounds are too loose to be relevant.
\end{itemize}


\item {\bf Large denoising rate, \mathversion{bold}{\it $0 < \tzeta \ll 1$}:\mathversion{normal}}  

In the opposite scenario, with small a $\tzeta$, we see that $e^{da}$ is going to be close to $\frac{d_xd_h}{16}$. 
\begin{itemize} [leftmargin=1em]
\item {\it Not-so-large data moments:} 
Whenever the data moments are not too large, their influence is almost surely mitigated because of the presence of large corruptions induced by small $\tzeta$ (see \eqref{eq:loss-da}).
This is the `good' DA regime -- where the aim is to perturb the data without necessarily destroying {\it all} the class-specific information. 
As suggested by the interplay in \eqref{eq:convda}, there might be good combinations of $d_h$, $B$, $\gamma$ and $N$ that will lead to good decay of expected gradients 
while also generating very useful representations -- these trends have been analyzed empirically in \cite{vincent2010stacked, erhan2010does}.
\item {\it Large data moments:} 
If the data moments are quite large ($\approx 1$), the convergence is almost entirely controlled by $\tzeta$, 
which in turn will be faster for thinner networks with large $B$ (the only other hyper-parameters playing a non-trivial role within this setting).
Here, the trends are similar to the large data moments regime discussed earlier with small $\tzeta$ mainly because $e^{da}$ cannot be controlled anymore. 
The worst scenario is when $\mu_\x \approx 1$, $\tau_\x \approx 1$, in which case the improvement resulting from a small last term in $e^{da}$ will be compensated by the second term, 
naturally leading to a loose bound.
\end{itemize}


\item {\bf Size of constraint set, \mathversion{bold}{Choice of $w_m$}:\mathversion{normal}} 

For small $w_m$, the projected gradients are expected to have small magnitude, 
which is intuitive because the magnitude of the projected gradients will be bounded by the diameter of the constraint set. 
Notice that \eqref{eq:convda} also predicts this behavior via the structure of $e^{da}$, which is quadratic in $w_m$ and decreases as $w_m$ decreases.  
On the other hand, as one increases the set size $w_m$, for a given network size and $\tzeta$, we are clearly forced to increase $N$ and/or $B$ correspondingly. 
Since the choice of $w_m$ is independent of data statistics and other hyper-parameters, 
all the earlier statements (for the three other cases) about the distributional and learning constants including network sizes will come into play as $w_m$ increases.
\end{itemize} 


Beyond these prototypical settings, for small stepsizes $\gamma$, $e^{da}_\gamma$ will be as large as possible making the bound tighter. 
Clearly, the influence of both data moments and $\tzeta$ on the decay of gradients will be mitigated by increasing $B$ and $N$ alone, but, 
as suggested in \cite{hardt2015train}, this will be at the cost of poorer generalization. 
The above discussed regimes are some of the widely used ones, but the interplay of all the terms used in Theorem \ref{thm:expgradda} and \eqref{eq:convda} is more involved. 
There are certain situations where \eqref{eq:convda} (and \eqref{eq:conv1nn}, as well, from Theorem \ref{thm:expgrad}) will be loose. 
For instance, across all regimes, the case of $\mu_\x \approx 1$ and $\tau_\x \approx 1$ seems to be pathological. 
In such a case, the data dimensions may be un-informative and highly correlated to begin with.
Similarly, as we keep increasing $d_x$ and $d_h$ (or $d_y$ for $1$-NN) to very large values, at some point, the bounds will become loose but depending on $\gamma$, $\tzeta$ and data moment, 
it may be the case that the learned representations are still meaningful.
Further, $e^{da}_\gamma$ decreases as $\gamma$ increases, and when the stepsize becomes too large (beyond reasonable values), 
the resulting bounds are not very informative even if empirically the architecture may still work well. 
Finally, \eqref{eq:convda} is less useful for the unconstrained case. We note that, our framework does handle this, 
but the corresponding analysis may not provide any new insights on the interplay of structure and learning beyond what is predicted by Theorem \ref{thm:expgradda}.


Recall the discussion about faster convergence time versus generalization performance from Section \ref{sec:prelim} and also \cite{hardt2015train}.
Using the above {\it guidelines} one can ensure smallest possible $N$ for good generalization.
For completeness, the following result presents the decay of expected gradients for DA RSG using constant stepsize with 
monotonically increasing $\mathbb{P}_R(\cdot)$ (the setting from Corollary \ref{thm:expgrad-mono}) with $p^k_R \leq p^{k+1}_R \forall k$. 
\begin{corollary}[{\bf DA with Monotonic $\mathbb{P}_R(\cdot)$}] \label{thm:expgradda-mono}
Consider the DA RSG with $\W \in [-w_m,w_m]^{d_h\times d_x}$. 
Let $e^{da}$ be given by \eqref{eq:convdaparam-eda} and $\uda$ denote the Lipschitz constant of $\nabla_{\W} f(\W)$ with loss from \eqref{eq:loss-da}.
Using constant stepsize $\gamma<\frac{2}{\uda}$ with $p^k_R \leq p^{k+1}_R \forall k$ and denoting $e^{da}_\gamma = 1 - \frac{\uda}{2}\gamma$, the expected projected gradients is 
\begin{equation} \label{eq:convda-mono}
\mathbb{E}_R (\| \nabla_{\W} \tf(\W^R)\|^2) \leq \frac{p^N_RD_f}{\gamma e^{da}_\gamma} + \frac{e^{da}}{B} \left( 1+\frac{p^N_RN}{e^{da}_\gamma} \right)
\end{equation}
and the optimal stepsize is $\sqrt{\frac{2Bf(\W^1)}{\uda e^{da}N}}$
\end{corollary}

{\it Remarks:}
Under the special case where $\mathbb{P}_R(\cdot)$ is uniform, the bound in \eqref{eq:convda-mono} reduces to \eqref{eq:convda}. 
Also as observed earlier with Corollary \ref{thm:expgrad-mono}, 
one can ensure that $R \approx N$ in tandem with a tighter (and the smallest possible) bound for the given $N$ and $B$ by choosing $\mathbb{P}_R(\cdot)$ as in \eqref{eq:exampstop}, 
where more mass is accounted by the last few iterations. 
Using \eqref{eq:exampstop}, \eqref{eq:convda-mono} reduces to (recall that $\vartheta > 1$)
\begin{equation} \label{eq:convda-mono1}
\mathbb{E} (\| \nabla_{\W} \tf(\W^R)\|^2) \leq \frac{\vartheta D_f}{N\gamma e^{da}_\gamma} + \frac{e^{da}}{B} \left( 1+\frac{\vartheta}{e^{da}_\gamma} \right)
\end{equation}
The cost of enforcing such a $\mathbb{P}_R(\cdot)$ is just a minor modification on DA learning procedures (see Alg. \ref{alg:da}). 
Several other observations made earlier about the viable and interesting choices of $\mathbb{P}_R(\cdot)$ (see Remarks of Corollary \ref{thm:expgrad-mono}, and Section \ref{sec:choosingstop}) 
still hold true for the above result, and hence for noise-injection based autoencoder learning in general. 
Finally, the following result presents the computational complexity of DA RSG,
similar to Theorem \ref{thm:conv1nn-mono} which computes the number of gradient iterations $N$ required for estimating a $(\epsilon,\delta)$ solution
(see Definition \ref{thm:epsdel}) for single-layer RSG.
Here, we assume that $SC \approx BN$ where $S$ is the sample size (training dataset size) and $C$ denotes the number of epochs.
Note that the same assumption was used in the Remarks of Theorem \ref{thm:conv1nn-mono} to discuss some example sample sizes. 

\begin{theorem}[{\bf DA RSG computational Complexity}] \label{thm:convda}
Consider a DA RSG learned with constant stepsize $\gamma$, denoising rate $\tzeta$ and uniform stopping iteration. 
To compute a $(\epsilon,\delta)$-solution, we need
\begin{equation}\label{eq:itersampda}
B \geq \frac{2 e^{da}}{\epsilon \delta^{1/T}} \left( 1+\frac{1}{e^{da}_\gamma} \right) \text{and for such a choice of batchsize} \hspace{2mm}  
N \geq \max\left\{ \frac{SC}{B}, \frac{2d_x}{\gamma e^{da}_\gamma \epsilon \delta^{1/T}} \right\}
\end{equation}
where $C$ is the number of backpropagation epochs and $S$ is the number of unsupervised training instances available.
\end{theorem}

{\it Remarks:} The above result provides a recipe for setting the batchsize and maximum number of iterations for the DA RSG. 
As the network size increases, $e^{da}$ increases which in turn implies larger batchsize $B$, and larger input size $d_x$ increases the maximum iterations $N$.
Observe that $f(\W^1)$ is replaced by $d_x$ there by making \eqref{eq:itersampda} computable in practice.
Once $B$ and $N$ are fixed using \eqref{eq:itersampda}, the $T$ different random stopping iterations corresponding to the $T$ different DA RSGs can be chosen independently.
This result will be used in one of the steps of the design procedures presented in Section \ref{sec:disc}.


\section{Multi-layer Networks}
\label{sec:multi}

We now extend our analysis to multi-layer networks. 
Using Theorem \ref{thm:expgradda} as a starting point, which analyzes a single hidden layer network using feature denoising, 
we first consider a $L$-layered multi-layer network that performs {\it layer-wise pretraining} {\em before} backpropagation based supervised finetuning.
As discussed in Section \ref{sec:prelim} while discussing the $L$-NN loss in \eqref{eq:loss-mulnn}, 
this two-fold learning mechanism corresponds to the classical deep learning regime where large amounts of unsupervised data is used to initialize the network transformations.
The resulting bound allows us to incorporate feature dropout in both the input and hidden layers during the supervised tuning stage, 
and eventually leads to our final result on the most general case of dropout learning for $L$-NNs.
A consequence of this general result is a clear and intuitive relation between dropout based supervised learning, layer-wise pretraining and other structural and distributional parameters.
Alg. \ref{alg:multi-nn} presents the random stopping stochastic gradients for multi-layer networks, referred to as multi-layer RSG.


\subsection{Layer-wise Pretraining}
\label{sec:multipretrain}

The $L-1$ hidden layers are pretrained using box-constrained denoising autoencoders from Alg. \ref{alg:da}, 
resulting in estimates of the unknowns $\W^1,\ldots,\W^{L-1}$ denoted by $\W^{1,1},\ldots,\W^{1,L-1}$ 
(the superscripts $(1,\cdot)$ simply correspond to the first iteration of $L$-NN learning, see description of $L$-NNs from Section \ref{sec:prelim}).
These estimates, along with a random estimate of $\W^L$, denoted by $\W^{1,L}$, will then initialize the $L$-NN. 
The final estimates are then learned via supervised back-propagation using $\{\y\}$s and the loss \eqref{eq:loss-mulnn}.
For notational convenience we collectively denote the final estimate $\W^{R,1},\ldots,\W^{R,L}$ simply as $\W^R$ in the results. 
Since the pretraining stage uses box-constraints, we are still interested in the expected projected gradients accumulated across all the layers.
The following result shows the decay of these expected projected gradients.
$D_f$ here denotes the initial deviation of the objective from \eqref{eq:loss-mulnn} {\it after} the $L-1$ layers have been pretrained.
Similar to $e^{da}$ from Theorem \ref{thm:expgradda}, $e^m_l$s for $l=1,\ldots,L$ encode the structural and learning hyper-parameters of the network. 
We assume that all the hidden layers are pretrained to the same degree i.e., each of them is pretrained to a given $(\alpha,\delta_{\alpha})$ solution (see Definition \ref{thm:epsdel}).


\begin{algorithm}
\caption{Multi-layer Randomized Stochastic Gradients (Multi-layer RSG) -- Dropout with/without layer-wise pretraining}\label{alg:multi-nn}
\begin{algorithmic}
\REQUIRE $d_1,\ldots,d_L$, $B$, $N$, $\gamma_1^k,\ldots,\gamma_L^k$, $\mathbb{P}_R(\cdot)$, $\mathcal{X}$, $w^1_m,\ldots,w^{L-1}_m$, $\W^{1,L}$, $\zeta_0,\ldots,\zeta_{L-1}$ \\  
\ENSURE $\W^{R,l} \in \mathbb{R}^{d_{l+1} \times d_l}$ for $l = 1,\ldots,L-1$
\bindent
\FOR{$l = 1,\ldots,L-1$} 
\IF{pretrain} 
\STATE $\W^{1,l} \leftarrow$ Algorithm \ref{alg:da} 
\ELSE 
\STATE$\W^{1,l} \leftarrow$ Random Initialization \ENDIF \ENDFOR
\STATE $R \sim \mathbb{P}_R(\cdot)$; \quad $\mathcal{I}^l = {\bf 1}_{d_{l+1} \times d_{l}}$ for $l = 1,\ldots,L-1$
\FOR{$k = 1, \ldots R-1$}  
\STATE $\{\x^i,\y^i\}\sim\mathcal{X}$
\IF{dropout} 
\STATE $z^l_1,\ldots,z^l_{d_x} \sim Bernoulli(\zeta_l)$ \quad $l = 0,\ldots,L-1$ 
\STATE $\mathcal{I}^l_{\cdot,k} = 0$ $\forall$ $k = \{i \in 1,\ldots,d_{l}; z^l_i=0 \}$ 
\ENDIF
\STATE $\x \leftarrow \x * \z^0$ \quad $\h^l \leftarrow \h^l\ \circ \z^l$ 
\STATE $\eta^i := \{\x^i,\y^i\}$
\STATE $\W^{k+1,L} \leftarrow \W^{k,L} - \mathcal{I}^L \circ \frac{\gamma_L^k}{B}\sum_{i=1}^{B} \nabla_{\W^l}\mathcal{L}(\eta^i;\W^{k,l})$
\FOR{$l = L-1,\ldots,1$}
\STATE $\nabla^{pq}_{\W^l}\mathcal{L}(\eta^i;\W^l) = h^{l-1}_q h^{l}_p(1-h^{l}_p) \langle P^{\cdot p}_{\W^{l+1}}(\eta^i;\W^{k+1,l+1}), \W^{k+1,l+1}_{\cdot p} \rangle$ \hspace{1mm} $\forall p$,$q$ 
\STATE $\W^{k+1,l} \leftarrow P_{\W^l} (\W^{k,l} - \mathcal{I}^l \circ \frac{\gamma_l^k}{B}\sum_{i=1}^{B} \nabla_{\W^l}\mathcal{L}(\eta^i;\W^{k,l}))$ \ENDFOR \ENDFOR
\eindent
\end{algorithmic}
\end{algorithm}


\begin{theorem}[{\bf Multi-layer Network}] \label{thm:expgradmulnn}
Consider a multi-layer RSG with {\it no} dropout from Alg. \ref{alg:multi-nn} learned via box-constrained layer-wise pretraining from Alg. \ref{alg:da} 
and supervised backpropagation with constant stepsizes $\gamma^l\forall l$.
Let $e^m_l = \frac{\gamma^l}{4}d_{l-1}d_{l}d_{l+1}w^l_m$ for $l=1,\ldots,L-1$ and $e^m_L = \frac{13d_{L-1}d_L\gamma^2_L}{256}$.
Whenever $\gamma_l < \frac{20}{\alpha d_{l+1}w^l_m}$, and the hidden layers are pretrained for a $(\alpha,\delta_{\alpha})$ solution (from Definition \ref{thm:epsdel}), we have 
\begin{equation} \label{eq:convmulnn}
\mathbb{E}_R \| \nabla_{\W} \tf(\W^R) \|^2  \leq \frac{1}{e^m_\gamma} \left( \frac{D_f}{N} + \frac{1}{B} (e^m_L + \alpha\sum_{l=1}^{L-1} e^m_l) \right)
\end{equation}
where $e^m_\gamma = \min \left\{ \gamma_L - \frac{13}{16}(\gamma_L)^2, \gamma_l - \frac{\alpha d_{l+1}w^l_m}{20}(\gamma_l)^2 \right\}$ for $l=1,\ldots,L-1$
\end{theorem}


{\it Remarks:} 
The structure of the bound in \eqref{eq:convmulnn} is, in principle, similar to those from \eqref{eq:conv1nn} and \eqref{eq:convda}.
Hence, the trends suggested by the interplay of the hyper-parameters -- the deviation $D_f$, the network depth $L$ and sizes ($d_0,\ldots,d_L$) including stepsizes, $N$ and $B$ 
for learning $L$-NN are clearly similar to those observed from the Theorem \ref{thm:expgrad} and Corollary \ref{thm:expgradda} (from \eqref{eq:conv1nn} and \eqref{eq:convda} respectively). 
Specifically, the discussions about denoising rates and data statistics, and their role in parameter convergence will still apply here for the layer-wise pretraining stage. 
It is interesting to see that the output layer influence (the second term in \eqref{eq:convmulnn}) needs to be completely compensated by the batchsize. 
Note that this bias is, in principle, similar to the terms in \eqref{eq:conv1nn} and \eqref{eq:convda}.
Also \eqref{eq:convmulnn} uses the same $(\alpha,\delta_{\alpha})$ for all layers i.e., all the $L-1$ hidden layers are pretrained to the same $(\alpha,\delta_{\alpha})$ solution. 
Instead, one can use different $\alpha_l$s for each of these layers. 
The modified decay of gradients bound, which follows from simple adjustments in the steps involved in constructing \eqref{eq:convmulnn} (see Section \ref{sec:proof11}) is,
\begin{equation} \begin{aligned} \label{eq:convmulnn-gen}
\mathbb{E}_R \| \nabla_{\W} \tf(\W^R) \|^2  \leq \frac{1}{e^m_\gamma} \left( \frac{D_f}{N} + \frac{1}{B} (e^m_L + \sum_{l=1}^{L-1} \alpha_le^m_l) \right) \\
\text{with}\quad\quad e^m_\gamma = \min \left\{ \gamma_L - \frac{13}{16}(\gamma_L)^2, \gamma_l - \frac{\alpha_ld_{l+1}w^l_m}{20}(\gamma_l)^2 \right\}
\end{aligned} \end{equation}
This setting is of interest when comparing two networks (in Section \ref{sec:compare}) and later in Section \ref{sec:disc} where we discuss the design choice problem.
Because of the presence of composition of multiple layers, the interplay of network structure and decay of expected projected gradients is much more involved 
than for $1$-NN or DA. 
The easiest way to interpret them is, clearly, by visualizing the trends, as will be presented in Section \ref{sec:vis}, but following the presentation in Section \ref{sec:pretrain}, 
we discuss some typical $L$-NN settings of interest before moving on to dropout.


\subsection{Goodness of Pretraining}
\label{sec:goodpretrain}

We see from Definition \ref{thm:epsdel} and Theorem \ref{thm:expgradmulnn} that $\alpha$ and $\delta_{\alpha}$ control the goodness of layer-wise pretraining.
Hence, the influence of the $L-1$ DAs, and the relevant hyper-parameters (from Theorem \ref{thm:expgradda}), are subsumed within $\alpha$, $\delta_{\alpha}$ and $D_f$.
For fixed network lengths and stepsizes, $e^m_l$ are constants and encode the variance of gradients within the $l^{th}$ layer and the corresponding free parameters.
As implied by the terms in \eqref{eq:convmulnn}, $\alpha \approx 0$ will ensure that the backpropagation is not influenced by the structure of these $L-1$ pretrained layers. 
With such a small $\alpha$, consider two different learning regimes depending on the deviation of the pretrained network $D_f$. 
\begin{itemize} [leftmargin=1em]
\item {\bf Small \mathversion{bold}{$D_f$}\mathversion{normal}:} 
If the deviation is small, then the supervised tuning is mostly confined to mapping $\h^{L-1}$ to the outputs $\y$.
This is the {\it ideal} multi-layer NN setting where the (learned) higher level representations are already very good (even before tuning them to the corresponding $\y$s) 
in predicting the output labels. 
Here, \eqref{eq:convmulnn} predicts that both $B$ and $N$ may be reasonably small, and under the assumption that $SC \approx BN$ (which was used in the Remarks of Theorem \ref{thm:conv1nn-mono}), 
this implies that a small number of labeled instances are sufficient. 
However, forcibly restricting the backpropagation to only learning the last layer is not necessarily good, 
and one may instead want to allow for all $\h^l$s ($l= 1,\ldots,L-1$) to change if needed \citep{vincent2010stacked, saxe2011random}. 
\item {\bf \mathversion{bold}{Large $D_f$}\mathversion{normal}:}
In the alternate setting with large $D_f$, although the pretrained $\h^l$s are good enough abstract representations of the inputs, the desired outputs are not well modeled yet. 
In such a case, aggressive tuning is needed, as suggested by \eqref{eq:convmulnn}, with large $N$ and $B$. 
\end{itemize}


Overall, \eqref{eq:convmulnn} implies that the goodness (or the lack thereof) of pretraining will be passed on to the tuning stage convergence via $\alpha$ and $D_f$.
Clearly, as network size increases, $e^m_l$s increase proportionally, thereby requiring large $N$ and batchsizes $B$. 
Recently, \cite{yosinski2014transferable} showed empirical evidence that aggressive pretraining (i.e., $\alpha \approx 0$), especially in higher layers, 
results in $\h^l$s that cannot be translated across multiple arbitrary tasks. 
Theorem \ref{thm:expgradmulnn} suggests the same -- to allow for aggressive tuning instead of aggressive pre-training, $\alpha$ and $D_f$ should not be small. 
So, if the goal is to generate representations that are {\it general enough} to be applicable to a wide-variety of related, but reasonably distinct tasks, 
then it makes sense to perform weak pretraining with large $\alpha \not\approx 0$, while increasing the supervised gradient iterations $N$.


\subsection{Comparing arbitrary networks}
\label{sec:compare}

\cite{yosinski2014transferable} and others have investigated empirically whether the lower level representations or the higher ones are better translatable across tasks.
From the convergence perspective, Theorem \ref{thm:expgradmulnn} provides a clear answer to this, including other open questions listed in Section \ref{sec:intro}, 
for instance, whether two given deep architectures are {\it equivalent} and when can one truncate the hidden layers without compromising the expected gradients (and eventual parameter estimation).
Very recently, \cite{wei2016network} addressed such equivalence and morphism properties of deep networks. 
To pursue these questions, consider a more general setting where the $L-1$ hidden layers are trained to $\alpha_l$ ($l=1,\ldots,L-1$) different goodness levels.
Such a setting with different $\alpha_l$s is only interesting when comparing two networks, and therefore, 
was not used for Theorem \ref{thm:expgradmulnn} and the discussion in Section \ref{sec:goodpretrain}.
Without loss of generality, let $\gamma^l = \gamma$, $w^l_m = w \forall l$ and $D_f \approx d_L$ (since $\y \in [0,1]^{d_L}$). 
With this setup, a deep network $\mathcal{D}$ is parameterized by the remaining hyper-parameters $(\alpha_l,d_l)$s, $B$, $N$ and the stopping distribution $\mathcal{P}_R(\cdot)$.


Consider learning two different deep networks $\mathcal{D}_1$ and $\mathcal{D}_2$ for datasets sampled from an underlying data generating distribution 
i.e., for a fixed $d_0$ and the data statistics ($\mu_\x$ and $\tau_\x$). 
Clearly, when comparing different networks that learn the same concept, it is reasonable to estimate the difference in objectives by comparing their generalization errors.
However, in addition, we look at the parameter estimation (and the resulting convergence) aspect as well. So, 
we instead use the bound in \eqref{eq:convmulnn} to decide whether $\mathcal{D}_1$ and $\mathcal{D}_2$ are {\it close enough} in terms of learning the underlying concept. 
While presenting the design principles (Section \ref{sec:disc}) and resulting evaluations on designing networks for a given task 
we discuss comparing the learning objectives and generalization error. We have the following definition.
\begin{definition}[{\bf Parameter convergence of two deep networks}] \label{thm:equiv}
Two arbitrarily different networks $\mathcal{D}_1$ and $\mathcal{D}_2$ learned on two datasets sampled from some data generating distribution are {\it equivalent} in terms of parameter convergence, 
denoted by $\mathcal{D}_1 \equiv_e \mathcal{D}_2$, whenever the bound in \eqref{eq:convmulnn} is the same for both. 
\end{definition}
Our discussion of which families of networks offer no better convergence guarantees over the other will be driven by the above definition. 
It should nevertheless be noted that, whenever \eqref{eq:convmulnn} is loose, such insights might not hold -- we discuss these issues in Section \ref{sec:disc}.


\subsubsection{Different networks of same depth} \label{sec:samedepth}

There are many different settings for $\alpha_l$s, $d_l$s and other hyper-parameters for which two equal depth networks will be equivalent (according to Definition \ref{thm:equiv}). 
The following result summarizes two such general families of networks that guarantee the same decay of expected gradients, pointing out a few special cases of interest. 
To restrict the comparison to pretraining and structural aspects, we fix $B$, $N$ and $\mathcal{P}_R(\cdot)$ across different networks (including the nature/type of activation functions used).
For this result, we fix the output layer and the hidden layer that feeds into it ($d_{L-1}$ and $d_L$), which will be relaxed later when comparing dropout networks in Section \ref{sec:samedepth-drop}.
\begin{corollary}[{\bf Equivalence of networks}] \label{thm:equiv-pretrain}
Consider the family of depth $L$ deep networks that learn datasets from an underlying distribution (i.e., given $d_0$, $\mu_\x$ and $\tau_\x$) 
using multi-layer RSG with no dropout and a given $B$, $N$, $\gamma_l$, $w^l_m$ and stopping iteration $R$. 
All such networks $\mathcal{D} := (\alpha_l,d_l)$, with $\alpha_ld_{l+1} < \frac{65}{4}$ for $l=1,\ldots,L-1$ and 
\begin{equation} \label{eq:equiv-pretrain}
\alpha_ld_{l-1}d_ld_{l+1} = \frac{1}{\Psi_l} \quad\text{for a given $\Psi_1,\ldots,\Psi_{L-1}$, $d_L$ and $d_{L-1}$} 
\end{equation}
are equivalent to each other according to Definition \ref{thm:equiv}. 
\end{corollary}


{\it Remarks:} 
The terms on the left hand side in \eqref{eq:equiv-pretrain} come from the structural hyper-parameters $e^m_l$ (see Theorem \ref{thm:expgradmulnn}). 
Hence, the constants $\Psi_1,\ldots,\Psi_{L-1}$ correspond, in some sense, to the contribution of each of $1$-layered networks that compose a $L$-NN to the eventual parameter convergence. 
Any network that belongs to the family of Corollary \ref{thm:equiv-pretrain} 
i.e., any choice of $\alpha_l$s and hidden layer lengths $d_l$s that satisfy \eqref{eq:equiv-pretrain}, 
for a given $B$, $N$, $d_0$ and $d_L$ will have the following decay of expected gradients (let $\gamma_l = \gamma$ and $w^l_m \ w \forall l$),
\begin{equation} \label{eq:expgradmulnn-equiv-pretrain}
\mathbb{E}_R \| \nabla_{\W} \tf(\W^R) \|^2 \leq \frac{1}{e^m_\gamma} \left( \frac{D_f}{N} + \frac{1}{B} \left( e^m_L + \frac{\gamma w}{4}\sum_{l=1}^{L-1} \frac{1}{\Psi_{l}}\right) \right)
\end{equation}
where both $e^m_\gamma = \gamma-\frac{13}{16}\gamma^2$ and $e^m_L = \frac{13d_{L-1}d_L\gamma^2}{256}$ are constants for all networks of this family. 
Thereby, $\Psi_1,\ldots,\Psi_{L-1}$ can be interpreted as the {\it tuning goodness criteria}, 
and \eqref{eq:equiv-pretrain} can be used to design the network i.e., {\it choose} hidden layer lengths to attain a given convergence level. 
Specifically, given $\Psi_l$s and the pretraining goodness criteria $\alpha_l$s, \eqref{eq:equiv-pretrain} can be used to recursively compute $d_l$s. 
Such strategies are discussed in detail in Section \ref{sec:disc} where \eqref{eq:equiv-pretrain}, along with its dropout counterpart from Section \ref{sec:samedepth-drop}, 
provide a systematic way to address the design choice problem from Section \ref{sec:design}.
Observe that $d_{L-1}$ and $d_L$ were fixed when discussing the equivalence of networks in the above result. 
This is because Corollary \ref{thm:equiv-pretrain} is driven by the layer-wise pretraining regime, where, 
the influence of $L-1$ pretrained layers is passed on to the tuning stage via the $L-1$ constraints in \eqref{eq:equiv-pretrain}.
This restriction is relaxed when discussing equivalence of dropout networks later in Section \ref{sec:samedepth-drop}.
We note that, beyond Corollary \ref{thm:equiv-pretrain}, there are several alternate settings of $(\alpha_l,d_l)$s that lead to equivalent networks 
but do not belong to the family characterized by the above result. 


Here, we list out one interesting network design `modulations' that one applies while seeking to identify the best deep network for a given problem, 
which provide the same level of convergence, independent of the generalization.
Let $\mathcal{D}_1 := (\alpha_{l,1},d_{l,1})$ be the original network, and $\mathcal{D}_2 := (\alpha_{l,2},d_{l,2})$ be the modulated one. 
\begin{itemize} [leftmargin=1em]
\item {\bf \mathversion{bold} Scaling $d_{l,1}$s and $\alpha_{l,1}$s \mathversion{normal}:} For any $\varphi>0$, $\mathcal{D}_1 \equiv_e \mathcal{D}_2$ 
\begin{equation} \begin{aligned} \label{eq:equiv-scale}
&d_{l,2} = \varphi d_{l,1} \quad\text{for}\quad l=1,\ldots,L-2 \quad\text{and}\quad d_{L-1,2} = d_{L-1,1} \\
\alpha_{1,2} = \frac{\alpha_{1,2}}{\varphi^2} \quad &\alpha_{l,2} = \frac{\alpha_{l,1}}{\varphi^3} \quad\text{for}\quad l=2,\ldots,L-2 \quad\text{and}\quad \alpha_{L-1,2} = \frac{\alpha_{L-1,1}}{\varphi}
\end{aligned} \end{equation}
These choices of $(\alpha_l,d_l)$s follow from Corollary \ref{thm:equiv-pretrain}.
Recall that there is very strong evidence for choosing large hidden layers \citep{bengio2009learning, hinton2010practical, bengio2012practical}. 
The identity in \eqref{eq:equiv-scale} suggests that depending on how the pretraining criteria are changed,
the newer network with changed hidden layer lengths might not necessarily improve the parameter convergence.
Alternatively, using \eqref{eq:equiv-scale} (and its parent \eqref{eq:equiv-pretrain}) the choices of $\alpha_l$ and $\varphi$ that do not guarantee improved convergence may be avoided if needed.
It is not reasonable to use very large or very small $\varphi$ (networks with too small hidden layers are unnatural and very restrictive), 
and the equivalence suggested by \eqref{eq:equiv-scale} is non-informative for such unreasonable (and uncommon) networks.
An interesting observation from \eqref{eq:equiv-scale} is that any change in small set of layers in the network can, in principle, be compensated with appropriate changes in the other layers. 
In other words, arbitrarily scaling up the layer lengths is not beneficial. A more closer discussion about this observation follows shortly. 
Some of these trends inferred from \eqref{eq:equiv-scale} seem intuitive, and the above set of results 
(Corollary \ref{thm:equiv-pretrain}, \eqref{eq:expgradmulnn-equiv-pretrain} and \eqref{eq:equiv-scale}) are, to our knowledge, the first 
to summarize such observations. 
\end{itemize}


Corollary \ref{thm:equiv-pretrain} fixes the hidden layer that feeds into the output layer (i.e., $d_{L-1}$ is fixed). 
Building upon the observations made earlier, i.e., the convergence (expected gradients) gains because of changes in some layers may, 
inadvertently, be compensated by changes in the rest, 
the following result removes this restriction on the $L-1^{th}$ layer. 
It presents the interesting scenario where one layer can, in principle, be a bottleneck for the entire network.
\cite{yosinski2014transferable} among others analyze the importance of lower layers and their learnability in producing representations that are generalizable. 
To that end, we consider the setting where $(\alpha_l,d_l)$s for the two networks $\mathcal{D}_1$ and $\mathcal{D}_2$ are chosen arbitrarily such that 
$\alpha_{l,1}d_{l+1,1} =\varrho_1 < \frac{65}{4}$ and $\alpha_{l,2}d_{l+1,2} = \varrho_2 < \frac{65}{4}$ for $l=1,\ldots,L-2$ (for some given $\varrho_1$ and $\varrho_2$).
Comparing the resulting \eqref{eq:convmulnn} for $\mathcal{D}_1$ and $\mathcal{D}_2$, we get the following
\begin{equation} \begin{aligned} \label{eq:equiv-onelayerbottle}
d_{L-1,2} &= \frac{1}{\Upsilon_2}(\Upsilon_1d_{L-1,1} + \Upsilon_3) \\
\Upsilon_1 = \frac{13}{256}d_L &+ \frac{\varrho_1}{4}d_{L-2,1} \quad \Upsilon_2 = \frac{13}{256}d_L + \frac{\varrho_2}{4}d_{L-2,2} \\ 
\Upsilon_3 = \sum_{l=1}^{L-2} &\left( {\varrho_1}{4} d_{l-1,1}d_{l,1} - {\varrho_2}{4} d_{l-1,2}d_{l,2} \right)
\end{aligned} \end{equation}


The above set of equalities follow by rearranging terms in \eqref{eq:convmulnn}. 
Although \eqref{eq:equiv-onelayerbottle} shows the relationship between the two $L-1^{th}$ hidden layers of $\mathcal{D}_1$ and $\mathcal{D}_2$, 
a similar set of equalities can be obtained for any other hidden layer. 
First, \eqref{eq:equiv-onelayerbottle} implies that, for a given $B$, $N$ and dataset, the convergence can be very sensitive to the choices of hidden layer lengths. 
The strong dependence of \eqref{eq:conv1nn}, \eqref{eq:conv1nn-drop}, \eqref{eq:convda} and \eqref{eq:convmulnn} on the layer lengths does provide some evidence for this sensitivity, 
but the above set of equalities summarize it in much finer detail. 
Second, \eqref{eq:equiv-onelayerbottle} says that two arbitrary equal depth networks with possibly different $\alpha_l$s and hidden layer lengths, 
can be `forced' to be equivalent to each other in terms of parameter convergence by simply choosing {\it one} of the layers according to \eqref{eq:equiv-onelayerbottle}.
Beyond the remarks for Corollary \ref{thm:equiv-pretrain}, this is further evidence that arbitrary (reasonably large) choices of hidden layers are not necessarily good. 


Further, \eqref{eq:equiv-onelayerbottle} predicts several of the observations made earlier in \cite{yosinski2014transferable, long2015learning}, 
where the authors present empirical evidence explaining the influence of lower versus higher layers on generalization. 
In particular, observe that \eqref{eq:equiv-onelayerbottle} simply says that any gain in parameter convergence by changing lower layers can, 
in principle, be undone by changing the higher layers -- in the worst case, one layer can act as the bottleneck.
Now, because of the structure of the loss function, the backpropagation algorithm moves the lower layer estimates slower than higher ones. 
Hence, the lower layers, from the view point of parameter convergence, are more translatable across tasks.
The specific rate of this translatability may be computed using the constants $\Upsilon_1$, $\Upsilon_2$ and $\Upsilon_3$ -- 
the higher they are, the more sensitive they are to predicting $\y$s, and hence less translatable. 
The argument for which layers regularize the network better may not be directly answerable by this discussion. 
The authors of \cite{erhan2010does} argue that pretraining is an unusual form of regularizer, while being an efficient warm-start for backpropagation. 
Definition \ref{thm:equiv}, Corollary \ref{thm:equiv-pretrain} and the resulting discussion in this section 
categorizes some such families of pretrained networks that provide the same warm-start behavior, 
which as one may expect, should facilitate addressing the deep network design choice problem (see Section \ref{sec:disc}).


\subsubsection{Fatter vs. Thinner vs. Taller vs. Shorter Networks:}\label{sec:multinetsize}

The observations made earlier, in Section \ref{sec:da-net-conv}, about thinner networks being preferable to fatter ones can be seen from $e^m_l$s in \eqref{eq:convmulnn}, 
which clearly suggests that the network depth should {\it not} be increased beyond necessary -- similar to Occam's Razor. 
Such a minimum depth will depend on the trade-off between convergence, here in the form of expected gradients from \eqref{eq:convmulnn}, 
and some performance measure like generalization \citep{hardt2015train}. 
To explore these insights more closely, we now consider the case where the two networks under comparison are of different depths, and loosely define four categories of $L$-NNs:
\begin{itemize} [leftmargin=1em] 
\item[] \quad\quad\quad\quad {\it Fat network:} Large $d_l$, Large $L$ \quad\quad\quad {\it Thin network:} Small $d_l$, Small $L$
\item[] \quad\quad\quad\quad {\it Tall network:} Small $d_l$, Large $L$ \quad\quad\quad {\it Short network:} Large $d_l$, Small $L$
\end{itemize}
Here $l$ corresponds to the hidden layers (input and output layer lengths are given).
Clearly, this is not a formal definition -- the terms {\it large} and {\it small} are not absolute, and will be used in the context of comparing two networks.
A simple observation from \eqref{eq:equiv-pretrain} and \eqref{eq:expgradmulnn-equiv-pretrain} is that, 
for a given dataset, $N$, $B$ and other learning parameters, fatter networks have a much slower decay of gradients compared to others. 
To compare a tall network $\mathcal{D}_t$ to a short one $\mathcal{D}_s$ (with $L_s < L_t$ and larger hidden layer lengths) consider the setting where all hidden layers are of the same length, 
denoted by $d_t$ and $d_s$ respectively (with $d_s > d_t$). With this definition we have the following result for their equivalence,


\begin{corollary}[{\bf Taller vs. Shorter networks}] \label{thm:tall-short}
Given a short network $\mathcal{D}_s := (\alpha,d_s)$ of depth $L_s$. Consider a tall network $\mathcal{D}_t := (\alpha,\frac{d_s}{1+\delta_{st}})$ of depth $L_t > L_s \geq 3$ with $\delta_{st}>1$.
Both networks learn datasets from the same distribution with a given $d_0$, $d_L$, $\mu_\x$ and $\tau_\x$, using multi-layer RSG with the same $B$, $N$, $\gamma$, $w$ and stopping iteration $R$. 
Let $\alpha d_s < \frac{65}{4}(1+\delta_{st})$. $\mathcal{D}_t$ is better than $\mathcal{D}_s$ in terms of $\equiv_e$ from Definition \ref{thm:equiv}, only if
\begin{equation} \label{eq:tall-short}
L_s < L_t \leq L_s + 3\delta_{st}(L_s - 3) + \Delta
\end{equation}
where $\Delta =  \delta_{st}(1+\delta_{st})^2 \left( \frac{e^m_1+e^m_{L-1}}{e^m_2} + \frac{e^m_L}{\alpha e^m_2} \right)$, 
and $e^m_1$, $e^m_2$, $e^m_{L-1}$ and $e^m_L$ are the structural constants of $\mathcal{D}_s$.
\end{corollary}
{\it Remarks:} This is an interesting result which says that in certain cases taller networks have faster convergence than shorter ones. 
Stacking up large number of layers is one of the extensively used design criteria to improve the generalization performance of deep networks. 
First, \eqref{eq:tall-short} implies that, whenever the depth of the larger network $\mathcal{D}_t$ is within a certain limit 
(and when the hidden layers of the $\mathcal{D}_t$ are smaller than $\mathcal{D}_s$), the taller network has a faster decay of gradients. 
This leads to a faster training time (smaller $N$) for $\mathcal{D}_t$, which using the result from \cite{hardt2015train}, implies that generalization improves. 
From the perspective of parameter convergence, this is the first clear justification for the necessity of adding more layers.
Second, \eqref{eq:tall-short} says that the gains in performance can only be retained if the depth is below a certain upper limit.
This is an interesting insight, with practical implications when applying deep networks on new tasks or changing the existing network for better performance. 
For instance, consider a depth $L_s=5$ network $\mathcal{D}_s := (0.01,100)$ learned on a dataset with $d_0=100$, $d_L=10$ (let $\gamma=0.5$ and $w=0.1$). 
A new network $\mathcal{D}_t$ with say $d_t = \frac{2d_s}{3}$) will be better in terms of parameter convergence only if its depth $L_t \lesssim 9$.


Apart from the results in the previous sections, it is clear from \eqref{eq:convmulnn}, that we can control the {\it convergence} of multi-layer nets 
(in terms of parameter estimation) by preceding the backpropagation tuning with layer-wise pretraining.
There is already strong empirical evidence for the generalization performance of pretraining \citep{erhan2010does, vincent2010stacked, saxe2011random, bengio2007greedy}, 
and Theorem \ref{thm:expgradmulnn} complements these studies with guaranteed convergence of gradients. 
The trends and trade-offs predicted in Sections \ref{sec:goodpretrain} and \ref{sec:multinetsize} can be related to the bounds on training times (derived from generalization \citep{hardt2015train}).
In Sections \ref{sec:vis} and \ref{sec:exps} we show more evidence for these bounds, 
deriving more subtle trends which are, for the most part, not observable in broad empirical studies by simulating the bounds.
We now adapt the result in Theorem \ref{thm:expgradmulnn} to the most general $L$-NN learning setting, leading to new and interesting observations about multi-layer deep networks. 


\section{Multi-layer with Dropout}
\label{sec:multidrop}

Using ReLUs and dropout, \cite{nair2010rectified, krizhevsky2012imagenet, szegedy2014going} and others have shown that whenever large amounts of labeled data are available, 
pretraining might not be {\em necessary}, and can be completely by-passed, to achieve good generalization, 
This can be partly seen from \eqref{eq:convmulnn} in Theorem \ref{thm:expgradmulnn} (and the corresponding remarks). 
The experiments reported in these studies clearly suggests a potential relationship between dropout and layer-wise pretraining.
To explore this interplay, the following result summarizes the convergence of expected projected gradients in the most general setting.
Here, the hidden layers are first pretrained (as in Section \ref{sec:multipretrain}), 
and the pretrained network is then tuned via supervised backpropagation {\it in tandem} with dropout induced in the input and hidden layers.
Observe that dropout is only used during this tuning phase. 
A special case of this general setting will be fully-supervised $L$-NN dropout learning with {\em no pretraining} 
and random initializations for $\{\W^1,\ldots,\W^L\}$ from \cite{srivastava2014dropout}.
As mentioned while introducing the loss in \eqref{eq:loss-mulnn}, $\zeta_{0},\ldots,\zeta_{L-1}$ denotes the dropout 
(i.e., w.p. $1-\zeta_l$ the activation is $0$: it is dropped) of the input and the $L-2$ hidden layers.
Since $\alpha$ controls the goodness of pretraining (see Theorem \ref{thm:expgradmulnn}), it is reasonable to expect some interaction between $\alpha$ and $\zeta$. 


\begin{theorem}[{\bf Pretraining vs. Dropout}] \label{thm:pretvsdrop-gen}
Given the dropout rates $\zeta_0,\ldots,\zeta_{L-1}$, for learning the $L$-layered network from Theorem \ref{thm:expgradmulnn} which is pretrained to a $(\alpha,\delta_{\alpha})$ solution, we have
\begin{equation}  \label{eq:mlnnpretdrop-gen}
\mathbb{E}_R \| \nabla_{\W} \tf(\W^R) \|^2 \leq \left( \frac{D_f}{Ne^m_\gamma\underline{\zeta}^2} 
+ \frac{1}{e^m_\gamma B} \left( \frac{\zeta_{\mathcal{B}}e^m_L}{\underline{\zeta}} + \alpha\zeta^2_{\mathcal{B}}e^m_{L-1} + \alpha\zeta^2_{\mathcal{B}} \sum_{l=1}^{L-2} e^m_l \right) \right) 
\end{equation}
where $\underline{\zeta} = \min_l \zeta_l$ and $\zeta_{\mathcal{B}} = \frac{\max_l \zeta_l}{\min_l \zeta_l}$ for $l=0,\ldots,L-1$
\end{theorem}
{\it Remarks:} This is our most general result.
Although, the concepts of layer-wise pretraining and dropout have been studied independently, to our knowledge, 
this is a useful (and possibly the first) result that explicitly combines these two regimes.
Specifically, \eqref{eq:mlnnpretdrop-gen} relates these regimes in a systematic way, 
and their {\it joint} influence on the convergence of the gradients, and to the `achievable' generalization. 
Recall that $e^m_1,\ldots,e^m_L$ encode the degrees of freedom of the $L$-NN.
Hence, the first term in \eqref{eq:mlnnpretdrop-gen} corresponds to the goodness-of-fit of the outputs, 
while the other terms represent the effective `reduction' in the number of free parameters because of pretraining.


An important hyper-parameter in Theorem \ref{thm:pretvsdrop-gen} is $\zeta_{\mathcal{B}}$, the ratio of the maximum and minimum dropout rates.
Clearly, $\zeta_{\mathcal{B}} \geq 1$, and as the variance of $\zeta_l$s increases, $\zeta_{\mathcal{B}}$ increases accordingly, requiring larger batchsizes and iterations. 
We refer to $\zeta_{\mathcal{B}}$ as the {\it dropout bottleneck ratio}, or simply the bottleneck ratio, 
since it corresponds to the variability in the fraction of unknowns (i.e., the fraction of dropped units) being updated across the $L$ layers.
The last three terms of \eqref{eq:mlnnpretdrop-gen} show that, $\zeta_{\mathcal{B}}$, $\underline{\zeta}$ (the smallest dropout rate across all layers) 
and the pretraining goodness $\alpha$ are critical in reducing this number of free parameters (and consequently, the induced bias from the stochastic gradients) after pretraining. 
For a fixed dataset (and $B$, $N$), the bottleneck ratio can be used as a surrogate to how `bad' the estimates $\W$ may be, or more specifically, 
how much variability there might be across different runs of multi-layer RSG. 
In the special case where $\zeta_l = \zeta$ for $l=0,\ldots,L-1$, \eqref{eq:mlnnpretdrop-gen} reduces to 
\begin{equation}  \label{eq:mlnnpretdrop-gen-cons}
\mathbb{E}_R \| \nabla_{\W} \tf(\W^R) \|^2 \leq \left( \frac{D_f}{Ne^m_\gamma\zeta^2} 
+ \frac{1}{e^m_\gamma B} \left( \frac{e^m_L}{\zeta} + \alpha e^m_{L-1} + \alpha \sum_{l=1}^{L-2} e^m_l \right) \right) 
\end{equation}
but as we show later in Corollary \ref{thm:pretvsdrop}, a tighter bound can be derived for this constant dropout setting.


Note that \eqref{eq:mlnnpretdrop-gen} is not the most general form of the decay bound for pretraining versus dropout.
Instead of using $\underline{\zeta}$ and $\zeta_{\mathcal{B}}$, as was done in Theorem \ref{thm:pretvsdrop-gen}, one can construct the bound by directly using the dropout rates $\zeta_0,\ldots,\zeta_L$. 
Such a bound is harder to interpret and involves an influence from all the different hyper-parameters, and has little to offer over \eqref{eq:mlnnpretdrop-gen} 
(see Section \ref{sec:proof13} for a discussion).
To better interpret \eqref{eq:mlnnpretdrop-gen}, while deriving interesting trends and inferences about pretraining versus dropout in the context of convergence, 
we further simplify Theorem \ref{thm:pretvsdrop-gen} by using the same dropout rate across all the layers i.e., $\zeta_l = \zeta$ for $l=0,\ldots,L-1$.
The following result summarizes the new bound. 
This is tighter than simply imposing $\zeta_l = \zeta$ in \eqref{eq:mlnnpretdrop-gen}, 
because the relaxation in the proof of Theorem \ref{thm:pretvsdrop-gen} (see Section \ref{sec:proof13}) to make the result more interpretable can now be removed. 


\begin{corollary}[{\bf Pretraining vs. Dropout ($\mathversion{bold} \zeta_l = \zeta \mathversion{normal}$)}] \label{thm:pretvsdrop}
Whenever the input and hidden layers have the same dropout rate $\zeta_l = \zeta$ for $l=0,\ldots,L-1$, 
for learning the $L$-layered network from Theorem \ref{thm:expgradmulnn} which is pretrained to a $(\alpha,\delta_{\alpha})$ solution, we have
\begin{equation}  \label{eq:mlnnpretdrop}
\mathbb{E}_R \| \nabla_{\W} \tf(\W^R) \|^2 \leq \frac{1}{e^m_\gamma} \left( \frac{D_f}{N\zeta^2} 
+ \frac{1}{B} \left( \frac{e^m_L}{\zeta} + \alpha e^m_{L-1} + \alpha\zeta\sum_{l=1}^{L-2} e^m_l \right) \right) 
\end{equation}
\end{corollary}
We see that \eqref{eq:mlnnpretdrop} has the same structure as \eqref{eq:mlnnpretdrop-gen} but is simpler.
Given the network structure ($L$ and $d_0,\ldots,d_L$) and stepsizes ($\gamma^1,\ldots,\gamma^L$), the hyper-parameter functions $e^m_1,\ldots,e^m_L$ are constants (see Theorem \ref{thm:expgradmulnn}). 
$D_f$ does depend on the goodness of pretraining, but will nevertheless be large because the loss is calculated 
over the extra output layer between the predicted and the observed $\y$s (see \eqref{eq:loss-mulnn}).
With the above result, consider the two standard learning mechanisms --- dropout learning with and without layer-wise pretraining. 


\begin{itemize} [leftmargin=1em]
\item {\bf Pretraining + Dropout:} \label{sec:pret+drop}
\begin{itemize} [leftmargin=1em]
\item {\bf Small \mathversion{bold}$\alpha$\mathversion{normal}:}
  If the network is pretrained to a reasonably small $\alpha$, the last two terms in \eqref{eq:mlnnpretdrop} are already as small as they can be, and will be dominated by
  the first two terms.
In this setting, for a given $B$ and $N$, the best choice for $\zeta$ will then be $\approx 1$ i.e., very minimal or dropout. 
It is known that this procedure works very well --- this is essentially ``good''
layer-wise pretraining followed by supervised fine-tuning \citep{hinton2006reducing, erhan2010does}.
The bound in \eqref{eq:mlnnpretdrop} presented a clear theoretical argument for the initial use of pretraining in deep networks 
from the perspective of convergence and eventually, generalization.
\item {\bf Large \mathversion{bold}$\alpha$\mathversion{normal}:}
Alternatively, if the pretraining is poor (i.e., large $\alpha$), and we still operate in the $\zeta \to 1$ regime, 
the fine-tuning process will update the full network in each iteration and may result in overfitting.
This is the fundamental argument that motivated the work on dropout learning \citep{srivastava2014dropout}. 
This argument was made empirically earlier in this work and others \citep{baldi2014dropout}, and our result in Corollary \ref{thm:pretvsdrop} provides a complete justification for it. 
The cost of using dropout though is slower convergence because the first and second terms will rapidly increase depending on $\zeta$. 
This is expected since dropout essentially adds `noise' to the solution path, 
forcing backpropagation to work with a subset of all activations \citep{wager2013dropout}, and overall, this involves more work.
Dropout dynamics studied in \cite{rifai2011adding, baldi2014dropout} also make this observation but {\it not} from the perspective of parameter convergence.
\end{itemize}
Independent of the precise dropout rate $\zeta$, \eqref{eq:mlnnpretdrop} clearly implies that pretraining always improves convergence (since $\alpha$ will reduce). 
\cite{abdel2013exploring, lee2009convolutional, srivastava2014dropout} have shown empirically that this is especially true for improving generalization.
The optimal choices of $\zeta$ are discussed later in this Section \ref{sec:droprates}.


\item {\bf Only Dropout:} \label{sec:only+drop}
When pretraining is not allowed, there is a complicated trade-off between the terms involving $\zeta$ in \eqref{eq:mlnnpretdrop}.
Specifically, the optimal $\zeta$ will depend on balancing the variance from the hidden layers (the last term) versus the goodness of 
approximation of the outputs (first and second terms).
For certain values of $D_f$ and $e^m_l$'s, the best $\zeta$ will be $\sim 0.5$, which was recently 
reported as the best rate as per dropout dynamics \citep{baldi2014dropout}.
The visualizations in Section \ref{sec:vis} will provide more insight into this interplay.
Large values of $B$ and $N$ will ensure that the bound is small, 
and a large $N$ will, in turn, demand a large training dataset size (refer to the setup from Theorems \ref{thm:conv1nn-mono}, \ref{thm:expgradda} and \ref{thm:expgradmulnn}).
Put another way, if large amounts of labeled data are available, one can by-pass pretraining completely and only perform supervised backpropagation 
forcing all the terms in \eqref{eq:mlnnpretdrop} to reduce with reasonably large $B$ and number of epochs.
This was the argument put forth by recent results in deep learning \citep{krizhevsky2012imagenet, szegedy2014going, ciregan2012multi} --- 
performing fully supervised dropout with very large amounts of training data where pretraining has been for the most part discarded completely.
\end{itemize}


\subsection{Computational Complexity of multi-layer RSG} \label{sec:sampmulnn}

Theorems \ref{thm:conv1nn-mono} and \ref{thm:convda} described the iteration complexity, i.e., the number of maximum allowable gradient iterations $N$, for single-layer and DA RSG respectively.
As observed in Theorem \ref{thm:pretvsdrop-gen} and Corollary \ref{thm:pretvsdrop}, 
the convergence bound for multi-layer dropout is complicated involving a multitude of learning and network hyper-parameters.
Nevertheless, under the reasonable assumption that $SC \approx BN$ where $S$ is the sample size (training dataset size) and 
$C$ denotes the number of epochs, the following result calculates the minimum number of training instances required to ensure a ($\epsilon, \delta$)
solution (see Definition \ref{thm:epsdel}) for multi-layer RSG with dropout. 
Note that the same assumption was used in the Remarks of Theorem \ref{thm:conv1nn-mono} to discuss some example sample sizes.

\begin{theorem}[{\bf Multi-layer RSG Computational Complexity}] \label{thm:sampmulnn-drop}
Consider a multi-layer RSG from Theorem \ref{thm:expgradmulnn} with input and hidden layer dropout rate of $\zeta$ and the network is pretrained to a $(\alpha,\delta_{\alpha})$ solution.
To obtain a $(\epsilon,\delta)$-solution, we need 
\begin{equation}\label{eq:sampmulnn-drop}
B \geq \frac{2 \Pi}{\epsilon^m_\gamma \epsilon \delta^{1/T}} \quad \text{and for such a choice of batchsize} \quad  
N \geq \max\left\{ \frac{SC}{B}, \frac{2d_L}{\epsilon^m_\gamma \zeta^2 \epsilon \delta^{1/T}} \right\}
\end{equation}
where $C$ is the number of backpropogation epochs, $S$ is the number of labeled training instances and $\Pi = \left( \frac{e^m_L}{\zeta} + \alpha e^m_{L-1} + \alpha\zeta\sum_{l=1}^{L-2} e^m_l \right)$
\end{theorem}

{\it Remarks:} As mentioned in the remarks of Theorem \ref{thm:convda}, larger networks entail large batchsizes.
Nevertheless, if the networks is pretrained to small $\alpha$ this requirement can be relaxed (see \eqref{eq:sampmulnn-drop}).
Similar to \eqref{eq:itersampda}, $D_f$ has been replaced by $D_L$ there by making the bounds in \eqref{eq:sampmulnn-drop} deployable in practice. 
The design procedures in Section \ref{sec:disc} will use this for computing the required $B$ and $N$.


\subsection{\bf Choosing the dropout rates}\label{sec:droprates}

The authors in \cite{baldi2014dropout} exhaustively analyzed the learning dynamics of the dropout algorithm. 
They suggest that, independent of the network lengths or size, 
$\zeta=0.5$ is optimal in the learning theoretic sense i.e., the resulting dropped representations are unbiased and asymptotically equivalent to the ideal `noise-injected' representations.
Our results in \eqref{eq:mlnnpretdrop-gen} and \eqref{eq:mlnnpretdrop} suggest that choosing $0.5$ dropout is good but {\it may not} always be ideal to guarantee the best decay of expected gradients.
Such an optimal $\zeta$ clearly depends on the network structure and data statistics. 
To analyze \eqref{eq:mlnnpretdrop} (which assumes $\zeta_l = \zeta$ for $l=1,\ldots,L-1$) more closely, 
we use the running example from Section \ref{sec:compare}, with constant stepsizes $\gamma^l = \gamma$, 
a fixed box-constraint across all layers $w^l_m = w \forall l$ (for some $w>0$) and assume $D_f \approx d_L$ (since $\y \in [0,1]^{d_L}$).
The following result presents the best $0 < \zeta < 1$ for the network with same hidden layer lengths $d_l = d$ for $l=1,\ldots,L-1$. 


\begin{corollary}[{\bf Optimal dropout rate}] \label{thm:bestdrop}
The best constant dropout rate $\zeta = \zeta_l ~~\forall l$ for learning a $L$-NN with a constant hidden layer length $d$, 
using multi-layer RSG from Alg. \ref{alg:multi-nn} with $\gamma^l = \gamma$, $w^l_m = w > 0 \forall l$ and $\ceil{\frac{1}{\kappa}}$ number of epochs, is given by
\begin{equation} \begin{aligned} \label{eq:bestdrop}
\text{median}\left\{0,\sqrt[3]{\frac{C_1}{C_2}},1\right\} \quad\quad&\text{where}\quad C_1 = \left( d_L + \frac{13d^2\gamma^2}{256\kappa} \right) \\
\quad\quad\text{and}\quad C_2 = \frac{\alpha\gamma w}{4\kappa} &\left( d_0d^2 + (L-3)d^3 + d^2d_L \right) 
\end{aligned} \end{equation}
\end{corollary}
{\it Remarks:}
We visualize these optimal rates in Section \ref{sec:vis} for several choices of the hyper-parameters involved in $C_1$ and $C_2$.
Clearly, a too small or a too large $\zeta$s may not be reasonable from the perspective of dropout dynamics \citep{baldi2014dropout}.
The interaction of $\alpha$ and $\zeta$, discussed above in \eqref{eq:bestdrop} formalizes several such observations.
A large $\alpha$ clearly pushes the best $\zeta$ to be closer to zero,  
i.e., the network will need to use large dropout rates when either (1) the pretraining is not effective or (2) the layers have not been pretrained to begin with. 
Consider a network with a reasonably large $d$ (compared to $d_L$) and number of epochs $\ceil{\frac{1}{\kappa}}$. 
In this setting, the constants $C_1$ and $C_2$ from \eqref{eq:bestdrop} can be approximated as 
\begin{equation} \label{eq:bestdrop-examp}
C_1 = \frac{13d^2\gamma^2}{256\kappa} \quad\text{and}\quad C_2 = \frac{\alpha\gamma w}{4\kappa} (L-3)d^3 \quad\text{and hence, the best}\quad \zeta \approx \sqrt[3]{\frac{0.2\gamma}{\alpha w(L-3)d}}
\end{equation}
For instance, for the example network $\mathcal{D}_s$ used in the remarks of Corollary \ref{thm:tall-short} with $d=100$, $\gamma=0.5$, $w=0.1$ and $\alpha=0.01$, 
using \eqref{eq:bestdrop-examp}, the optimal $\zeta$ is approximately $0.79$ 
i.e., in each iteration of the backpropagation, only $20\%$ of all units can be nullified and the updates can be restricted to the remaining $80\%$ of the whole network. 
When pretraining is poor (or not utilized) and $\alpha$ increases to, say $2$, from $0.01$ (the latter is the well pretrained network), 
the optimal $\zeta$ decreases to $0.13$ and updates must be restricted to a much smaller part of the whole network. 


\subsection{Comparing arbitrary networks -- with dropout}\label{sec:comparedrop}

In Section \ref{sec:compare}, we used Theorem \ref{thm:expgradmulnn} to analyze the equivalence (see Definition \ref{thm:equiv}) 
of arbitrary layer-wise pretrained multi-layer networks. 
Based on the discussion, equivalence results of equal length networks (Corollary \ref{thm:equiv-pretrain}) and arbitrary networks (Corollary \ref{thm:tall-short}) were obtained.
We replicate this set of results for dropout networks using Theorem \ref{thm:pretvsdrop-gen} and Corollary \ref{thm:pretvsdrop}.
Such results from Theorem \ref{thm:pretvsdrop-gen}, because of the presence of unequal dropout rates $\zeta_l$s, lead to more complicated trends, 
and the resultant interpretation may not necessarily provide any new additional insight towards addressing the central question of comparing 
and designing deep networks on data generated from a given distribution. 
Hence, we use the constant dropout setting from Corollary \ref{thm:pretvsdrop}. 
Similar to the setup from Section \ref{sec:compare}, consider learning two different deep networks for datasets sampled from some distribution, 
i.e., for a fixed $d_0$ and the data statistics ($\mu_\x$ and $\tau_\x$). 
We stick to the same general setting from Section \ref{sec:compare} where the $L-1$ hidden layers have different pretraining goodness criteria $\alpha_l$s.
Following Definition \ref{thm:equiv}, for dropout $L$-NNs $\mathcal{D}_1 \equiv_e \mathcal{D}_2$ whenever the bound in \eqref{eq:mlnnpretdrop} is the same for either of them.
Within this setting, we have the following result that characterizes two general families of networks that are equivalent.


\subsubsection{Different networks of same depth -- with dropout} \label{sec:samedepth-drop}

\begin{corollary}[{\bf Equivalence of dropout networks}] \label{thm:equiv-pretrain-drop}
Consider the family of depth $L$ deep networks that learn datasets from an underlying distribution (i.e., given $d_0$, $\mu_\x$ and $\tau_\x$) 
using multi-layer RSG with dropout $\zeta$ and a given $B$, $N$, $\gamma_l$, $w^l_m$ and stopping iteration $R$. 
All such networks $\mathcal{D} := (\zeta,\alpha_l,d_l)$, with $\alpha_ld_{l+1} < \frac{65}{4}$ for $l=1,\ldots,L-1$ and 
\begin{equation} \begin{aligned} \label{eq:equiv-pretrain-drop}
&\zeta\alpha_l d_{l-1}d_ld_{l+1} = \frac{1}{\Phi_l} \quad\quad \alpha_{L-1}d_{L-2}d_{L-1}d_{L} = \frac{1}{\Phi_{L-1}} \\
&\frac{d_{L-1}d_L}{\zeta} = \frac{1}{\Phi_L} \quad\quad \frac{d_L}{\zeta^2} = \frac{1}{\Phi_f} \quad\quad\text{for a given $\Phi_1,\ldots,\Phi_L$ and $\Phi_{f}$}
\end{aligned} \end{equation}
are equivalent ($\equiv_e$) to each other according to Definition \ref{thm:equiv}. 
\end{corollary}
{\it Remarks:} 
The setup of Corollary \ref{thm:equiv-pretrain-drop} is similar to Corollary \eqref{thm:equiv-pretrain}. 
Similar to $\Psi_l$s from \eqref{eq:equiv-pretrain}, $\Phi_l$s and $\Phi_f$ denote the contribution of the structural hyper-parameters $e^m_l$s to the convergence bound -- the tuning goodness criteria. 
Specifically, $\Phi_1,\ldots,\Phi_L$ correspond to the contribution of the $L$ different $1$-NNs that compose the multi-layer network, 
and $\Phi_f$ corresponds to the contribution from the initial deviation. 
Note that \eqref{eq:equiv-pretrain-drop} has a total of $L+1$ such constraints.
In the earlier non-dropout case from Corollary \eqref{thm:equiv-pretrain}, $\Psi_f$ and $\Psi_L$ type terms were not used because $d_l$ and $d_{L-1}$ were fixed. 
The above result removes this restriction. 
Recall the remarks of Corollary \ref{thm:equiv-pretrain} and \eqref{eq:expgradmulnn-equiv-pretrain} discuss how \eqref{eq:equiv-pretrain} can be used to design the network.
Similarly, the more general form in
\eqref{eq:equiv-pretrain-drop} can also be used for designing dropout networks for a given $d_0$ and the pretraining and tuning criteria $\alpha_l$s, $\Phi_l$s and $\Phi_f$.
Any family of networks satisfying \eqref{eq:equiv-pretrain-drop} will have the following decay of gradients
\begin{equation} \label{eq:expgradmulnn-equiv-pretrain-drop}
\mathbb{E} \| \nabla_{\W} \tf(\W^R) \|^2 \lesssim \frac{1}{e^m_\gamma} \left( \frac{1}{\Phi_fN} 
+ \frac{1}{B} \left( \frac{13\gamma^2}{256\Phi_L} + \frac{\gamma w}{4}\sum_{l=1}^{L-1} \frac{1}{\Phi_{l}}\right) \right)
\end{equation}
We defer the discussion on actually designing such families of networks to Section \ref{sec:disc}.


\subsubsection{Fatter vs. Thinner vs. Taller vs. Shorter Networks:}\label{sec:multinetsize-drop}

The discussion in the previous section focused on networks of the same depth. 
Similar to Section \ref{sec:multinetsize-drop} and Corollary \ref{thm:tall-short}, 
we have the following result for the taller versus shorter dropout networks, following the setup and discussion from Section \ref{sec:multinetsize}.
As observed with the non-dropout case in Section \ref{sec:multinetsize-drop}, fatter dropout networks have the slowest convergence. 
\begin{corollary}[{\bf Taller vs. Shorter dropout networks}] \label{thm:tall-short-drop}
Given a short dropout network $\mathcal{D}_s := (\zeta,\alpha,d_s)$ of depth $L_s$. 
Consider a tall dropout network $\mathcal{D}_t := (\zeta,\alpha,\frac{d_s}{1+\delta_{st}})$ of depth $L_t > L_s \geq 3$ with $\delta_{st}>1$.
Both networks learn datasets from the same distribution with a given $d_0$, $d_L$, $\mu_\x$ and $\tau_\x$, using multi-layer RSG with the same $B$, $N$, $\gamma$, $w$ and stopping iteration $R$. 
Let $\alpha d_s < \frac{65}{4}(1+\delta_{st})$. $\mathcal{D}_t$ is better than $\mathcal{D}_s$ in terms of $\equiv_e$ from Definition \ref{thm:equiv}, only if
\begin{equation} \label{eq:tall-short-drop}
L_s < L_t \leq L_s + 3\delta_{st}(L_s - 3) + \Delta
\end{equation}
where $\Delta = \delta_{st}(1+\delta_{st})^2 \left( \frac{e^m_1}{e^m_2} + \frac{e^m_{L-1}}{\zeta e^m_2} + \frac{e^m_L}{\alpha \zeta^2 e^m_2} \right)$, 
and $e^m_1$, $e^m_2$, $e^m_{L-1}$ and $e^m_L$ are the structural constants of $\mathcal{D}_s$.
\end{corollary}
{\it Remarks:} 
The earlier statement about taller versus shorter networks (see remarks of Corollary \ref{thm:tall-short-drop}) still applies here. In closing, we briefly summarize the main message. 
To our knowledge, \eqref{eq:tall-short-drop} is potentially the first result justifying the increase of the network depth,
in order to improve the convergence bound and consequently, the training time $N$, 
which, as pointed out for Corollary \ref{thm:tall-short-drop}, in tandem with \cite{hardt2015train}, implies a better generalization. 
Consider the example network used earlier in the remarks of Corollary \ref{thm:tall-short} but with a dropout rate $\zeta=0.5$, 
i.e., $\mathcal{D}_s := (0.5,0.01,100)$ with $L_s = 5$ learned on a dataset with $d_0=100$, $d_L=10$ (let $\gamma=0.5$ and $w=0.1$). 
A new dropout network $\mathcal{D}_t$ with say $d_t = \frac{2d_s}{3}$) will be better in terms of parameter convergence only if its depth $L_t \lesssim 10$.


Overall, the results in Sections \ref{sec:pretrain} and \ref{sec:multi} corroborate many existing observations about pretraining and dropout, 
and provides new tools to {\em guide} the learning procedure, in general.
They provide insights into the convergence behavior, and allow us to {\it explicitly} calculate the best possible settings for all hyper-parameters. 
After evaluating the bounds in Section \ref{sec:exps} and \ref{sec:vis}, we return to the design choice problem, motivated in Section \ref{sec:design}, 
providing design {\it guidelines} for deep networks to guarantee certain level of parameter convergence and generalization on new learning tasks.
Observe that the three algorithms presented in this work are only minor modifications over the classical autoencoder and backpropagation pipeline, and so, are very easy
to implement.


\section{Experiments}
\label{sec:exps}

\begin{figure}[!h]\centering
\subfloat[]{\includegraphics[width=47mm]{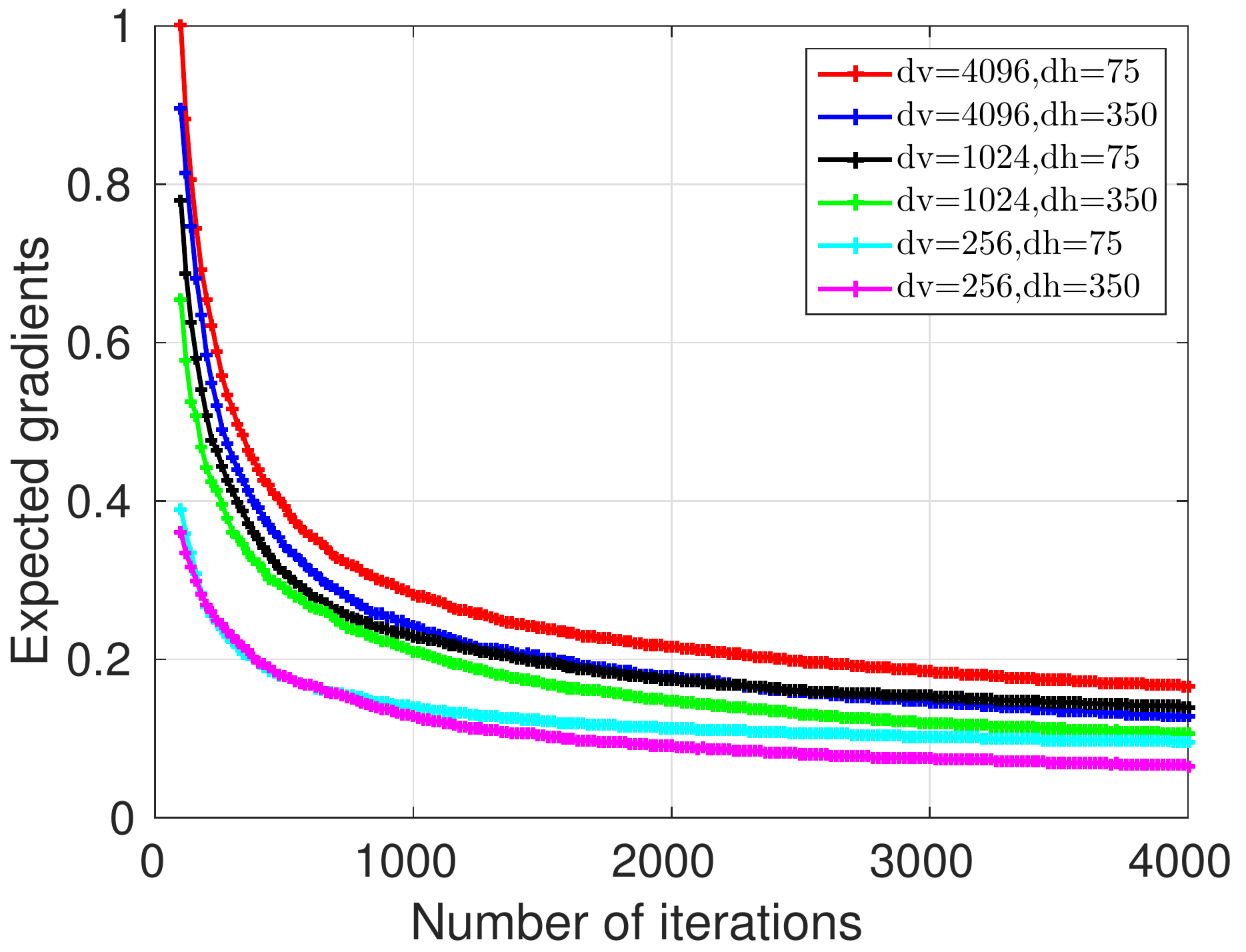}} \quad 
\subfloat[]{\includegraphics[width=47mm]{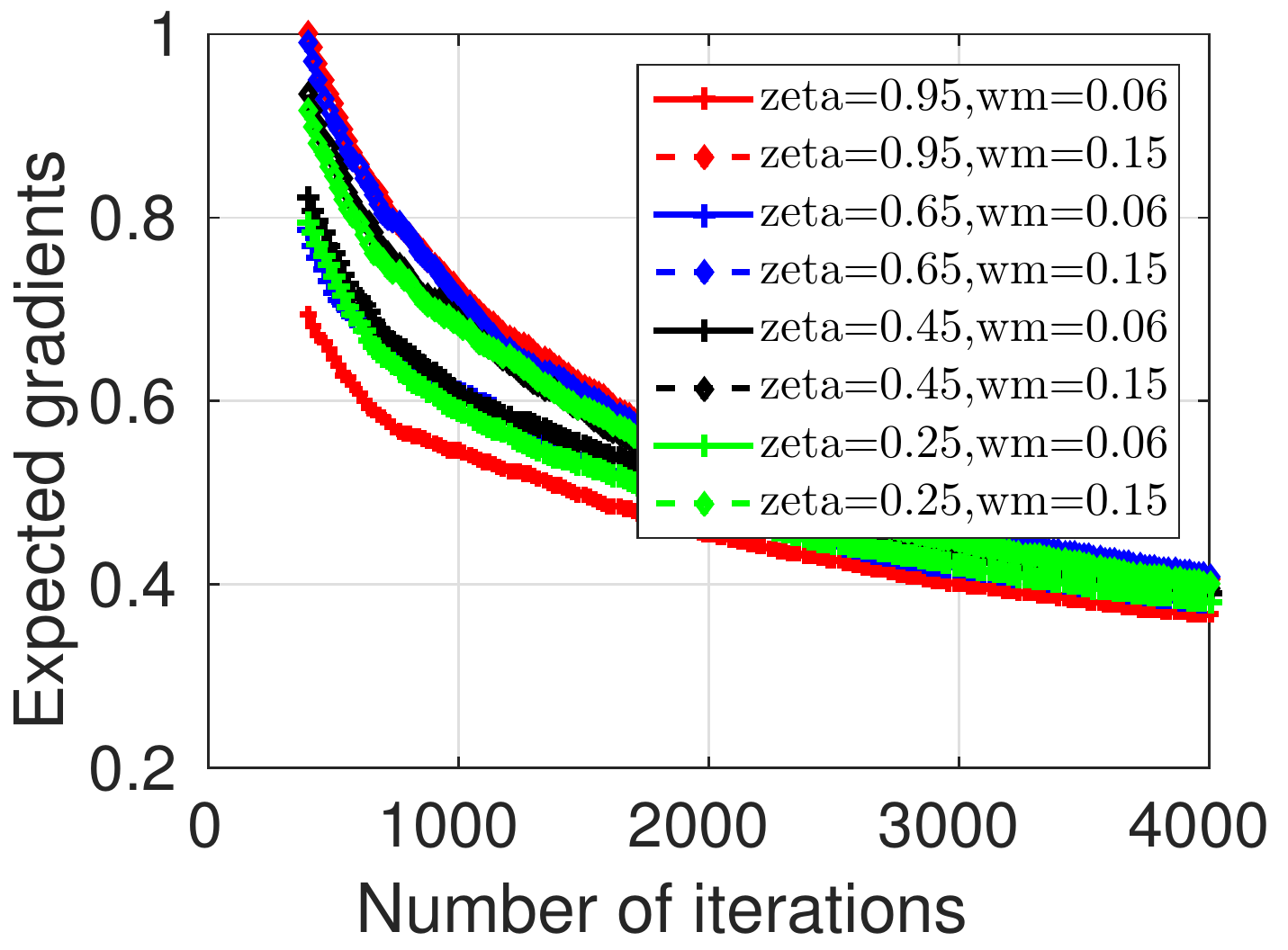}} \quad 
\subfloat[]{\includegraphics[width=47mm]{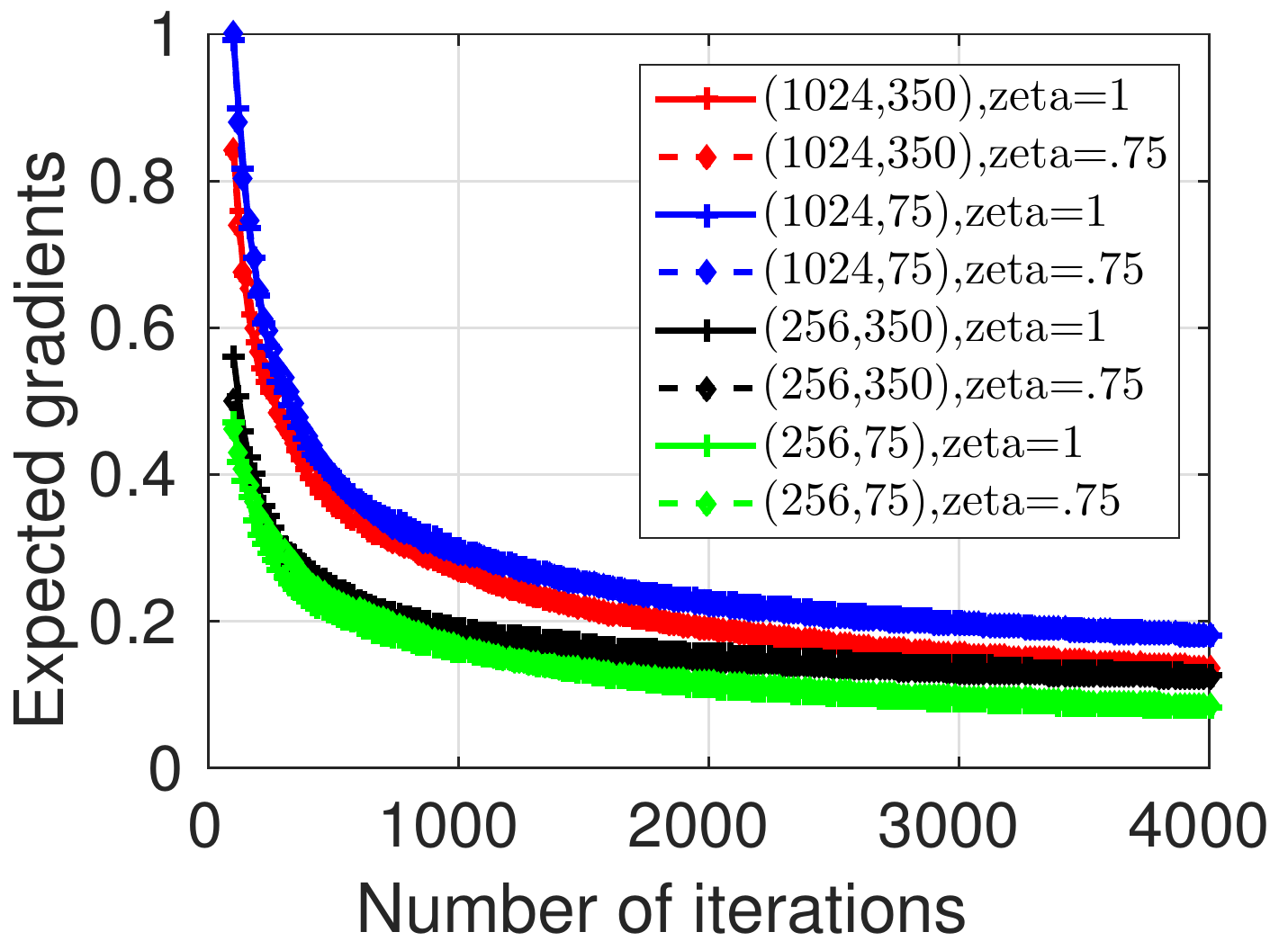}} \\
\subfloat[]{\includegraphics[width=47mm]{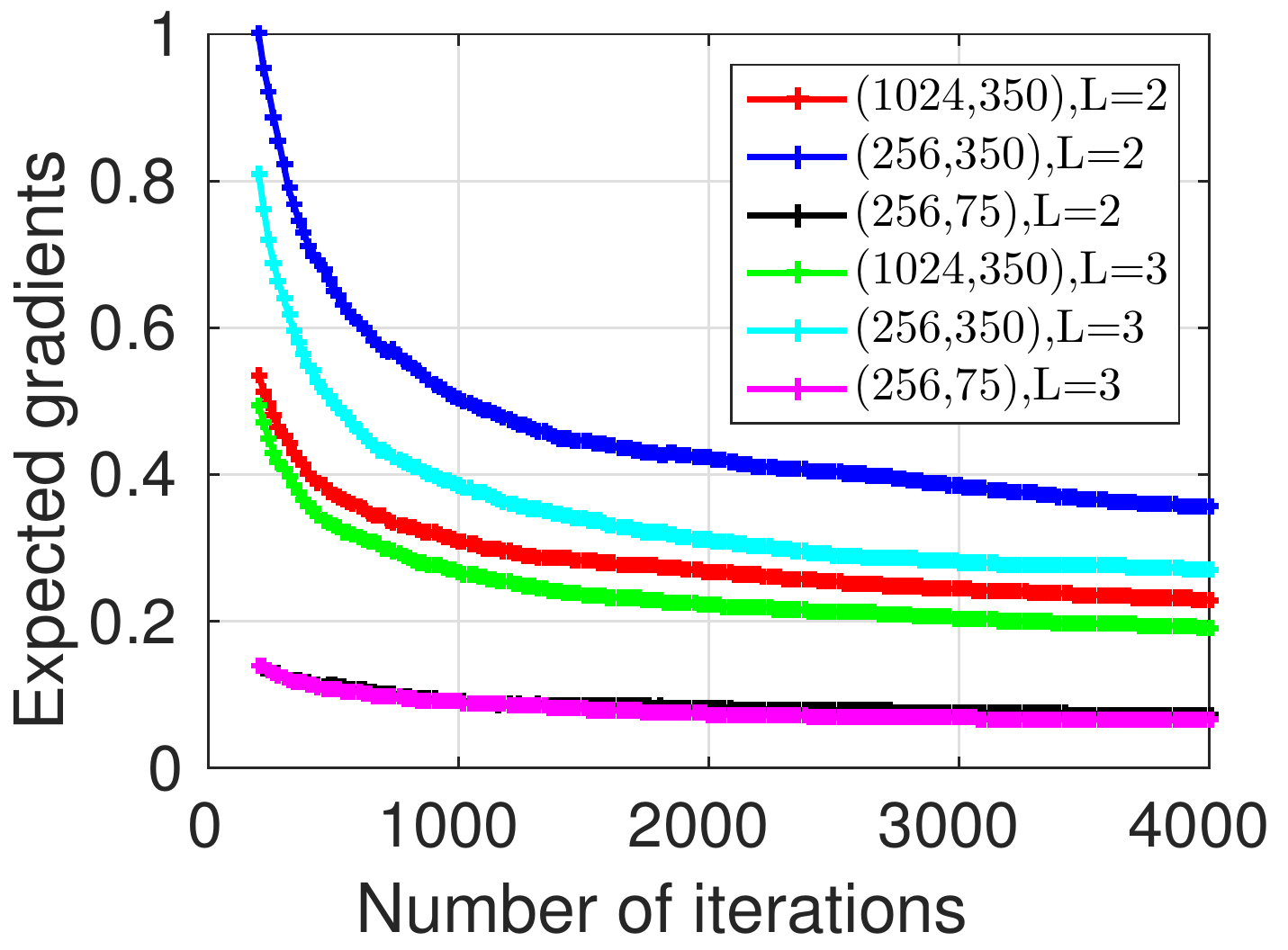}} \quad 
\subfloat[]{\includegraphics[width=47mm]{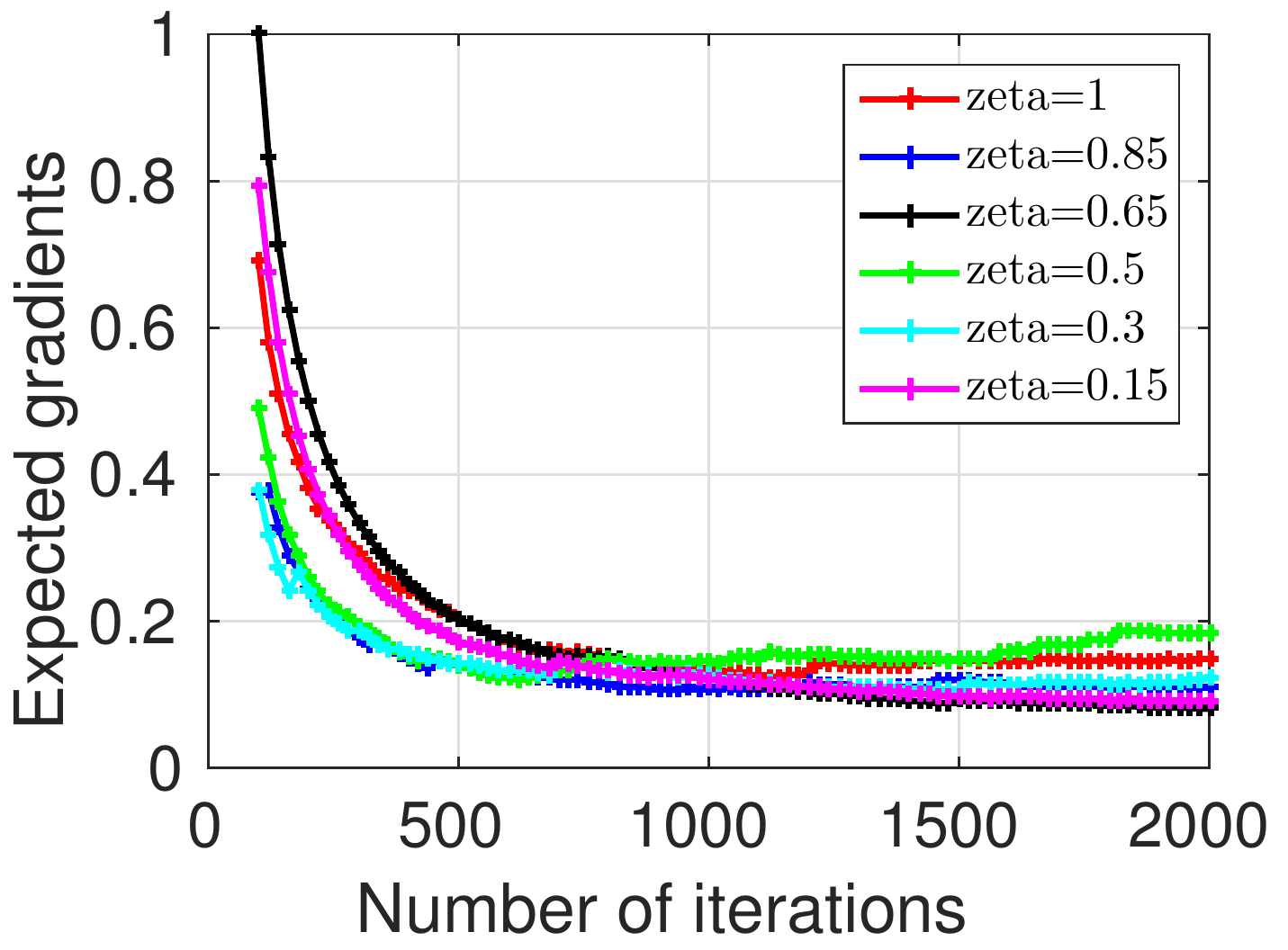}} \quad 
\subfloat[]{\includegraphics[width=47mm]{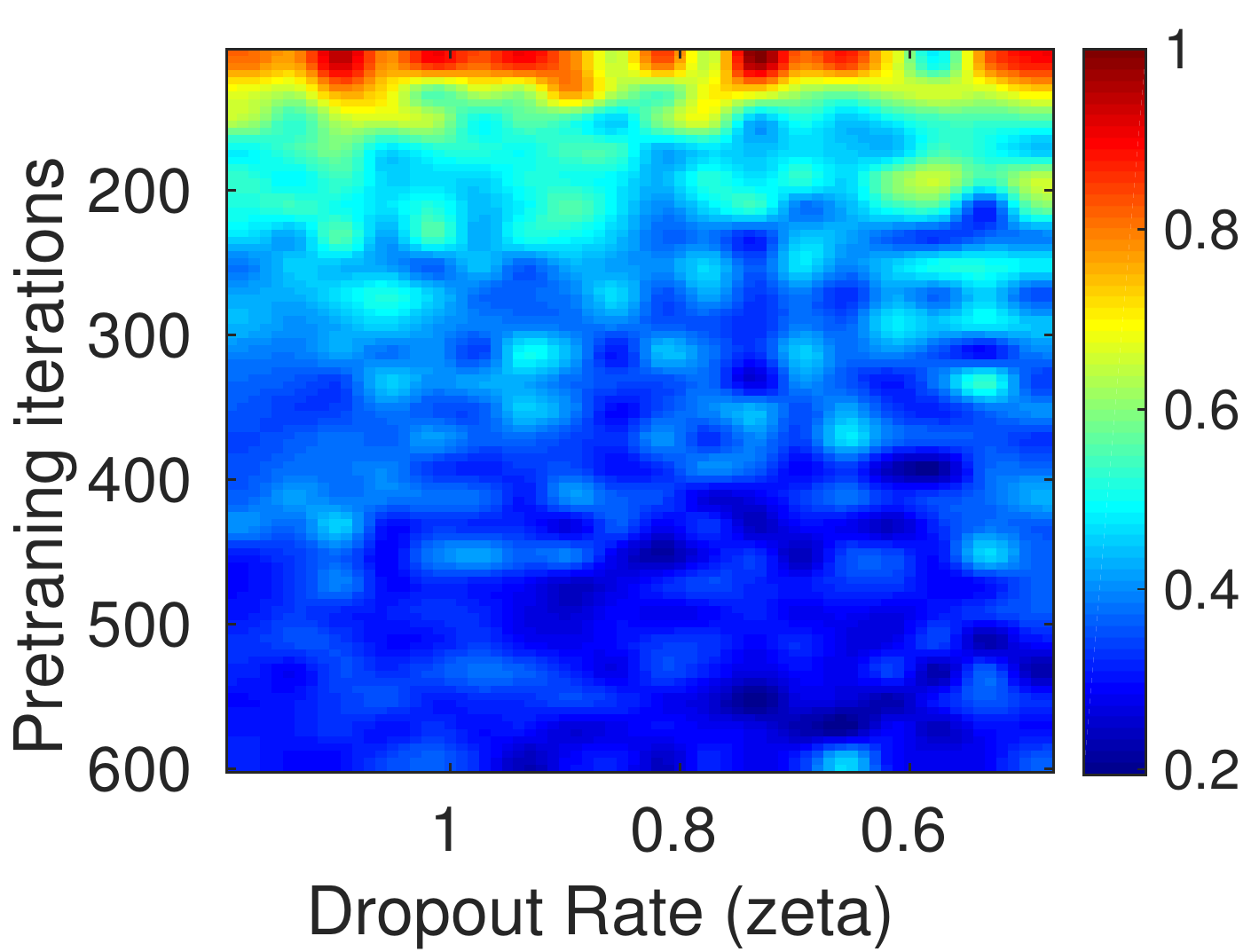}}
\caption{\footnotesize \label{fig:allexp_mnist} {\bf Decay of expected gradients ({\it mnist})} Expected gradients in pretraining (a,b) and multi-layer networks (c-e). 
(f) Pretraining vs. dropout. $B$ is set to $100$. All $y$-axes are scaled by corresponding maximum values.
The plots show that the trends predicted by the proposed framework are observed in practice -- the decay of expected gradients versus several hyperparameters.
Convergence is faster for smaller and thinner networks (a,c,d). 
Changes in visible layer length dominates the hidden layer lengths for DA (a). Dropout rate has less influence (e). Pretraining compensates for dropout (f).
Strong pretraining and dropout rate around $0.5$ gives fastest empirical gradients (dark blue region in (f)).} 
\end{figure}

\begin{figure}[!h]\centering
\subfloat[]{\includegraphics[width=47mm]{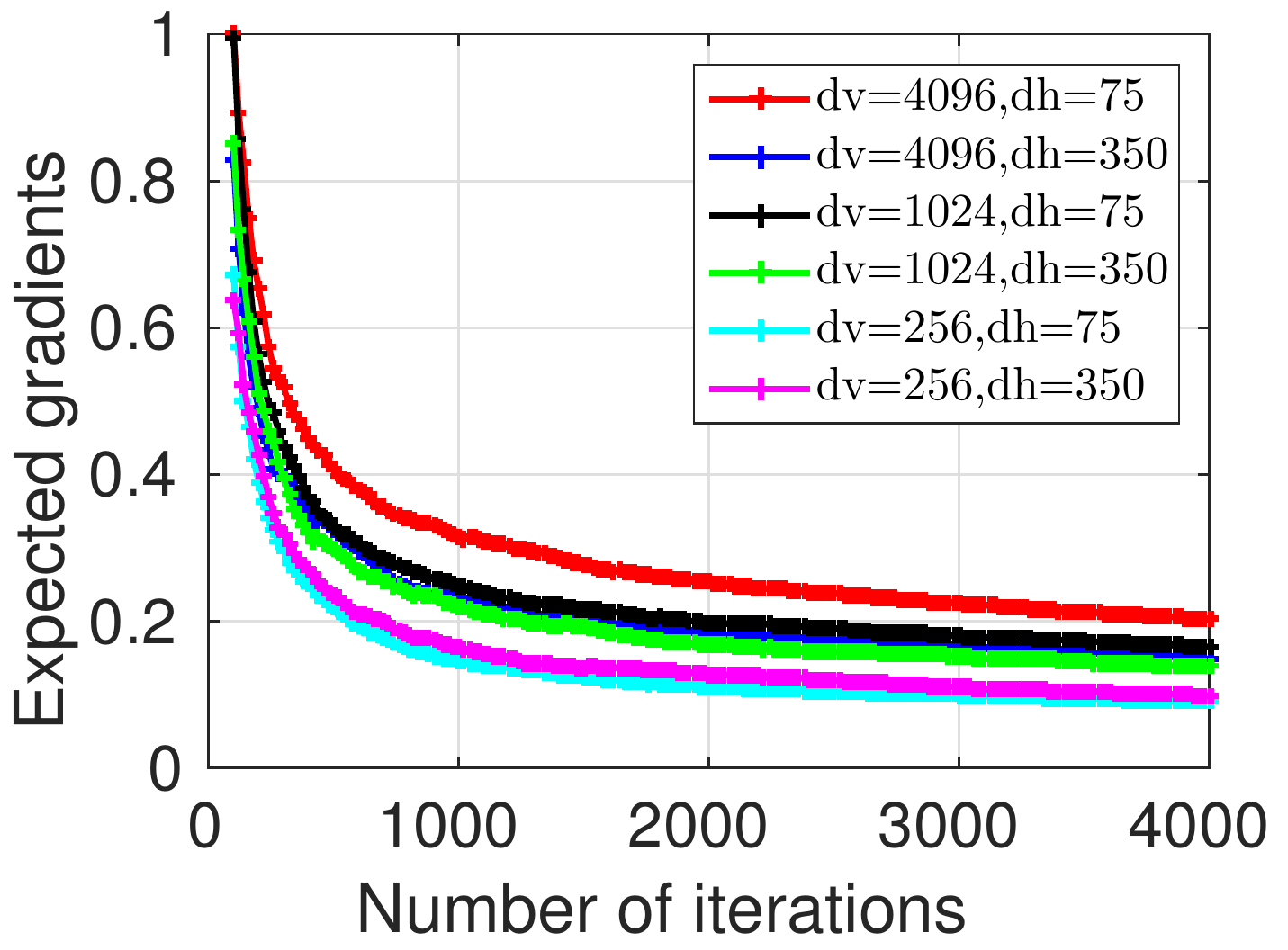}} \quad 
\subfloat[]{\includegraphics[width=47mm]{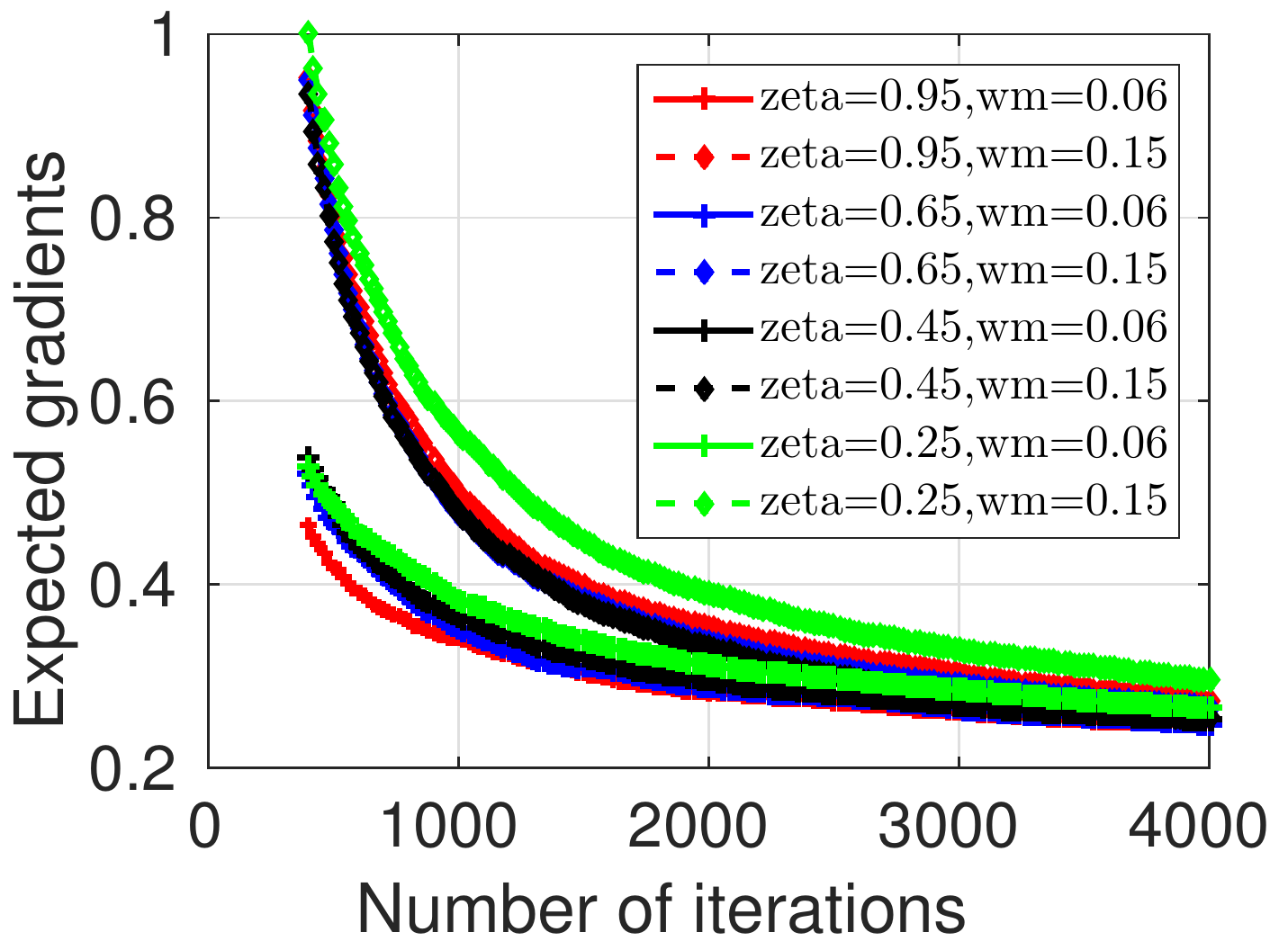}} \quad
\subfloat[]{\includegraphics[width=47mm]{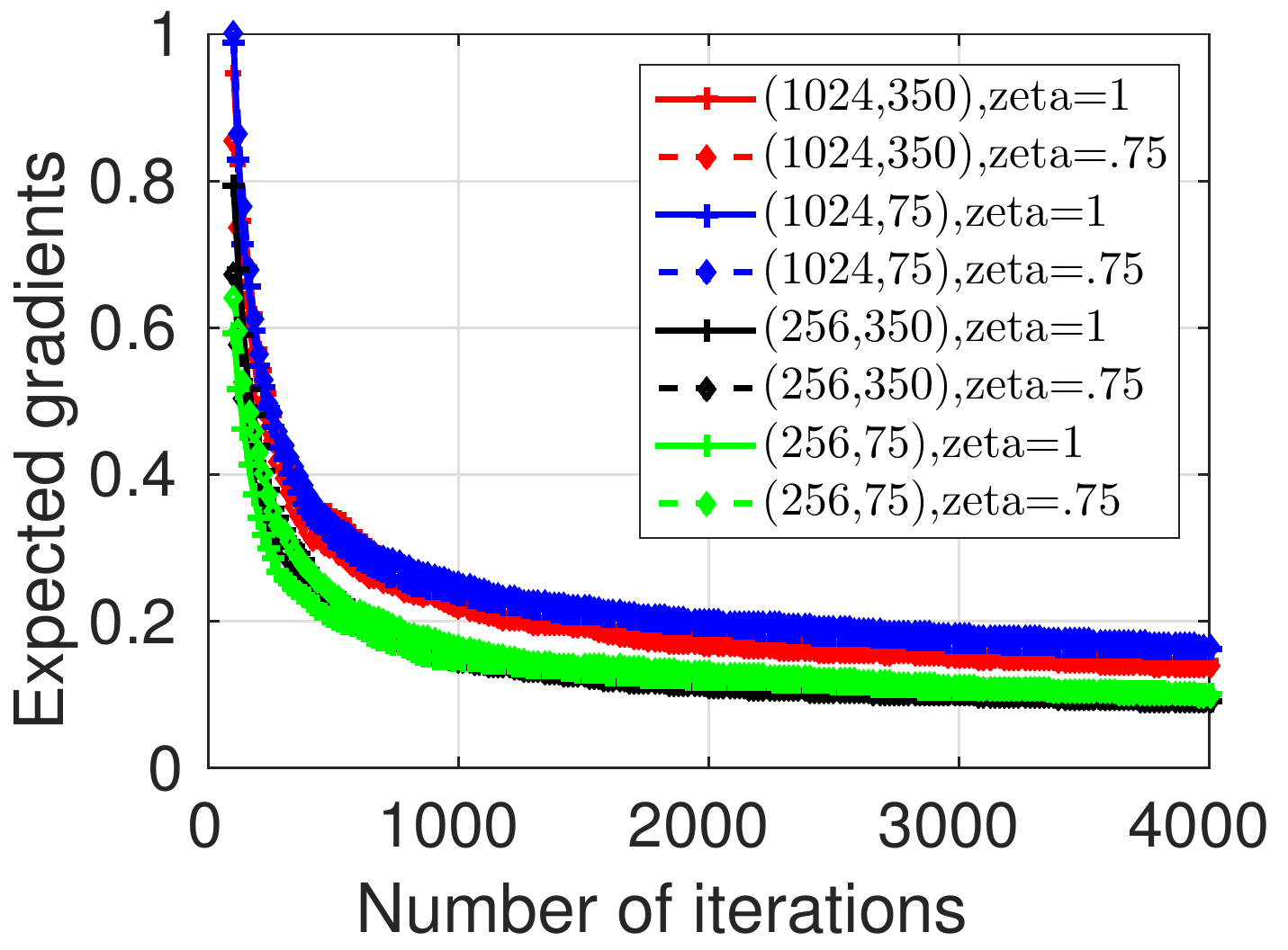}} \\ 
\subfloat[]{\includegraphics[width=47mm]{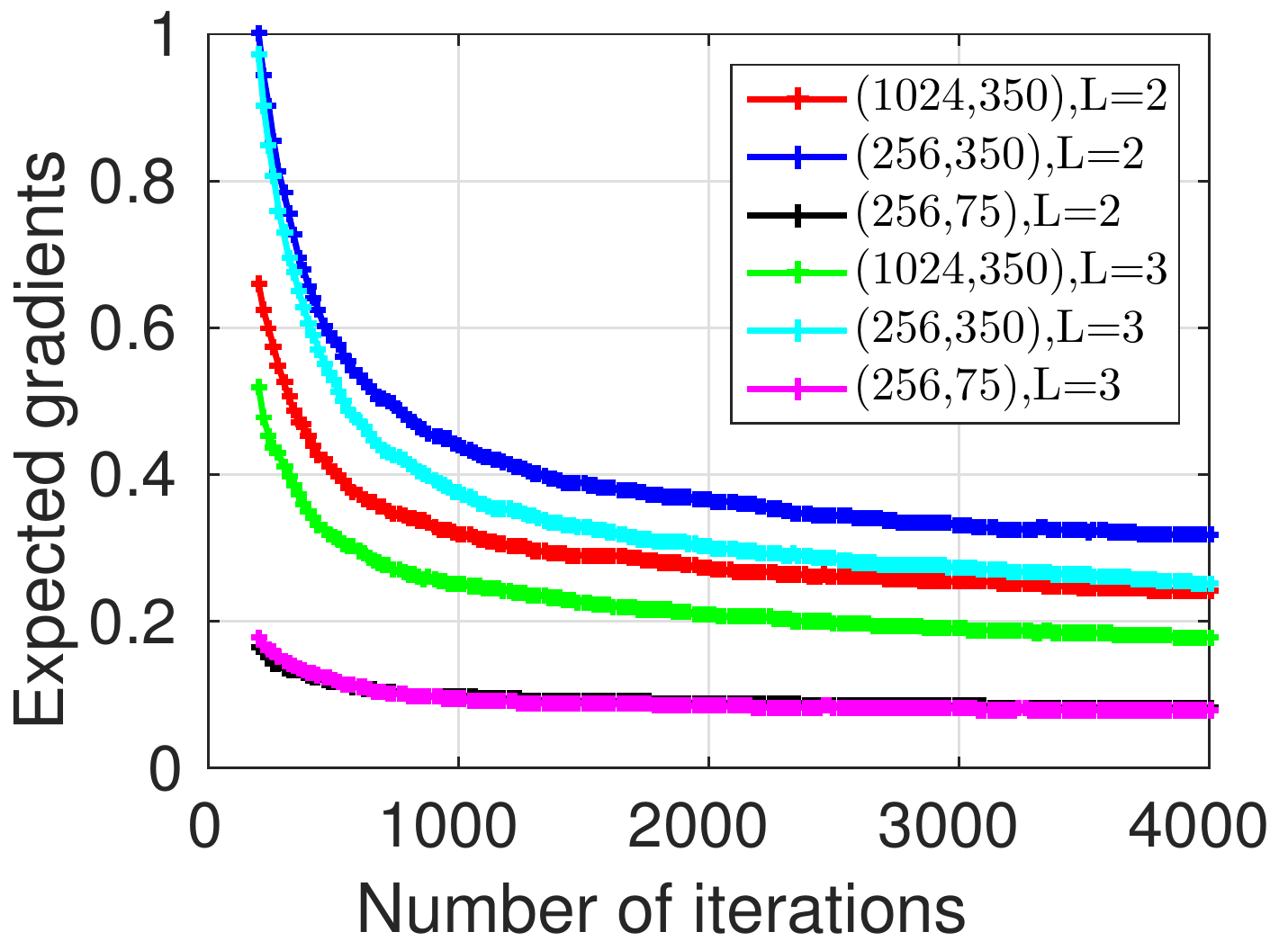}} \quad 
\subfloat[]{\includegraphics[width=47mm]{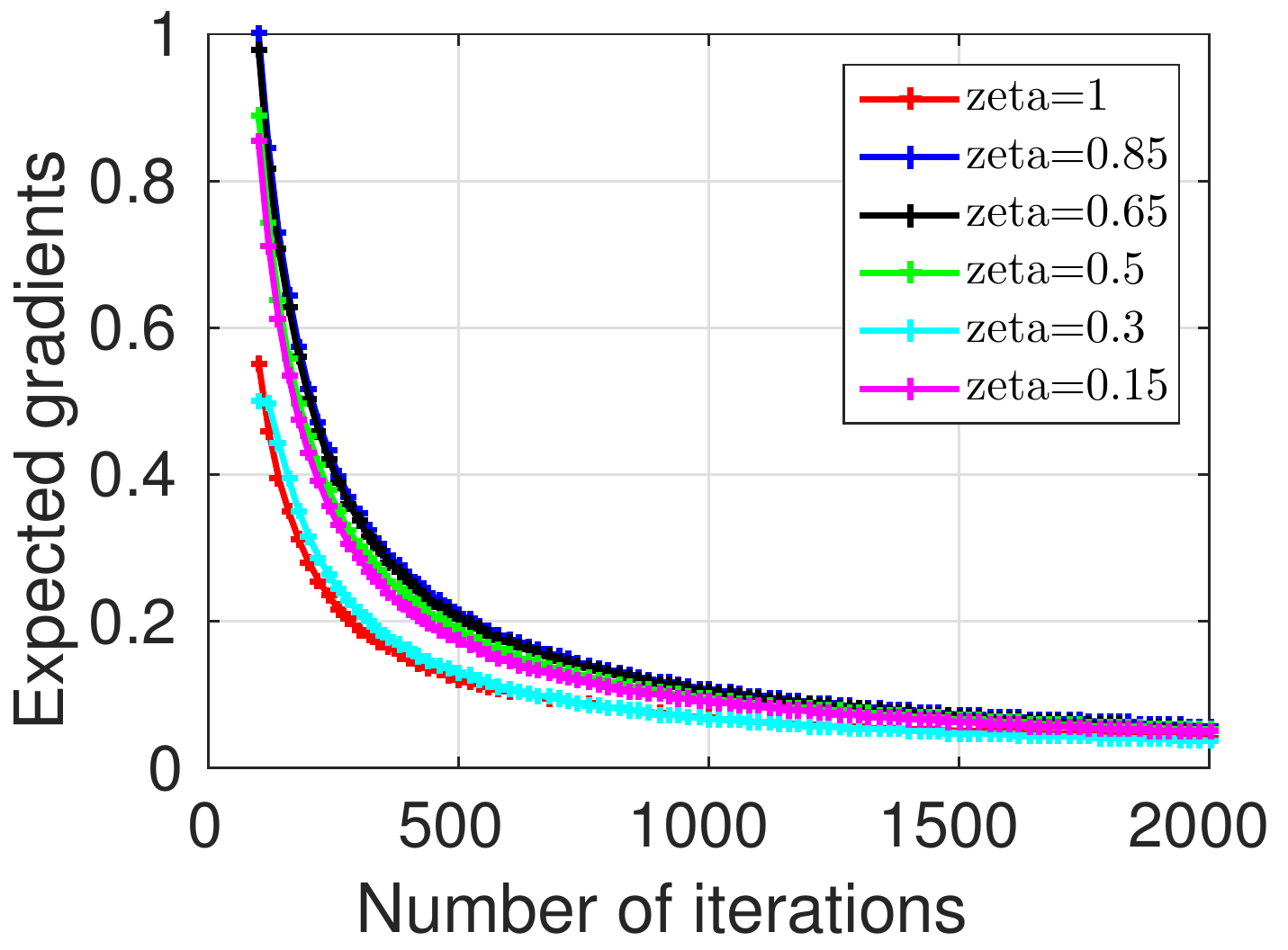}} \quad  
\subfloat[]{\includegraphics[width=47mm]{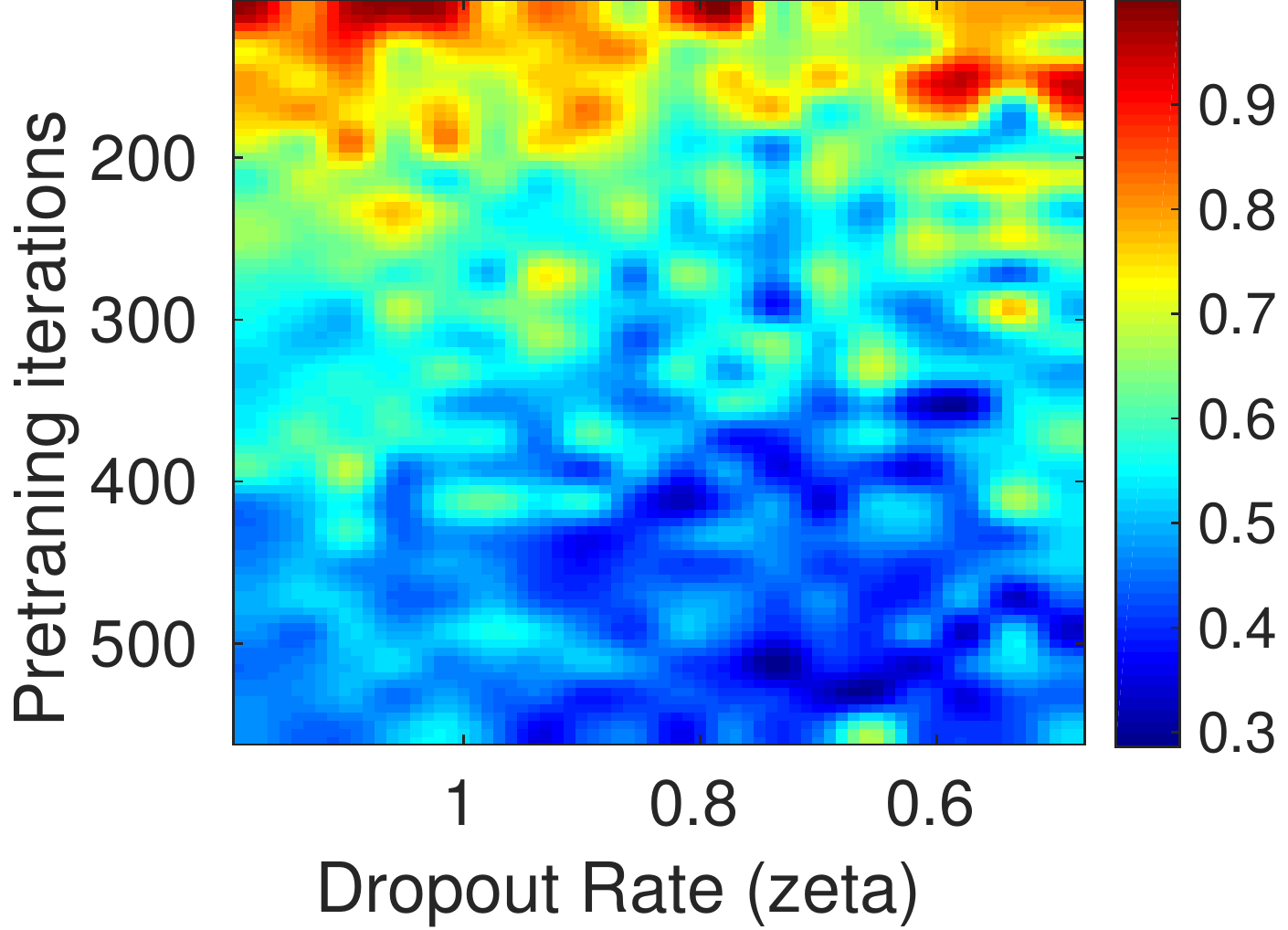}} 
\caption{\footnotesize \label{fig:allexp_cifar} {\bf Decay of expected gradients ({\it cifar})} Expected gradients in pretraining (a,b) and multi-layer net (c-e). 
(f) Pretraining vs. dropout. $B$ is set to $100$. All $y$-axes are scaled by corresponding maximum values.
The plots show that the trends predicted by the proposed framework are observed in practice -- the decay of expected gradients versus several hyperparameters.
Convergence is faster for smaller and thinner networks (a,c,d). 
Changes in visible layer length dominates the hidden layer lengths for DA (a). Dropout rate has less influence (e). Pretraining compensates for dropout (f).
Strong pretraining and dropout rate around $0.5$ gives fastest empirical gradients (dark blue region in (f)).}
\end{figure}

\begin{figure}[!h]\centering
\subfloat[]{\includegraphics[width=47mm]{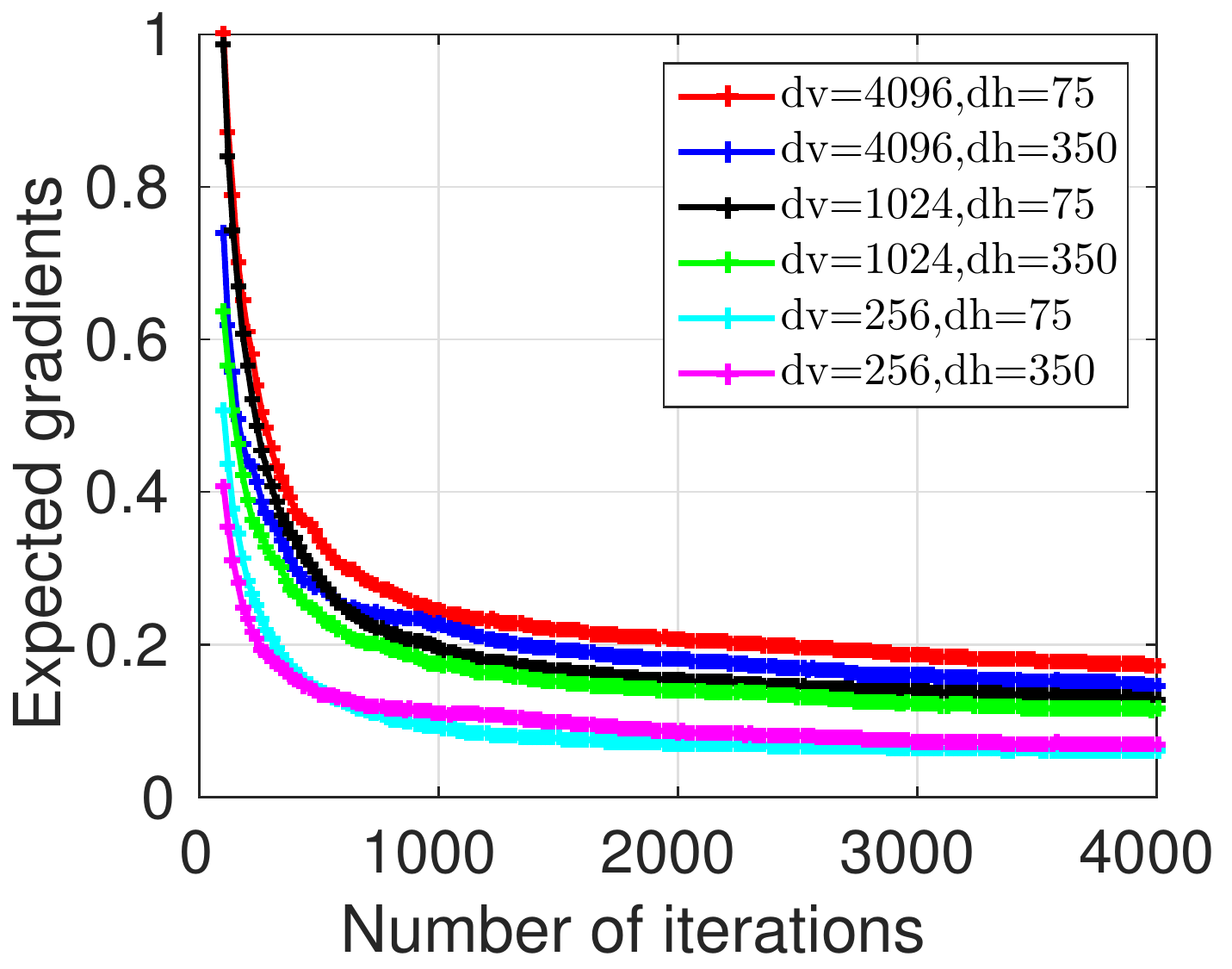}} \quad 
\subfloat[]{\includegraphics[width=47mm]{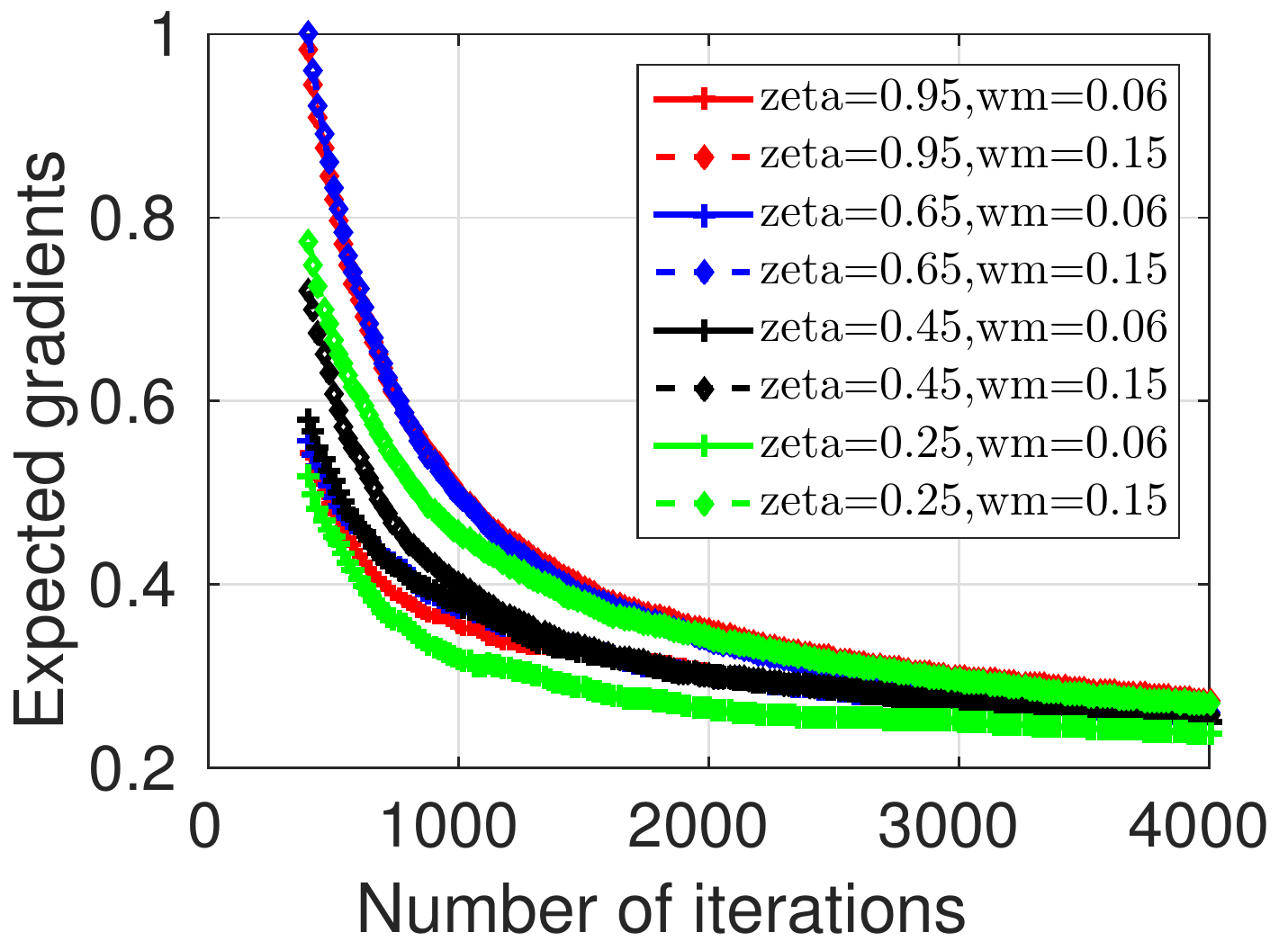}} \quad
\subfloat[]{\includegraphics[width=47mm]{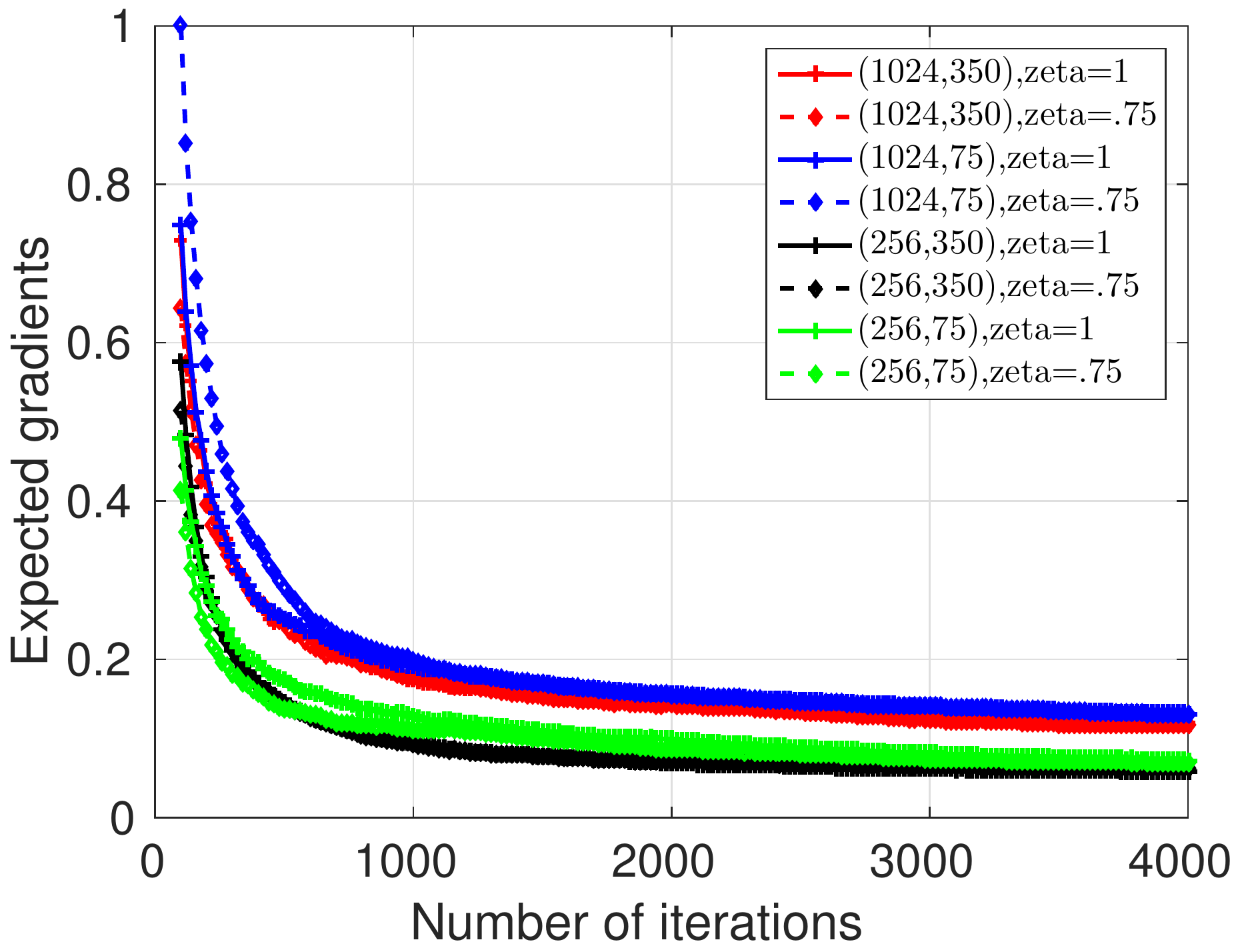}} \\
\subfloat[]{\includegraphics[width=47mm]{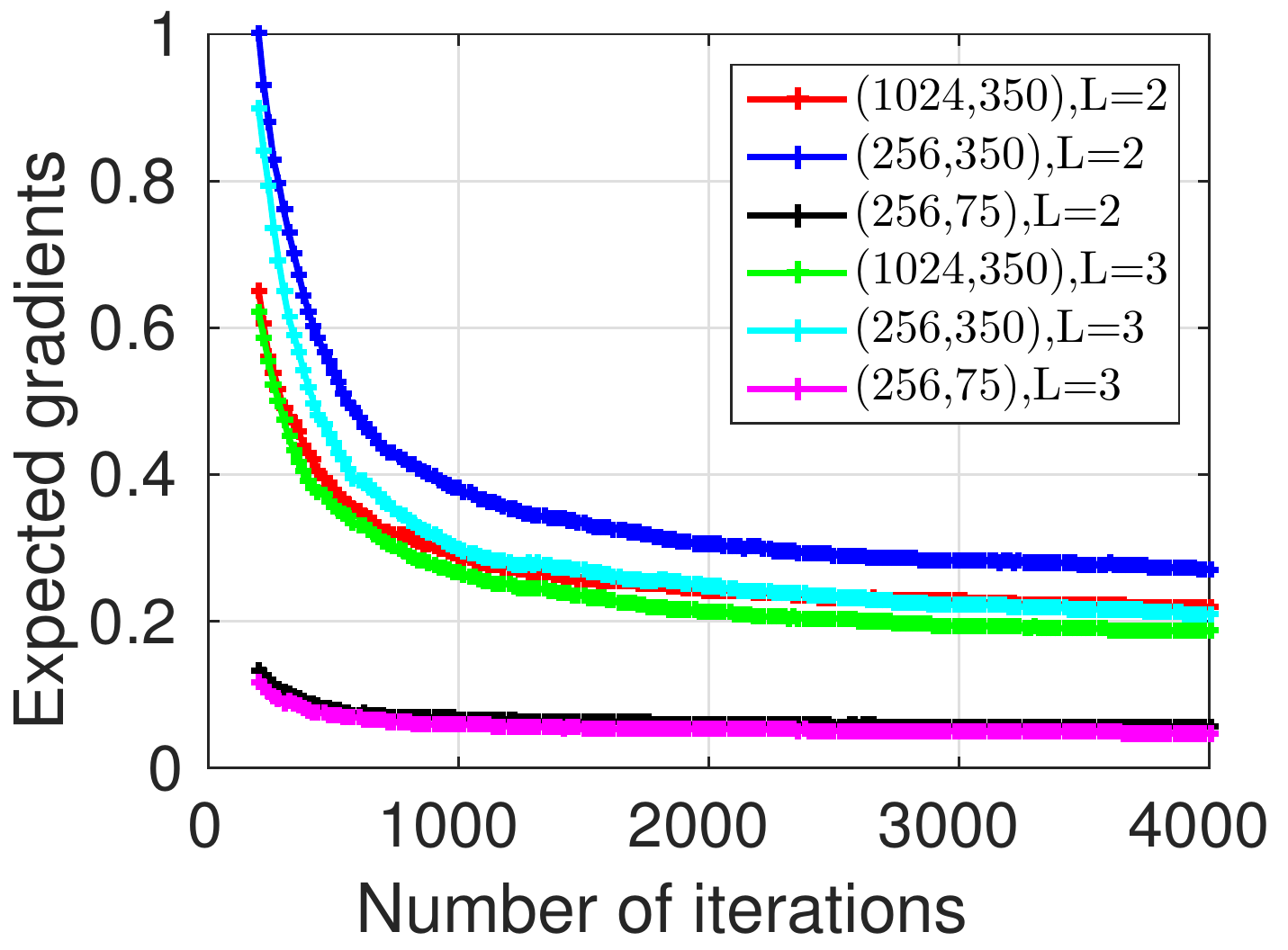}} \quad 
\subfloat[]{\includegraphics[width=47mm]{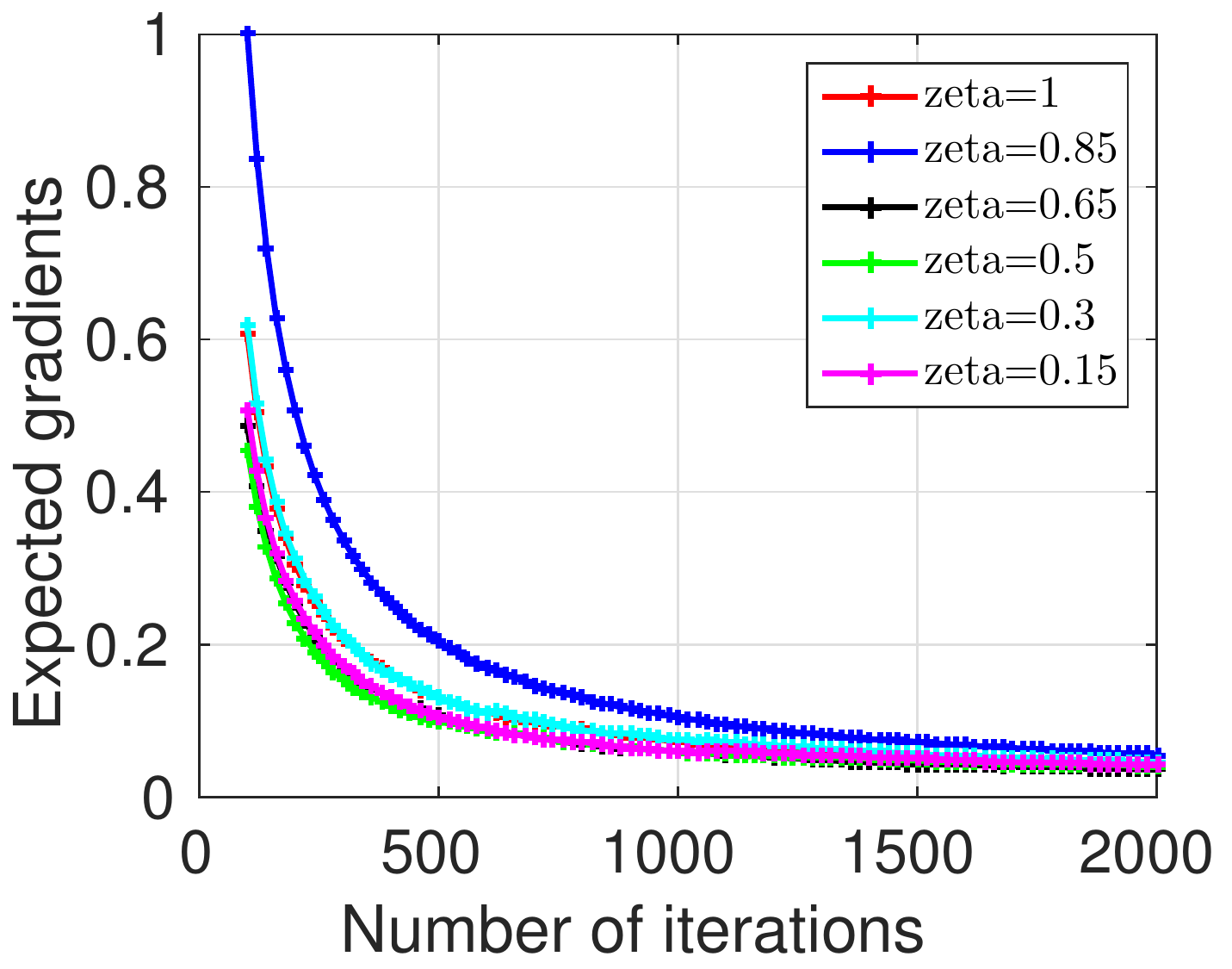}} \quad  
\subfloat[]{\includegraphics[width=47mm]{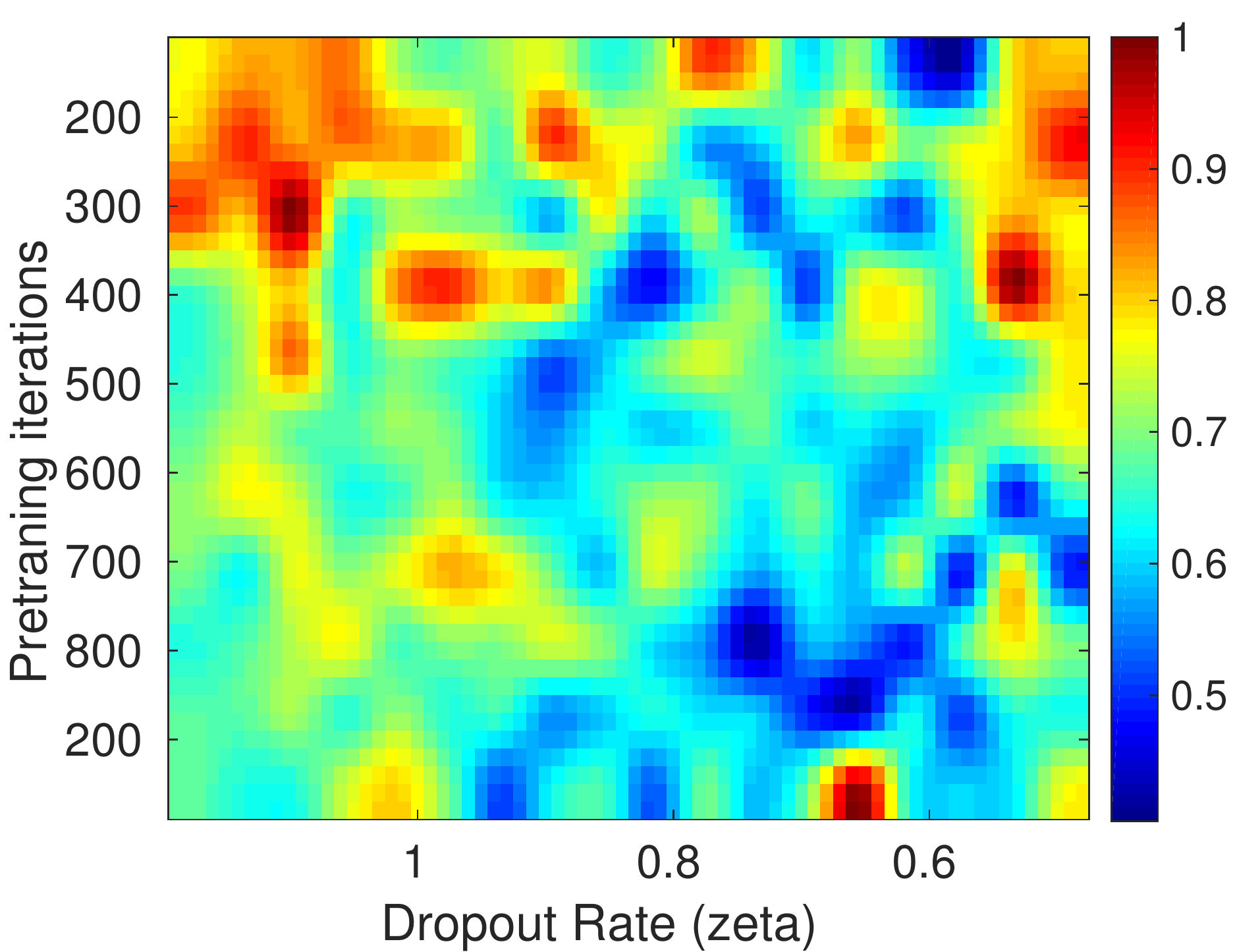}}
\caption{\footnotesize \label{fig:allexp_caltech} {\bf Decay of expected gradients ({\it caltech})} Expected gradients in pretraining (a,b) and multi-layer net (c-e). 
(f) Pretraining vs. dropout. $B$ is set to $100$. All $y$-axes are scaled by corresponding maximum values.
The plots show that the trends predicted by the proposed framework are observed in practice -- the decay of expected gradients versus several hyperparameters.
Convergence is faster for smaller and thinner networks (a,c,d). 
Changes in visible layer length dominates the hidden layer lengths for DA (a). Dropout rate has less influence (e). Pretraining compensates for dropout (f).
Strong pretraining and dropout rate around $0.5$ gives fastest empirical gradients (dark blue region in (f)).}
\end{figure}

\begin{figure}[!h]\centering
\subfloat[]{\includegraphics[width=47mm]{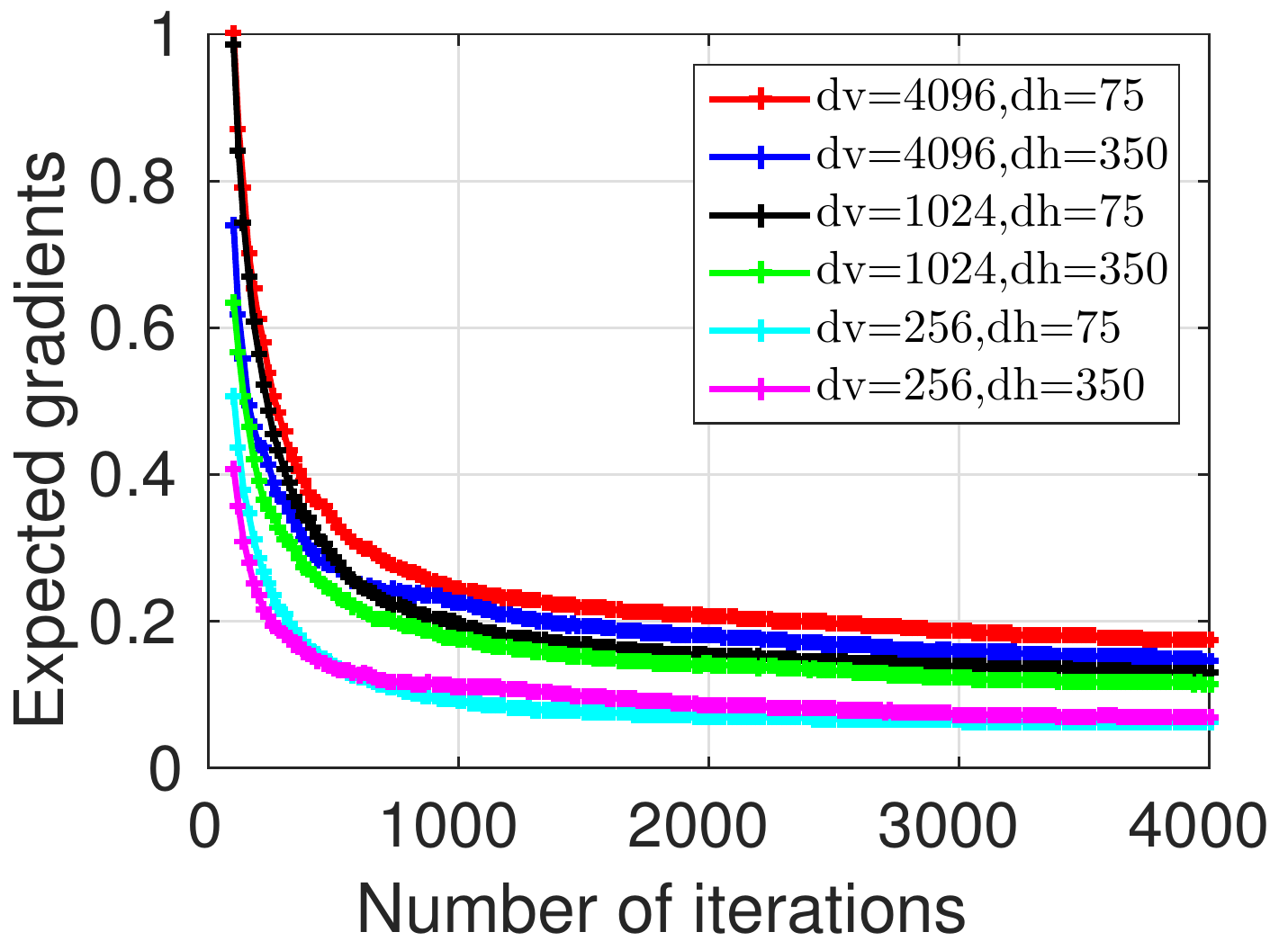}} \quad 
\subfloat[]{\includegraphics[width=47mm]{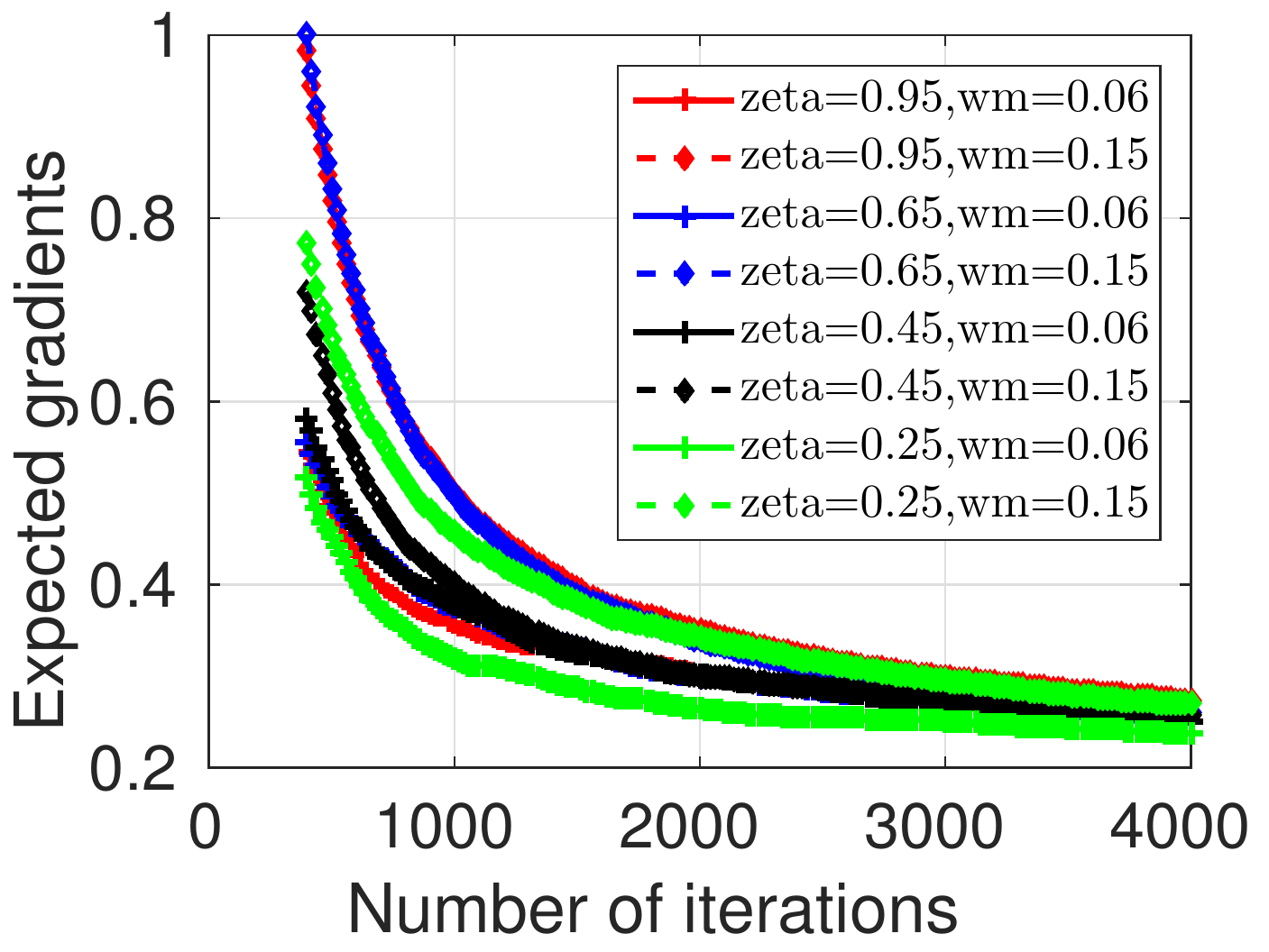}} \quad
\subfloat[]{\includegraphics[width=47mm]{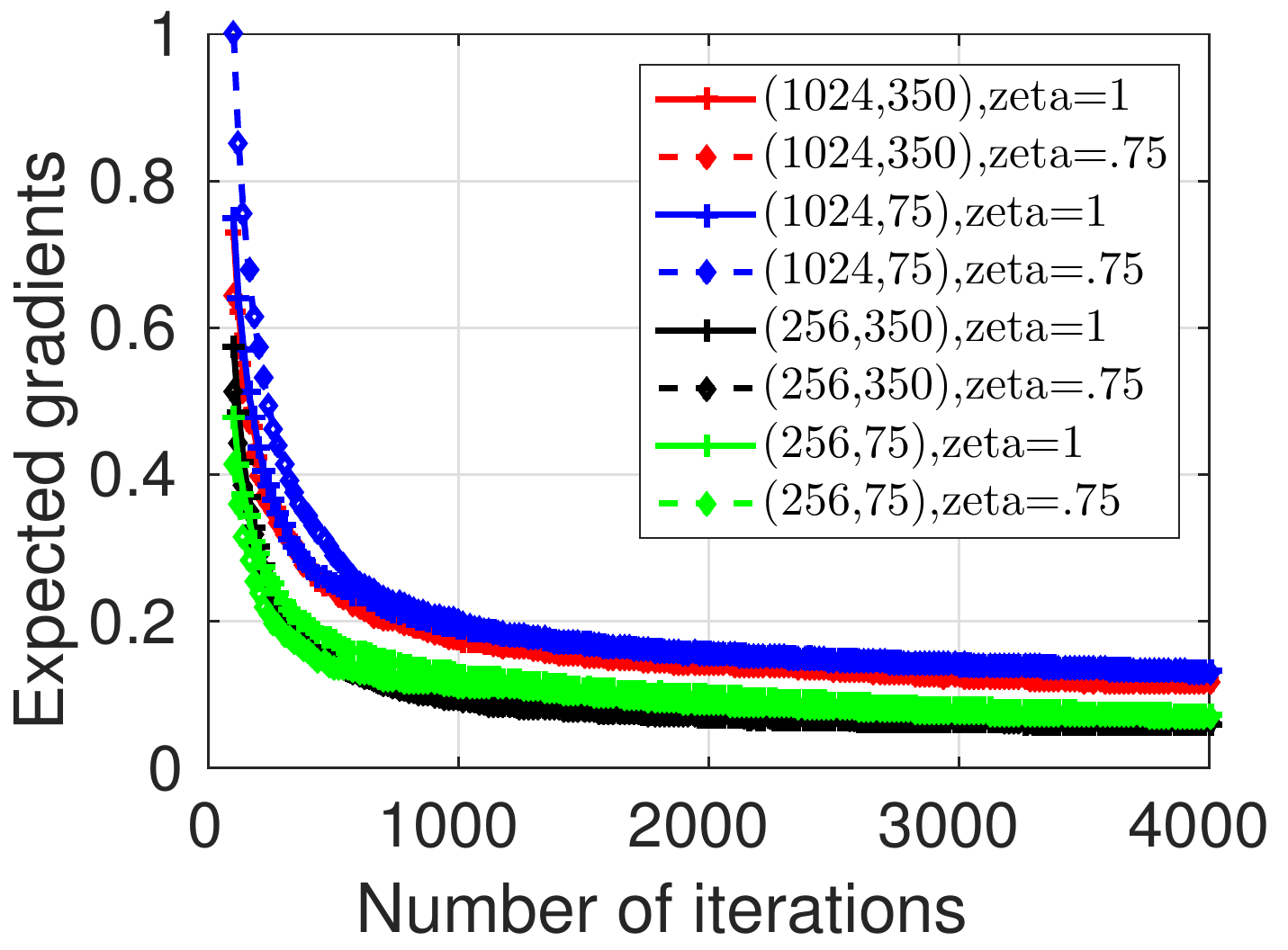}} \\
\subfloat[]{\includegraphics[width=47mm]{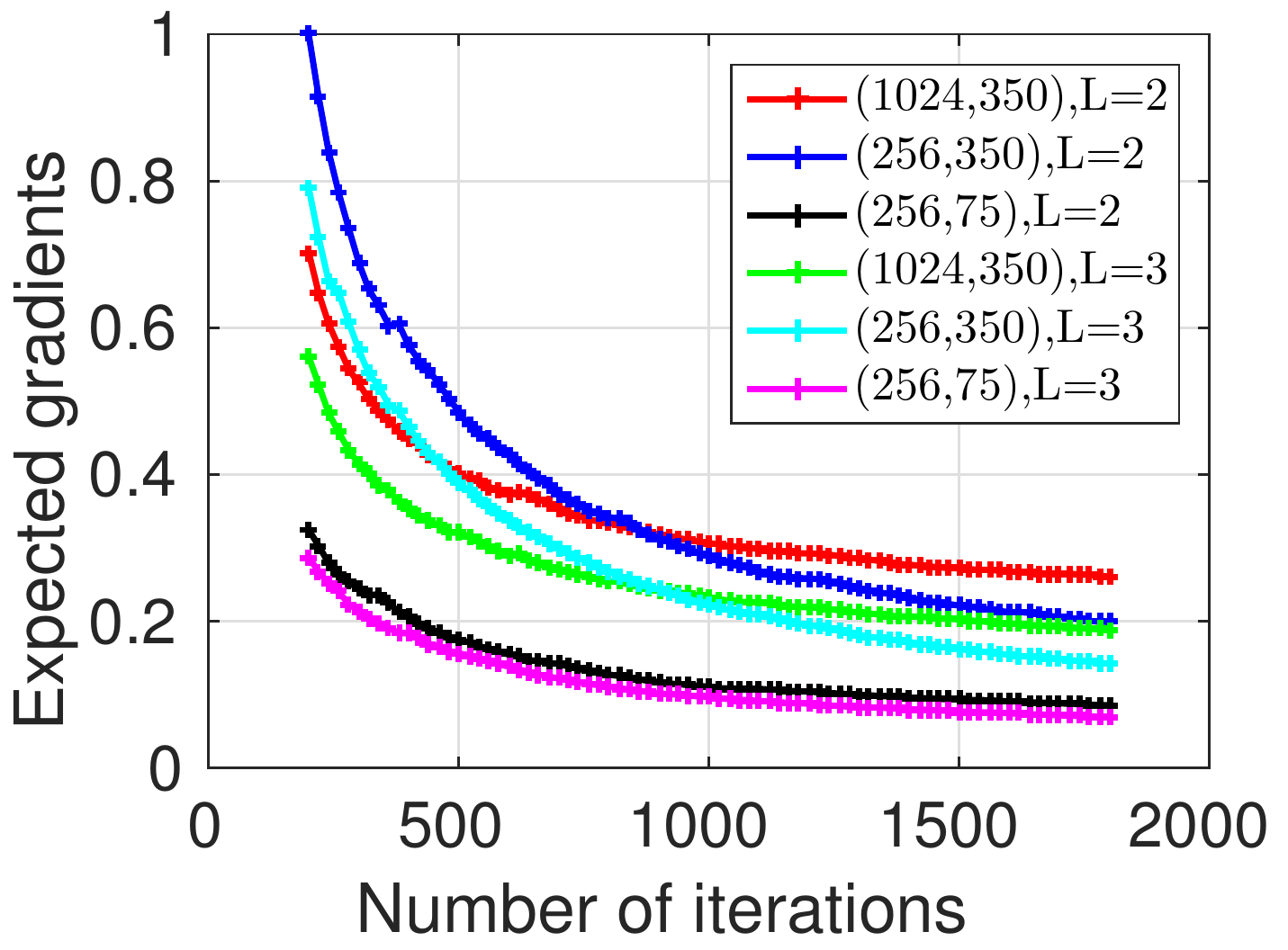}} \quad 
\subfloat[]{\includegraphics[width=47mm]{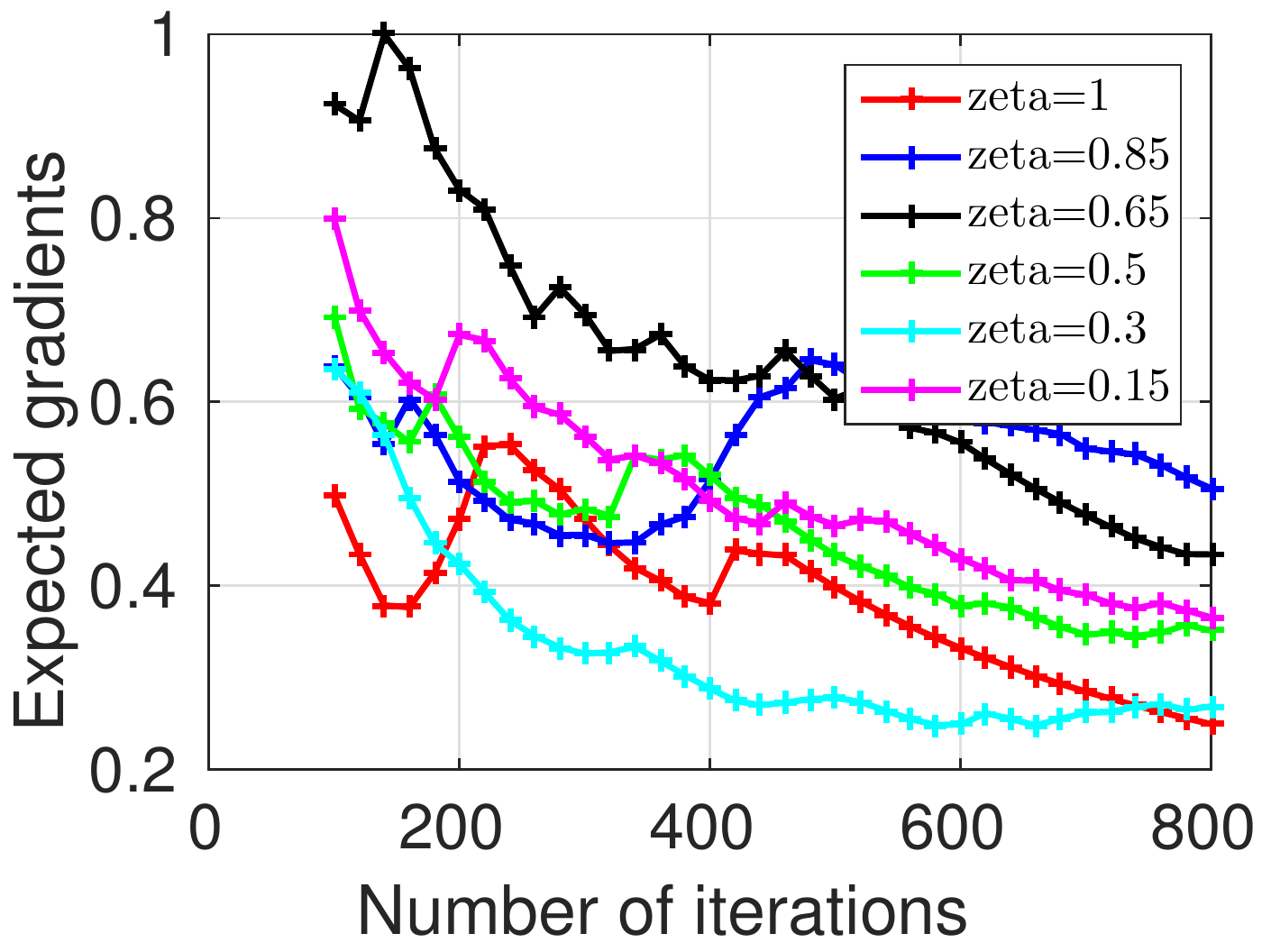}} \quad  
\subfloat[]{\includegraphics[width=47mm]{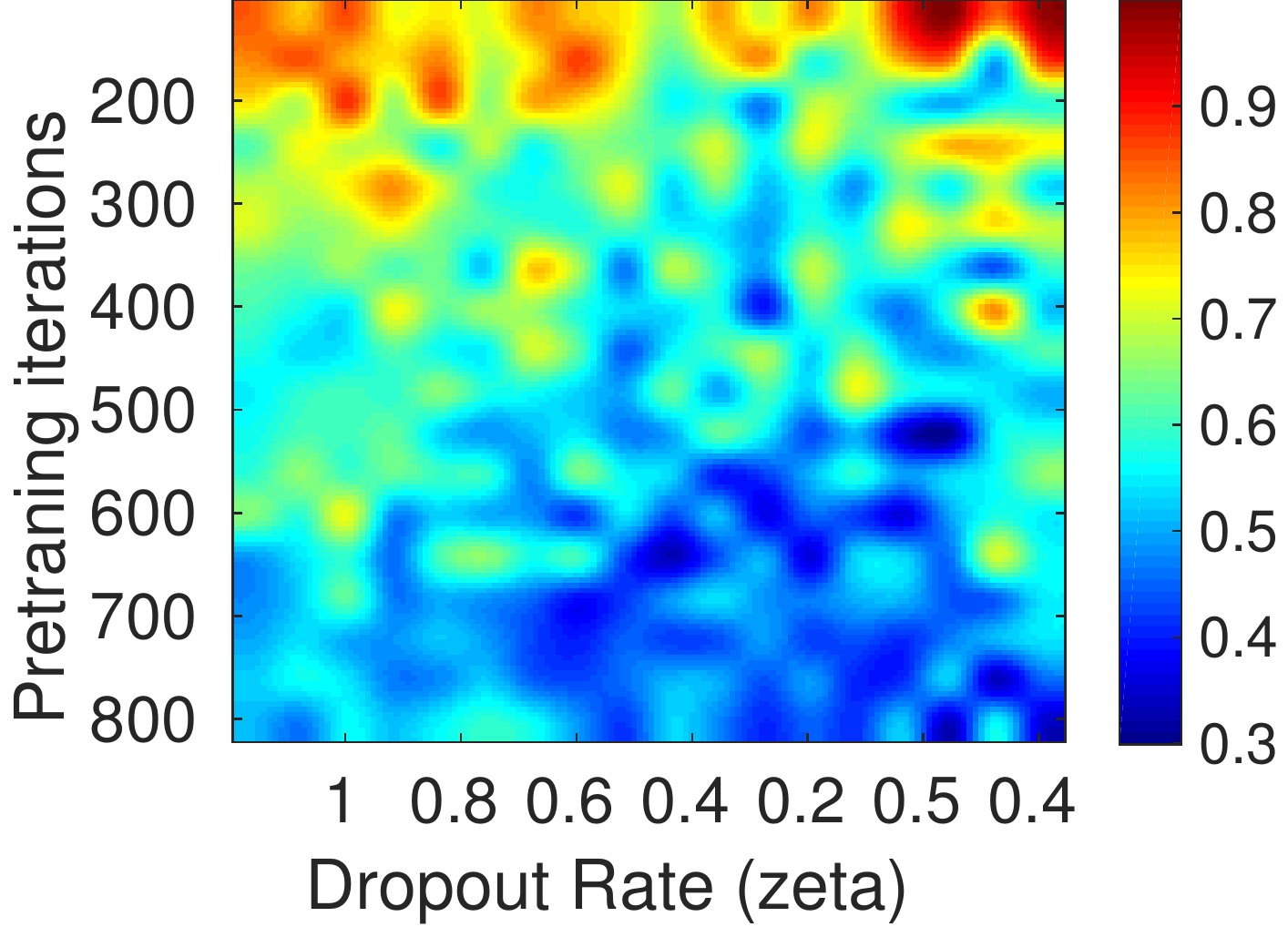}}
\caption{\footnotesize \label{fig:allexp_neuro} {\bf Decay of expected gradients ({\it neuro})} Expected gradients in pretraining (a,b) and multi-layer net (c-e). 
(f) Pretraining vs. dropout. $B$ is set to $100$. All $y$-axes are scaled by corresponding maximum values.
The plots show that the trends predicted by the proposed framework are observed in practice -- the decay of expected gradients versus several hyperparameters.
Convergence is faster for smaller and thinner networks (a,c,d). 
Changes in visible layer length dominates the hidden layer lengths for DA (a). Dropout rate has less influence (e). Pretraining compensates for dropout (f).
Strong pretraining and dropout rate around $0.5$ gives fastest empirical gradients (dark blue region in (f)).}
\end{figure}

\begin{figure}[!p]\centering
\subfloat[]{\includegraphics[width=47mm]{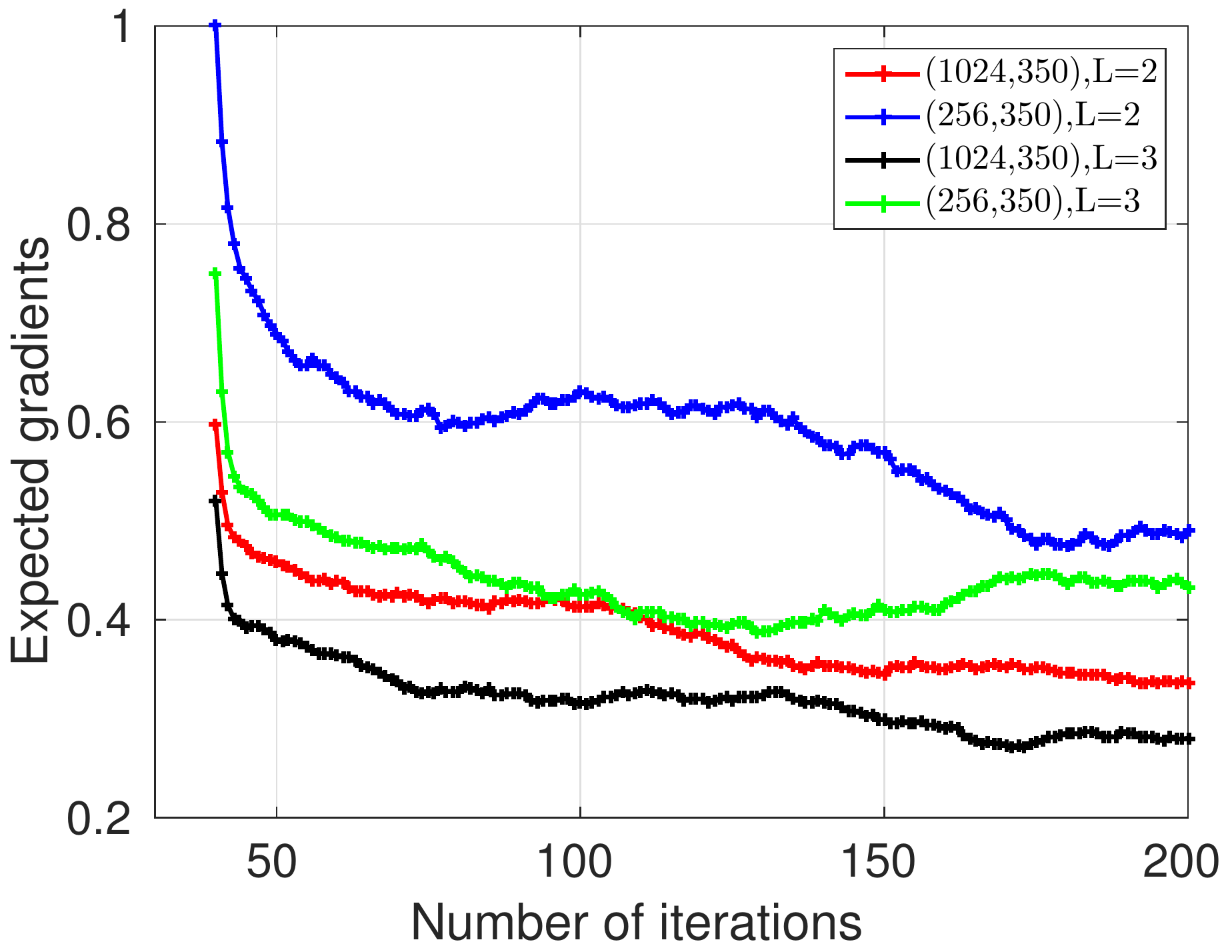}} \quad 
\subfloat[]{\includegraphics[width=47mm]{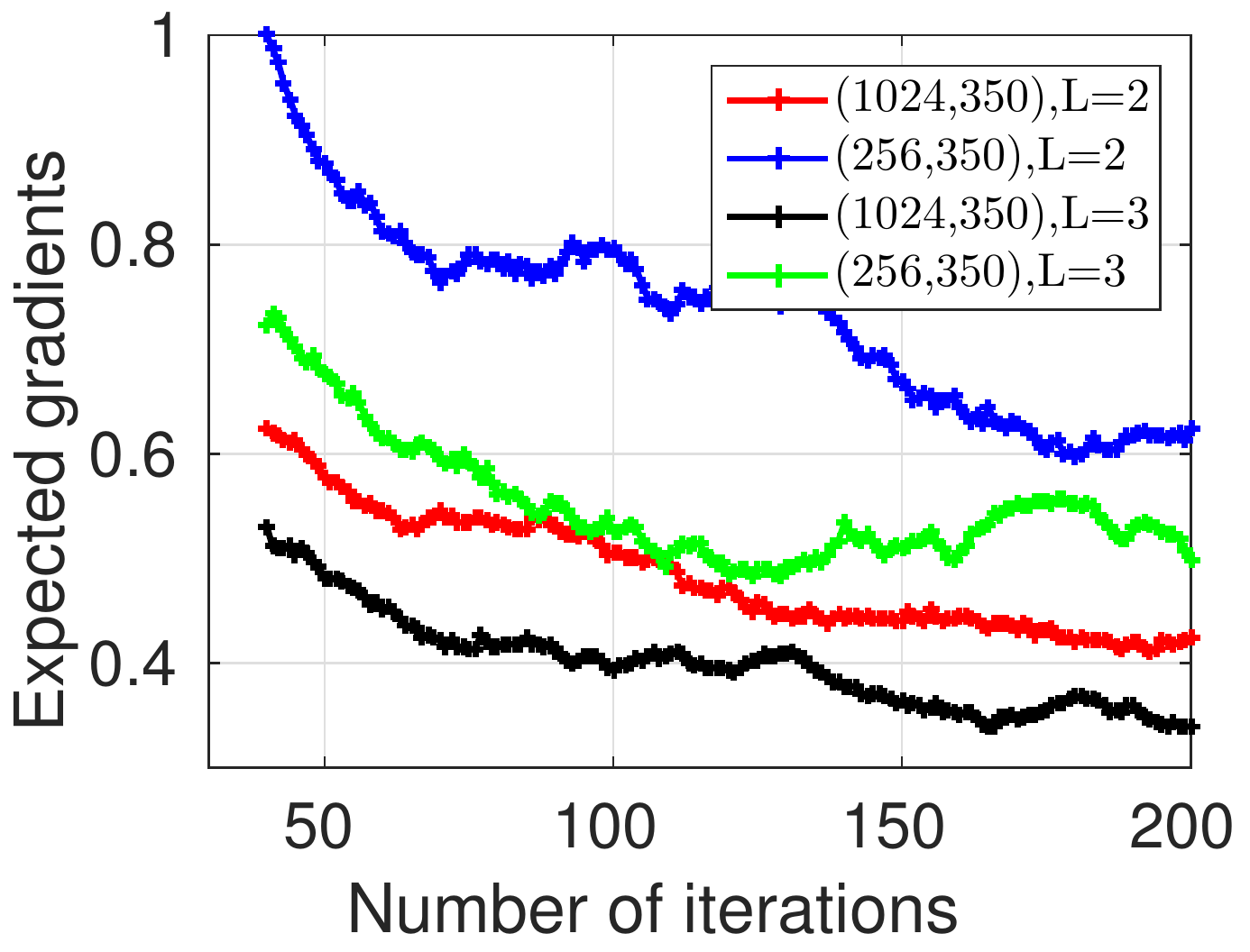}} \quad
\subfloat[]{\includegraphics[width=47mm]{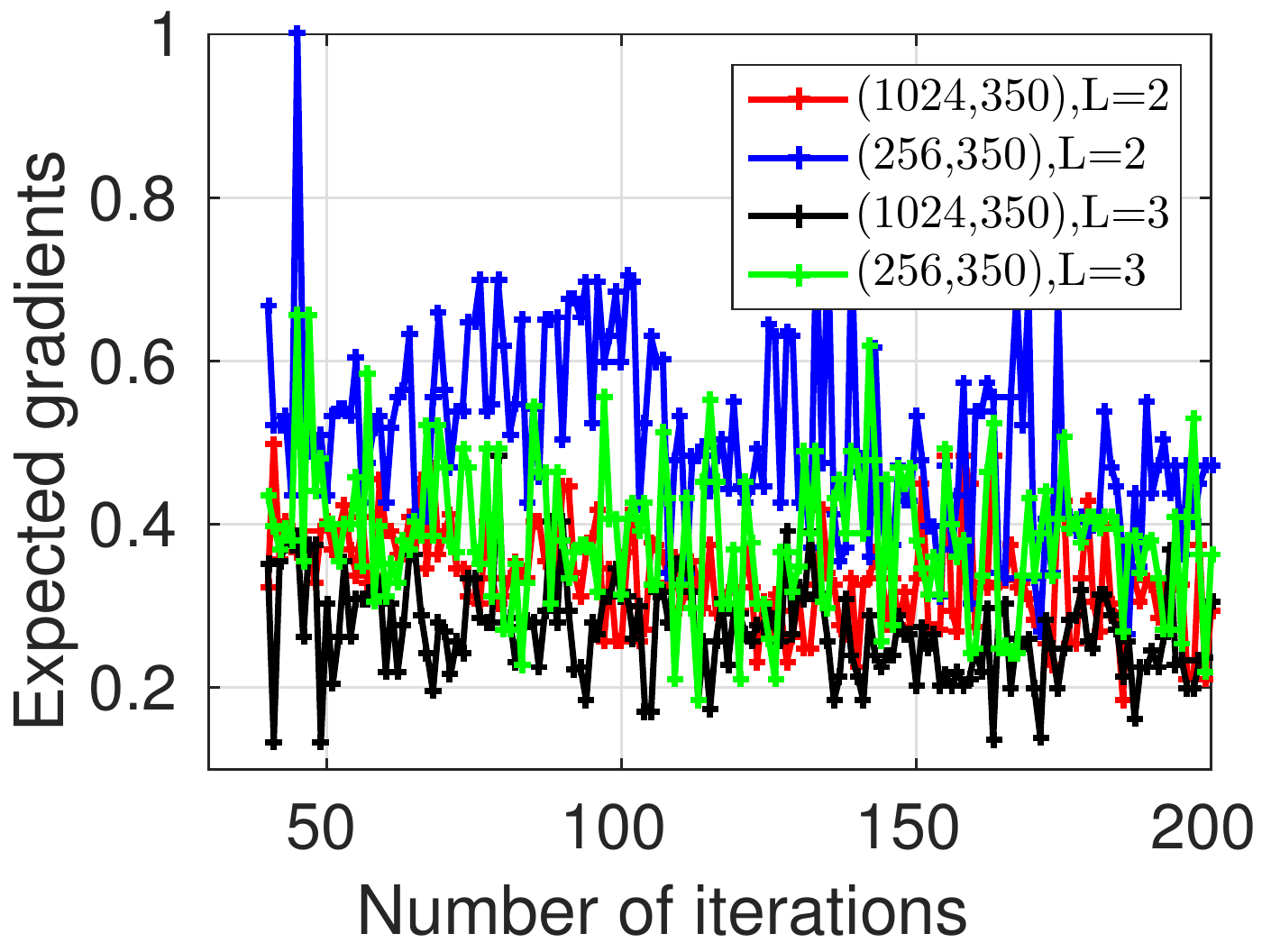}} \\
\subfloat[]{\includegraphics[width=47mm]{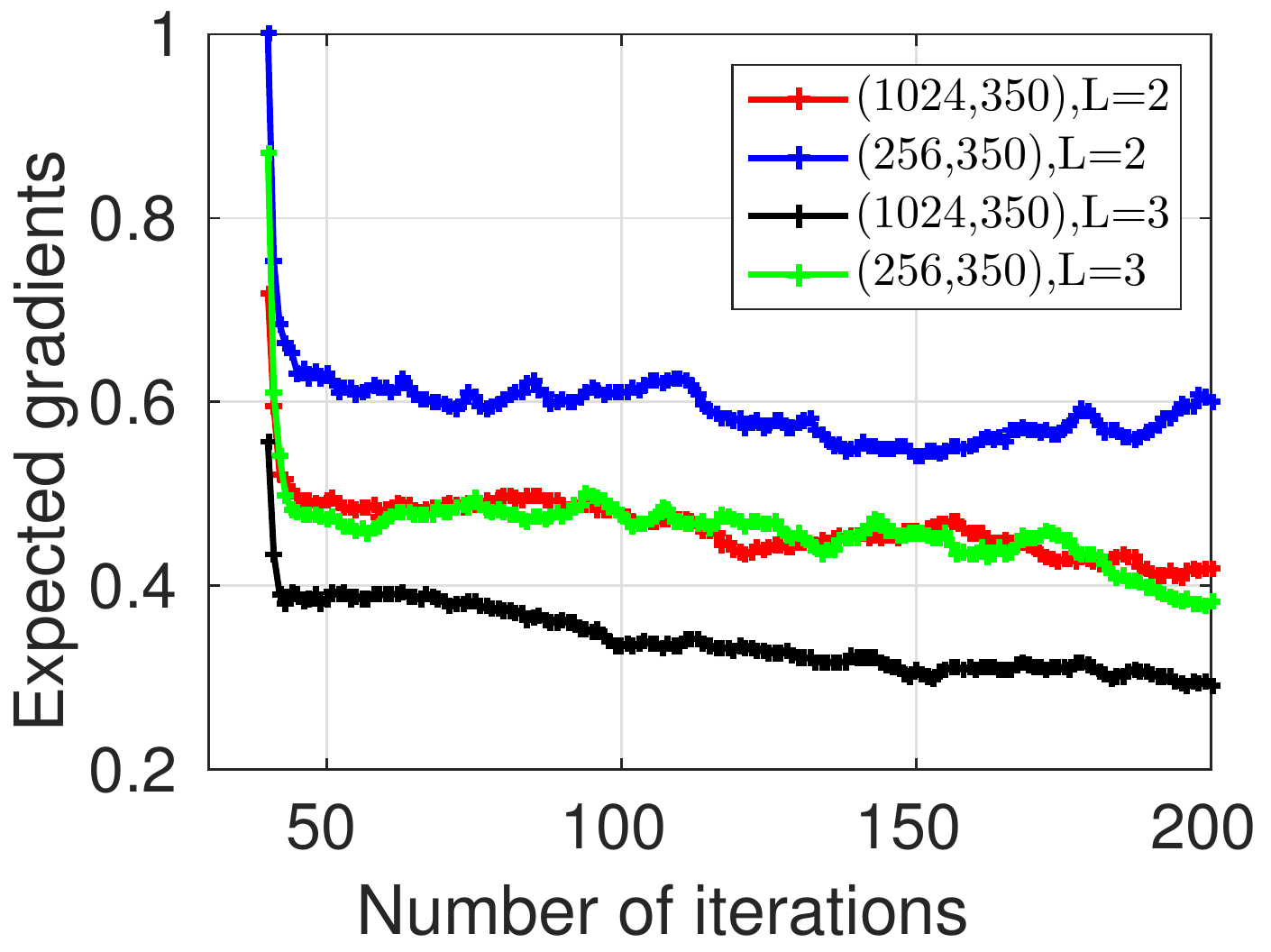}} \quad 
\subfloat[]{\includegraphics[width=47mm]{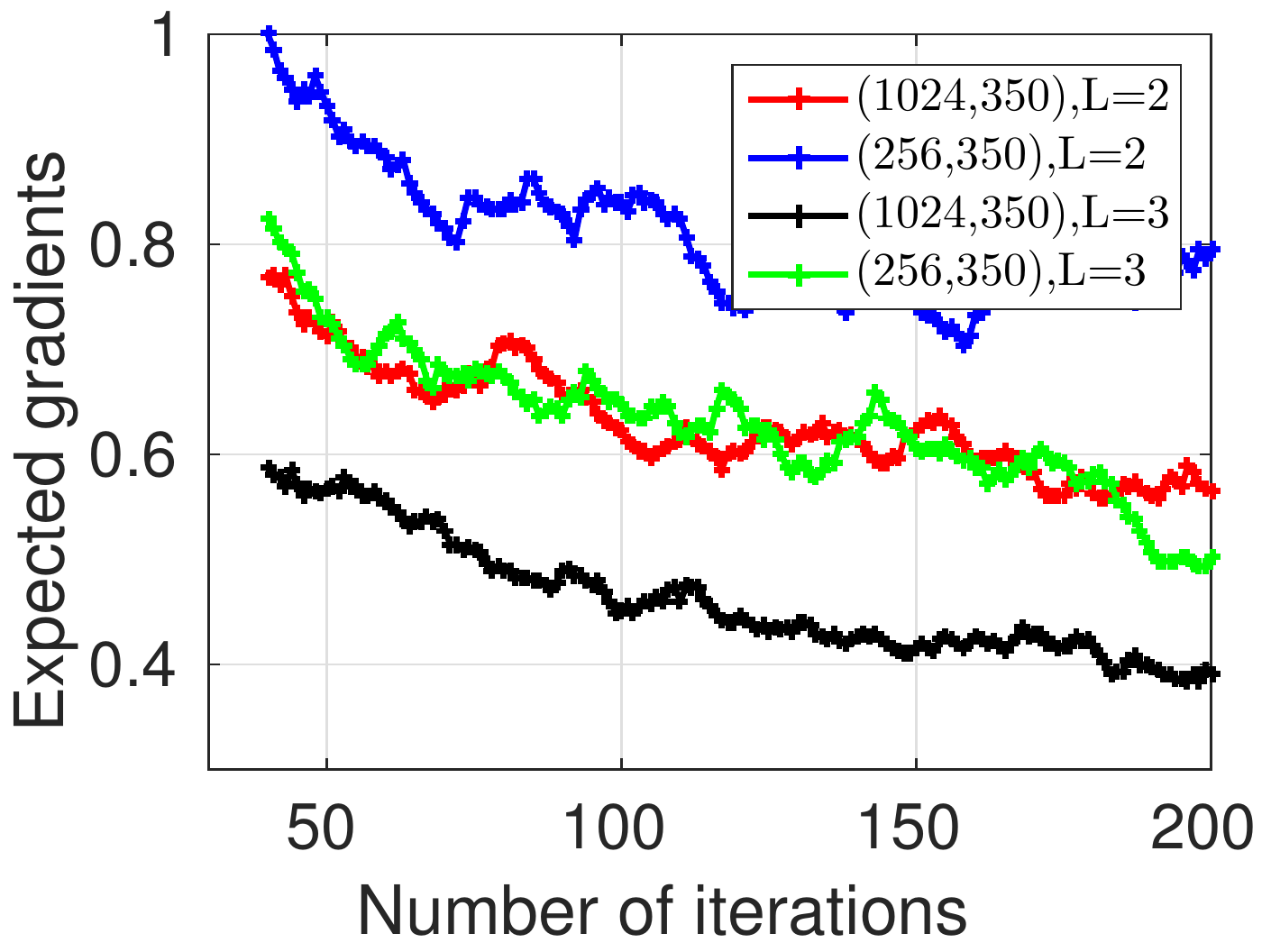}} \quad
\subfloat[]{\includegraphics[width=47mm]{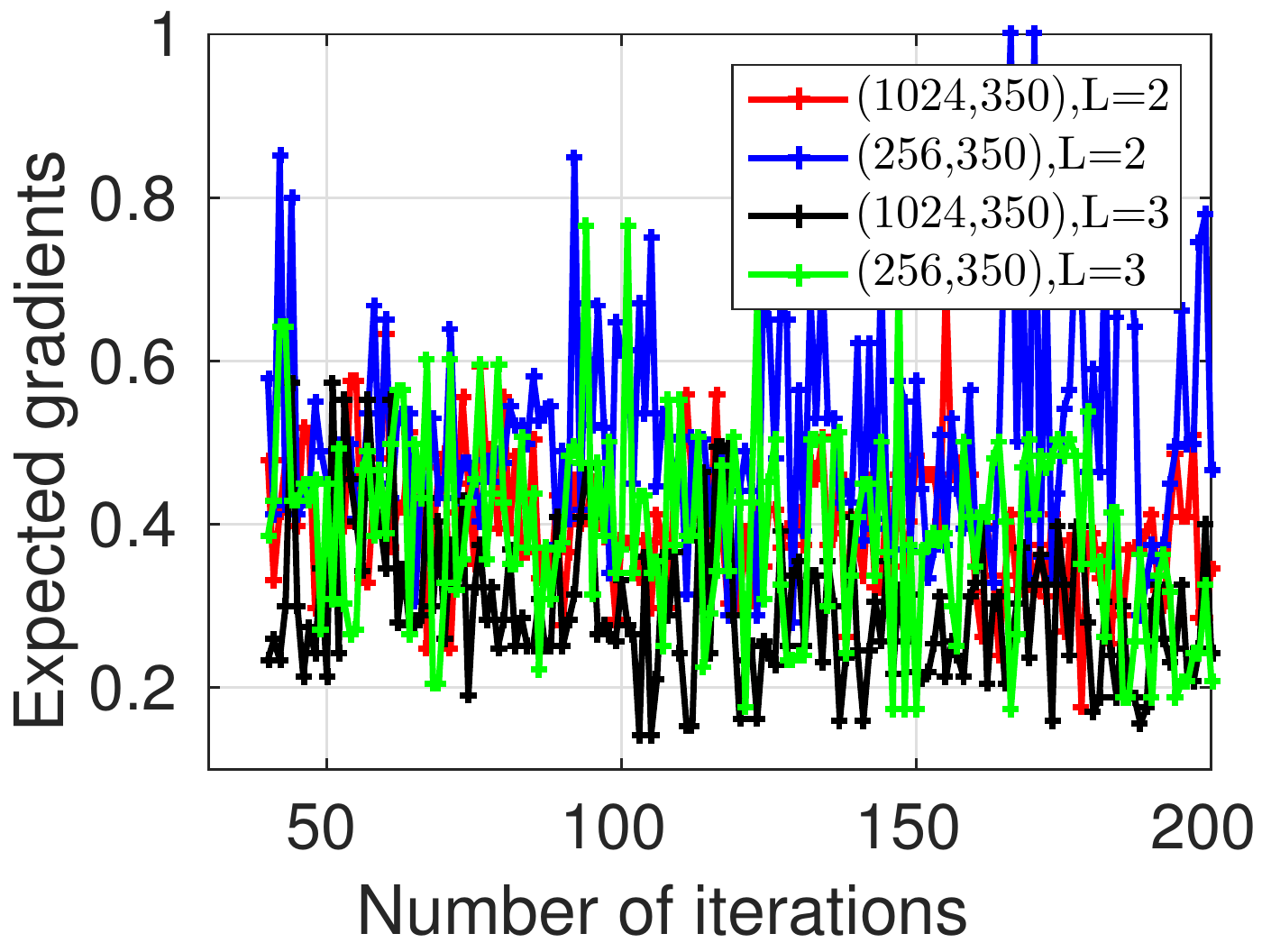}} \\
\subfloat[]{\includegraphics[width=47mm]{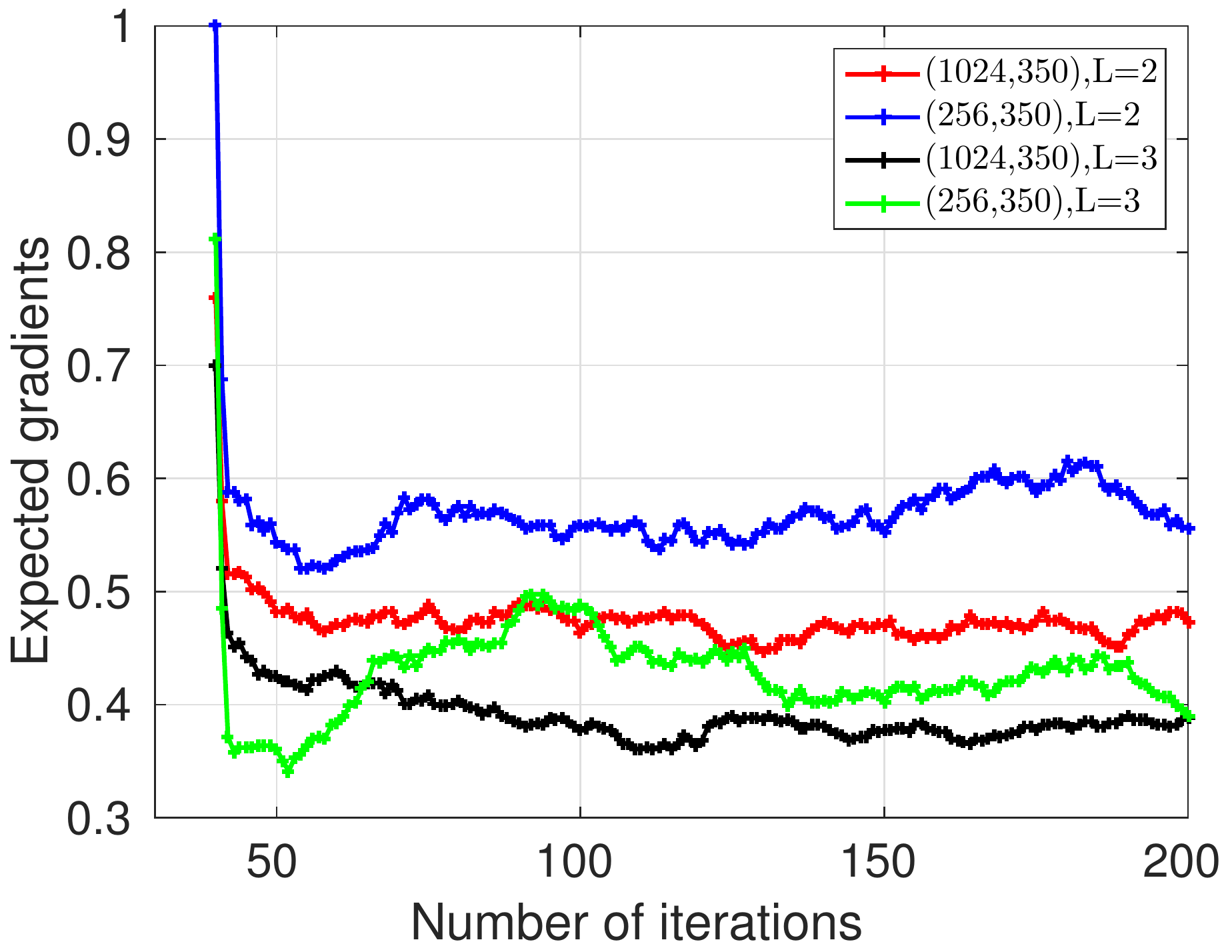}} \quad 
\subfloat[]{\includegraphics[width=47mm]{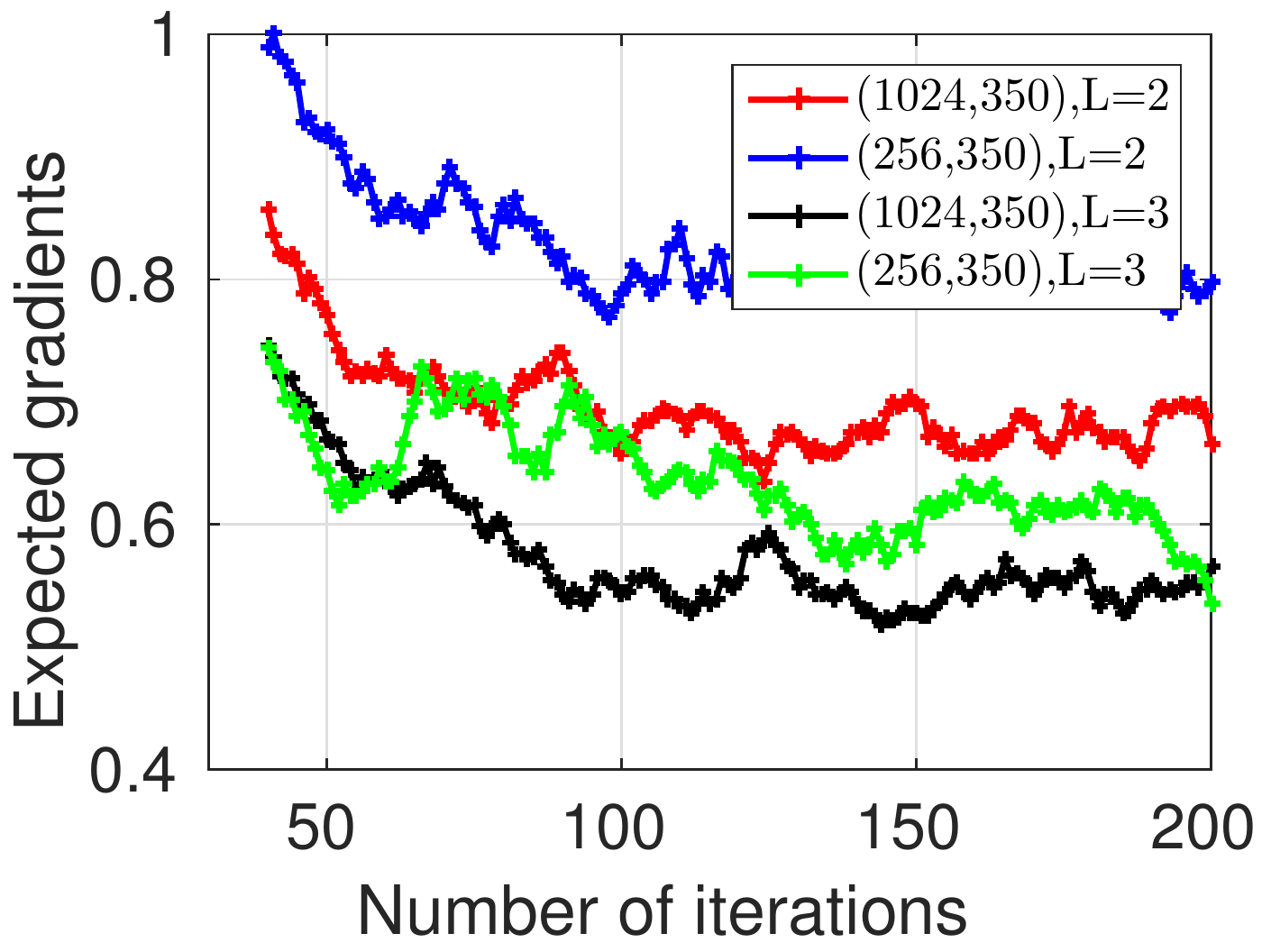}} \quad
\subfloat[]{\includegraphics[width=47mm]{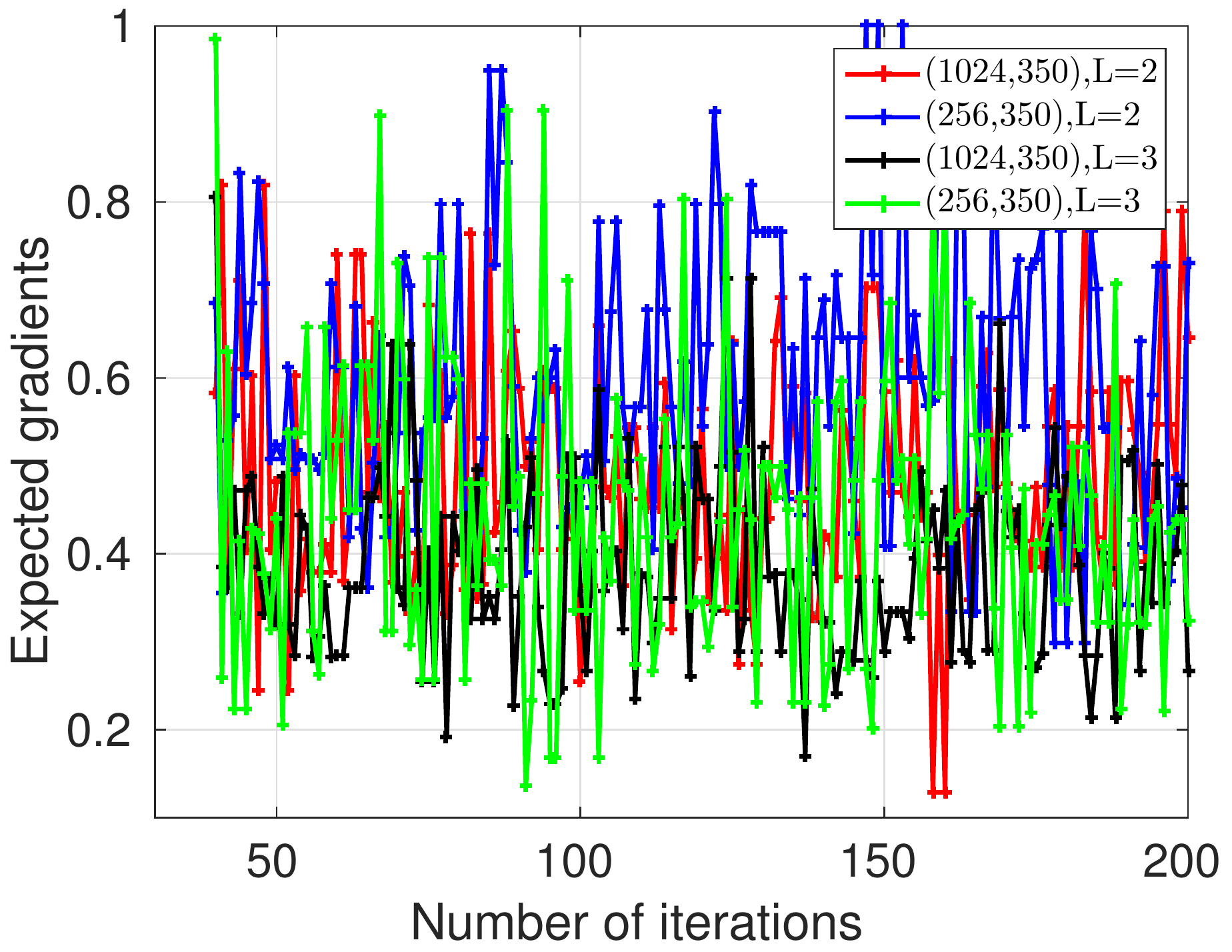}} \\
\subfloat[]{\includegraphics[width=47mm]{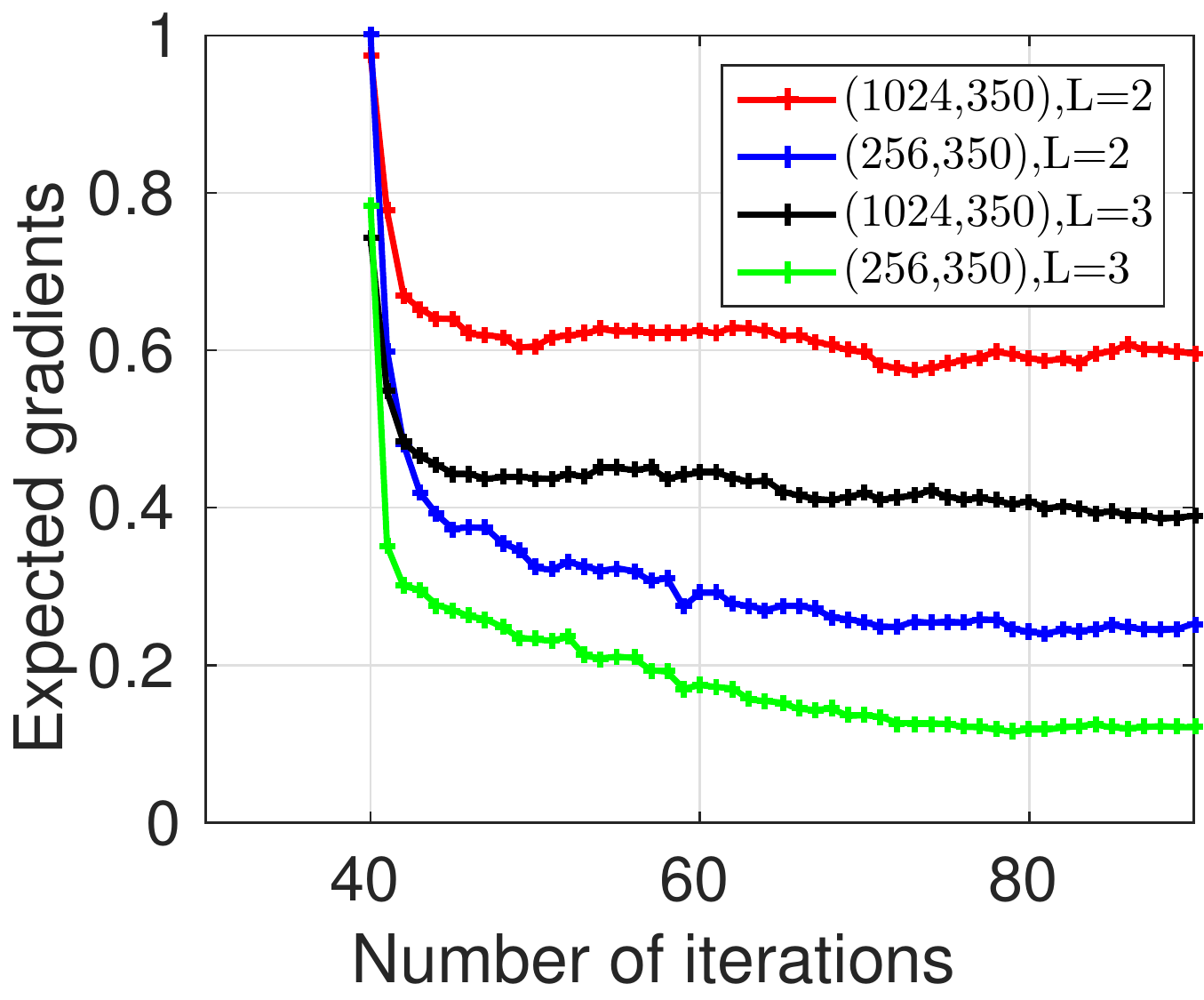}} \quad 
\subfloat[]{\includegraphics[width=47mm]{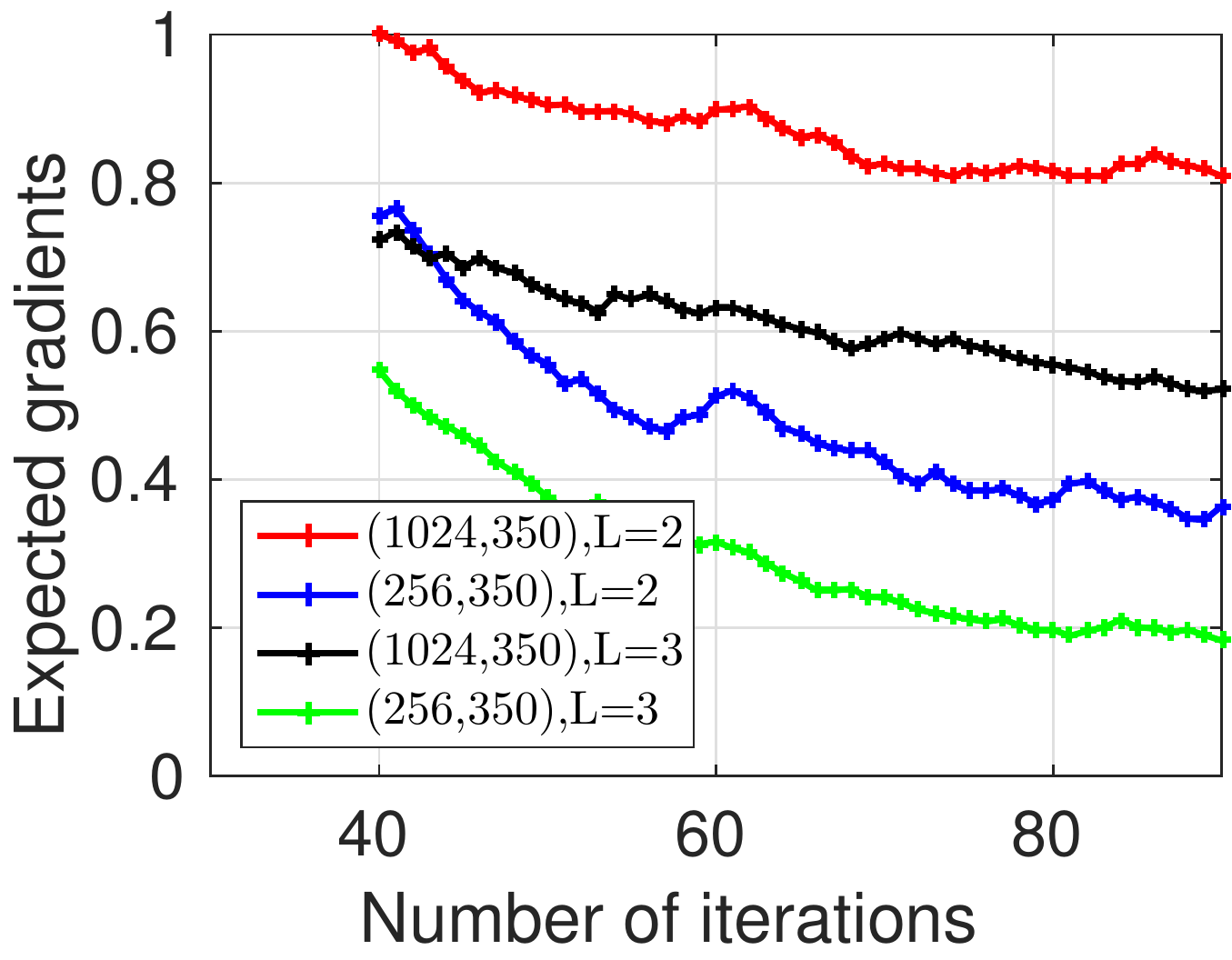}} \quad
\subfloat[]{\includegraphics[width=47mm]{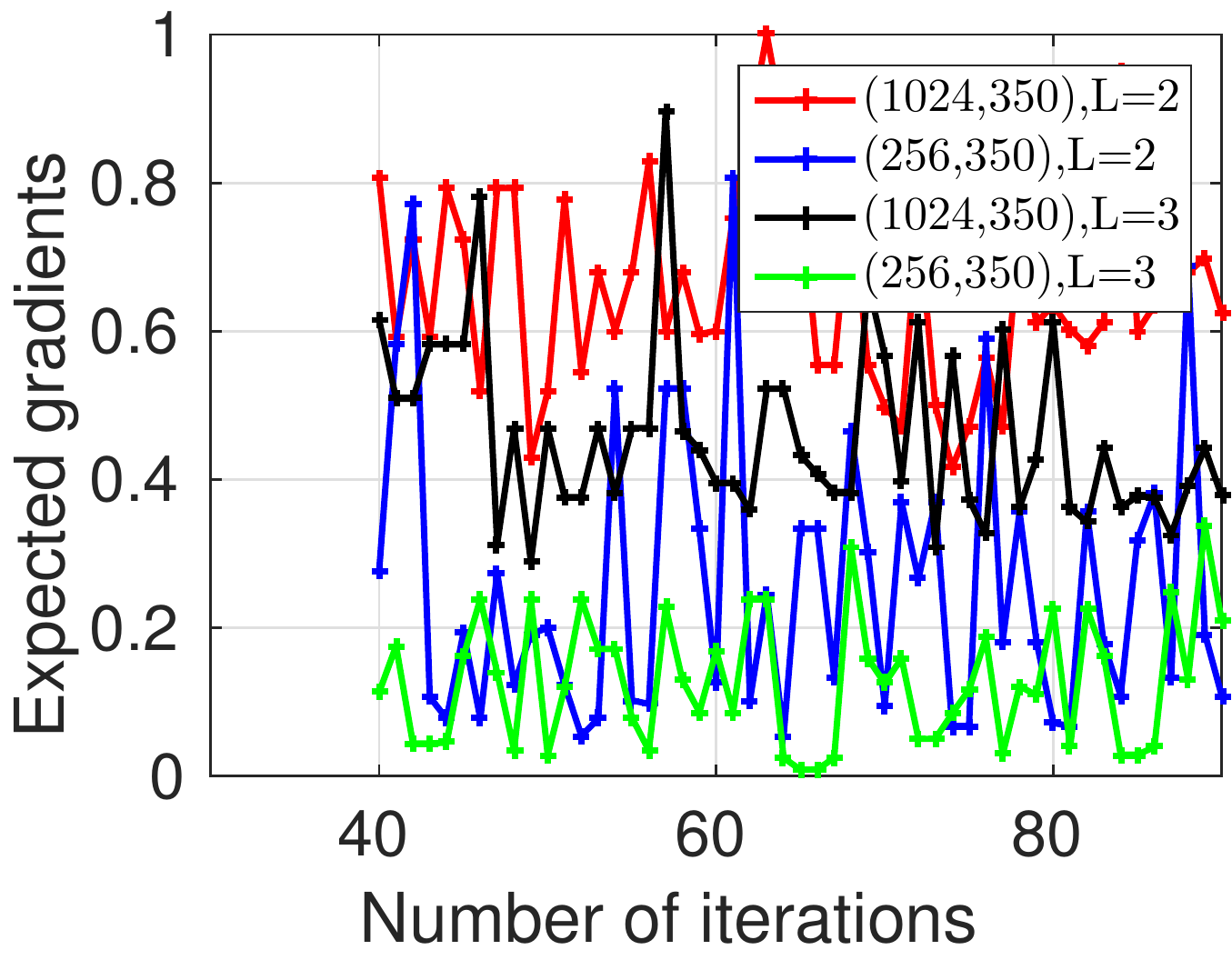}} 
\caption{\footnotesize \label{fig:multi_stop} {\bf Choice of stopping distribution:} Expected gradients computed using three different stopping iterations 
(corresponding to three columns) for $2$ and $3$ hidden layer networks. $B$ is set to $100$. 
The plots show that the whenever $\mathbb{P}_R$ is chosen reasonably (first and second columns; third column is delta stopping distribution), 
the decay trends are consistent, as predicted by the analysis.
Such plots can be used as surrogates for choosing when to stop the gradient updates -- an alternative for measuring performance on some validation set.} 
\end{figure}

\begin{figure}[!p]\centering
\subfloat[]{\includegraphics[width=47mm]{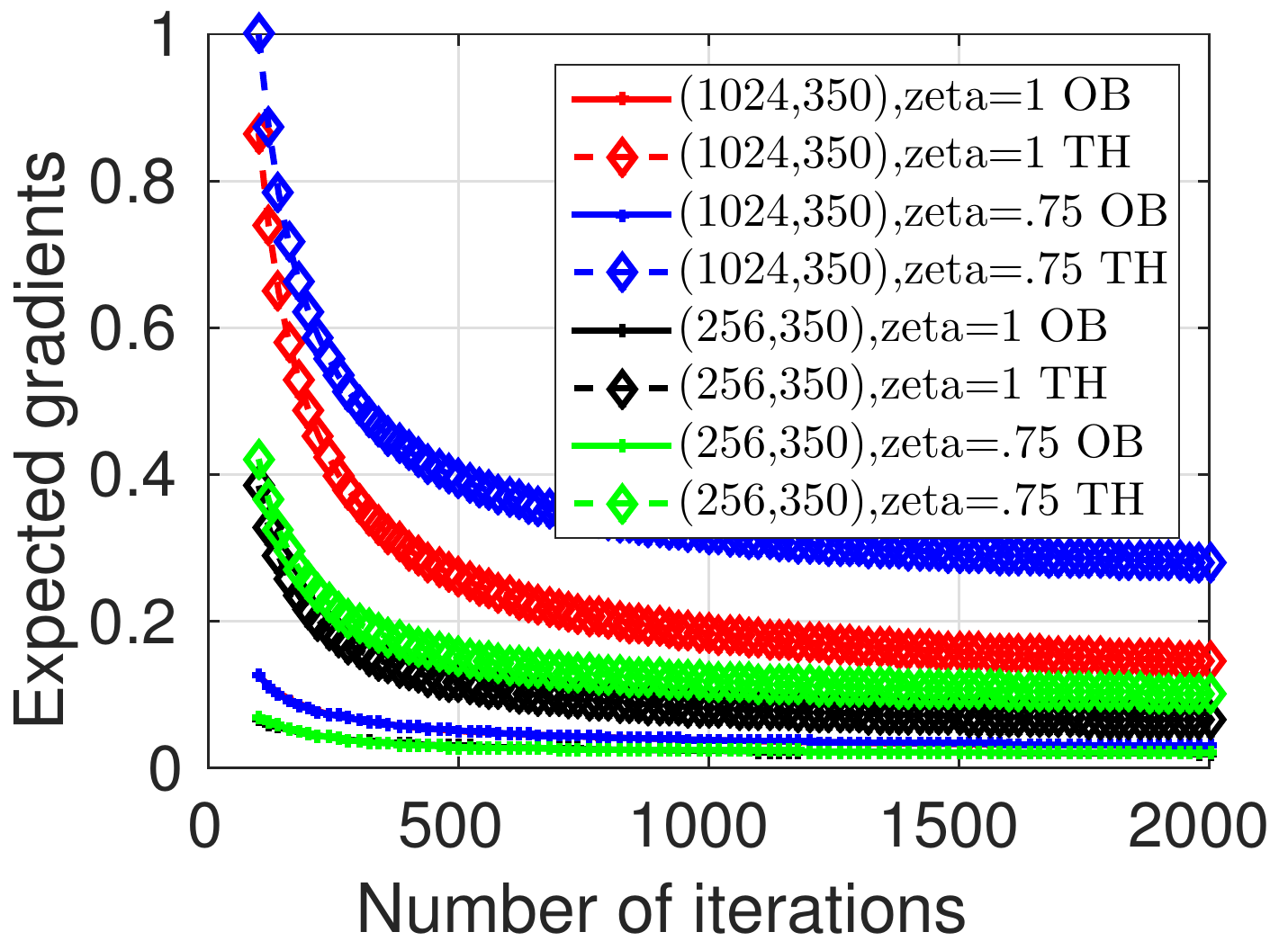}} \quad 
\subfloat[]{\includegraphics[width=47mm]{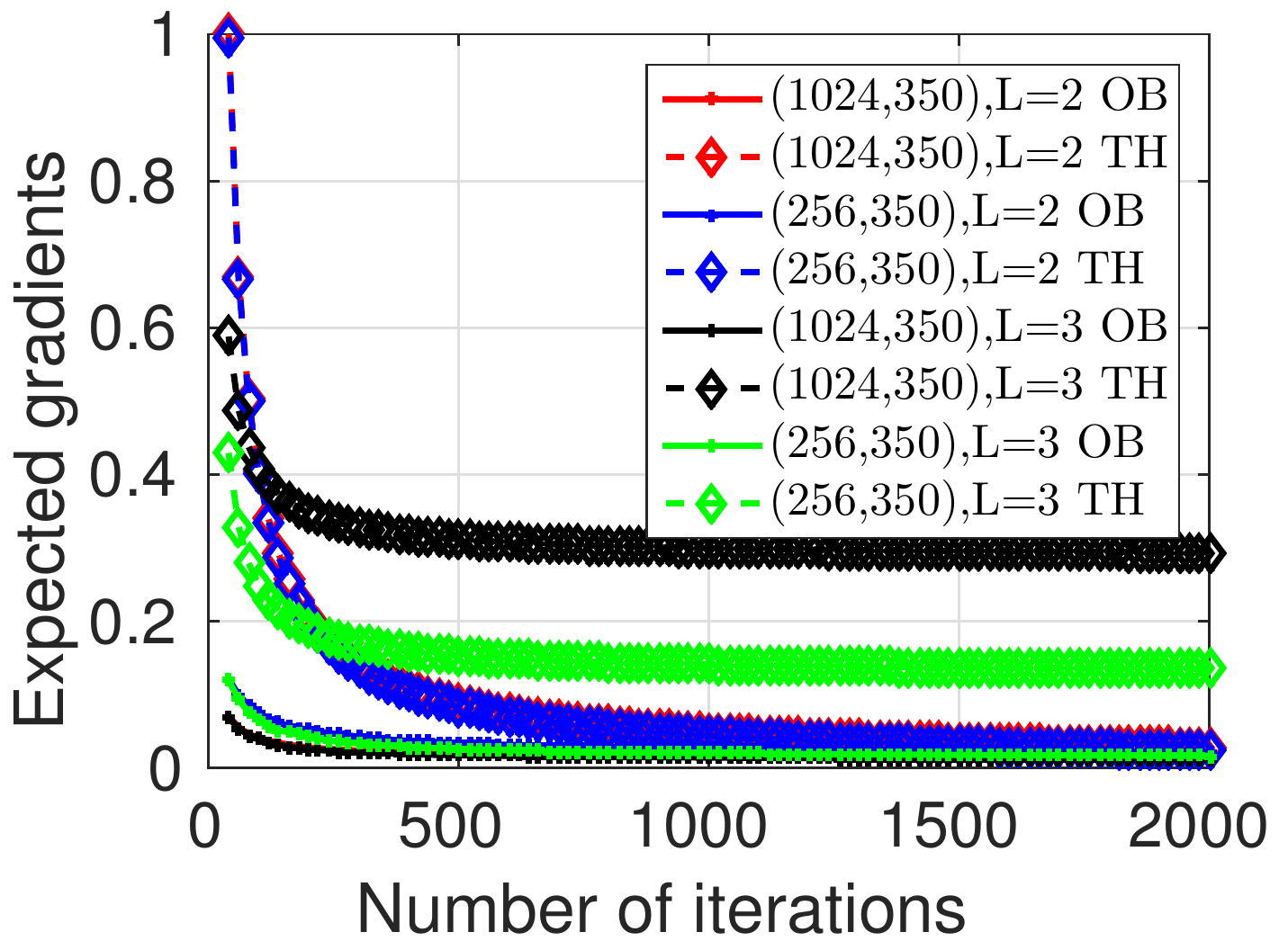}} \\
\subfloat[]{\includegraphics[width=47mm]{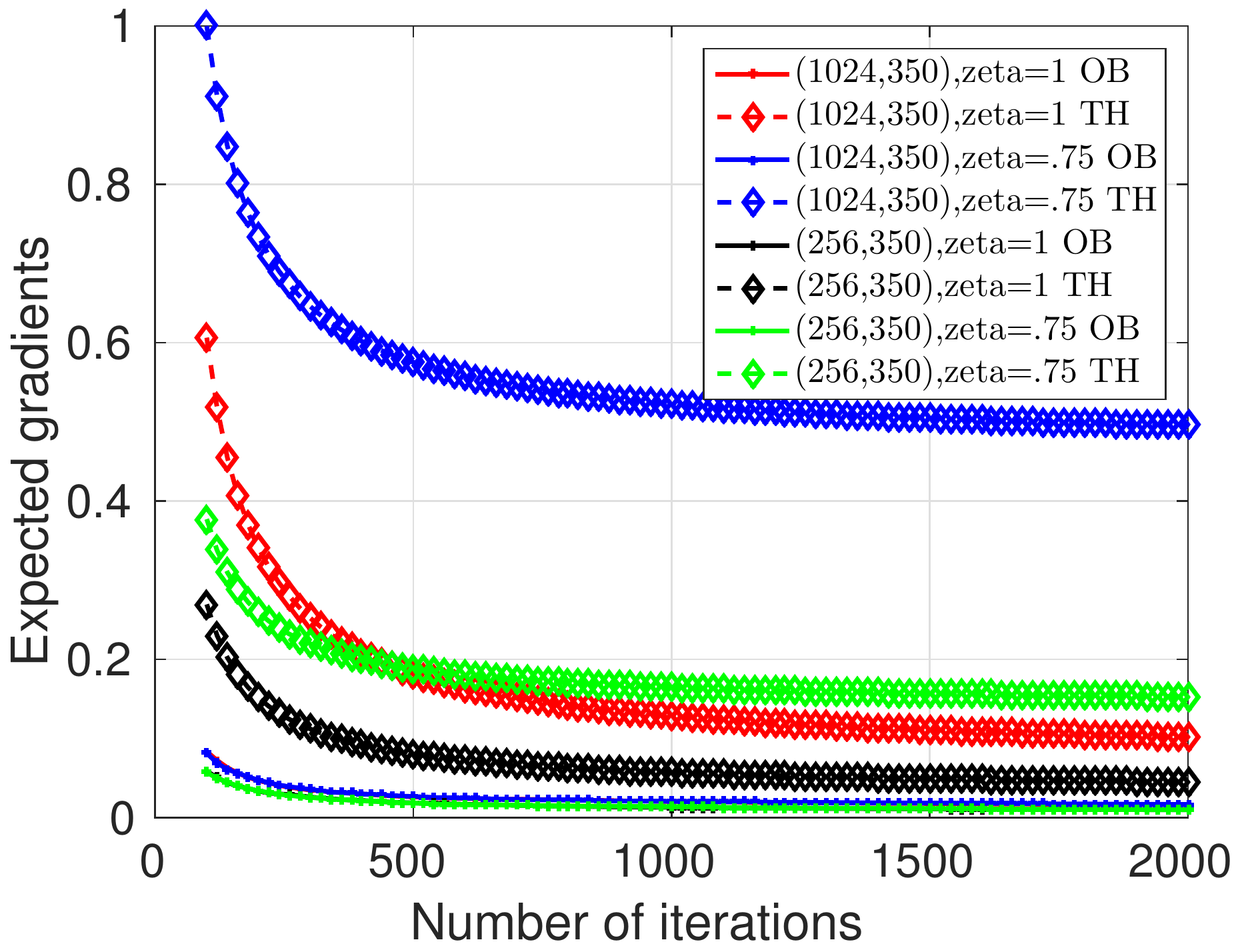}} \quad 
\subfloat[]{\includegraphics[width=47mm]{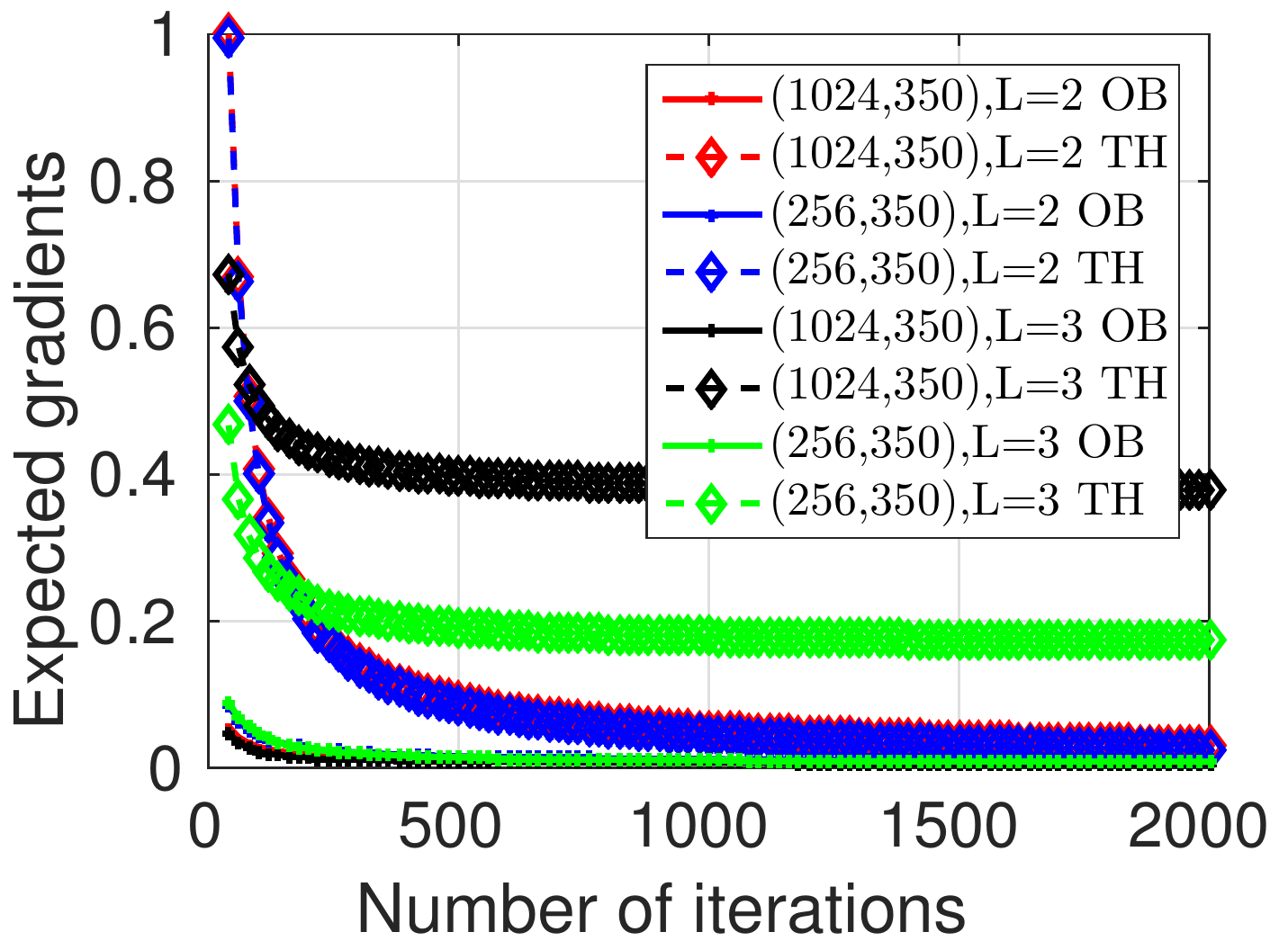}} \\
\subfloat[]{\includegraphics[width=47mm]{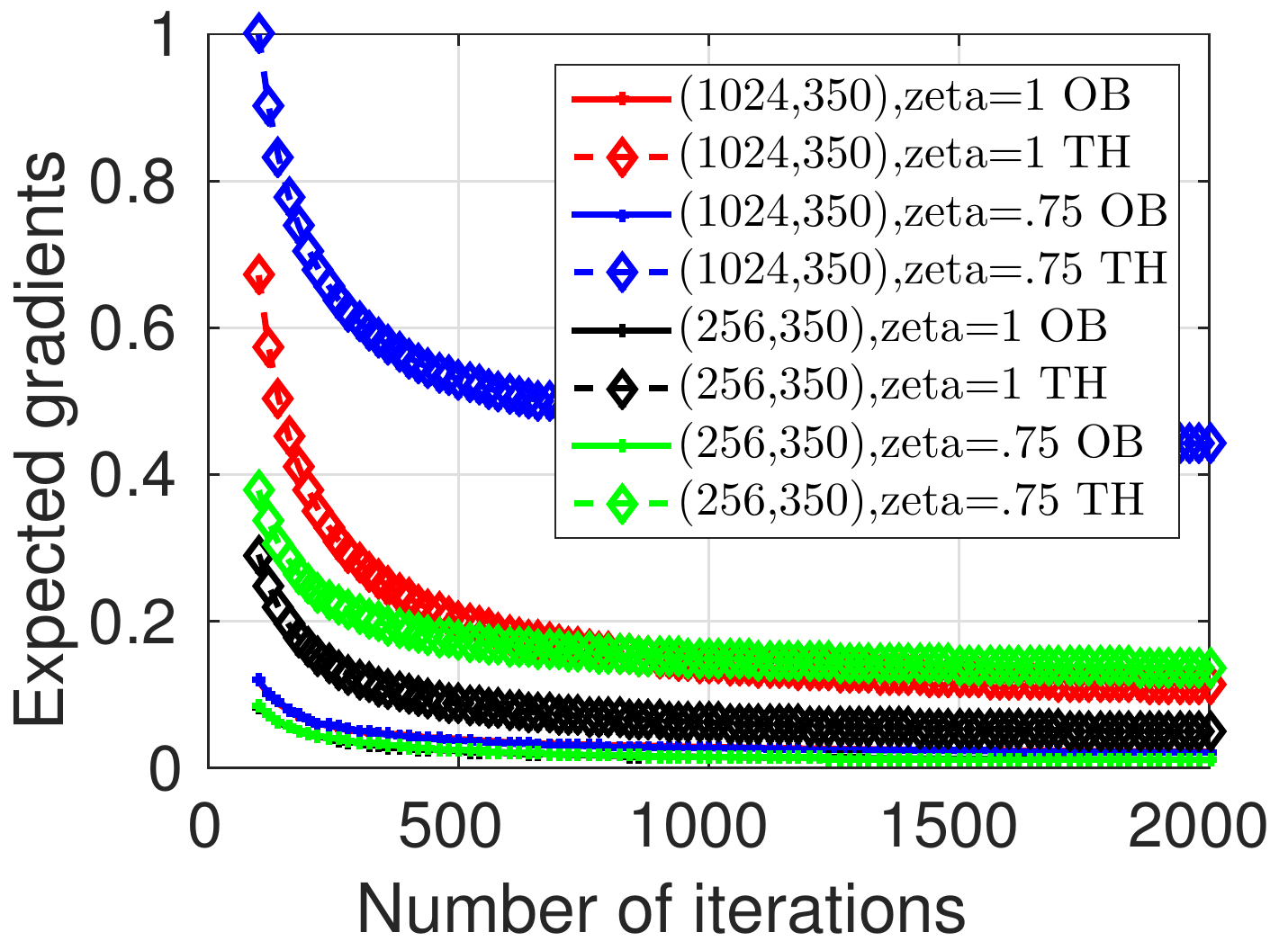}} \quad 
\subfloat[]{\includegraphics[width=47mm]{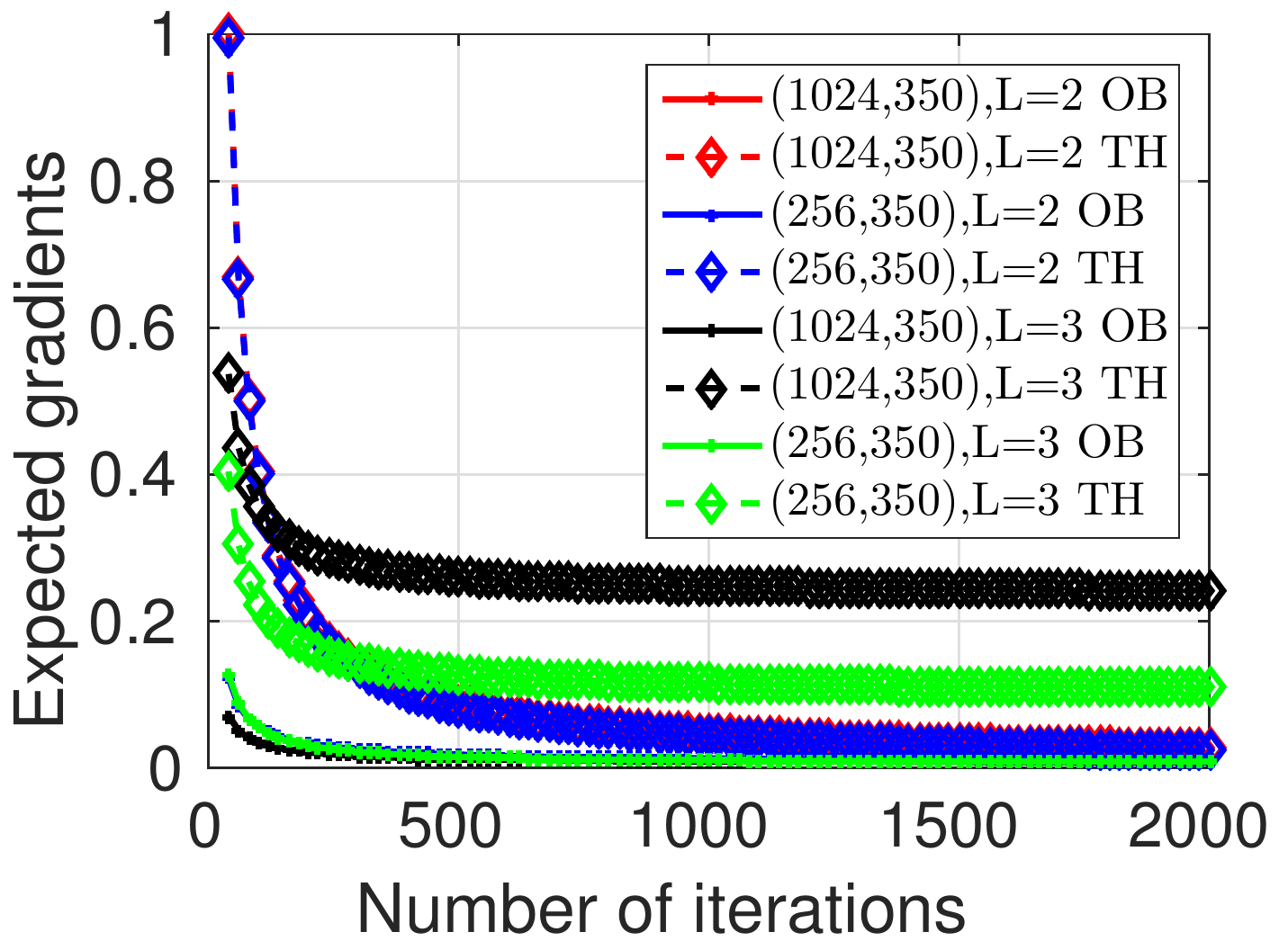}} \\
\subfloat[]{\includegraphics[width=47mm]{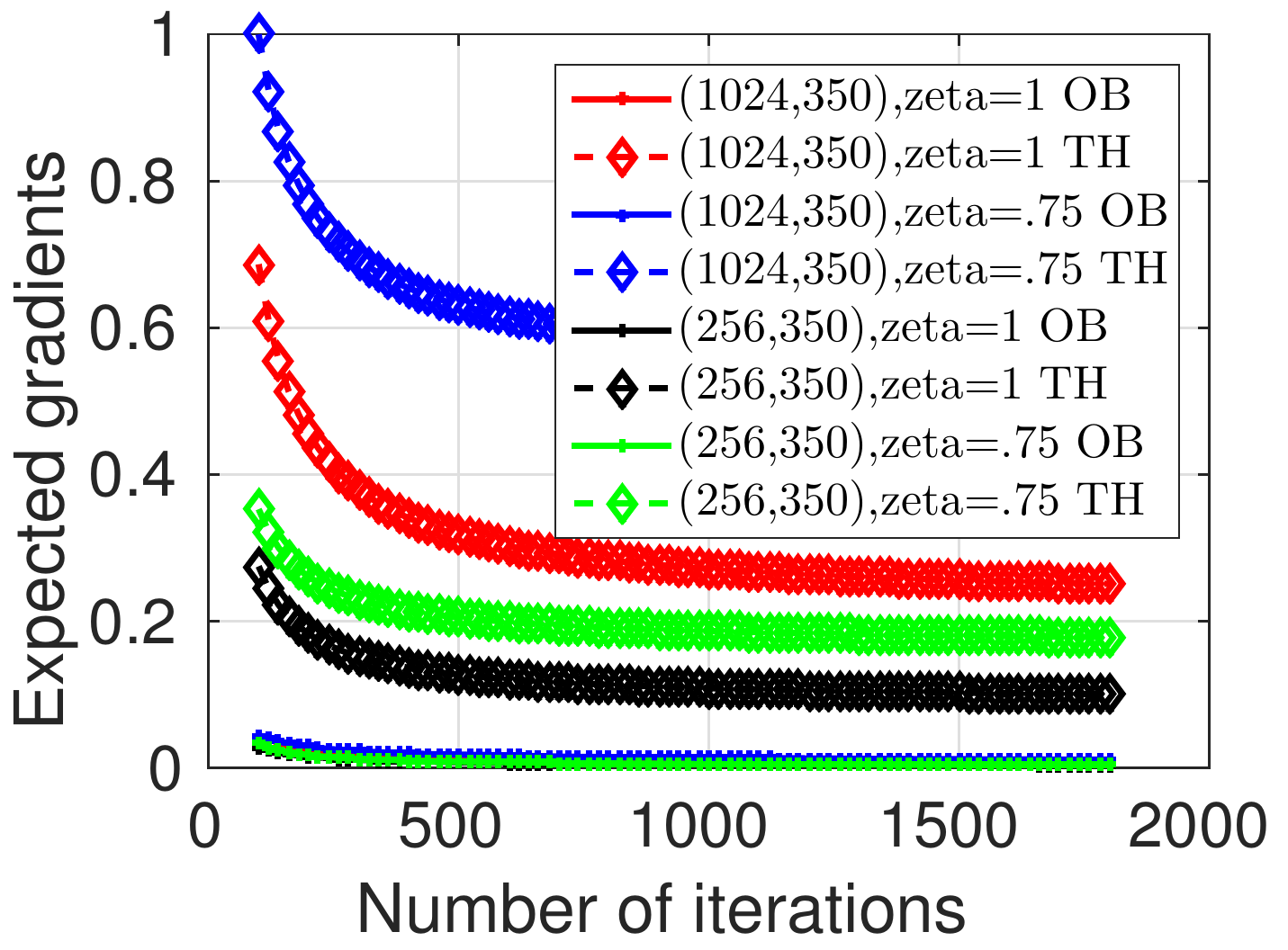}} \quad 
\subfloat[]{\includegraphics[width=47mm]{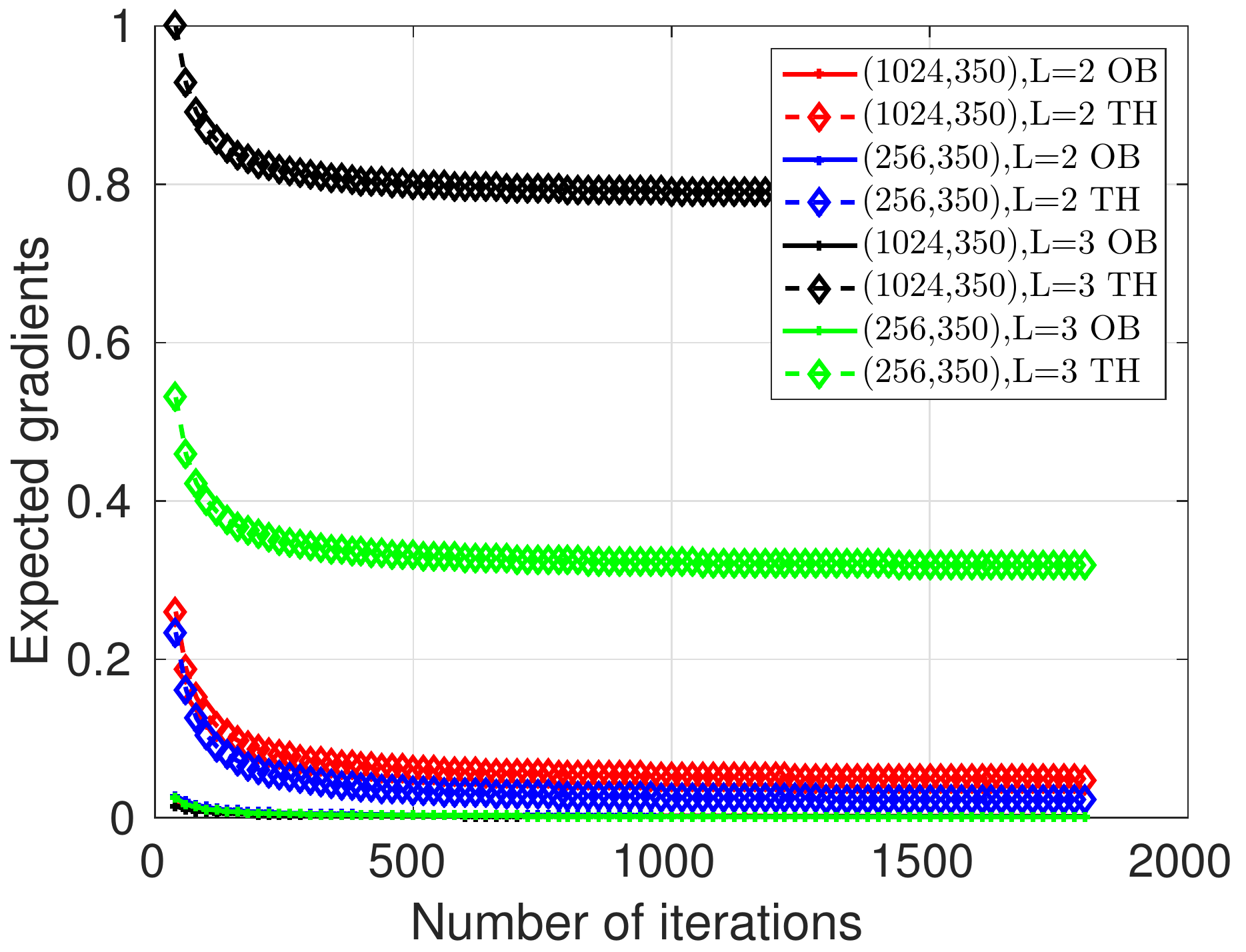}} 
\caption{\footnotesize \label{fig:the_vs_obs} {\bf Predicted vs. observed trends:} Predicted decay of expected gradients (dotted lines in the plots) vs. 
the observed decay (think lines) for the four datasets (one per each column). $B$ is set to $100$.
The plots clearly show the validity (and correctness) of our analysis.
The theoretical and empirical expected gradients decay trends are similar across several choices of hyperparameters and network architecture.}
\end{figure}

\paragraph{\bf Data:} \label{sec:data}
We empirically evaluated the bounds suggested by \eqref{eq:convda}, \eqref{eq:convmulnn} and \eqref{eq:mlnnpretdrop} using the following four different datasets. 
\begin{itemize} [leftmargin=1em]
\item {\bf\it Mnist}: The handwritten digits ($0-9$) dataset with $70000$ binary (black and white) images of size $28 \times 28$ \citep{lecun1998gradient}.
This dataset is well studied and was also used in the early papers on deep networks.
\item {\bf\it Cifar}: An object recognition dataset with $60000$ color images, each of size $32 \times 32$ \citep{krizhevsky2009learning}. 
  We use CIFAR-10 which has $10$ different classes. 
\item {\bf\it Caltech}: The CALTECH $101$ dataset includes images of objects belonging to $101$ categories, 
and approximately $40$ to $800$ images per category \citep{fei2007learning}.
A total of $5000$ images across the $101$ classes are used, which were rescaled to $32 \times 32$ pixels. 
\item {\bf\it neuro}: This dataset comprises of $3$D Magnetic Resonance Images (MRI) of $412$ subjects/patients, 
which were acquired as a part of Alzheimer's disease Neuroimaging Initiative (ADNI) for Alzheimer's disease (AD) research \citep{jack2008alzheimer}. 
The images belong to two classes -- healthy and diseased. 
Standard brain imaging pre-processing has been applied to these images (using Statistical Parametric Mapping toolbox, see \cite{ashburner2008spm8}).
Data corresponding to $10000$ most discriminating voxels ($3$D-pixels) 
were used in the experiments which were selected based on a $t$-test between healthy and non-healthy classes.
\end{itemize}
The datasets were rescaled appropriately for training networks with different input layer lengths ($d_0$). All features were normalized to $[0,1]$. 
Figures \ref{fig:allexp_mnist}, \ref{fig:allexp_cifar}, \ref{fig:allexp_caltech} and \ref{fig:allexp_neuro} 
summarize the behavior of expected gradients (or projected gradients) for the four datasets considered here, in order, using different architectures and learning parameters. 
Figure \ref{fig:multi_stop} shows the influence of the stopping distribution, and Figure \ref{fig:the_vs_obs} shows the gap in the predicted versus observed trends. 
The $y$-axis (and colorbar, whichever appropriate) in the plots has been scaled by the corresponding maximum values to allow interpretation of the trends rather than the actual values. 
The batchsize $B = \min\{\ceil{0.1N},100\}$ for all the experiments, and the expected gradients is simply the empirical average of gradients computed over the last $\min\{N,50\}$ iterations 
(recall that $N$ is the maximum number of allowable iterations).
For the four datasets, the data statistics $(\mu_\x,\tau_\x)$ are given by $(0.13,0.11)$ (mnist), $(0.43,0.30)$ (cifar), $(0.52,0.35)$ (caltech) and $(0.14,0.09)$ (neuro).




\paragraph{\bf Convergence versus $d_x$, $d_h$, $w_m$, $\zeta$:} \label{sec:convexps}
Figures \ref{fig:allexp_mnist}--\ref{fig:allexp_neuro}(a) show the expected projected gradients of DA pretraining versus network sizes (hidden and input layer lengths). 
As predicted in \eqref{eq:conv1nn} and \eqref{eq:convda}, the convergence is slower for large networks, 
and visible and hidden layer lengths have unequal influence on the convergence (see remarks of Theorem \ref{thm:expgradda}).
Here, $w_m=1/\sqrt{d_vd_h}$ for all networks. 
The influence of denoising rate and $w_m$ is shown in Figures \ref{fig:allexp_mnist}--\ref{fig:allexp_neuro}(b) ($d_h = d_v$ for these networks).
As suggested by \eqref{eq:convda}, the convergence is faster for small $w_m$s.
It is interesting that across all cases, the choice of $\zeta$s has almost a negligible influence. 
Figures \ref{fig:allexp_mnist}--\ref{fig:allexp_neuro}(c), show the interaction of network sizes vs $\zeta$, and as observed in Figures \ref{fig:allexp_mnist}--\ref{fig:allexp_neuro}(a), 
the networks lengths dominate the convergence with the visible layer length being the most influential factor. 
The data moments of caltech and cifar are larger than mnist and neuro, and as suggested by \eqref{eq:convdaparam-eda} and \eqref{eq:convda}, 
this results in a stronger influence of $\zeta$, $w_m$ and the network sizes on the decay of gradients -- which can be seen from Figures \ref{fig:allexp_mnist}--\ref{fig:allexp_neuro}(b,c).
Figures \ref{fig:allexp_mnist}--\ref{fig:allexp_neuro}(d) shows the influence of the network depth with $w_m=1/\sqrt{d_vd_h}$ and $\zeta=0.5$.
Clearly, the expected gradients are influenced more by the layer lengths than the number of layers itself, as predicted by \eqref{eq:convmulnn}.


\paragraph{\bf Does dropout compensate pretraining?} \label{sec:droppretexps}
Figures \ref{fig:allexp_mnist}--\ref{fig:allexp_neuro}(e) show the effect of changing the dropout rate in a $3$-layered network. 
The influence of $\mu_\x$, $\tau_\x$ and $w_m$ can be seen clearly where the trends are much stronger for caltech and cifar (Figures \ref{fig:allexp_cifar}, \ref{fig:allexp_caltech}(e)) than the rest.
Although the overall effect is small, the convergence is slower for very small (red curve) and very large (cyan, magenta curves) $\zeta$s (see \eqref{eq:mlnnpretdrop}), 
and the trends are noisy for the neuro dataset (Figure \ref{fig:allexp_neuro}(e)).
This is not surprising because, unlike the other computer vision datasets, 
neuro data includes co-registered features with very small data moments and small changes in feature values correspond to class difference.
Figures \ref{fig:allexp_mnist}--\ref{fig:allexp_neuro}(f) evaluates Theorem \ref{thm:pretvsdrop}, by comparing pretraining (rows) to dropout (columns).
Across all datasets, expected gradients clearly decrease as pretraining iterations increase, while for small and large dropouts, the convergence is slower.
The best regimes, as predicted by \eqref{eq:mlnnpretdrop}, are large pretraining with small dropout 
(bottom-right region on Figures \ref{fig:allexp_mnist}--\ref{fig:allexp_neuro}(f)) and $\zeta \approx 0.5$ with less pretraining (center of the image). 


\paragraph{\bf Influence of $\mathbb{P}_R(\cdot)$:} \label{sec:stopexps}
The different choices of $\mathbb{P}_R(\cdot)$ used in Section \ref{sec:single} for Theorem \ref{thm:expgrad} 
and Corollaries \ref{thm:expgrad-noncon} and \ref{thm:expgrad-mono} are evaluated in Figure \ref{fig:multi_stop}.
The rows correspond to the four datasets while the columns correspond to using different $\mathbb{P}_R(\cdot)$. 
For the networks in the first column, $R$ is uniformly distributed over the last $15$ iterations. 
The second column used a Gamma distribution peaking at the last iteration and decreasing monotonically from the last to first iteration 
(an instance of the $\mathbb{P}_R(\cdot)$ used in \eqref{eq:conv1nn-mono}). 
The last column calculated expected gradients over a randomly chosen iteration from the last $15$ iterations.
Beyond the fact that the decay in the last column is noisier than the others, these types of plots are useful in choosing the best $\mathbb{P}_R(\cdot)$ from a dictionary of distributions.


\paragraph{\bf Predicted versus observed trends:} \label{sec:pred_vs_obs}
The discrepancy between the trends predicted by our results relative to the empirical observations are shown in Figure \ref{fig:the_vs_obs}.
Each row is a dataset. The first column shows pretraining and the second one corresponds to pretraining + supervised tuning. 
Clearly, the predicted trends (dotted lines) follow the observed ones (thick lines). 
Overall, across all the datasets and architectures, the predicted bounds from the results in the paper are larger than the observed ones, but the trends nevertheless are consistent. 
There is a bias in the predicted bounds, coming from the terms (that do not contain $N$) involving the network degrees of freedom (i.e., network lengths). 
The estimate of their contribution is larger, suggesting that there is some redundancy.


\paragraph{\bf Gradient norms and Local optima:} \label{sec:gradnormexps}
Recall that it is NP Hard to check local optimality even for simple nonconvex problems \citep{murty1987some}. 
As this may be important for a practitioner, we address it empirically. 
Specifically, after performing $N$ iterations (suggested by our theoretical results), we randomly sampled points around our final estimate $\W^R$ (i.e., perturbed estimate). 
We observed that the fraction of such points that were better (by $0.02\%$ of the objective value) than our estimate $\W^R$ rapidly decreases as $N$ increases. 
These fractions represent the probability that $\W^R$ empirically satisfies first order local optimality conditions. 
Overall, these evaluations, when considered together with observations from \cite{bengio2009learning, erhan2010does, baldi2014dropout} 
provide empirical evidence that the results in Section \ref{sec:single}--\ref{sec:multi} are sensible.


\subsection{Visualizing trends}
\label{sec:vis}

\begin{figure}[!h]\centering
\subfloat[]{\includegraphics[width=47mm]{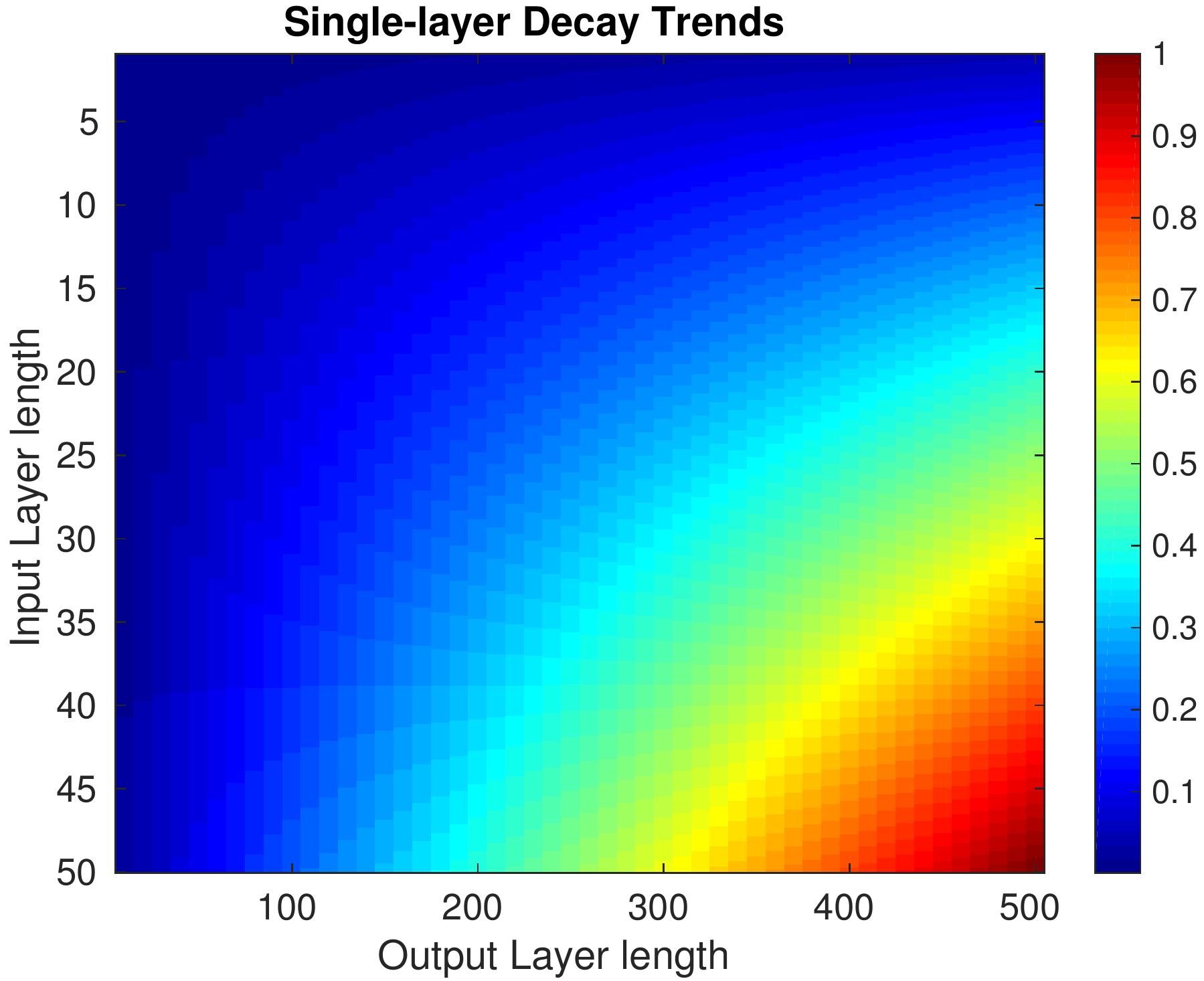}} \quad 
\subfloat[]{\includegraphics[width=47mm]{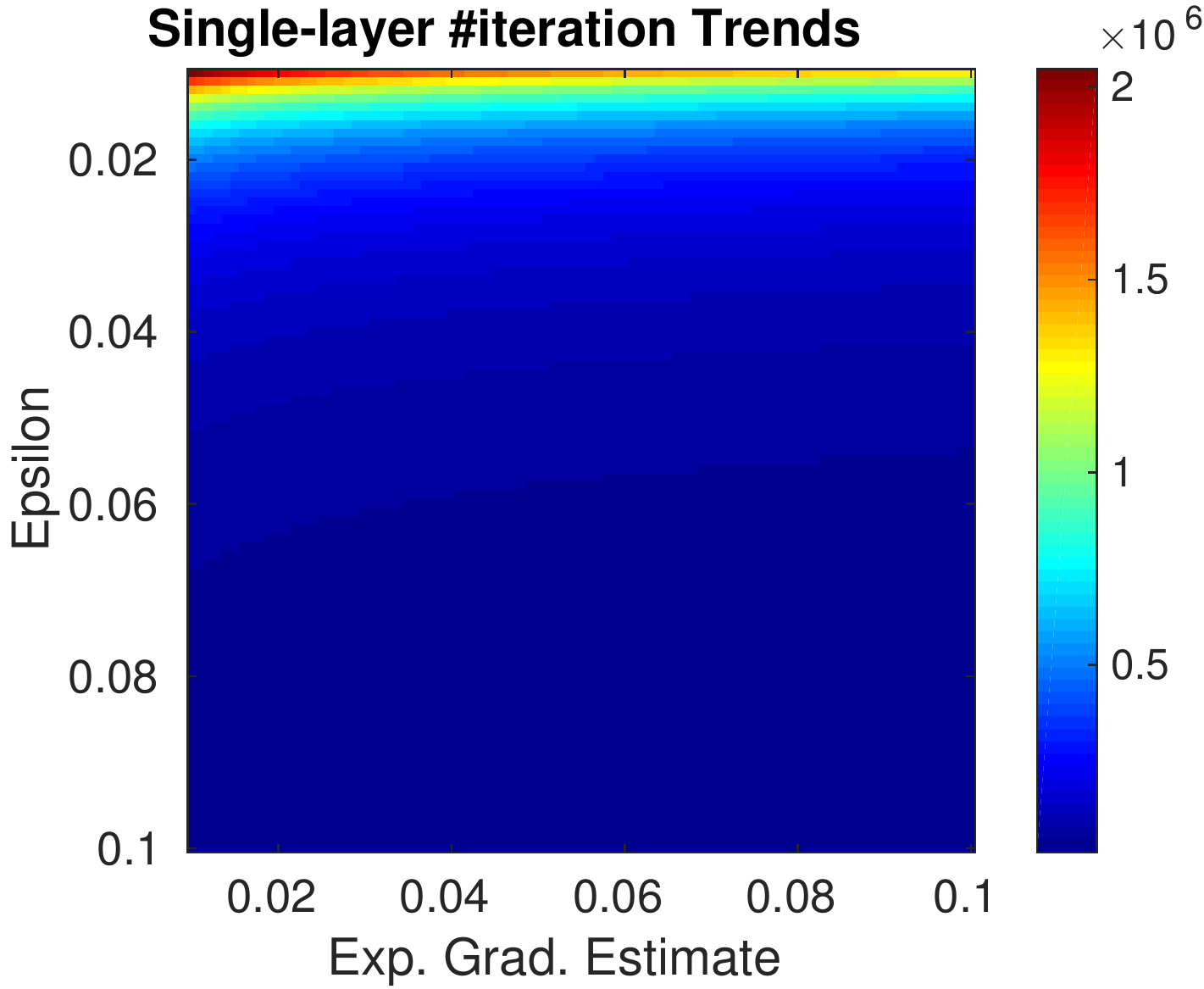}} \\ 
\subfloat[]{\includegraphics[width=47mm]{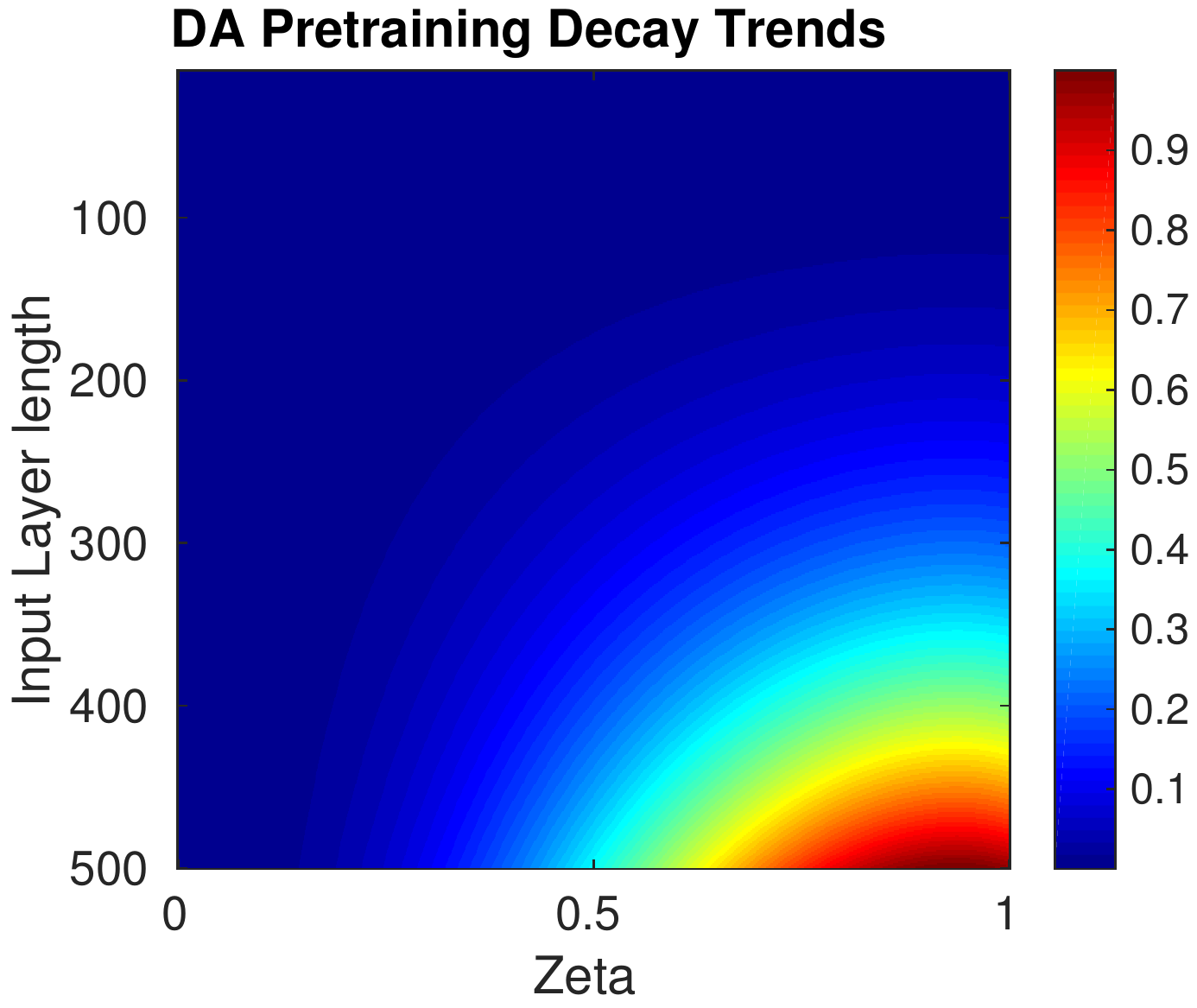}} \quad 
\subfloat[]{\includegraphics[width=47mm]{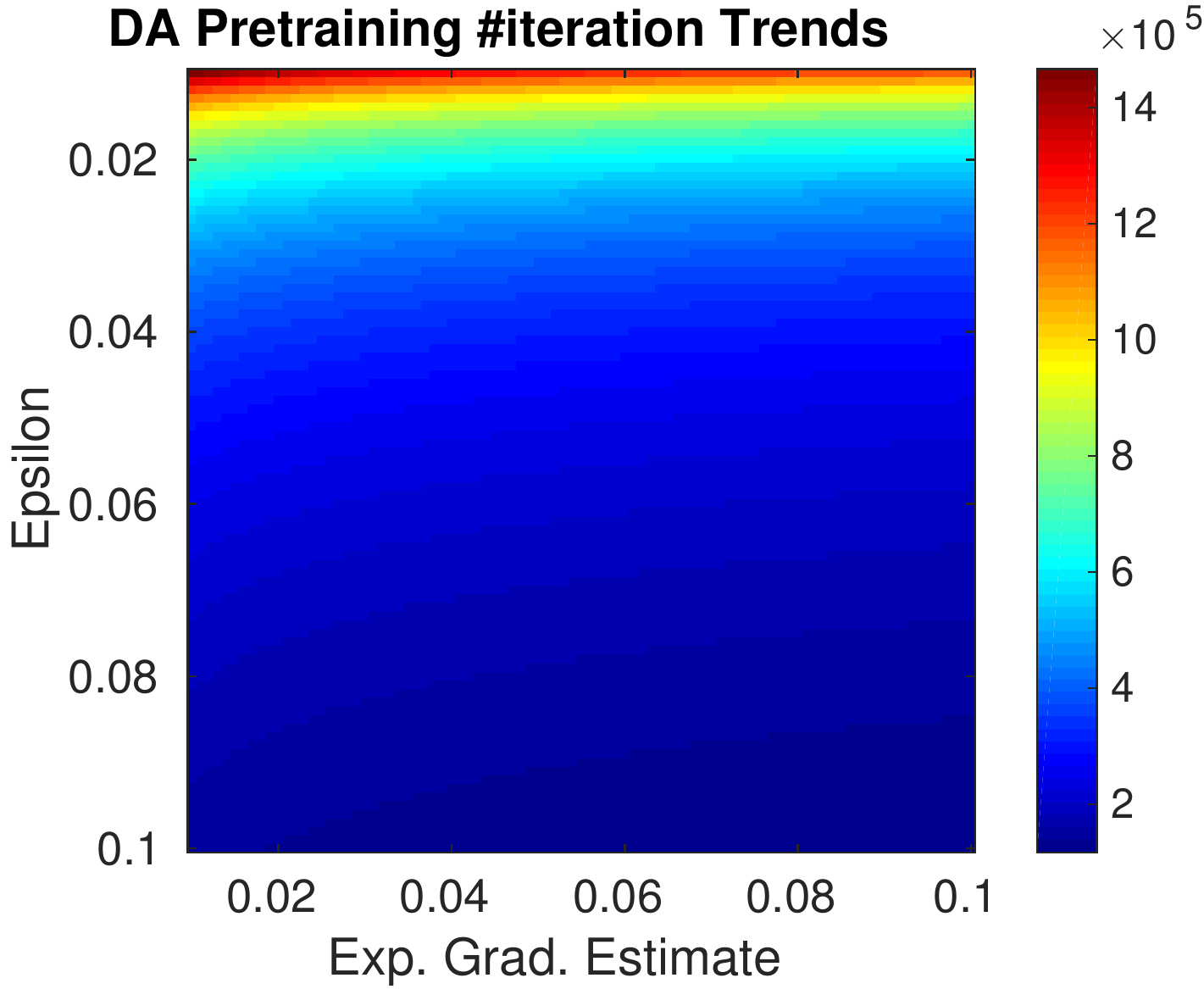}} 
\caption{\footnotesize \label{fig:visual_single} {\bf Decay trends of Single-layer Networks:} Theoretical trends of $1$-NN (top row) and DA (bottom row). 
Columns show decay of gradients and number of iterations.
(a,c) show that expected trend that network size influences the gradient decay.
(b,d) show that $\epsilon$ plays a far more influential role than $\delta$, implying that small changes in $\epsilon$ require adjusting $N$ appropriately to maintain the same level of convergence.
}
\end{figure}

\begin{figure}[!h]\centering
\subfloat[]{\includegraphics[width=47mm]{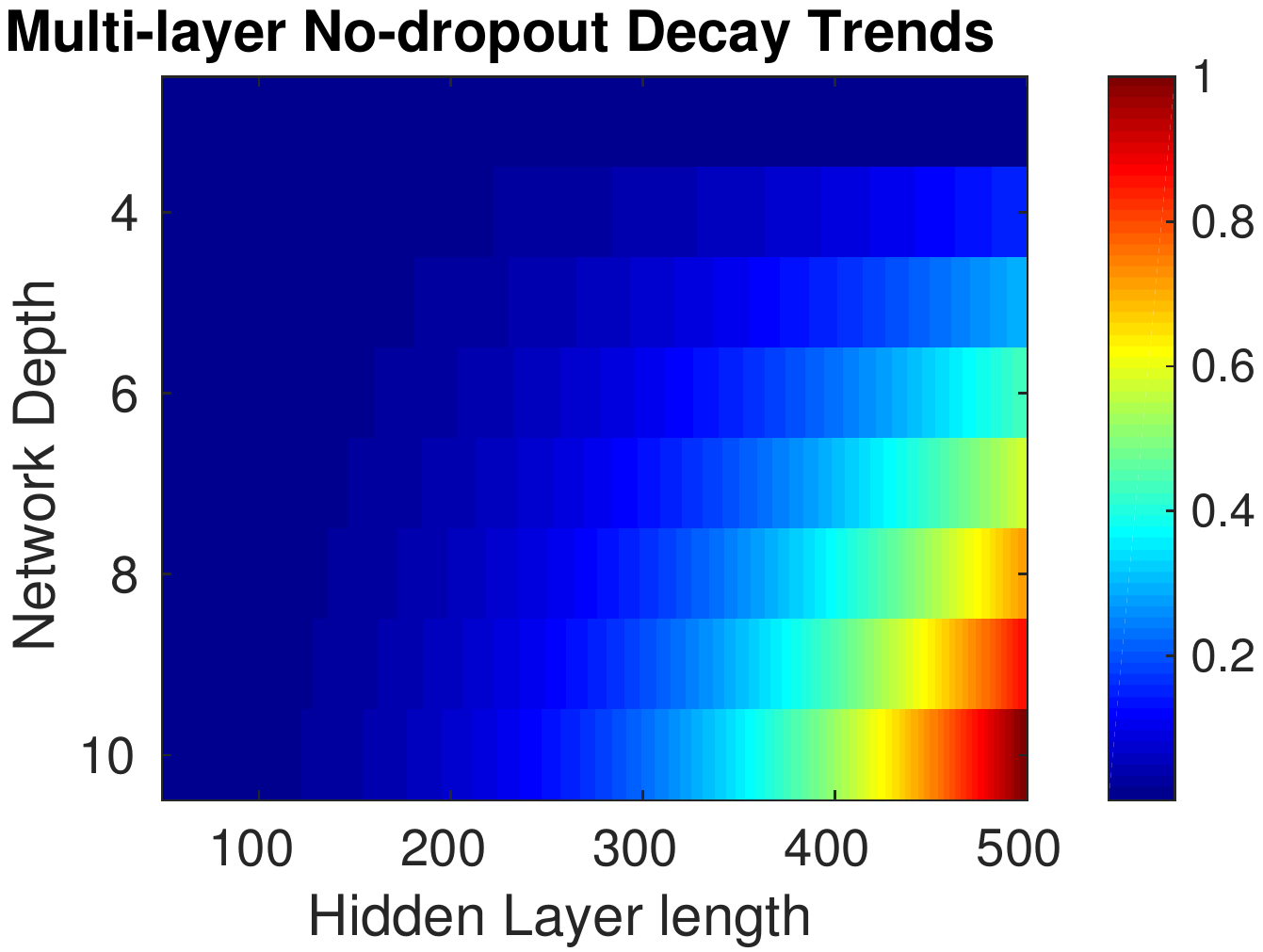}} \quad 
\subfloat[]{\includegraphics[width=47mm]{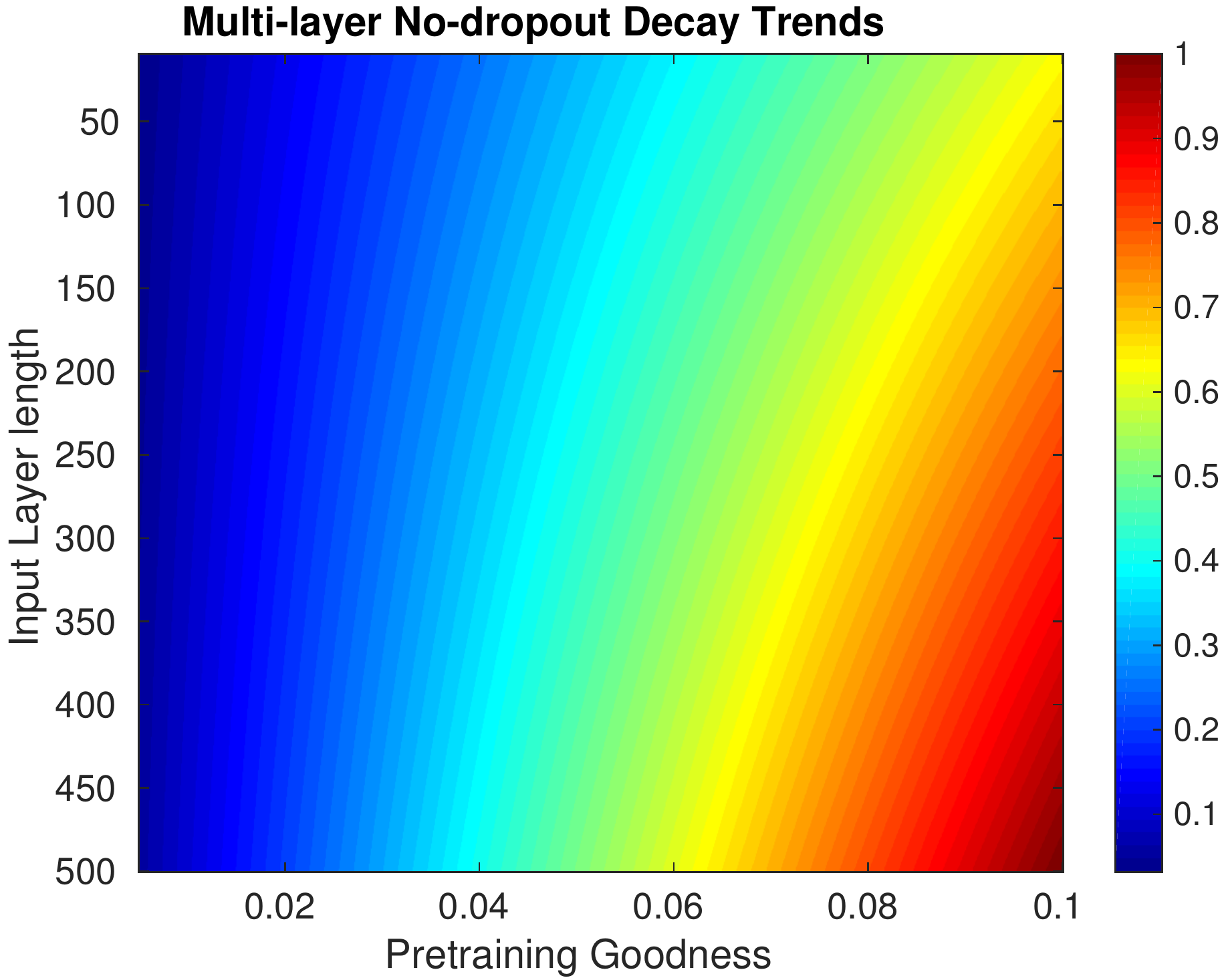}} \quad 
\subfloat[]{\includegraphics[width=47mm]{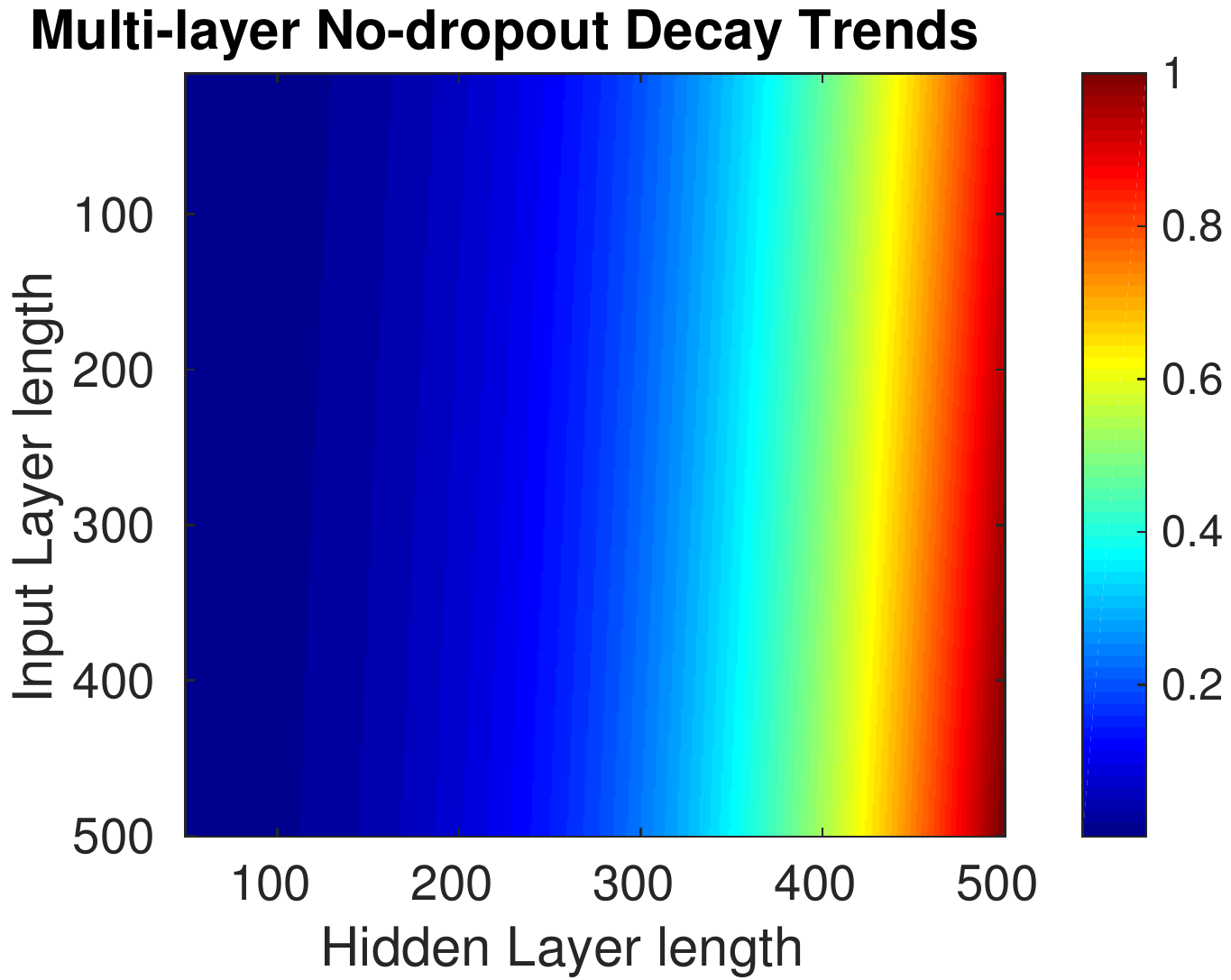}} \\ 
\subfloat[]{\includegraphics[width=47mm]{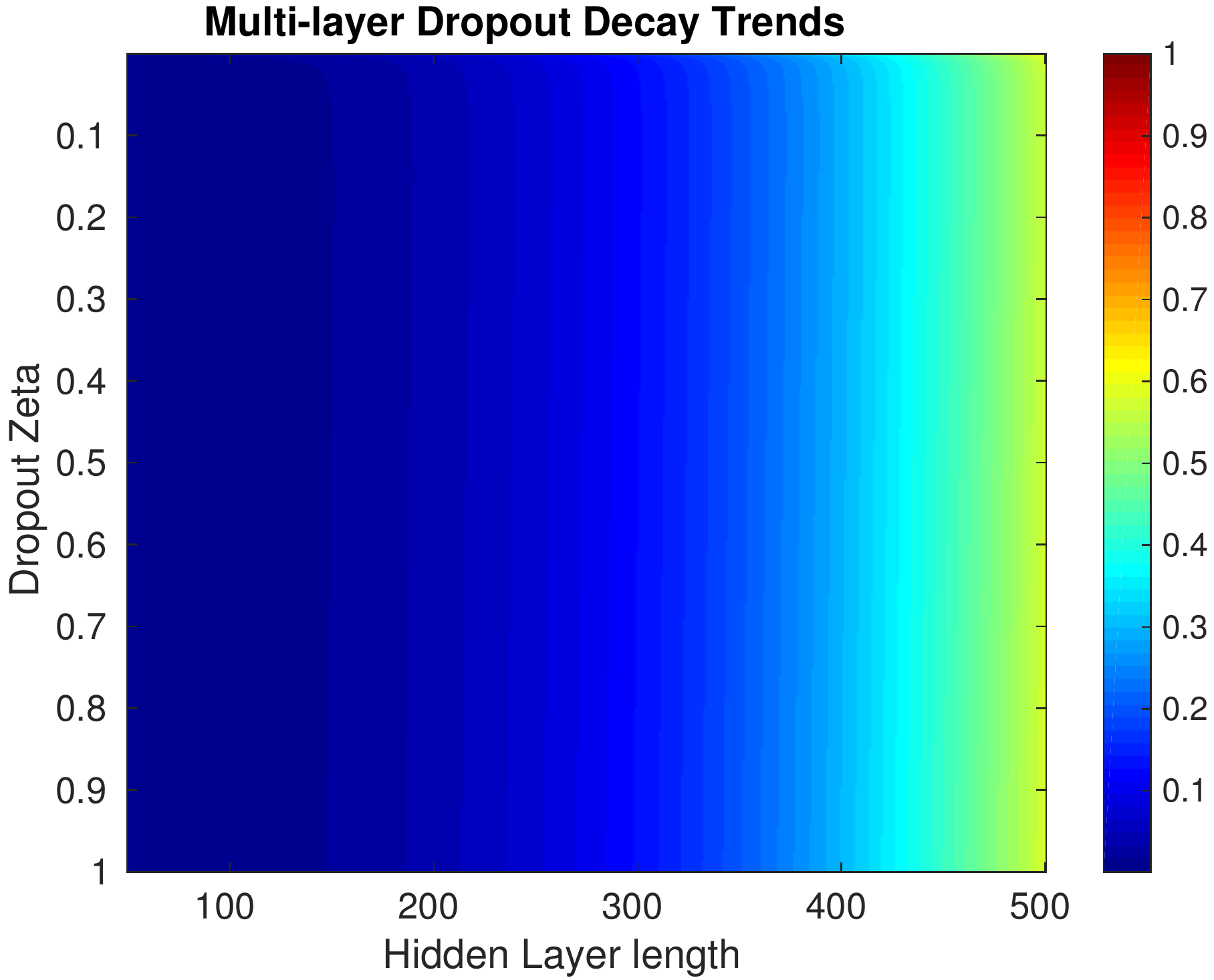}} \quad 
\subfloat[]{\includegraphics[width=47mm]{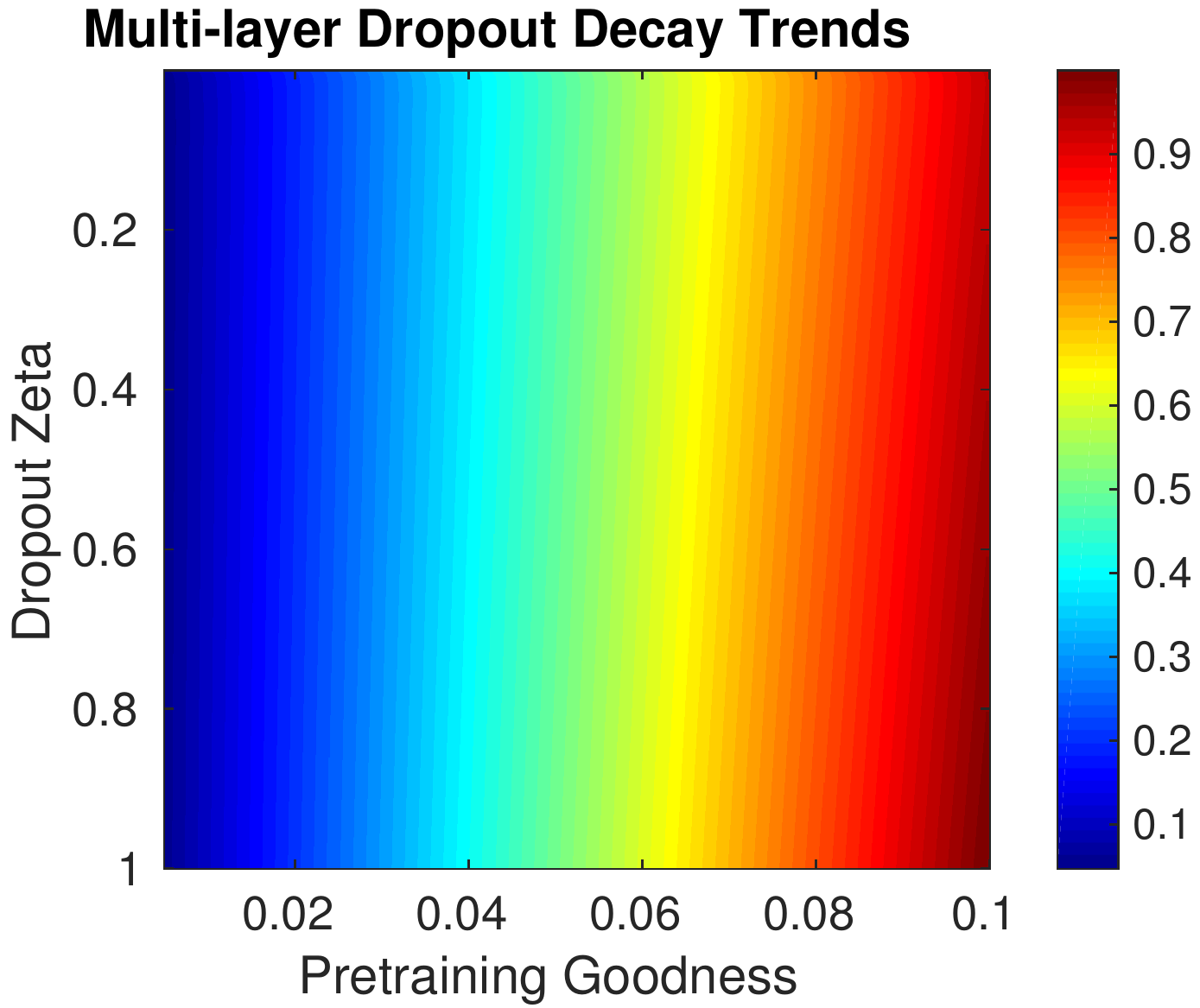}} \quad 
\subfloat[]{\includegraphics[width=47mm]{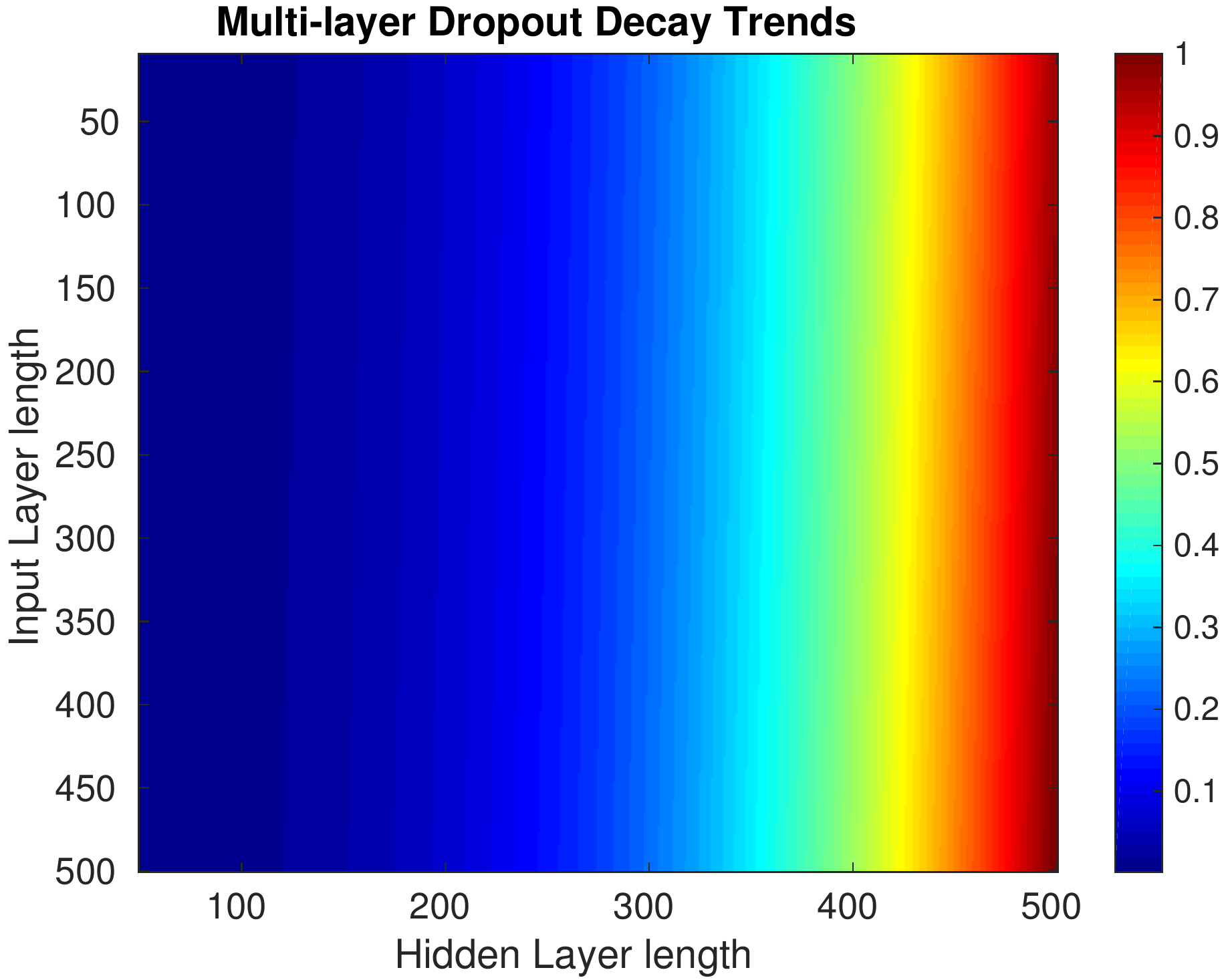}} 
\caption{\footnotesize \label{fig:visual_multi} {\bf Decay trends of Multi-layer Networks:} Theoretical trends of $L$-NNs without (top row) and with (bottom row) dropout. 
  (a) shows influence of network depth. (b) visualizes the necessity of pretraining. (e) shows that dropout rate has smaller influence than pretraining goodness.
  (c,f) show that hidden layer lengths are more influential than input layer length.}
\end{figure}

\begin{figure}[!h]\centering
\subfloat[]{\includegraphics[width=47mm]{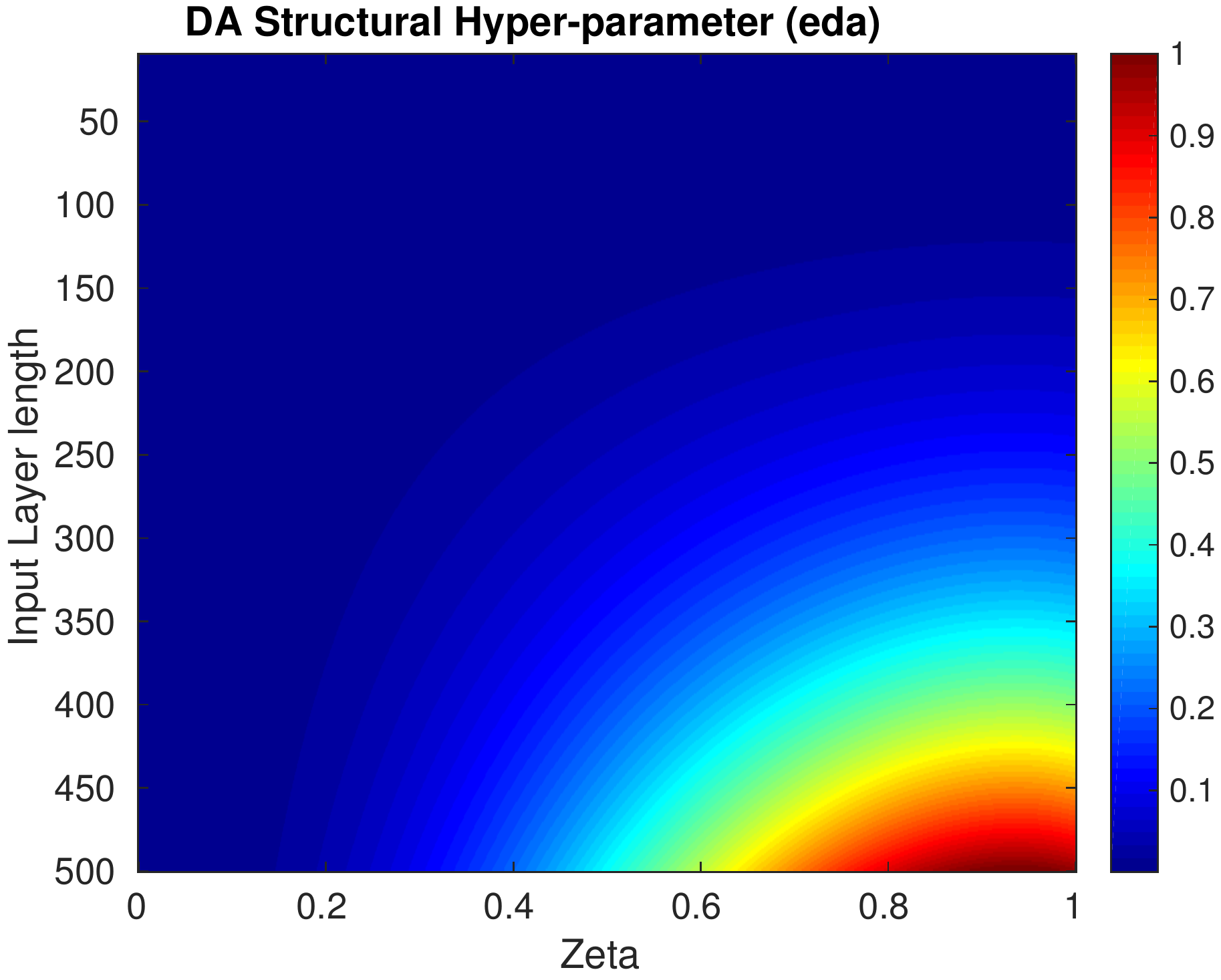}} \quad 
\subfloat[]{\includegraphics[width=47mm]{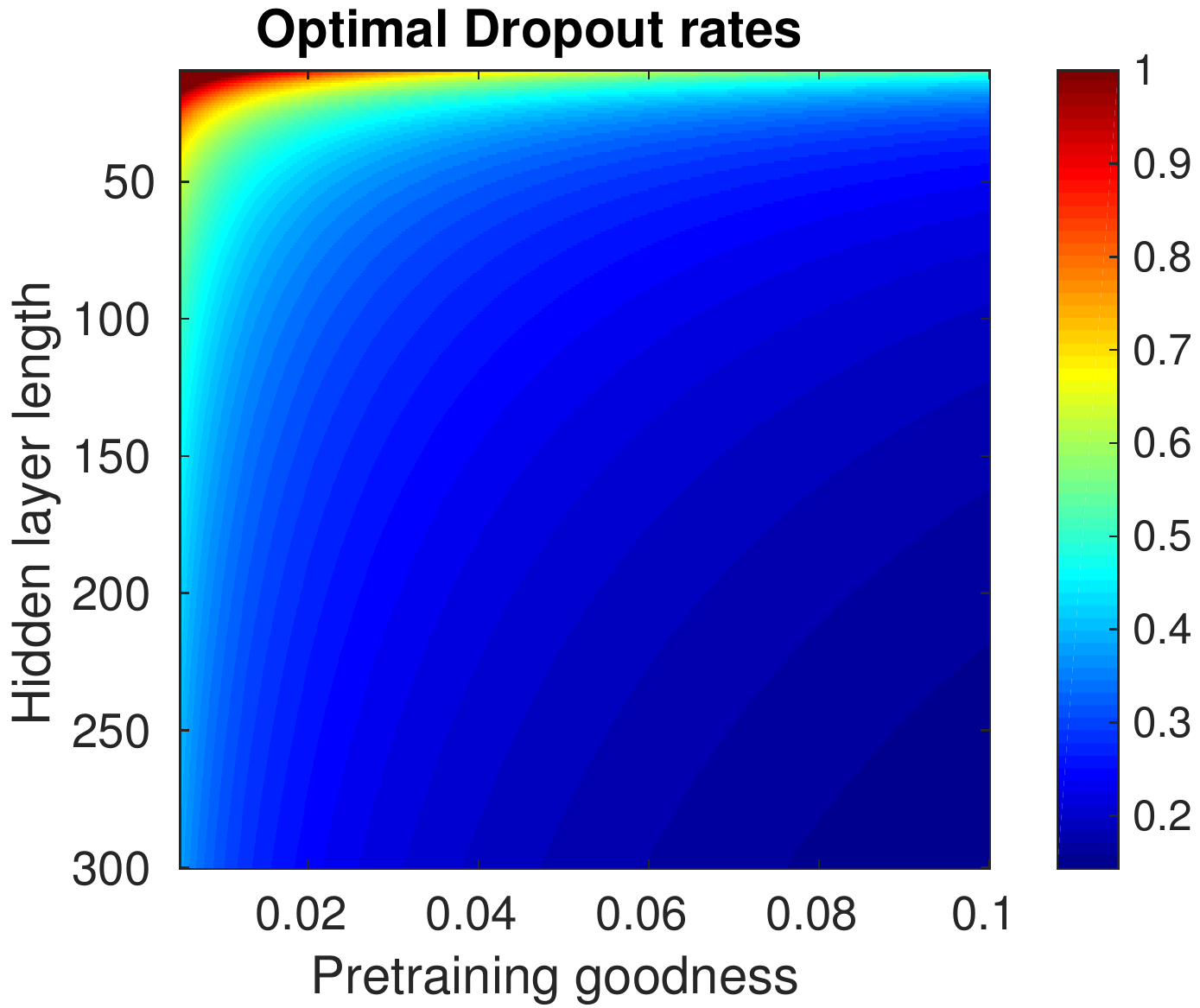}} 
\caption{\footnotesize \label{fig:visual_den_drop} {\bf Denoising and Dropout Trends:} (a) Influence of denoising rates by plotting $e^{da}$.
  $\tzeta$ smaller than $0.5$ si the good regime, however, the input data length has much stronger influence (b) Optimal dropout rates. Too large hidden layer lengths ask for stronger dropout rate.}
\end{figure}

The interplay of all the structural and learning parameters shown by the decay bounds in \eqref{eq:conv1nn}, 
\eqref{eq:convda} and \eqref{eq:convmulnn} are complicated to interpret towards designing new networks.
Although the experiments presented above provide empirical support for these results, the decay trends were measured mainly as a function of $N$.
To better understand the bounds, Figures \ref{fig:visual_single}--\ref{fig:visual_den_drop} visualize the trends predicted. 
Note that here we simulate the bounds derived earlier in the technical results from Sections \ref{sec:single}--\ref{sec:multidrop}, 
unlike the results presented above which were based on the networks learned via Alg. \ref{alg:single-nn}--\ref{alg:multi-nn}.
Except Figures \ref{fig:visual_single}(b,d) and \ref{fig:visual_den_drop}(b) which show the gradient iterations and optimal dropout rates respectively, 
all the other images are scaled to the corresponding maximum values since they are essentially computing the decay bound. 
The hyper-parameters $N$, $B$, $\gamma$ and $w$ are fixed for constructing Figures \ref{fig:visual_single}--\ref{fig:visual_den_drop}, 
and so, increasing $N$ may resolve some of the outcomes/issues that will be pointed out. 


\begin{itemize} [leftmargin=1em]
\item Figure \ref{fig:visual_single}(a) shows the trivial trend where the decay bound increases as the input and output layer lengths increase.
Small denoising rate i.e., large $\tzeta$s in \eqref{eq:convda} lead to larger bounds, which is shown in Figure \ref{fig:visual_single}(c).
Observe that for a reasonably small $d_x$, the influence of the denoising rate is almost negligible (top half of Figure \ref{fig:visual_single}(c)). 
The iteration estimates from Figures \ref{fig:visual_single}(b,d) are interesting. Recall the parameters $\epsilon$ and $\delta$ from Definition \ref{thm:epsdel}. 
With repeated runs of the random stopping stochastic gradients, $1-\delta$ denotes the probability that the maximum allowable norm of the gradients at the stopping iteration is $\epsilon$. 
Figures \ref{fig:visual_single}(b,d) show that $\epsilon$ plays a far more influential role than $\delta$. 
Specifically, this implies that even small changes in $\epsilon$ require adjusting $N$ appropriately to maintain the same level of convergence. 
\item The top and bottom rows in Figure \ref{fig:visual_multi} show the trends predicted by the bounds for non-dropout and dropout networks respectively. 
Figure \ref{fig:visual_multi}(a) shows the strong influence of the network depth and the hidden layer lengths. This is not surprising from the structure of the bound in \eqref{eq:convmulnn}.
Figure \ref{fig:visual_multi}(b) clearly shows the necessity of pretraining, especially in networks with large layer lengths. 
It is interesting to see that large hidden layer lengths completely dominate any changes in the dropout rate (Figure \ref{fig:visual_multi}(d)).
Figure \ref{fig:visual_multi}(e) shows that the dropout rate 
has a much smaller influence than the pretraining goodness (see discussion below about Figure \ref{fig:visual_den_drop}(b)).
Figures \ref{fig:visual_multi}(c,f) show that hidden layer lengths are more influential than $d_0$.
This may be seen from \eqref{eq:convmulnn} and \eqref{eq:mlnnpretdrop} because the depth parameter $L$ inherently puts more weight on the hidden layer constants $e^m_l$.  
\item The optimal denoising rate will be given by the solutions of \eqref{eq:convdaparam-eda}. 
Figure \ref{fig:visual_den_drop}(a) plots $e^{da}$ as the denoising rate ($x$-axis) and $d_x$ ($y$-axis) are changed. 
$\tzeta$ smaller than $0.5$ seems to be the `good' regime, however, $d_x$ has a much stronger influence. 
For very large $d_x$, the ranges of $e^{da}$ do not seem to be influenced by $\zeta$ (see the top half vs. bottom half of Figure \ref{fig:visual_den_drop}(a)).
This is in agreement with the interpretations made earlier in Section \ref{sec:da-net-conv}.
\item Figure \ref{fig:visual_den_drop}(b) shows the optimal dropout rate $\zeta$ (from \eqref{eq:bestdrop}) as $\alpha_l$s and $d_l$s are changed appropriately. 
The interpretations made earlier in the remarks of Corollary \ref{thm:bestdrop} can be seen here, e.g., $\zeta$ is small whenever $\alpha_l$s is large. 
Further, \eqref{eq:bestdrop} seems to indicate that very large hidden layer lengths need large dropout rates independent of how good the pretraining is. 
If one marginalizes the influence of pretraining (i.e., summing up columns of Figure \ref{fig:visual_den_drop}(b)), 
with no extra information of the network depth and also, the layer lengths are reasonable (or large), the {\it go to} choice of dropout is simply $0.5$.
The authors in \cite{baldi2014dropout} proved that $0.5$ dropout rate is optimal for dropout dynamics, and Figure \ref{fig:visual_den_drop}(b) shows 
that such a rate is really the default choice even from the perspective of convergence.
While presenting design strategies in Section \ref{sec:design-revisit}, we show that other choices are optimal when extra information about data and network structure are allowed to be used.
\end{itemize}




\section{Discussion}
\label{sec:disc}

Building upon the evaluations in Section \ref{sec:exps} where the technical results from Sections \ref{sec:single}--\ref{sec:multi} are assessed empirically,
we now turn to the design choice problem presented in Section \ref{sec:design} for constructing design strategies or {\it ground rules} to drive the construction and learning of deep networks. 
We approach this problem in a few different ways -- motivated mostly by the arguments in Section \ref{sec:design}. 
Once the network structure is set up, we then compute the minimum required number of training instances and the batchsize 
using other learning hyper-parameters and input data statistics -- all of which come from Theorems \ref{thm:expgradda}, \ref{thm:expgradmulnn}, \ref{thm:pretvsdrop-gen}, 
Corollaries \ref{thm:equiv-pretrain}, \ref{thm:bestdrop}, \ref{thm:equiv-pretrain-drop}, 
and the corresponding discussion in Sections \ref{sec:da-net-conv}, \ref{sec:goodpretrain}, \ref{sec:compare}, \ref{sec:droprates} and \ref{sec:comparedrop}. 
Corollary \ref{thm:equiv-pretrain-drop} guides the overall design procedure. 


\subsection{Revisiting the design choice problem}
\label{sec:design-revisit}

By design of deep network, we mean the task of choosing all the network hyper-parameters (see Section \ref{sec:design} for a discussion). 
For convenience, we first list the key hyper-parameters involved in different stages of the network design in Table \ref{tab:allhyper}. 
The pretraining hyper-parameters (the denoising rates for the $L-1$ DAs, the pretraining batchsize and iterations) are not included in this list because they are inherently controlled by $\alpha_l$s.
Once $\alpha_l$s are chosen, these can be picked appropriately based on the discussion in Section \ref{sec:da-net-conv}.
Depending on the task at hand, we are at liberty to change the output layer length $d_L$, if needed.
Specifically, for classification or regression on a {\it fixed} set of outcomes $\y$, it is reasonable to fix $d_L$ together with $d_0$.
On the other hand, when the networks are used for learning representations, allowing $d_L$ to be chosen by the design procedure is more sensible. 
So, we include $d_L$ within the design choice as well. 

\begin{table}[!h] \centering 
\caption{\footnotesize \label{tab:allhyper} The hyper-parameters involved in the design choice of multi-layer RSG (from Alg. \ref{alg:multi-nn})}
\vspace{2mm}
\begin{tabular}{|c|c|} \hline 
\multicolumn{2}{|l|}{\bf Structural Hyper-parameters} \\ \hline
$L$ & Network depth \\ 
$d_0,\ldots,d_L$ & Input, hidden and output layers length \\ \hline
\end{tabular}
\vspace{3mm}

\begin{tabular}{|c|c|} \hline 
\multicolumn{2}{|l|}{\bf Learning Hyper-parameters} \\ \hline
$\gamma^1,\ldots,\gamma^L$ & Stepsize choices \\
$w_m^1,\ldots,w_m^L$ & Box-constraints for each DA \\
$\zeta_1,\ldots,\zeta_{L-1}$ & Dropout rates for $L-1$ hidden layers \\
$\ceil{\frac{1}{\kappa}}$ & Number of epochs \\
$N^1_u,\ldots,N^{L-1}_u$ & $L-1$ DA iterations \\ 
$N_s$ & Backpropagation iterations \\ 
$S_u$ and $S_s$ & Unlabeled and Labeled instances \\ \hline
\end{tabular}
\vspace{3mm}

\begin{tabular}{|c|c|} \hline 
\multicolumn{2}{|l|}{\bf Goodness Criteria} \\ \hline
$(\alpha_1,\delta_{\alpha_1}),\ldots(\alpha_{L-1},\delta_{\alpha_{L-1}})$ & Pretraining goodness for $L-1$ hidden layers \\
$\Psi_1,\ldots,\Psi_{L-1}$ & Convergence goodness (Non-dropout network)\\
$\Phi_1,\ldots,\Phi_L$ and $\Phi_f$ & Convergence goodness (Dropout network) \\
$(\epsilon,\delta)$ & Goodness of solution/estimate \\ \hline
\end{tabular} \end{table} 


The motivation in Section \ref{sec:design}, especially the multiple sites setup in Section \ref{sec:resource}, 
deals with the comparability of different network designs. We make a few remarks about the criteria
and the availability (or the lack thereof) of unsupervised data before describing the procedure. 
We see from empirical observations that $\W$s can be interpreted as ``filters'' transforming the given signals. 
For instance, the transformation maps in deep Boltzmann machines can be interpreted as representation `stubs' which the network tries to identify in the inputs, 
e.g., the binary $\W$s learned with MNIST data \citep{nair2010rectified}.
In convolutional networks, $\W$s directly correspond to the filter maps that are sequentially applied on the transformed inputs \citep{kavukcuoglu2010learning, lee2009convolutional}. 
Therefore, it is reasonable that we ask that the dimensions of these filters $\W$s or hidden layer lengths $d_l$s be the same
to ensure that networks are comparable via the transformation maps. 
We refer to this setting by saying that the network design needs to be ``transferable''.
Alternatively, if one is more concerned with generalization, then such a restrictive control on $d_l$s is not necessary, 
and in certain cases may also lead to sub-optimal networks. 
Further, for each of these settings, depending on the availability of unlabeled data, 
the multi-layer RSG in Alg. \ref{alg:multi-nn} may either include a layer-pretraining procedure or may not. 
So, we have a total of four different design procedures.


At a high level, the convergence and pretraining goodness criteria are used to bound the hidden layer lengths (using input and output layers), 
which are then used to decide the learning hyper-parameters.  
For convenience, the constraints from \eqref{eq:equiv-pretrain} and \eqref{eq:equiv-pretrain-drop} are summarized in Table \ref{tab:constraints}
(see the remarks for Corollaries \ref{thm:equiv-pretrain} and \ref{thm:equiv-pretrain-drop}). 
Based on Table \ref{tab:constraints}, Tables \ref{tab:pret} presents the design procedure for datasets with unlabeled data whereas Table \ref{tab:drop} presents the fully-supervised design.
The inputs to Tables \ref{tab:pret} and \ref{tab:drop} are the pretraining and convergence goodness criteria (see Table \ref{tab:allhyper}), 
the desired depth of the network, number of epochs allowed and the input data.
The specific steps within these procedures are self-explanatory -- using these inputs, the procedures first decide the network structure followed by the appropriate learning hyper-parameters.
Wherever appropriate, some free hyper-parameters like $\gamma$ and $w^m_l$s are listed in Tables \ref{tab:pret} and \ref{tab:drop} which can be set to reasonable values.
Note that we do not claim that these strategies are ideal -- instead, 
they are the {\it simplest} ones directly implied by our results with no additional assumptions.


\begin{table}[!h] \centering 
  \caption{\footnotesize \label{tab:constraints} Design constraints from \eqref{eq:equiv-pretrain} and \eqref{eq:equiv-pretrain-drop} in
    Corollaries \ref{thm:equiv-pretrain} and \ref{thm:equiv-pretrain-drop}}
\vspace{2mm}
\begin{tabular}{|lll|} \hline
\multicolumn{3}{|c|}{} \\
{\bf U.1} & $\log(\zeta) + \log(d_0) + \log(d_1) + \log(d_2) \qquad\qquad\qquad\qquad\qquad\qquad$ & = $\log(\frac{1}{\Psi_1\alpha_1})$ \\
{\bf U.2} & $\log(\zeta) + \qquad\qquad \log(d_1) + \log(d_2) + \log(d_3) \qquad\qquad\qquad\qquad$ & = $\log(\frac{1}{\Psi_2\alpha_2})$ \\
\qquad $\vdots$ & \qquad\qquad\qquad\qquad\qquad\qquad\qquad\qquad$\vdots$ & $\vdots$ \\
{\bf U.L-2} & $\log(\zeta) + \qquad\qquad\qquad\quad \log(d_{L-3}) + \log(d_{L-2}) + \log(d_{L-1}) \qquad$ & $= \log(\frac{1}{\Psi_{L-2}\alpha_{L-2}})$ \\
{\bf U.L-1} & $-\log(\zeta) + \quad\qquad\qquad\qquad\qquad\qquad\qquad\quad + \log(d_{L-1}) + \log(d_L)$ & $= \log(\frac{1}{\Phi_L})$ \\ 
{\bf U.L} & $-2\log(\zeta) + \quad\qquad\qquad\qquad\qquad\qquad\qquad\qquad\qquad\qquad + \log(d_L)$ & $= \log(\frac{1}{\Phi_f})$ \\
\multicolumn{3}{|c|}{} \\ \hline
\multicolumn{3}{|c|}{} \\ 
{\bf S.1} & $\log(\zeta) + \log(d_0) + \log(d_1) + \log(d_2) \qquad\qquad\qquad\qquad\qquad\qquad$ & = $\log(\frac{1}{\Phi_1})$ \\
{\bf S.2} & $\log(\zeta) + \qquad\qquad \log(d_1) + \log(d_2) + \log(d_3) \qquad\qquad\qquad\qquad$ & = $\log(\frac{1}{\Phi_2})$ \\
\qquad $\vdots$ & \qquad\qquad\qquad\qquad\qquad\qquad\qquad\qquad$\vdots$ & $\vdots$ \\
{\bf S.L-2} & $\log(\zeta) + \qquad\qquad\qquad\quad \log(d_{L-3}) + \log(d_{L-2}) + \log(d_{L-1}) \qquad$ & $= \log(\frac{1}{\Phi_{L-2}})$ \\
{\bf S.L-1} & $-\log(\zeta) + \quad\qquad\qquad\qquad\qquad\qquad\qquad\quad + \log(d_{L-1}) + \log(d_L)$ & $= \log(\frac{1}{\Phi_L})$ \\
{\bf S.L} & $-2\log(\zeta) + \quad\qquad\qquad\qquad\qquad\qquad\qquad\qquad\qquad\qquad + \log(d_L)$ & $= \log(\frac{1}{\Phi_f})$ \\
\multicolumn{3}{|c|}{} \\ \hline
\end{tabular} \end{table} 


\begin{table}[!h] \centering 
\caption{\footnotesize \label{tab:pret} Deep network design -- $S_u$ and $S_s$ available}
\begin{tabular}{|c|c|} \hline
\multicolumn{2}{|c|}{} \\
\multicolumn{2}{|l|}{Given the goodness criteria $\alpha_l$s, $\delta_{\alpha_l}$s, $\Psi_l$s, $\Phi_L$, $\Phi_f$, $\epsilon$ and $\delta$ from Table \ref{tab:allhyper}} \\
\multicolumn{2}{|l|}{\quad depth $L$, input length $d_0$, number of epochs $\ceil{\frac{1}{\kappa}}$} \\
\multicolumn{2}{|c|}{} \\
\multicolumn{2}{|l|}{If $d_L$ and $\zeta$ are not fixed} \\ 
\multicolumn{2}{|l|}{\quad If $d_l$s need to be transferable, use $d_L = \ceil{\frac{65}{4\alpha_{L-1}}}$;} \\ 
\multicolumn{2}{|l|}{\quad\quad else pick any $d_L < \ceil{\frac{65}{4\alpha_{L-1}}}$;} \\
\multicolumn{2}{|l|}{\quad solve the linear system {\bf U.1, \ldots, U.L} from Table \ref{tab:constraints}} \\
\multicolumn{2}{|l|}{If either $d_L$ or $\zeta$ is given/fixed} \\ 
\multicolumn{2}{|l|}{\quad solve the linear system {\bf U.1, \ldots, U.L} from Table \ref{tab:constraints}} \\
\multicolumn{2}{|l|}{If $d_L$ and $\zeta$ are fixed} \\ 
\multicolumn{2}{|l|}{\quad solve the linear system {\bf U.1, \ldots, U.L-1} from Table \ref{tab:constraints}} \\
\multicolumn{2}{|c|}{} \\
\multicolumn{2}{|l|}{{\it Pretraining:} For each of the $l=1,\ldots,L-1$ DAs} \\
\multicolumn{2}{|l|}{\quad Select denoising rates, $w_m$ and stepsizes appropriately (No restriction)} \\
\multicolumn{2}{|l|}{\quad Compute $N^l_u$ from \eqref{eq:itersampda} for $(\alpha_l,\delta_{\alpha_l})$ solution} \\
\multicolumn{2}{|c|}{} \\
\multicolumn{2}{|l|}{{\it Backprop using Alg.} \ref{alg:multi-nn}:} \\ 
\multicolumn{2}{|l|}{Compute $e^m_1,\ldots,e^m_L$} \\
\multicolumn{2}{|l|}{Choose $\gamma$ and $w^m_1,\ldots,w^m_L$ appropriately (No restriction)} \\
\multicolumn{2}{|l|}{If $\zeta \approx 0$ or $\approx 1$} \\
\multicolumn{2}{|l|}{\quad use \eqref{eq:bestdrop} with $d=\frac{1}{L-1}\sum_{l=1}^{L-1}d_l$, or simply set $\zeta=0.5$} \\
\multicolumn{2}{|l|}{Estimate required $N_s$ using \eqref{eq:sampmulnn-drop} for $(\epsilon,\delta)$ solution} \\ 
\multicolumn{2}{|c|}{} \\ \hline
\end{tabular} \end{table} 


\begin{table}[!h] \centering 
\caption{\footnotesize \label{tab:drop} Deep network design -- $S_u$ not available}
\begin{tabular}{|c|c|} \hline
\multicolumn{2}{|c|}{} \\
\multicolumn{2}{|l|}{Given the goodness criteria $\Phi_l$s, $\Phi_f$, $\epsilon$ and $\delta$ from Table \ref{tab:allhyper}} \\
\multicolumn{2}{|l|}{\quad depth $L$, input length $d_0$, number of epochs $\ceil{\frac{1}{\kappa}}$} \\
\multicolumn{2}{|c|}{} \\
\multicolumn{2}{|l|}{If $d_L$ and $\zeta$ are not fixed} \\ 
\multicolumn{2}{|l|}{\quad If $d_l$s need to be transferable; pick $\zeta=0.5$} \\ 
\multicolumn{2}{|l|}{\quad\quad else choose any $0<\zeta<1$} \\
\multicolumn{2}{|l|}{\quad solve the linear system {\bf S.1, \ldots, S.L} from Table \ref{tab:constraints}} \\
\multicolumn{2}{|l|}{If either $d_L$ or $\zeta$ is given/fixed} \\ 
\multicolumn{2}{|l|}{\quad solve the linear system {\bf S.1, \ldots, S.L} from Table \ref{tab:constraints}} \\
\multicolumn{2}{|l|}{If $d_L$ and $\zeta$ are fixed} \\ 
\multicolumn{2}{|l|}{\quad solve the linear system {\bf S.1, \ldots, S.L-1} from Table \ref{tab:constraints}} \\
\multicolumn{2}{|c|}{} \\
\multicolumn{2}{|l|}{{\it Backprop using Alg.} \ref{alg:multi-nn}:} \\ 
\multicolumn{2}{|l|}{Compute $e^m_1,\ldots,e^m_L$. Estimate $\alpha_l$s with random network initialization} \\
\multicolumn{2}{|l|}{Choose $\gamma$ and $w^m_1,\ldots,w^m_L$ appropriately (No restriction)} \\
\multicolumn{2}{|l|}{If $\zeta \approx 0$ or $\approx 1$} \\
\multicolumn{2}{|l|}{\quad use \eqref{eq:bestdrop} with $d=\frac{1}{L-1}\sum_{l=1}^Ld_l$ and $\alpha = \frac{1}{L-1}\sum_{l=1}^{L-1}\alpha_l$} \\
\multicolumn{2}{|l|}{\quad or simply set $\zeta=0.5$} \\
\multicolumn{2}{|l|}{Estimate required $S_s$ using \eqref{eq:sampmulnn-drop} for $(\epsilon,\delta)$ solution} \\ 
\multicolumn{2}{|c|}{} \\ \hline
\end{tabular} \end{table} 


\subsubsection{\mathversion{bold} Interpreting and Choosing $\alpha_l$s, $\delta_{\alpha_l}$s, $\epsilon$ and $\delta$ \mathversion{normal}} 
\label{sec:choice-goodness-1}

These goodness criteria influence the large deviation estimates of the pretraining and dropout learning and their choice is vital towards the structure of the network.  
Recalling the definition of $(\epsilon,\delta)$-solution from Definition \ref{thm:epsdel}, and further 
using Theorem \ref{thm:expgradmulnn}, we have the following constraints on these criteria.
\begin{equation} \begin{aligned} \label{eq:not-psi-phi}
& 0 < \alpha_l < 1, \quad 0 < \delta_{\alpha_l} < 1 \quad\text{for}\quad l=1,\ldots,L-1 \\
& 0 < \epsilon < 1, \quad 0 < \delta < 1
\end{aligned} \end{equation}
$(\alpha_1,\delta_{\alpha_1}), \ldots, (\alpha_{L-1},\delta_{\alpha_{L-1}})$ determine the large deviation pretraining estimates of the $L-1$ hidden layers -- 
see \eqref{eq:convmulnn-gen}, Theorem \ref{thm:expgradmulnn}, Theorem \ref{thm:pretvsdrop} and the corresponding remarks. 
$(\epsilon,\delta)$ governs the large deviation estimate of the supervised backpropagation. 
Since these four sets of criteria come from large deviation estimates (see Definition \ref{thm:epsdel}), these are straightforward to setup: 
\begin{compactitem}
\item Choose $\alpha_l$s and $\epsilon$ such that $0 < \alpha_l, \epsilon \ll 1$ 
\item Choose $\delta_{\alpha_l}$s and $\delta$ such that $0 \ll \delta_{\alpha_l}, \delta < 1$
\end{compactitem}


\subsubsection{\mathversion{bold} Interpreting and Choosing $\Psi_l$s, $\Phi_l$s and $\Phi_f$ \mathversion{normal}} \label{sec:choice-goodness-2}

We find that $\Psi_l$s, $\Phi_l$s and $\Phi_f$ govern the contribution of each of the $1$-NNs that compose the $L$-NN to the decay of projected gradients -- 
see Corollaries \ref{thm:equiv-pretrain}, \ref{thm:equiv-pretrain-drop} and the corresponding remarks. 
Unlike the goodness criteria from Section \ref{sec:choice-goodness-1} which are directly related to the overall large deviation gradient norm of the network 
(and so easier to interpret and choose), the meta-parameters $\Psi_l$s, $\Phi_l$s and $\Phi_f$ are not straight forward to set up.
They come out of the terms in the decay bounds -- see \eqref{eq:equiv-pretrain} and \eqref{eq:equiv-pretrain-drop} -- and one will want them to be 
as small as possible which in turn leads to small expected gradients.
Using \eqref{eq:equiv-pretrain} and \eqref{eq:equiv-pretrain-drop} have,
\begin{equation} \begin{aligned} \label{eq:psi-phi}
& \Psi_l > 0 \quad\text{for}\quad l=1,\ldots,L-1 \\
& \Psi_l > 0 \quad\text{for}\quad l=1,\ldots,L \quad\text{and}\quad  \Psi_f > 0
\end{aligned} \end{equation}
Recall that given $\Psi_l$s, $\Phi_l$s and $\Phi_f$, the network is guaranteed to achieve the convergence level from 
\eqref{eq:expgradmulnn-equiv-pretrain} and \eqref{eq:expgradmulnn-equiv-pretrain-drop} (the quantities are not upper bounded).


A simple way to interpret them is by thinking of them, loosely, as {\it significance levels}. 
If the pre-specified threshold is large, the test is liberal
and on the other hand, a smaller threshold makes the test very conservative.
Too small values for $\Psi_l$s, $\Phi_l$s and $\Phi_f$ lead to large networks, which requires large $N$, batchsize and sample sizes (too liberal). 
On the other hand, we also want to avoid prohibitively smaller networks that may not generalize (too conservative). 
Since $\Psi_l$s, $\Phi_l$s and $\Phi_f$ are proportional to $\frac{1}{d^{-3}}$ where $d$ denotes the average hidden layer length (see Table \ref{tab:allhyper}), 
their {\it typical} ranges seems to be closer to $0$, and therefore, without loss of generality, we restrict ourselves to $[0,0.01]$.

\begin{itemize} [leftmargin=1em]
\item {\bf \mathversion{bold} $0 < \Psi_l, \Phi_l, \Phi_f \ll 1$ \mathversion{normal}:}
The smaller these criteria are, the larger the hidden and output layer lengths will be, which directly follows the equalities {\bf U.l} and {\bf S.l} 
(or \eqref{eq:equiv-pretrain} and \eqref{eq:equiv-pretrain-drop}). 
Following the overwhelming evidence for large hidden layers \citep{bengio2009learning, hinton2010practical, bengio2012practical},
one may clearly choose $\Psi_l$, $\Phi_l$s and $\Phi_f$ to be small ($\rightarrow 0$). 
Such small values will make the network {\it liberal} i.e., given sufficient training data and $N$, the network may model any complex concept but will 
be computationally very expensive.
\item {\bf \mathversion{bold} $0 \ll \Psi_l, \Phi_l, \Phi_f < 1$ \mathversion{normal}:}
However, Corollaries \ref{thm:tall-short} and \ref{thm:tall-short-drop} showed that the taller networks may not always generalize better than shorter ones. 
In this alternate regime where the $\Psi_l$, $\Phi_l$s and $\Phi_f$ are reasonably larger than $0$, the hidden layer lengths may be small 
i.e., the network is {\it conservative} and can achieve faster convergence with small training size and $N$ but may not be able to learn complex concepts. 
\item {\bf \mathversion{bold} $\Psi_l\alpha_l \approx \Phi_l$ \mathversion{normal}}
As described earlier, $\alpha_l$s are easier to choose compared to $\Psi_l$s, $\Phi_l$s and $\Phi_f$.
Further, the linear systems, {\bf U.l}s and {\bf S.l}s from Table \ref{tab:allhyper} suggest an interesting relation between these criteria
\begin{equation} \label{eq:alpha-phi-psi}
\alpha_l\Psi_l \approx \Phi_l \quad l=1,\ldots,L-2
\end{equation}
Since $\alpha_l \ll 1$, the good regimes of $\Phi_l$s are much smaller than $\Psi_l$s. 
\end{itemize}
This list of regimes gives a clear strategy to choose the appropriate $\Psi_l$s and $\Phi_l$s. 
Using these prototypical ranges for $\alpha_l$s, $\Psi_l$s, $\Phi_l$s and $\Phi_f$, 
we present a few example designs in the next section to demonstrate the practicality of our design procedures from Tables \ref{tab:pret} and \ref{tab:drop}, 


\subsubsection{Example designs:} \label{sec:example-designs}

\begin{itemize} [leftmargin=1em]
\item {\bf Example 1.} {\it $d_0 = 100$, $10$ epochs, $L=5$, No $S_u$:} Designs from Table \ref{tab:drop} are applicable here.
\begin{itemize}
\item {\it $d_L = 5$, $\zeta = 0.5$:} \\ 
\begin{equation} \begin{aligned} &\Phi_1 = \Phi_2 = \Phi_3 = 10^{-5} \quad \Phi_5 = 10^{-3} \\
&\quad\quad \implies d_1 = 100, d_2 = 20, d_3 = 100, d_4 = 100 \end{aligned} \end{equation}
\item {\it $d_L = 5$:} \\
\begin{equation} \begin{aligned} &\Phi_1 = \Phi_2 = \Phi_3 = 10^{-5} \quad \Phi_5 = 10^{-3} \quad \Phi_f = 0.01 \\
&\quad\quad \implies \zeta = 0.25, d_1 = 50, d_2 = 80, d_3 = 100, d_4 = 50 \end{aligned} \end{equation}
\item {\it No $d_L$, No $\zeta$:} \\
\begin{equation} \begin{aligned} &\Phi_1 = \Phi_2 = \Phi_3 = 10^{-5} \quad \Phi_5 = 10^{-3} \quad \Phi_f = 0.01 \\
&\quad\quad \implies \quad\text{with $\zeta =0.5$,}\quad d_1 = 50, d_2 = 80, d_3 = 100, d_4 = 50, d_5 = 25 \end{aligned} \end{equation}
\end{itemize}


\item {\bf Example 2.} {\it $d_0 = 100$, $10$ epochs, $L=5$:} The designs from Table \ref{tab:pret} are applicable here.
\begin{itemize}
\item {\it $d_L = 5$, $\zeta = 0.5$:} \\
\begin{equation} \begin{aligned} &\Psi_1 = \Psi_2 = \Psi_3 = 10^{-3} \quad \alpha_1 = \alpha_2 = \alpha_3 = 10^{-2} \quad \Phi_5 = 10^{-3} \\
&\quad\quad \implies d_1 = 100, d_2 = 20, d_3 = 100, d_4 = 100 \end{aligned} \end{equation}
\item {\it $d_L = 5$:} \\
\begin{equation} \begin{aligned} &\Psi_1 = \Psi_2 = \Psi_3 = 10^{-3} \quad \alpha_1 = \alpha_2 = \alpha_3 = 10^{-2} \quad \Phi_5 = 10^{-3} \quad \Phi_f = 0.01 \\
&\quad\quad \implies \zeta = 0.25, d_1 = 50, d_2 = 80, d_3 = 100, d_4 = 50 \end{aligned} \end{equation}
\item {\it No $d_L$, No $\zeta$:} \\
\begin{equation} \begin{aligned} &\Psi_1 = \Psi_2 = \Psi_3 = 10^{-3} \quad \alpha_1 = \alpha_2 = \alpha_3 = 10^{-2} \quad \Phi_5 = 10^{-3} \quad \Phi_f = 0.01 \\
&\quad\quad \implies \quad\text{with $\zeta =0.5$,}\quad d_1 = 50, d_2 = 80, d_3 = 100, d_4 = 50, d_5 = 25 \end{aligned} \end{equation}
\end{itemize}


\item {\bf Example 3.} {\it $d_0 = 100$, $10$ epochs, $L=5$, No dropout:} Designs from Table \ref{tab:pret} are applicable here with $\zeta=1$.
\begin{itemize}
\item {\it $d_L = 5$:} \\
\begin{equation} \begin{aligned} &\Psi_1 = \Psi_2 = \Psi_3 = 10^{-4} \quad \alpha_1 = \alpha_2 = \alpha_3 = 10^{-2} \quad \Phi_5 = 10^{-3} \\
&\quad\quad \implies d_1 = 200, d_2 = 50, d_3 = 100, d_4 = 200 \end{aligned} \end{equation}
\item {\it No $d_L$:} \\
\begin{equation} \begin{aligned} &\Psi_1 = \Psi_2 = \Psi_3 = 10^{-4} \quad \alpha_1 = \alpha_2 = \alpha_3 = 10^{-2} \quad \Phi_5 = 10^{-3} \quad \Phi_f = 0.1 \\
&\quad\quad \implies d_1 = 100, d_2 = 100, d_3 = 100, d_4 = 100, d_5 = 10 \end{aligned} \end{equation}
\end{itemize}
\end{itemize}

\begin{figure}[!h]\centering
\subfloat[]{\includegraphics[width=140mm]{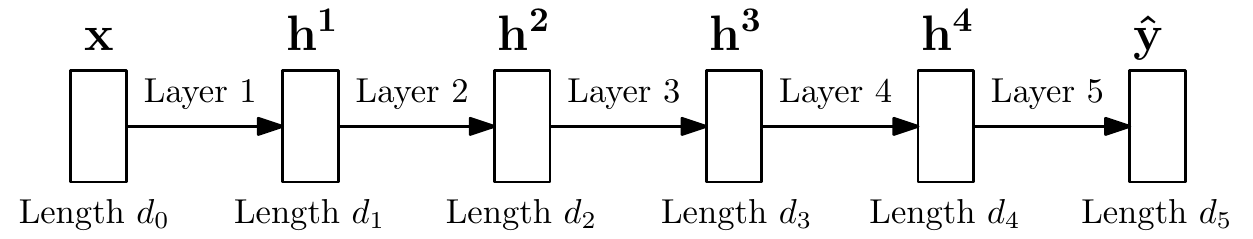}} \\ 
\subfloat[]{\includegraphics[width=47mm]{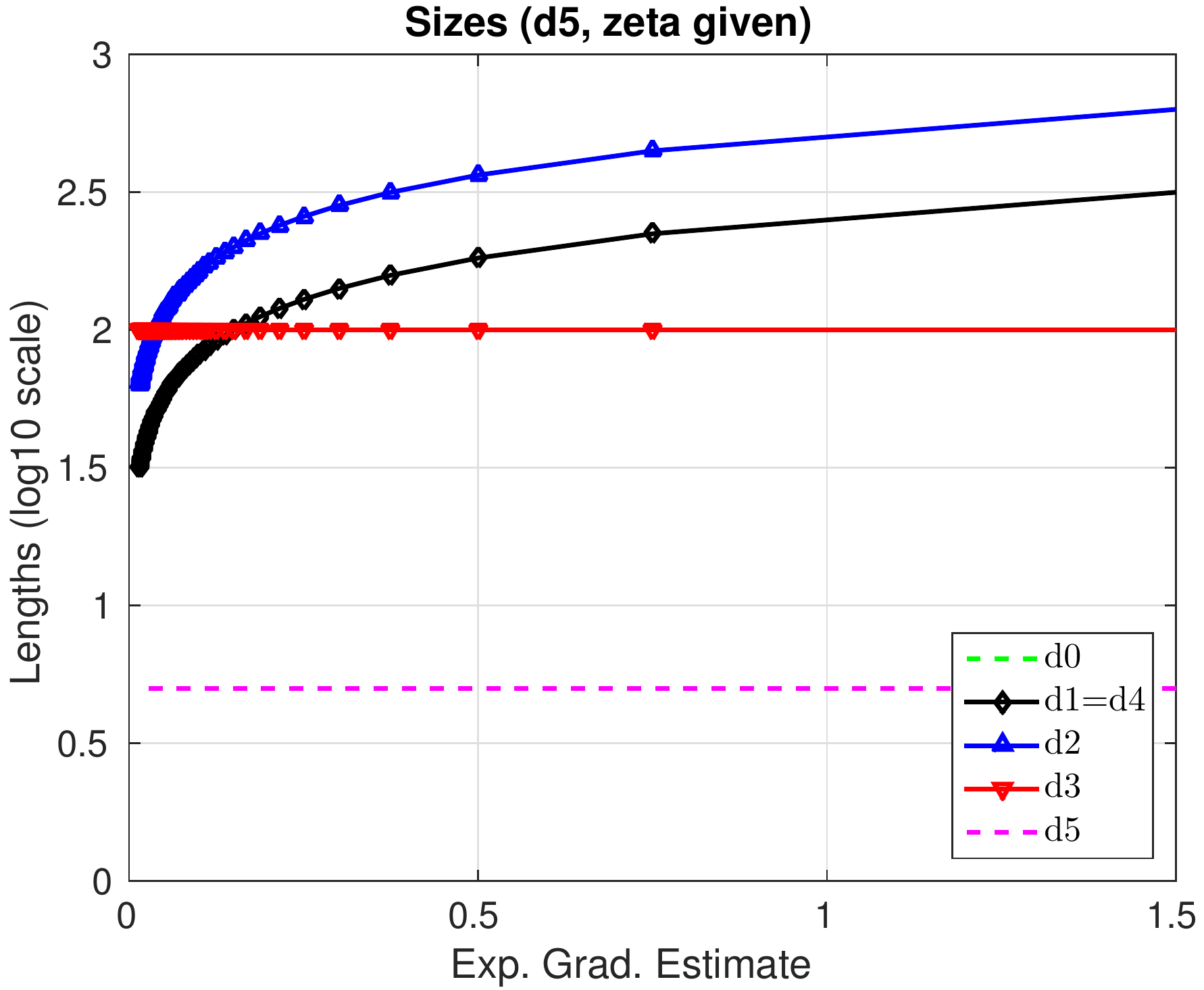}} \quad 
\subfloat[]{\includegraphics[width=47mm]{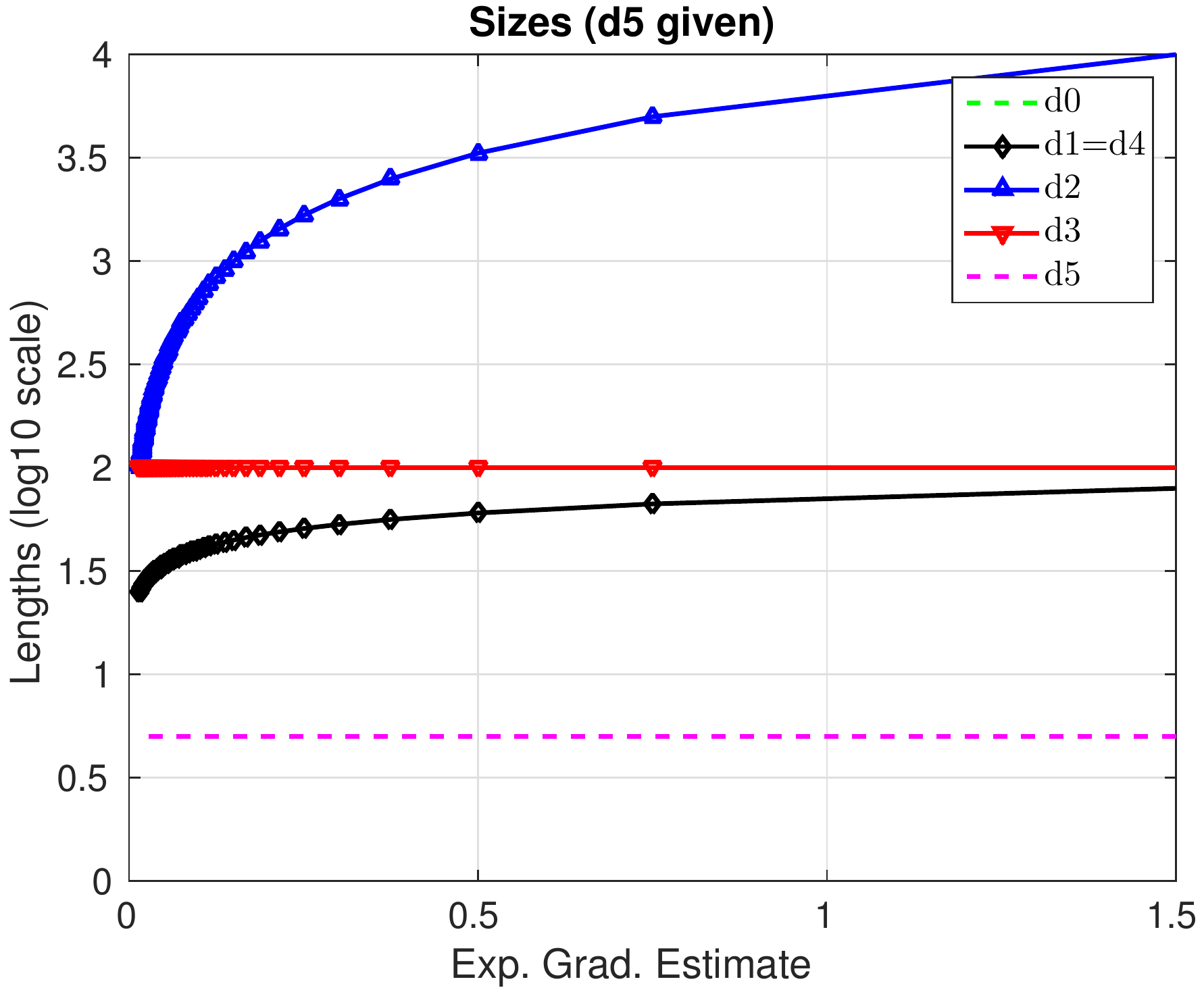}} \quad
\subfloat[]{\includegraphics[width=47mm]{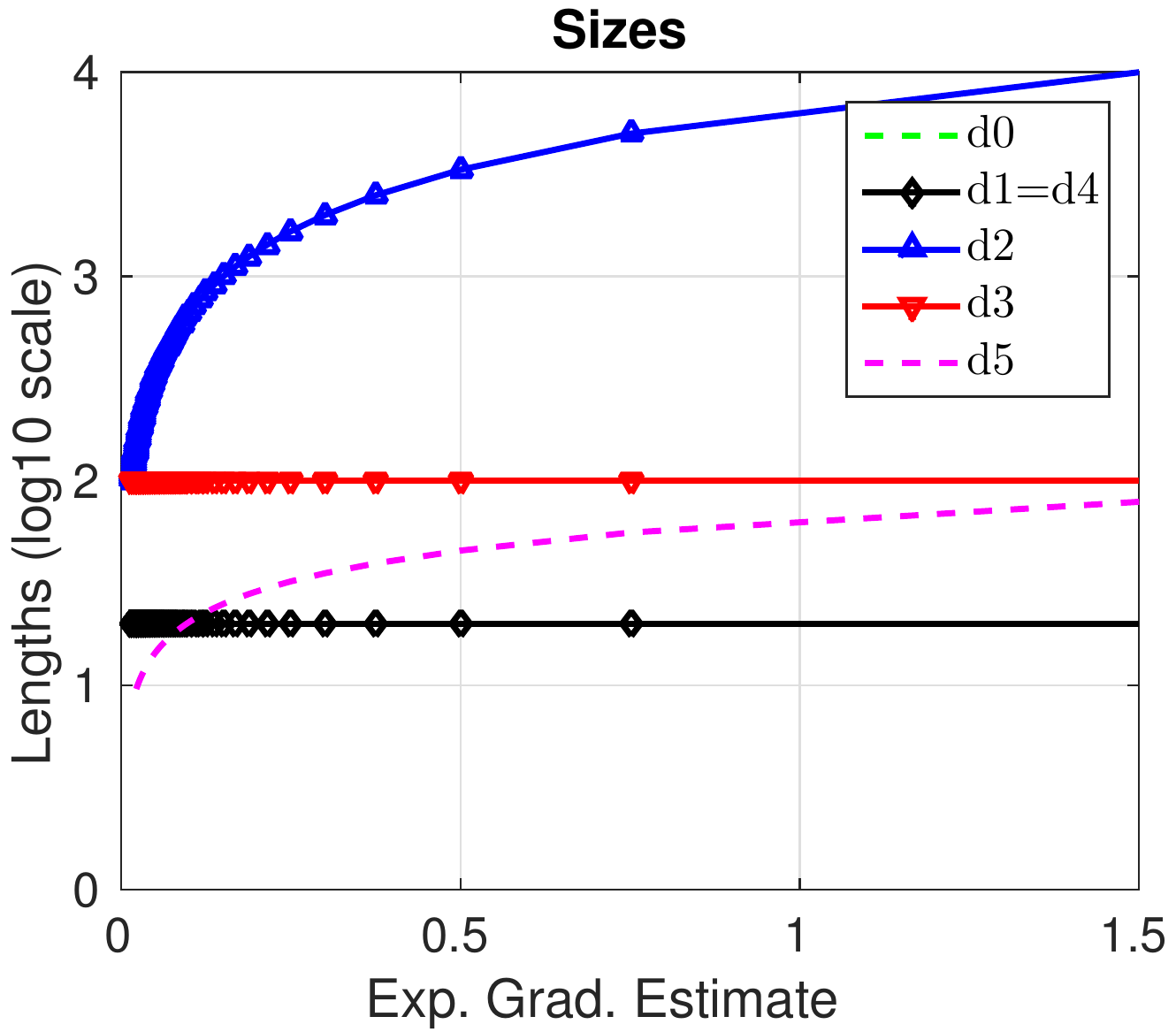}} \\
\subfloat[]{\includegraphics[width=47mm]{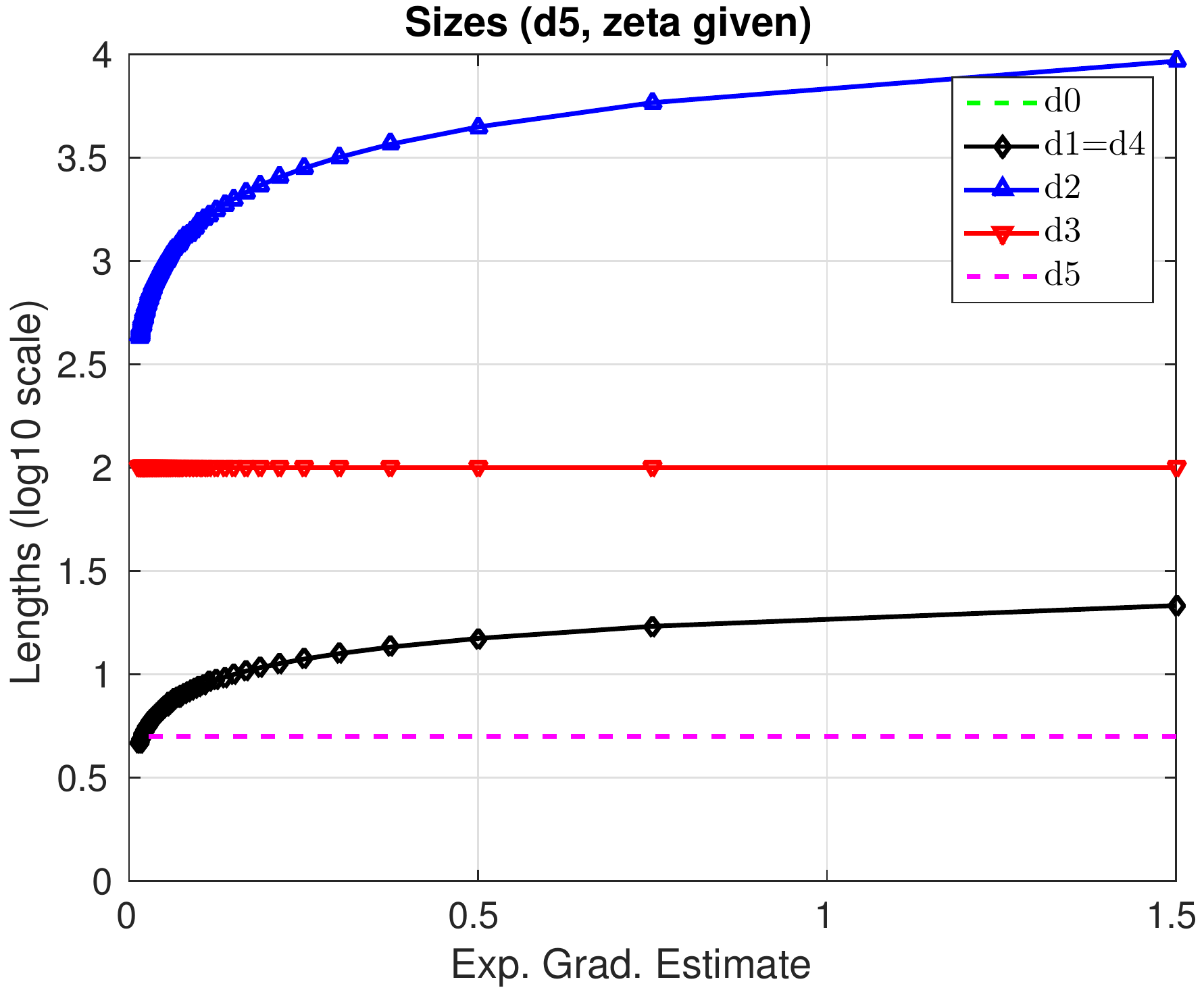}} \quad 
\subfloat[]{\includegraphics[width=47mm]{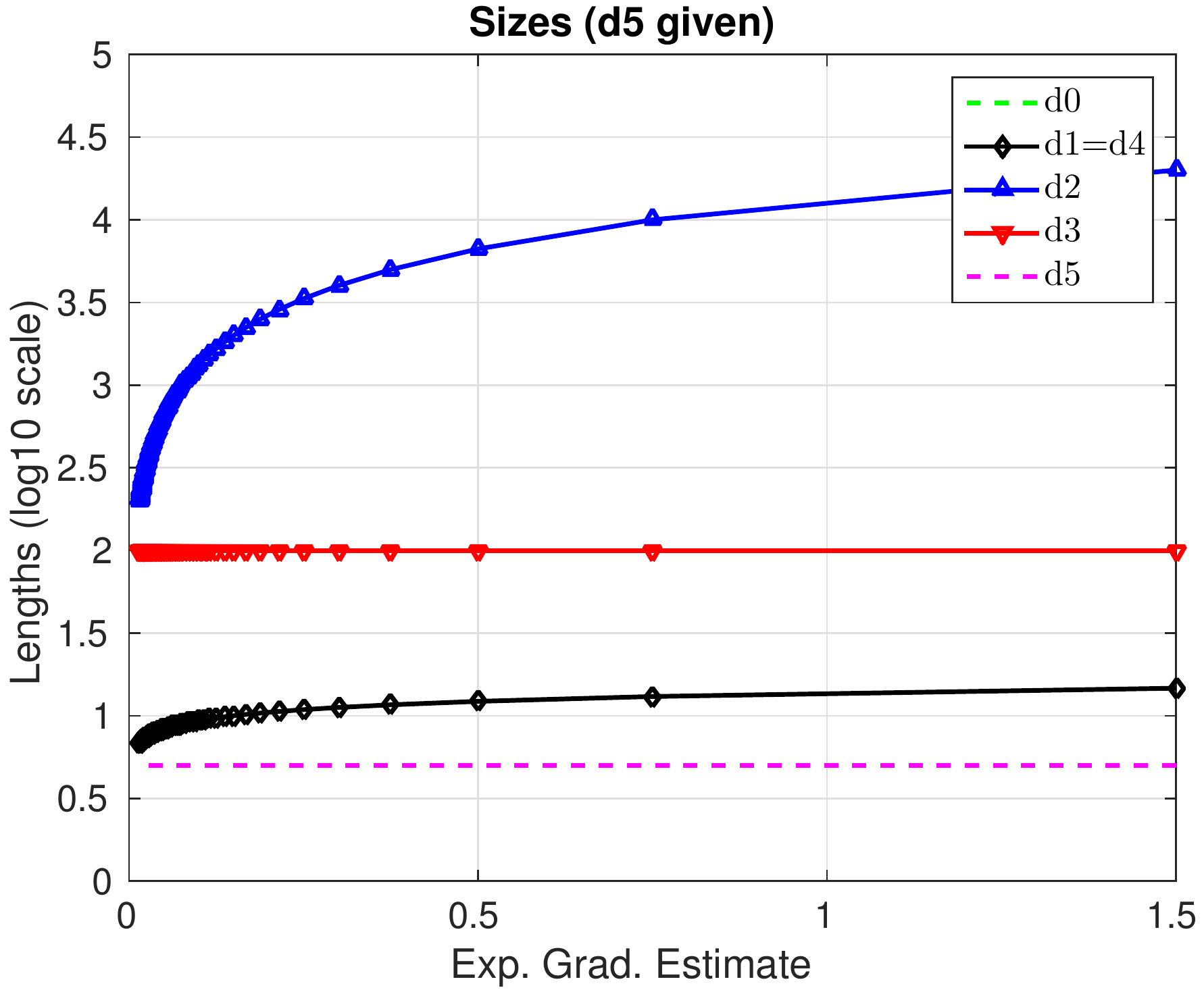}} \quad  
\subfloat[]{\includegraphics[width=47mm]{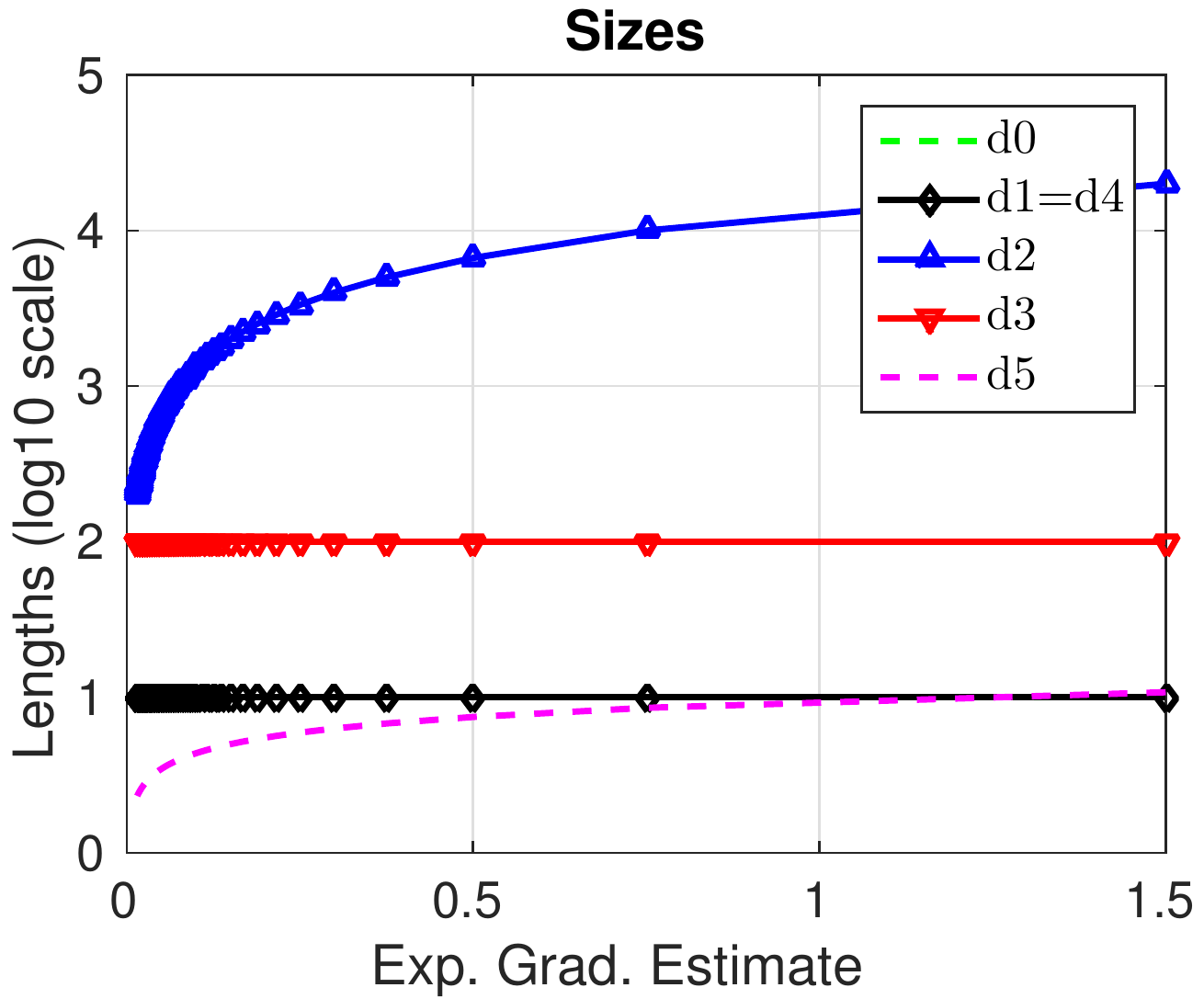}}
\caption{\footnotesize \label{fig:ex_designs} {\bf Designs (Network Lengths)} Example hidden and output layer lengths using Table \ref{tab:drop} design of a $L=5$ network (shown at the top). 
Columns correspond to $d_5$, $\zeta$ known; $\zeta$ unknown and both unknown respectively. 
$x$-axis shows $\Phi_1 = \Phi_2 = \Phi_3$. Top row uses $\Phi_5 = \sqrt{\Phi_1}$ and bottom row uses $\Phi_5 = \sqrt[3]{\Phi_1}$. $\Phi_f=10\Phi_5$.
$y$-axis shows layer lengths in $\log_{10}$ scale.
} 
\end{figure}

Note that the later half of the design procedure (from Tables \ref{tab:pret} and \ref{tab:drop}) is not shown in these examples.
The examples here assume that some of the $\Phi_l$s and $\Psi_l$s are the same, although there is no such restriction. 
Specifically, the ranges of $\Phi_L$ and $\Phi_f$ plays a role in deciding the regimes of higher layer lengths. 
Figure \ref{fig:ex_designs} shows more evidence of this behavior including some good strategies to choose $\Phi_L$ and $\Phi_f$
based on (the rest of) the criteria. 
The designs in Figure \ref{fig:ex_designs} construct a network of depth $L=5$ (shown in the top row), with the 
second row using $\Phi_5 = \sqrt{\Phi_1}$ and the bottom row using $\Phi_5 = \sqrt[3]{\Phi_1}$.
When $\Phi_5 = \sqrt[3]{\Phi_1}$ the hidden layer lengths have high variance compared to $\Phi_5 = \sqrt{\Phi_1}$. 
This variance is much higher when $d_5$ (the output layer length) is fixed while allowing $\zeta$ to change (middle column in Figure \ref{fig:ex_designs}). 

\noindent {\bf Shiny-based Interactive Webpage for design choices:}
The design procedures from Tables \ref{tab:pret} and \ref{tab:drop} are packaged in an accompanying HTML document, using R-based \textsc{Shiny} interactive webpages,
for both with and without pretraining regimes (i.e., unsupervised data available and the fully-supervised setting).
Apart from the network architectures, this interactive webpage computes the batchsize $B$ and maximum number of gradient iterations $N$ required to setup the multi-layer RSG
(as demanded by the results constructed in this work).
The ranges of the allowable hyper-parameters in these applets (like input and output layer lengths) can be modified by editing the corresponding R Markdown code. 
This web interface can also be found at \href{https://vamsiithapu.shinyapps.io/designchoicessup/}{https://vamsiithapu.shinyapps.io/designchoicessup/} and
\href{https://vamsiithapu.shinyapps.io/designchoicesunsup/}{https://vamsiithapu.shinyapps.io/designchoicesunsup/} for the fully-supervised and pretraining followed by fine-tuning cases respectively.

\subsection{Multi-center Studies: Network design, Resource Allocation and Savings} \label{sec:resource-revisit}

The plots in Figure \ref{fig:ex_designs} and the network lengths in Section \ref{sec:example-designs}
show some evidence of how the designs in Table \ref{tab:pret} and \ref{tab:drop} will be deployed in practice. 
Building upon these results, we now tackle the design choice problem for neuro data considered earlier in Section \ref{sec:exps} and Figure \ref{fig:allexp_neuro}.
We first construct several sets of designs for this dataset -- both transferable and not (see Section \ref{sec:design-revisit}), perform 
learning and then compare their expected gradients trends to assess the usefulness of the proposed design procedures.
Once this is done, we then evaluate the multi-center setting described in Section \ref{sec:resource} where multiple sets of learning models 
that are comparable -- are constructed on different (but related) datasets across multiple sites.
Specifically, for a two-center setting we present resource savings using the proposed design procedures. 


\noindent \paragraph{\bf Setup:} \label{sec:case-study-eval}
For a given set of goodness criteria (similar to those used in Example 2, Section \ref{sec:example-designs}), three different deep networks 
are designed using Tables \ref{tab:pret} and \ref{tab:drop} respectively.
These three designs correspond to a transferable design (i.e., hidden layer lengths are same) and changing dropout rates
Once the designs are computed, each is then trained using $10$ different stopping iterations (where the stopping iteration is sampled uniformly from the last $50$ iterations). 
This leads to $20$ different networks for one set of goodness criteria -- a total of $5$ different sets of goodness criteria are used.
Note that all these $100$ learned networks are trained on the {\it same} neuro dataset.
The goal then is to quantify the discrepancy within and across these $5$ sets of deep networks.
Clearly, for a fixed goodness criteria, we expect the designs to achieve the same level of empirical expected gradients and similar generalization.
On the other hand, we expect the empirical parameter convergence to show some interesting trends as the goodness criteria vary.


\subsubsection{Comparing designs with a {\it given} goodness criteria} 
\label{sec:case-study-discrim}

The simplest choice for a direct discrepancy measure comparing two models is to compute the test set error (generalization), 
or some form of a stability measure as reported in \cite{hardt2015train}.
The design procedures in Tables \ref{tab:pret} and \ref{tab:drop} are entirely motivated by parameter convergence, which is neither quantified by the test set error, nor the empirical stability. 
Also the empirical stability computes the difference in the risk computed on two datasets that differ at most in one instance, 
whereas we are interested in comparing networks which learn data generated from the same underlying distribution.
Hence we use the following measures.
\begin{itemize} [leftmargin=1em]
\item[] {\bf (D1)} the difference in the learning objective (or loss) \\
Since the goal is generalization in the context of parameter convergence, quantifying the absolute difference of the learning objective is a reasonable measure to evaluate
\item[] {\bf (D2)} the norm of the difference of gradients at the stopping iteration \\
Because of the use of stochastic gradients, 
it is sensible to quantify the discrepancy between networks by comparing the parameter neighborhood at the stopping iteration. 
If such neighborhoods are similar between the replicates, 
then the estimates converged to 'similar' regions of the parameter spaces.
The norm of the difference in gradients evaluates this measure which is only applicable for the transferable design where the network structure
(and the number of parameters) is the same across designs.
\end{itemize}

For certain networks, computing {\bf (D2)} is cheaper than {\bf (D1)}. 
As pointed out earlier in the results from Sections \ref{sec:single}--\ref{sec:multidrop}, 
any lower bound on $N$, will lead to a lower bound on the training time, which then leads to a sense of generalization achievable such that overfitting is avoided.
This is supported by \cite{hardt2015train}, and other empirical studies that have exhaustively studied stochastic gradients in deep networks \citep{bottou2010large, dauphin2014identifying}. 
Hence, these discrepancy measures will be evaluated against the gradient iteration count $N$.


\subsubsection{Results} \label{sec:case-study-results}

\begin{figure}[!h]\centering
\subfloat[]{\includegraphics[width=47mm]{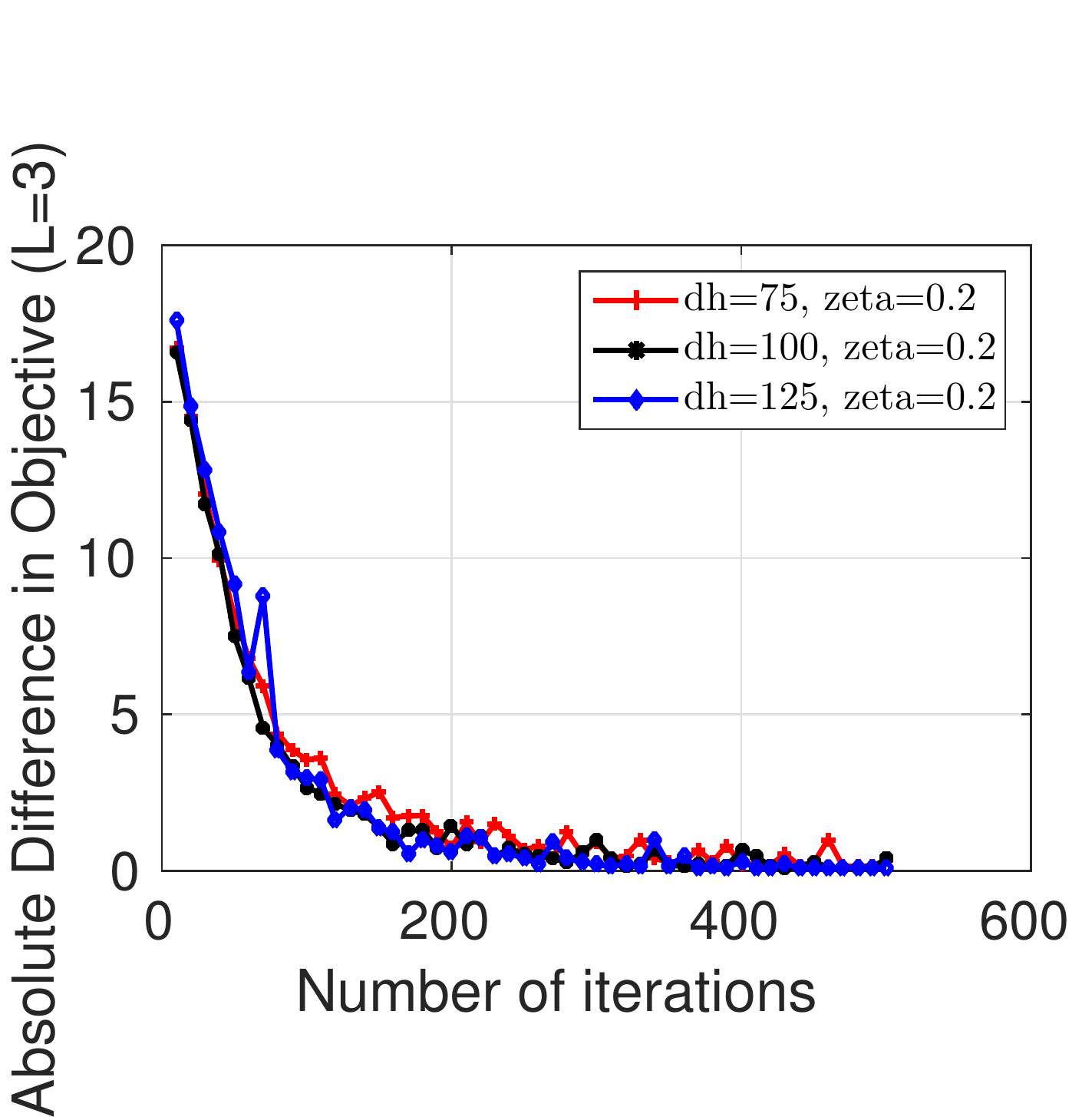}} \quad 
\subfloat[]{\includegraphics[width=47mm]{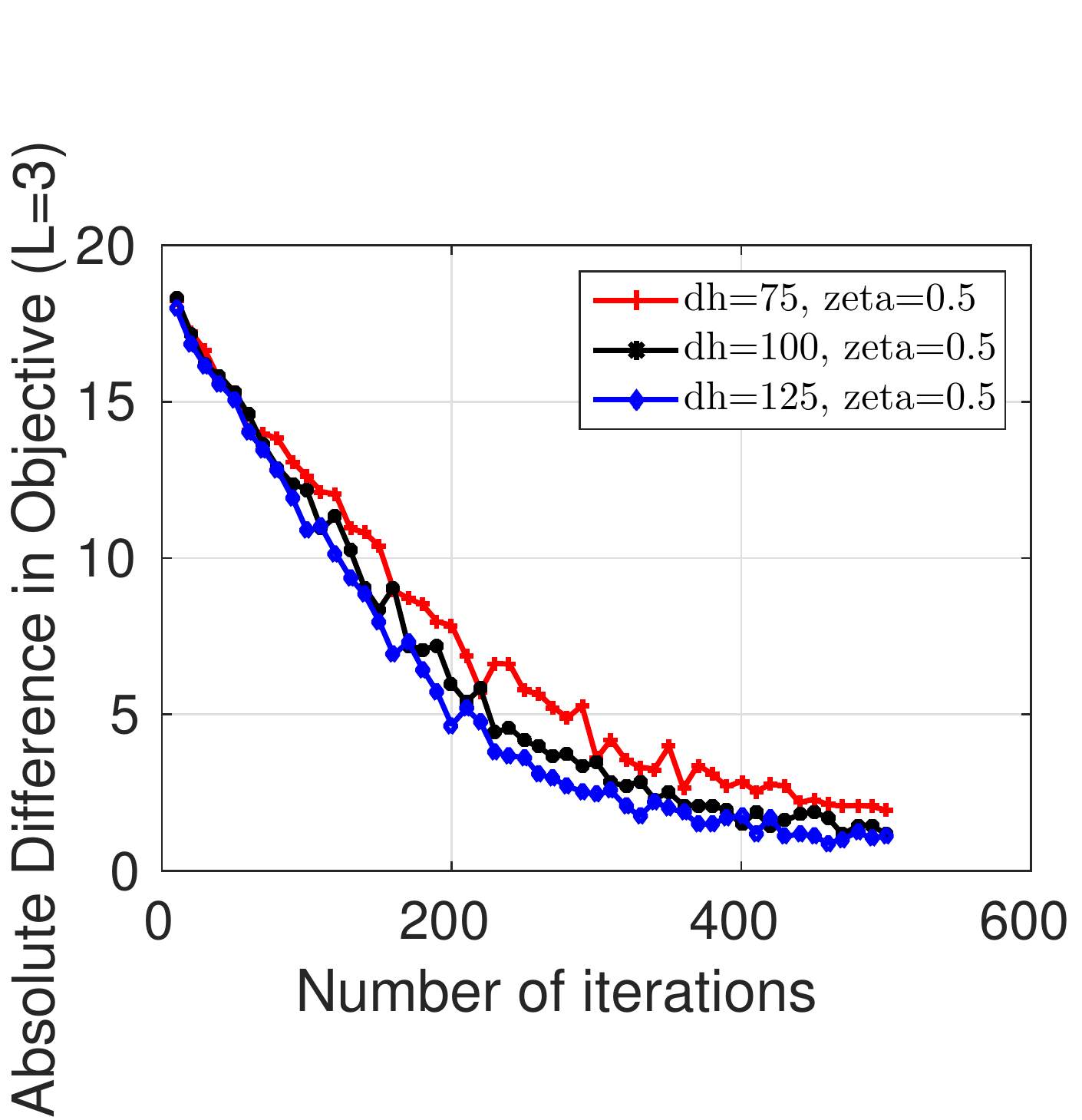}} \quad
\subfloat[]{\includegraphics[width=47mm]{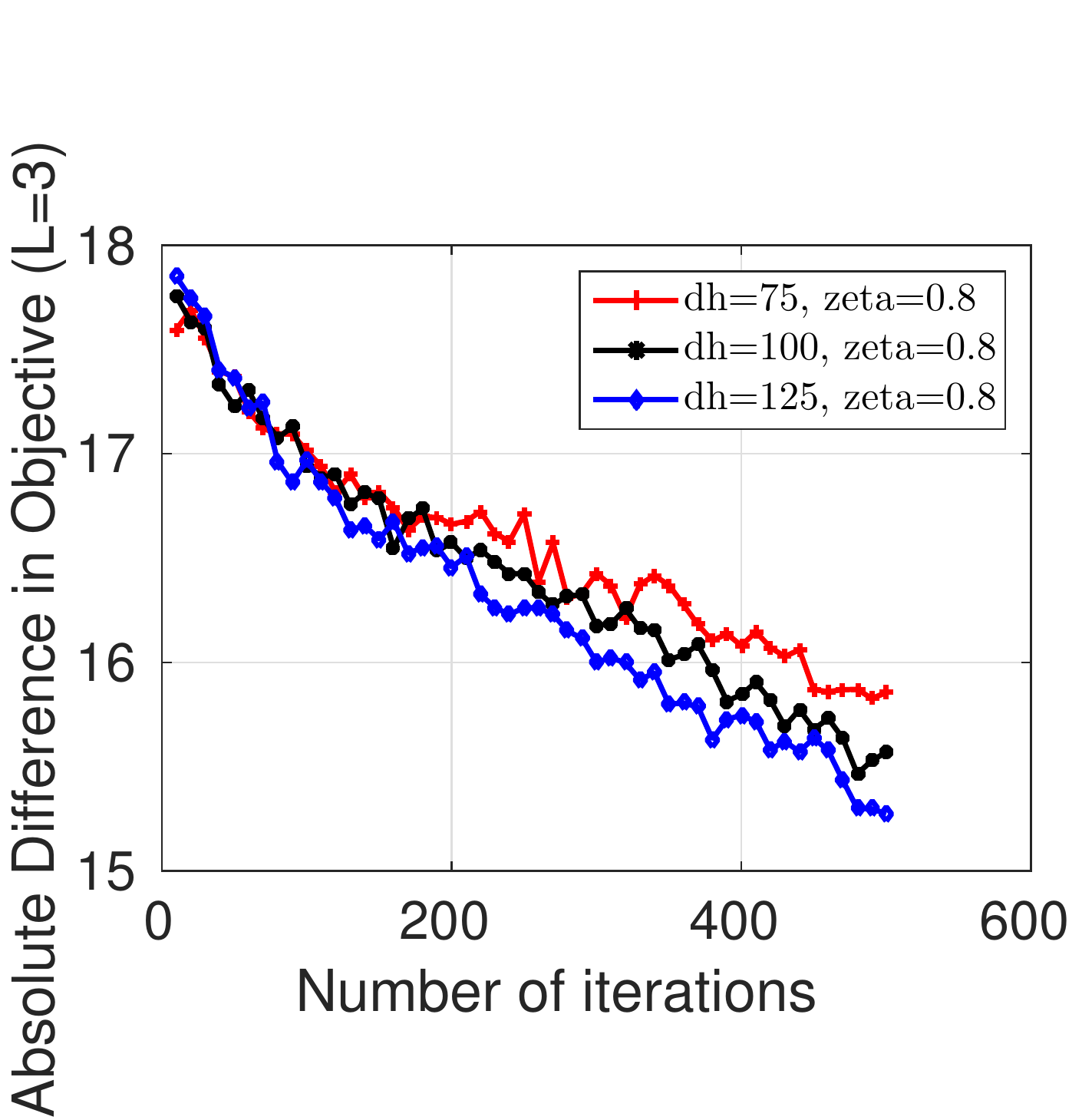}} \\
\subfloat[]{\includegraphics[width=47mm]{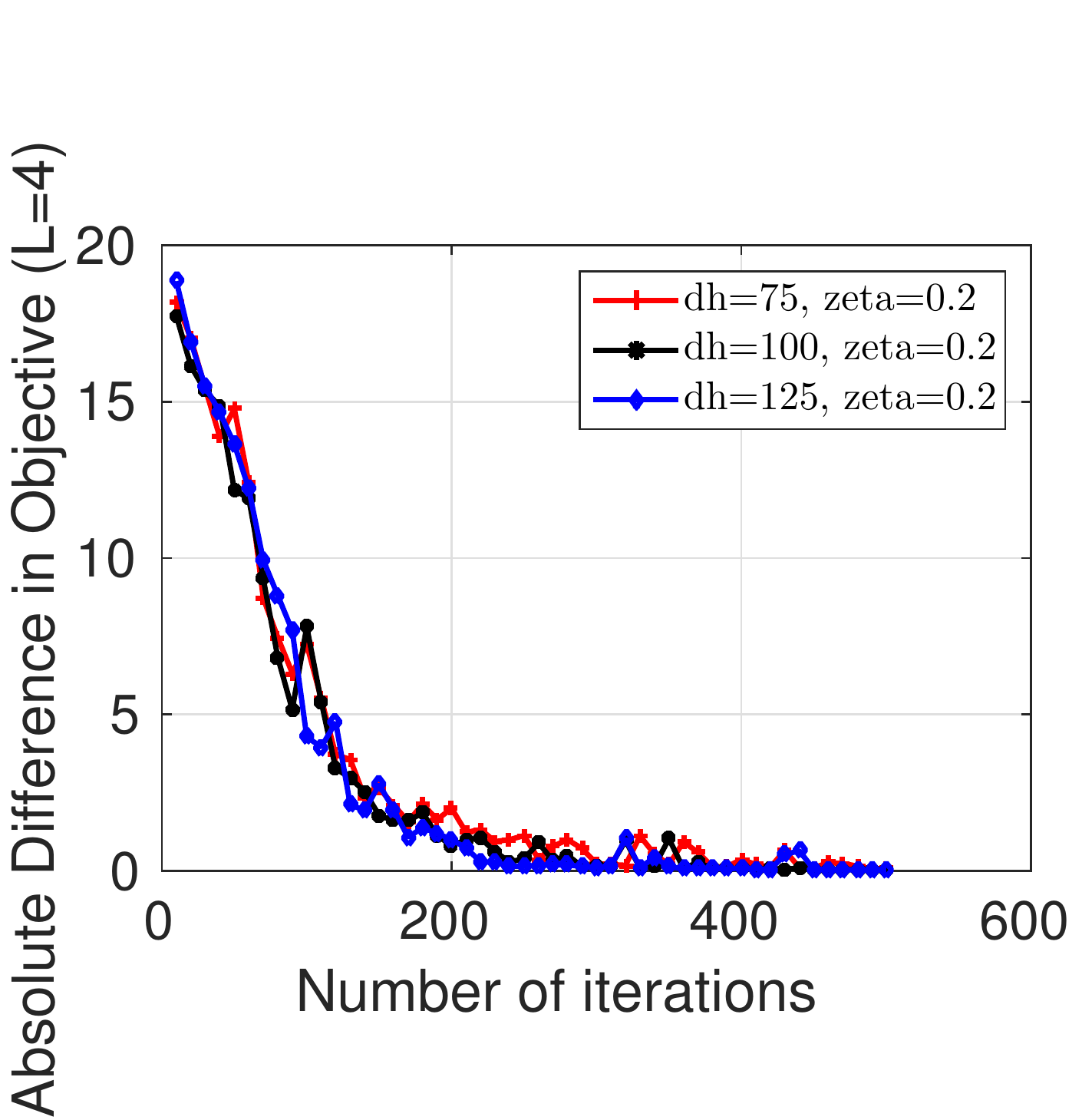}} \quad 
\subfloat[]{\includegraphics[width=47mm]{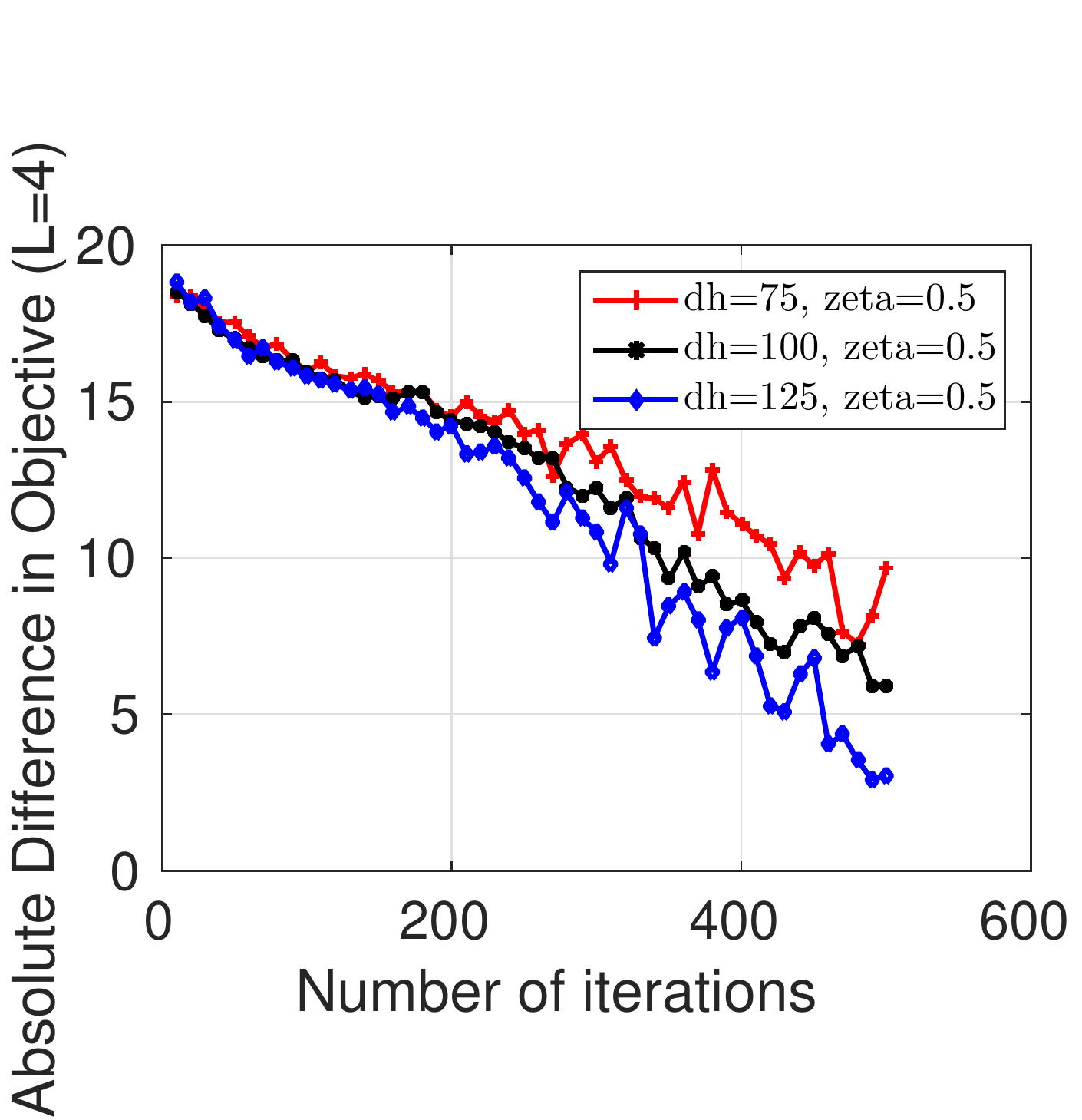}} \quad  
\subfloat[]{\includegraphics[width=47mm]{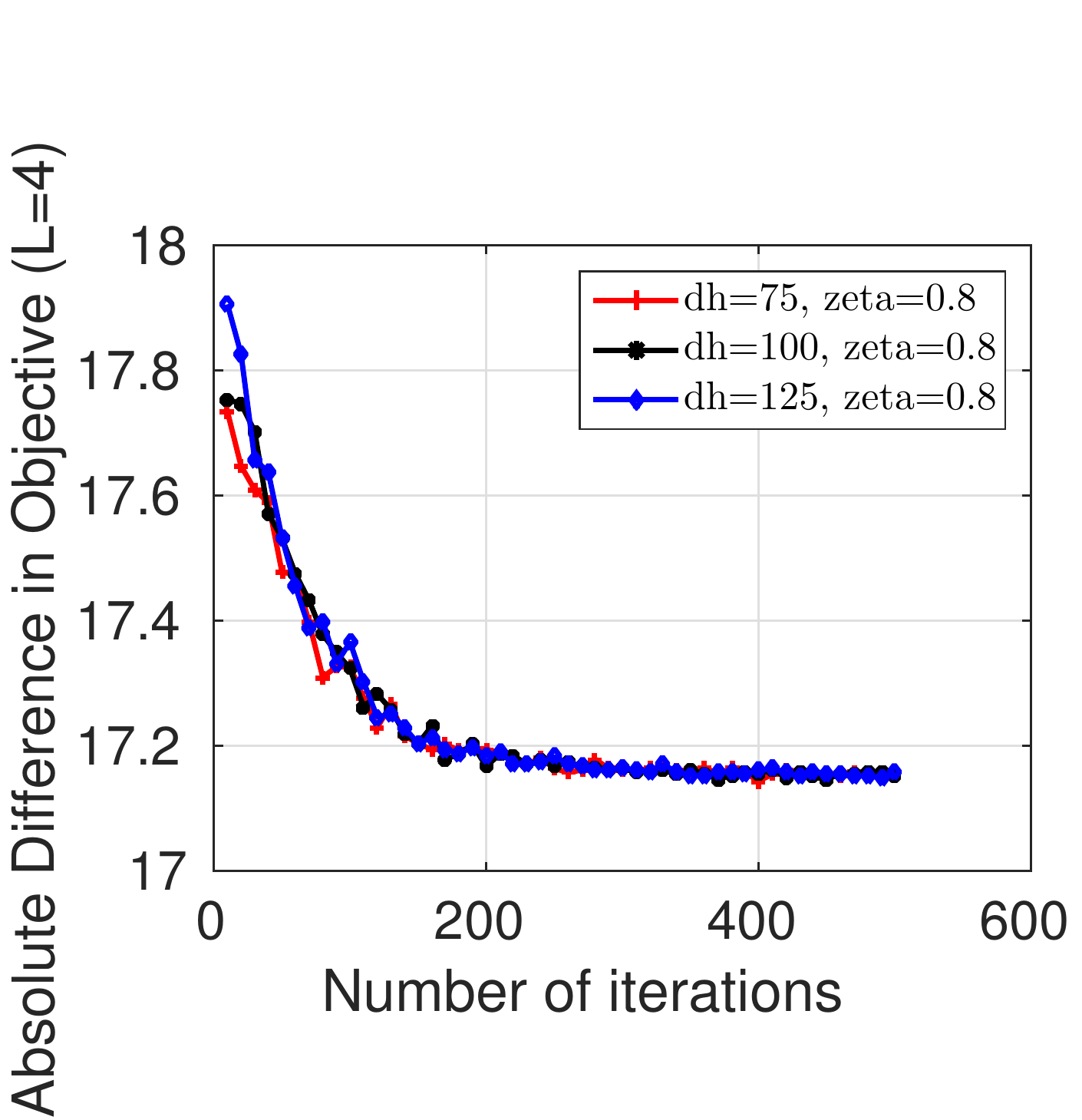}} 
\caption{\footnotesize \label{fig:objdiff_same} {\bf Difference in objective (Transferable design)} Difference in learning objective for depth $3$ (a,b,c) and depth $4$ (d,e,f) networks.
Columns represent different $d_l$s and $\zeta$s. The stopping iterations is chosen randomly from the last $\min(N,15)$ iterations.} 
\end{figure}

\begin{figure}[!h]\centering
\subfloat[]{\includegraphics[width=47mm]{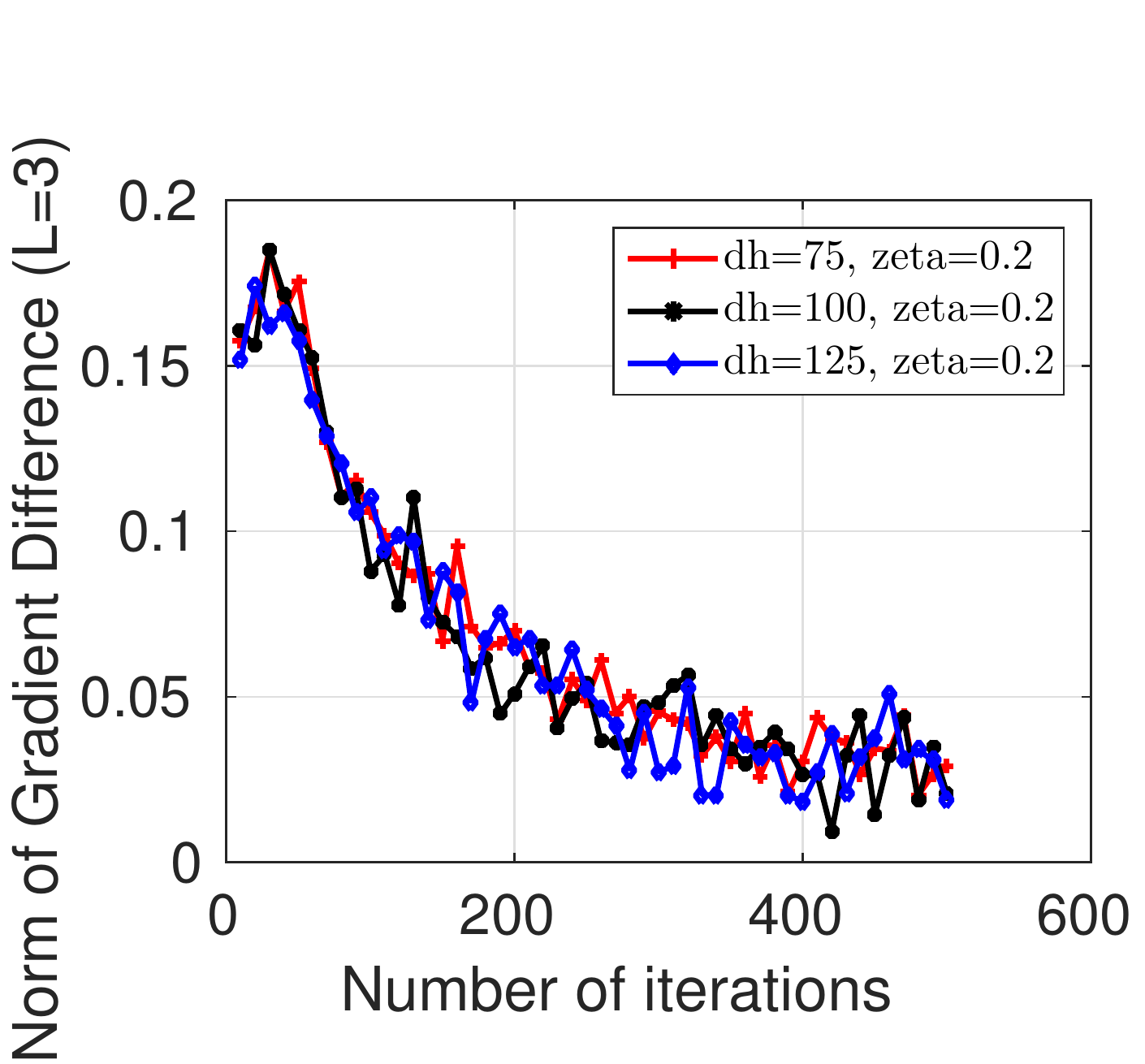}} \quad 
\subfloat[]{\includegraphics[width=47mm]{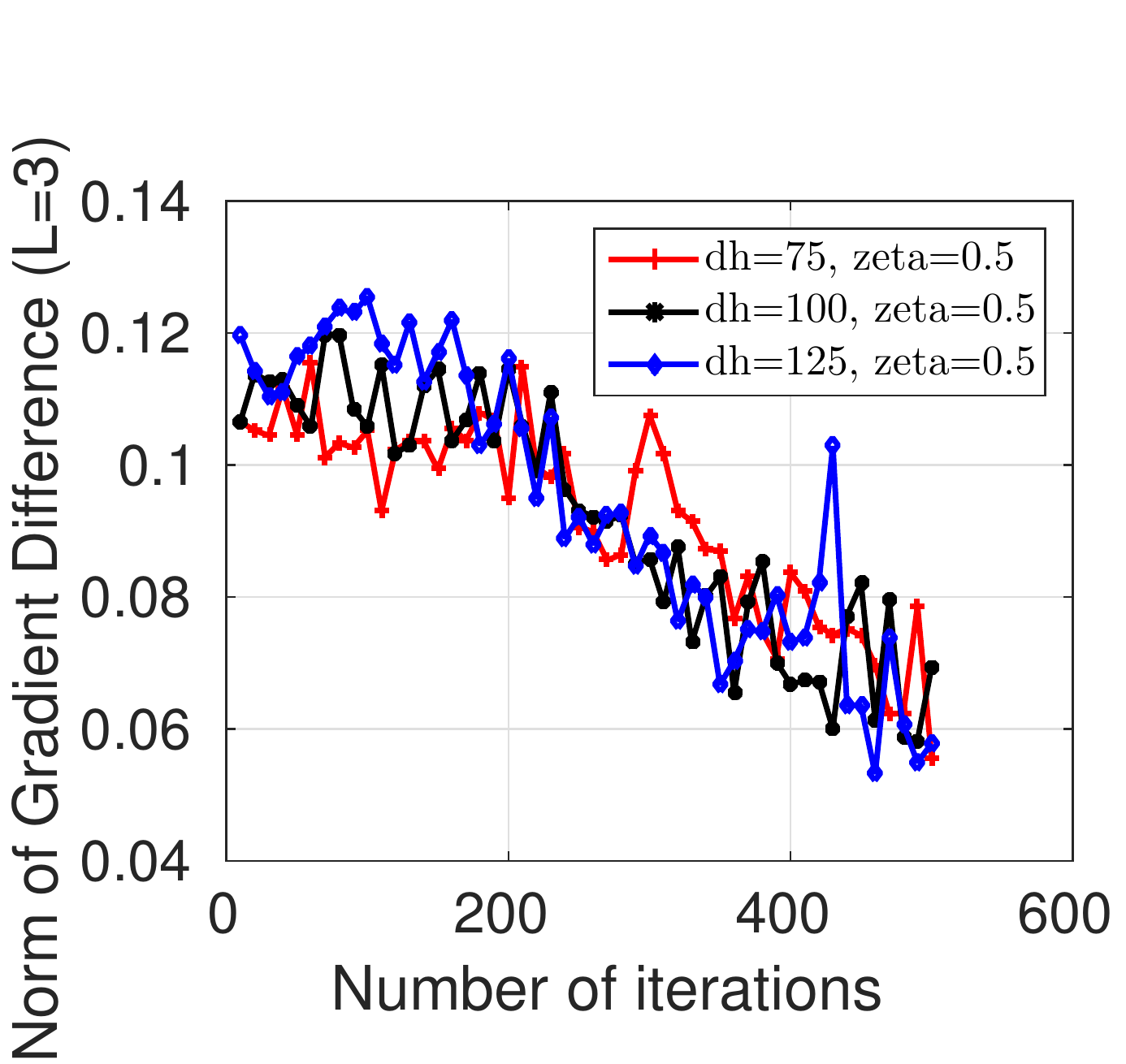}} \quad
\subfloat[]{\includegraphics[width=47mm]{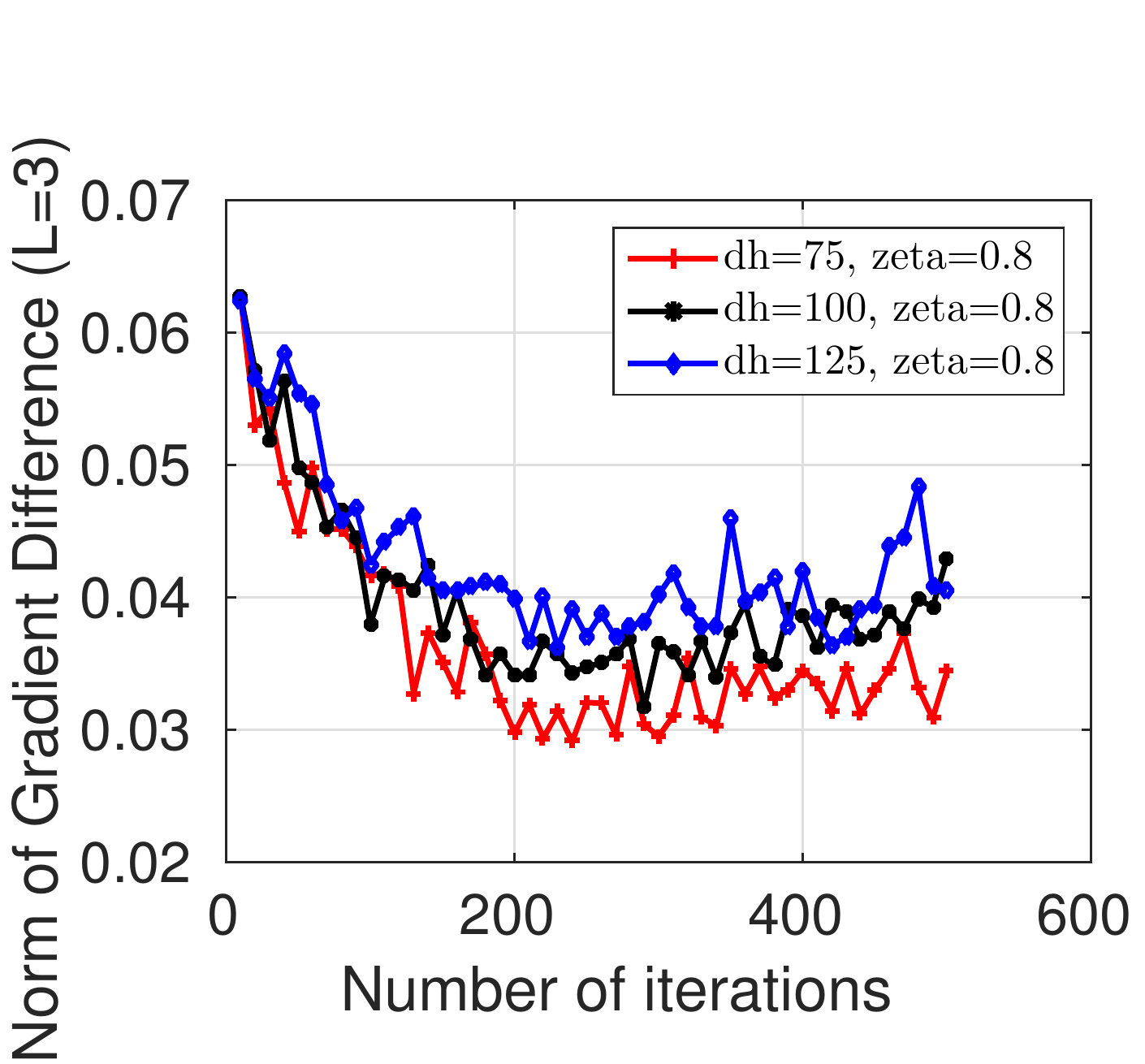}} \\
\subfloat[]{\includegraphics[width=47mm]{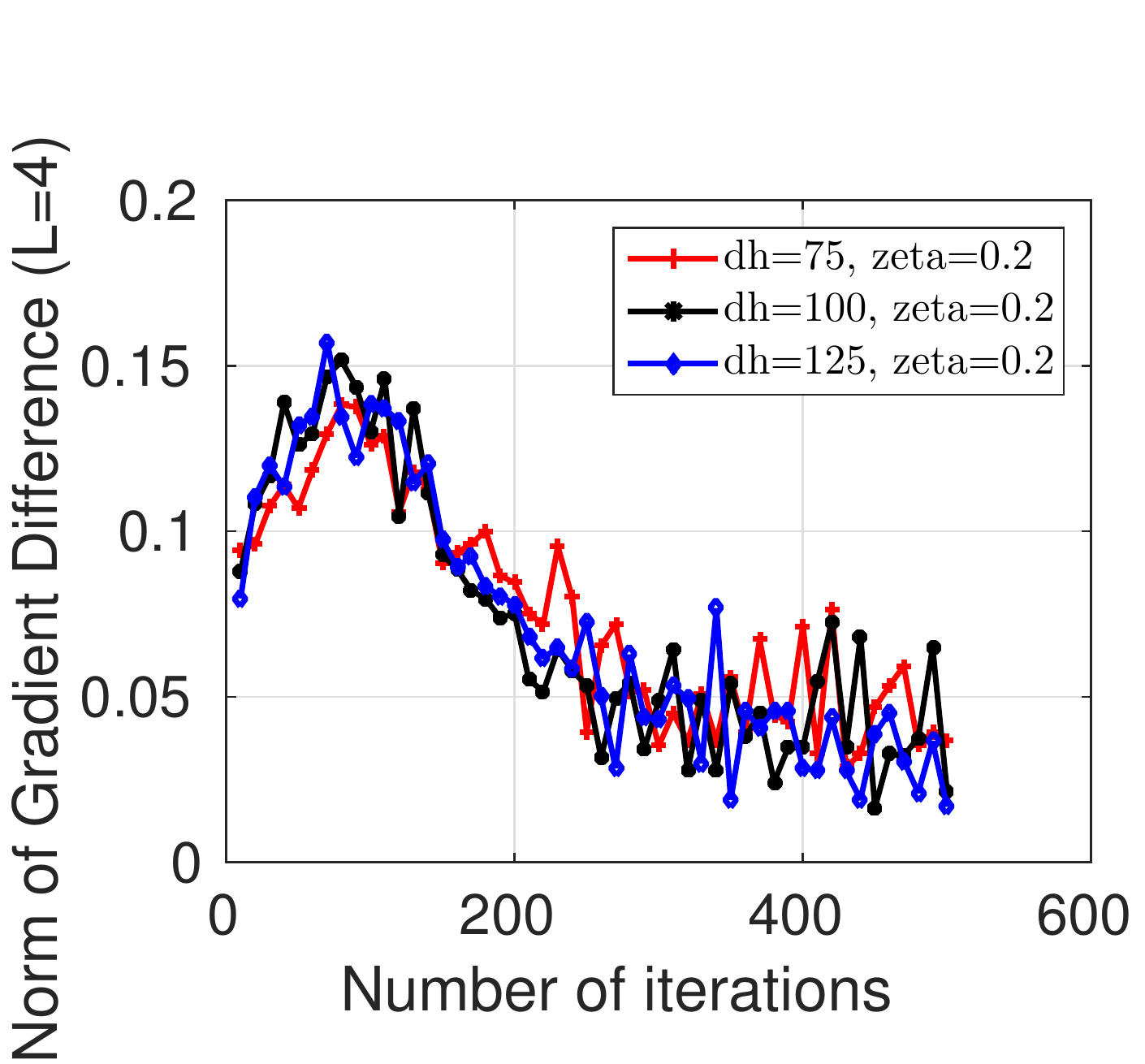}} \quad 
\subfloat[]{\includegraphics[width=47mm]{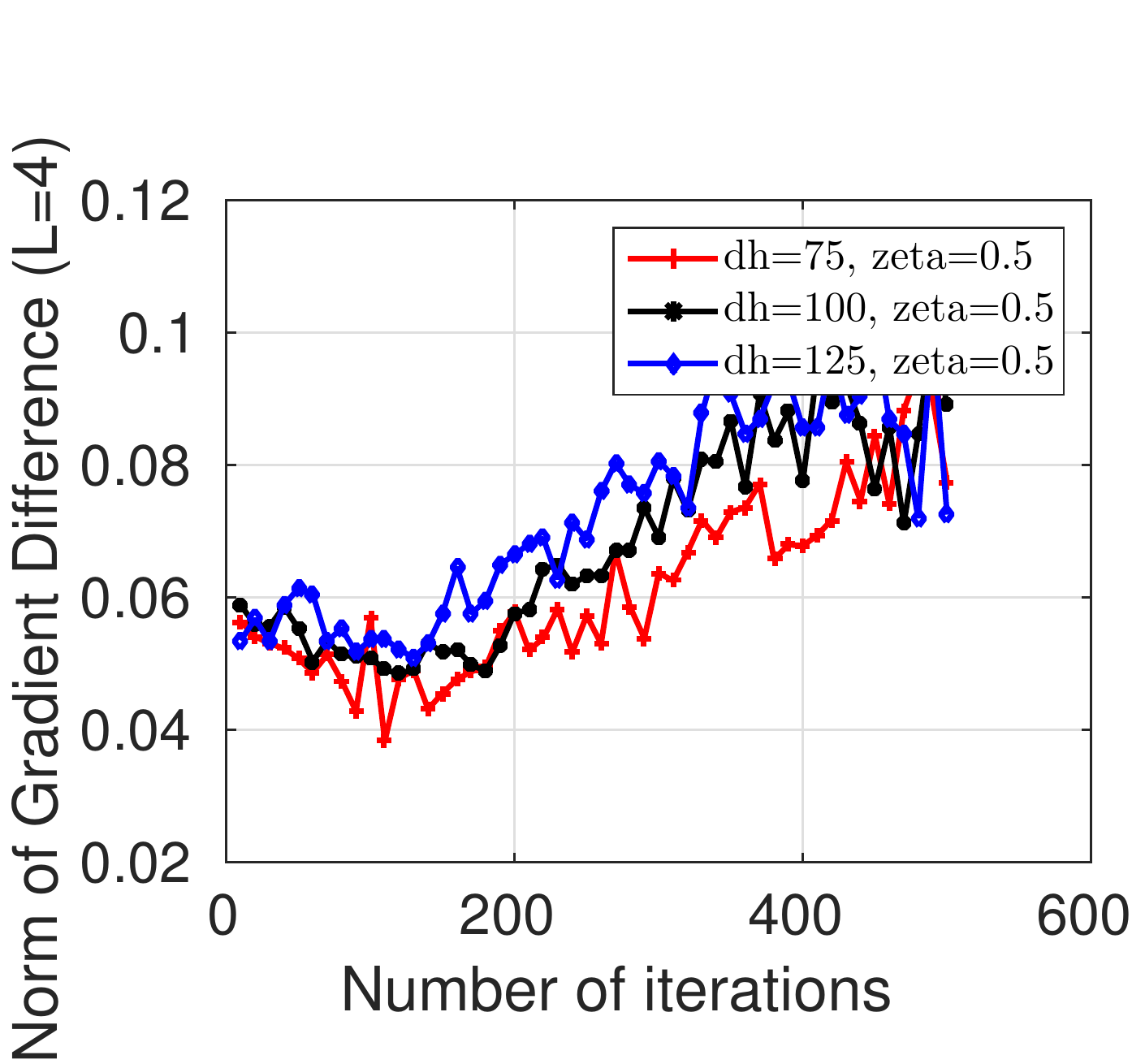}} \quad  
\subfloat[]{\includegraphics[width=47mm]{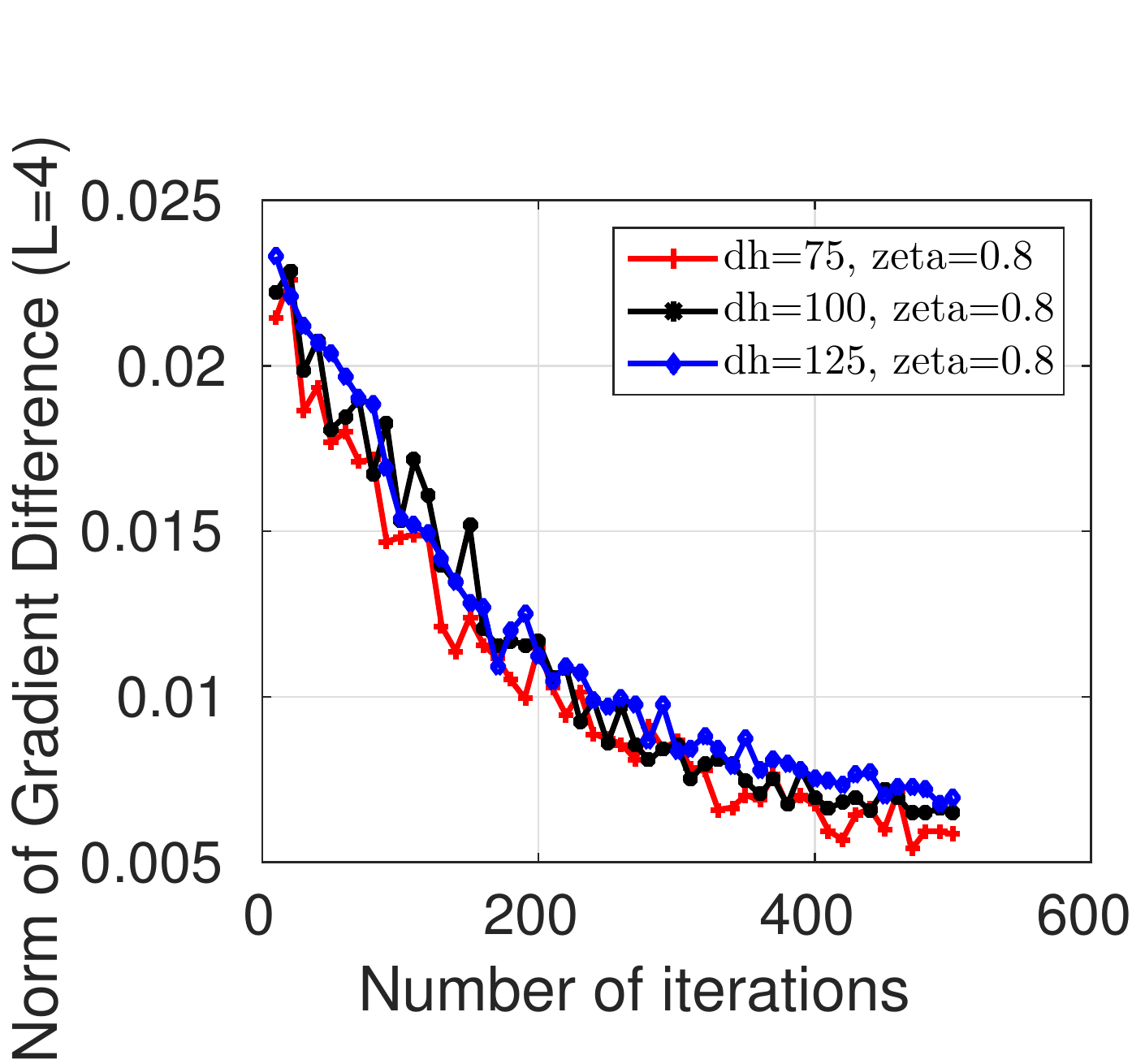}}
\caption{\footnotesize \label{fig:graddiff_same} {\bf Difference in Gradients (Transferable design)} The norm of the difference in the gradients computed at the stopping iteration 
(chosen randomly from the last $\min(N,15)$ iterations) for depth $3$ (a,b,c) and depth $4$ (d,e,f) networks. Columns represent different $d_l$s and $\zeta$s.} 
\end{figure}

\begin{figure}[!h]\centering
\subfloat[]{\includegraphics[width=47mm]{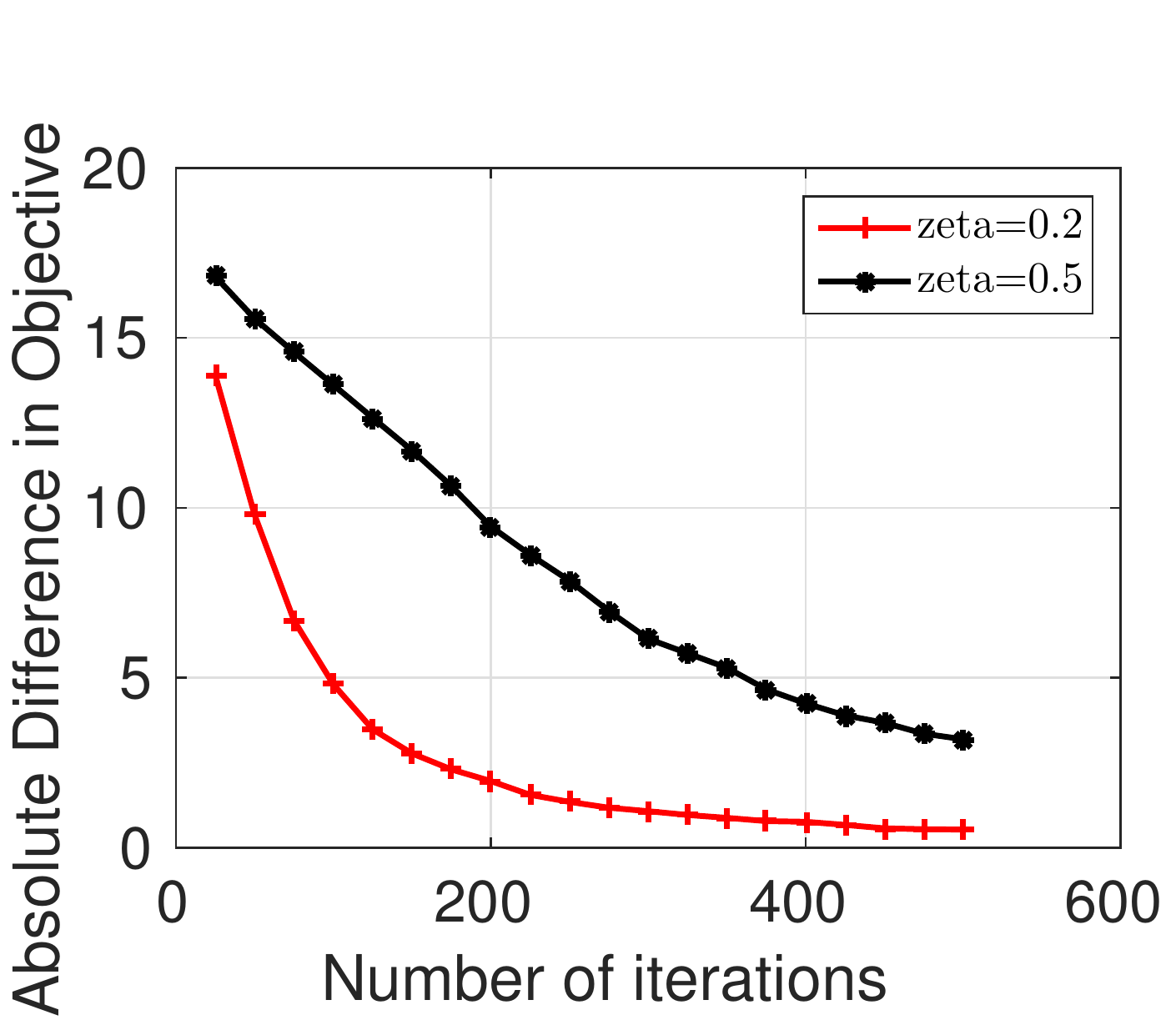}} \quad 
\subfloat[]{\includegraphics[width=47mm]{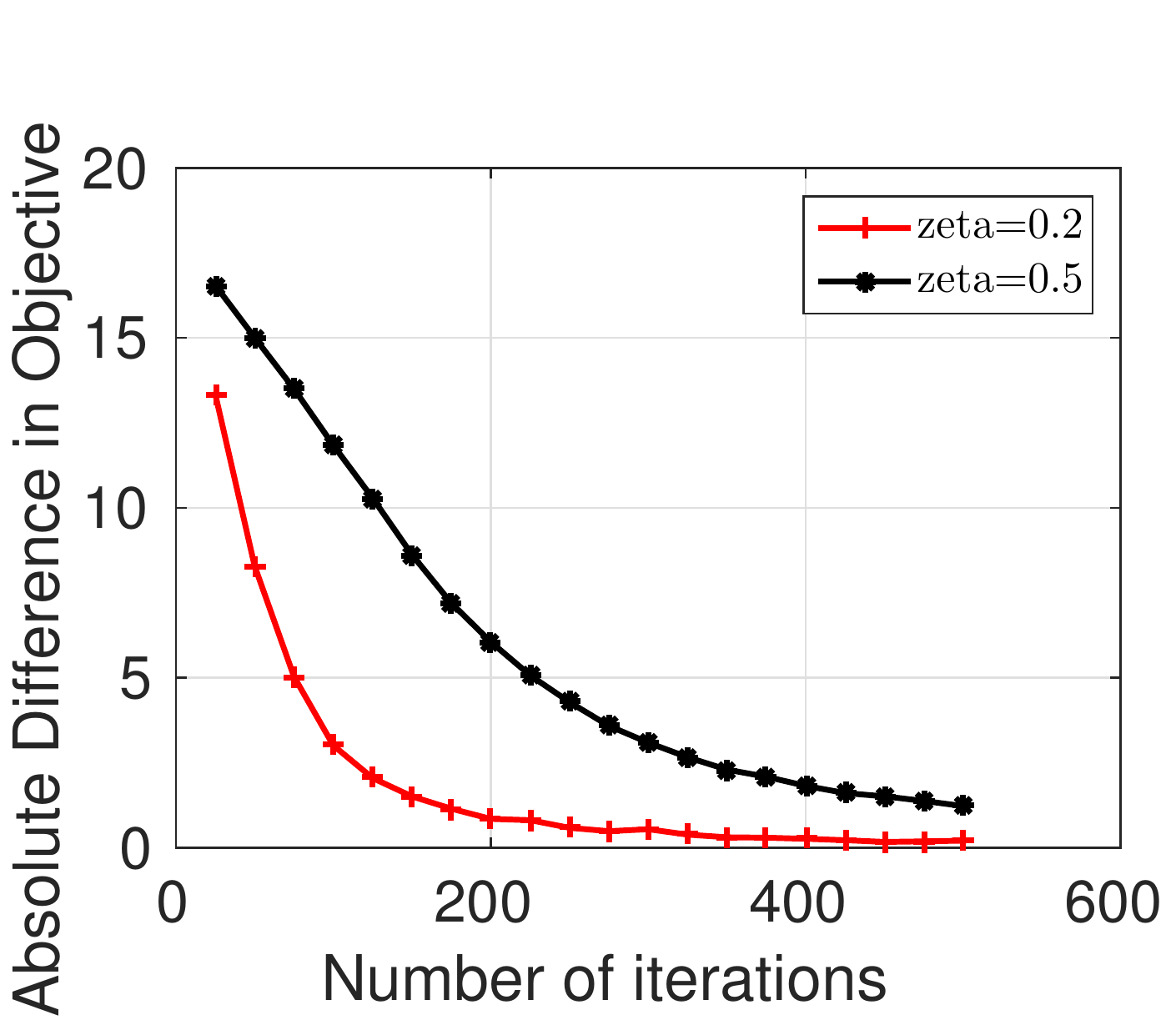}} \quad
\subfloat[]{\includegraphics[width=47mm]{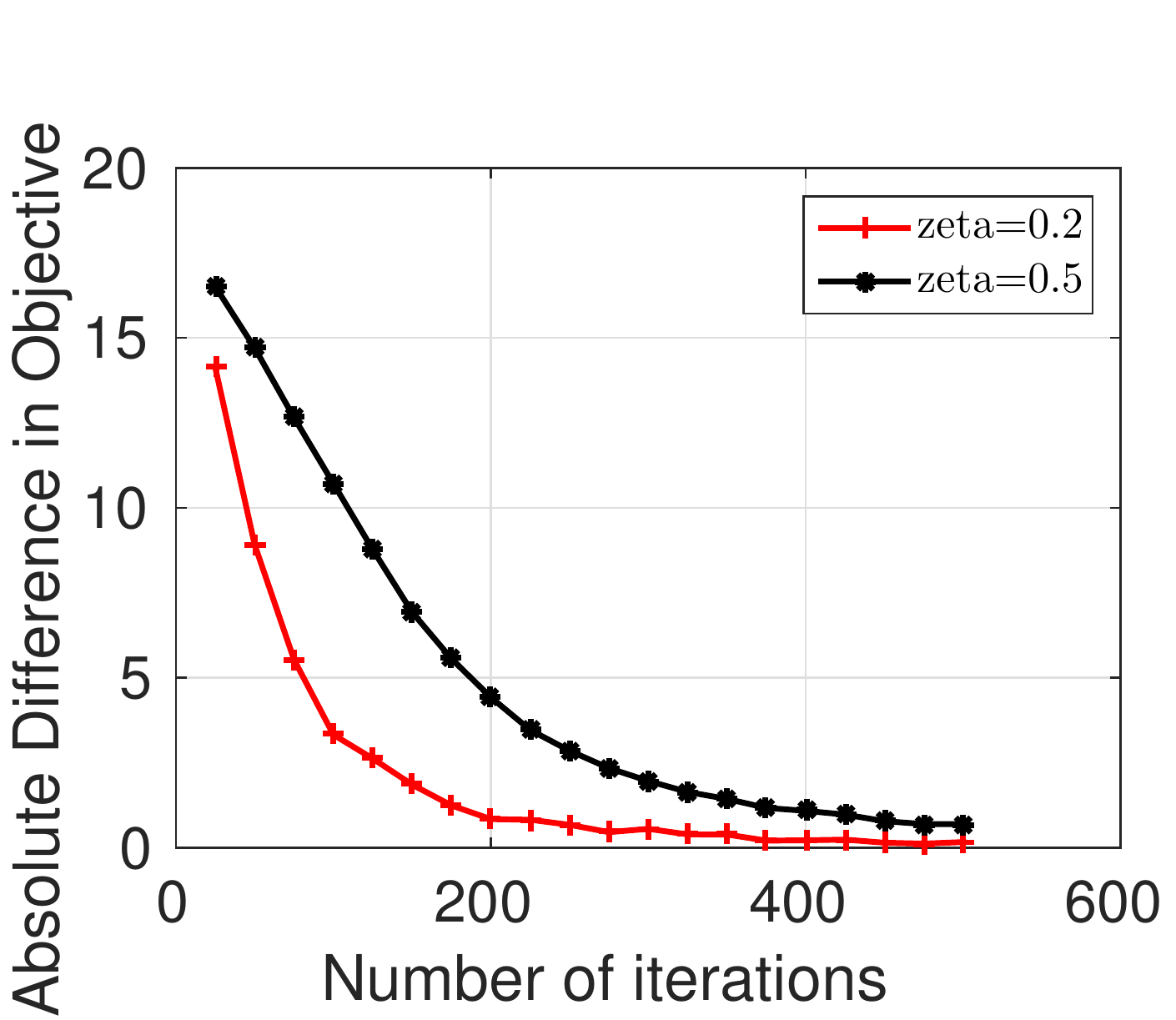}} \\
\subfloat[]{\includegraphics[width=47mm]{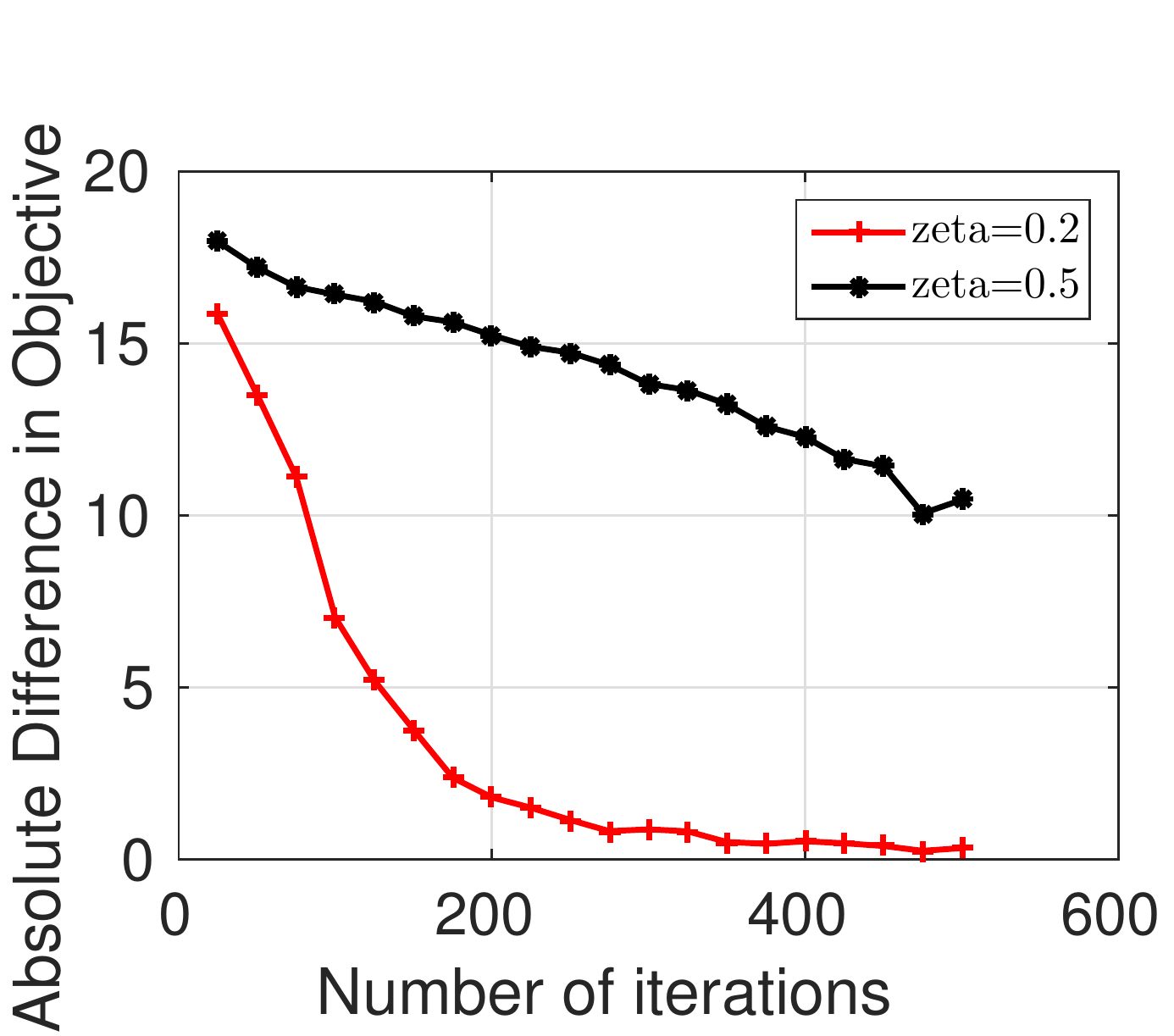}} \quad 
\subfloat[]{\includegraphics[width=47mm]{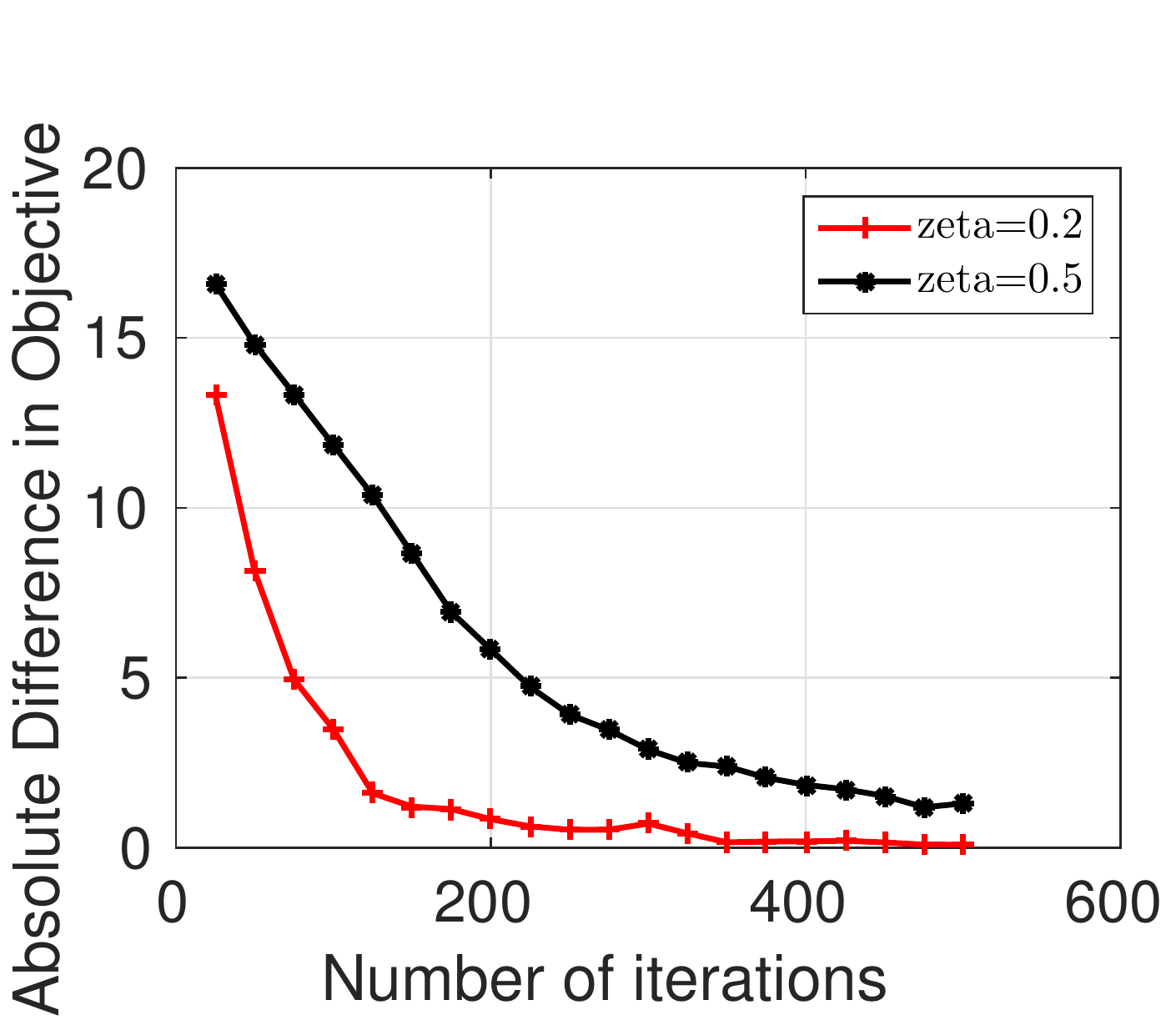}} \quad  
\subfloat[]{\includegraphics[width=47mm]{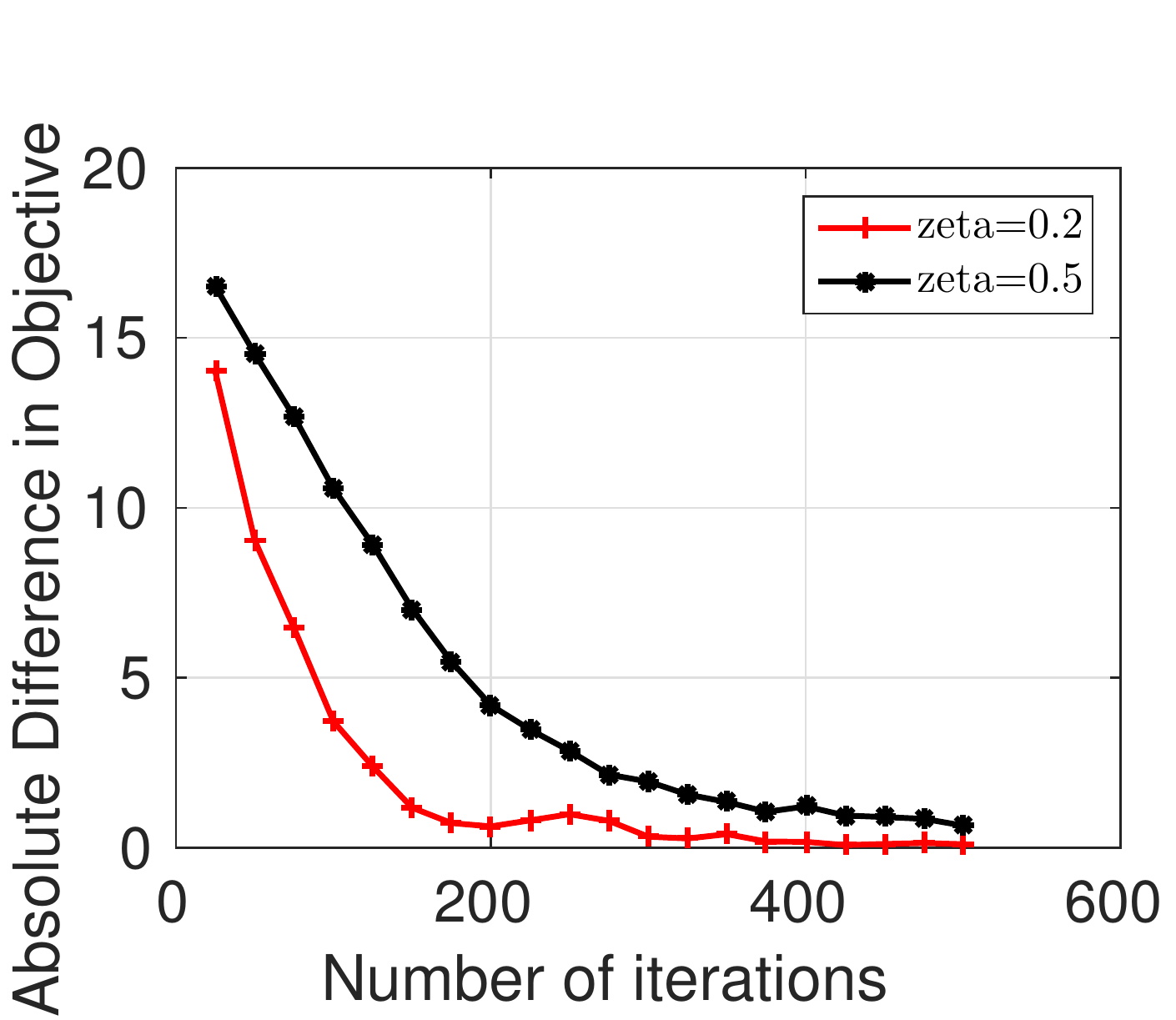}} 
\caption{\footnotesize \label{fig:objdiff_diff} {\bf Difference in objective (Non-transferable design)} Difference in learning objective for depth $3$ (a,b,c) and depth $4$ (d,e,f) networks.
Columns represent different $d_l$s and $\zeta$s. The stopping iterations is chosen randomly from the last $\min(N,15)$ iterations.} 
\end{figure}

Figures \ref{fig:objdiff_same} and \ref{fig:graddiff_same} show the results of transferable designs (same network structure) for depth $3$ and $4$, 
and Figure \ref{fig:objdiff_diff} shows the case where hidden layer lengths are not transferable allowing for fatter and thinner networks for the same depth. 
Unlike the plots in Section \ref{sec:exps}, the $y$-axes in these plots are not scaled and the absolute difference is presented. 

\begin{itemize} [leftmargin=1em]
\item {\bf Same hidden layer designs:} 
Figure \ref{fig:objdiff_same} shows the average difference of the learning objective across the two sets of networks, 
while Figure \ref{fig:graddiff_same} shows the norm of difference in gradients. 
Clearly, these discrepancy measures decrease as the iterations $N$ increase. 
However, as shown by Figures \ref{fig:objdiff_same}(c,f) compared to Figures \ref{fig:objdiff_same}(a,b,d,e), 
the objective at stopping iterations are different whenever the dropout rate is small (i.e., large $\zeta$) -- an indicator of overfitting.
Larger depth networks (bottom row versus the top row in Figure \ref{fig:objdiff_same}) seem to have a higher disagreement among the estimates learned, 
which is expected because the iterations needs to be increased as network depth increases. 
The norm of difference in gradients shown in Figure \ref{fig:graddiff_same} seems to decrease as $N$ increases, but saturates for certain designs (Figure \ref{fig:graddiff_same}(e)).
\item {\bf Different hidden layers designs:}
A similar set of trends are observed for non-transferable designs from Figure \ref{fig:objdiff_diff} which permit differences in hidden layer lengths. 
Note that a total of $10$ different sets of networks are designed for each set, each of which is learned with $10$ stopping iterations 
i.e., the comparisons in Figure \ref{fig:objdiff_diff} are between two sets of $100$ learned networks.
The dropout rate still seems to play a key role, with differences between the two sets of networks increasing as the dropout rate decreases (red versus black curves).
As observed earlier, the bottom row with ($L=4$) has larger discrepancy measures compared to the top row ($L=3$).
\end{itemize}
Although for the networks used in Figures \ref{fig:objdiff_same}--\ref{fig:objdiff_diff} the dropout rate and the output length are fixed, 
a similar set of trends can be produced any other alternate designs from Tables \ref{tab:pret} and \ref{tab:drop}.


\subsection{Two-center resource savings} \label{sec:2centersave}

\begin{figure}[!h]\centering
\includegraphics[width=80mm]{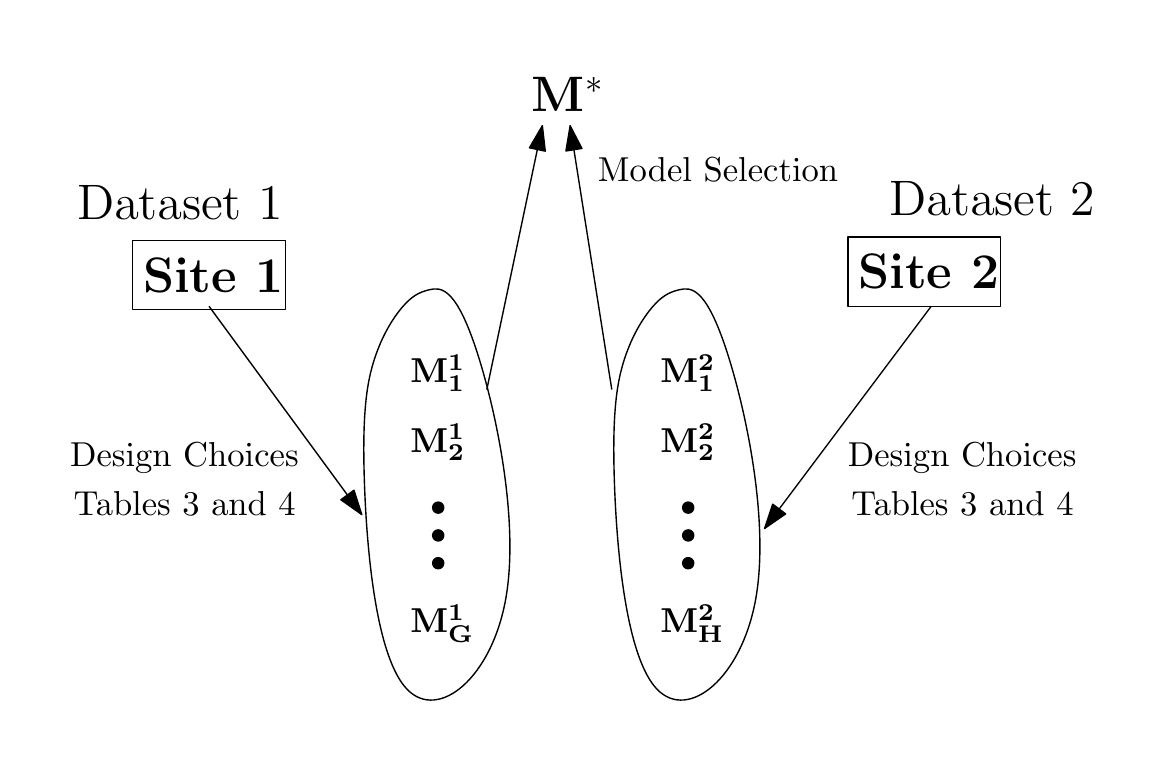} 
\caption{\footnotesize \label{fig:resource} {\bf Two-center learning} An example setup of learning the best model from datasets collected at two data acquisition sites.
  The datasets may have some covariate shift. Models are learned independently at each center/site and the best one may be chosen among them (as in meta-analysis).} 
\end{figure}

Building upon the validation experiments in Sections \ref{sec:case-study-discrim} and \ref{sec:case-study-results}, 
we now look at the two center scenario where the goal is to design multiple {\it comparable} models for datasets coming in from two sites 
-- see the discussion from Section \ref{sec:resource-revisit}.
Figure \ref{fig:resource} shows the two center scenario.
Whenever the two set of models $M^1_1,\ldots,M^1_G$ and $M^2_1,\ldots,M^2_H$ are designed using Table \ref{tab:pret} and \ref{tab:drop}, 
the corresponding discussion and results from Section \ref{sec:case-study-results} ensure that they are comparable.


To see the resource benefits assume that the models $M^1_1,\ldots,M^1_G$ and $M^2_1,\ldots,M^2_H$ are rich with large hidden layer lengths and large depth.
Tables \ref{tab:pret} and \ref{tab:drop} provide the smallest possible designs for the given goodness criteria. 
Specifically, this is either in terms of the size of the network (i.e., number of parameters to be learned) which will influence storage/memory requirements, 
or the minimum number of iterations required which summarizes the cost (and time) allocated to learn the models.
To quantify this resource saving aspect of the design presented here, 
consider the setting where the models are learned on a workstation or a cloud platform. 
Every ``compute hour'' or calculation has a financial cost associated with it. 
Even in the modest setting where, for a given set of goodness criteria, the design choices reduce the number of computations for learning the models by $10\%$, 
and the dollar cost is say $\$1$ per hour, the computational savings can be substantial. 
For example, if $G = H = 50$, the financial resources need to be allocated to learn the models is reduced by at least a factor of $90$. 
Although a more systematic empirical study is required to further explore this aspect of savings in resource allocation, 
the above hypothetical example provides some justification for the applicability of Tables \ref{tab:pret} and \ref{tab:drop} from the monetary perspective.




\subsection{The open questions -- Revisited}
\label{sec:openques}

Apart from the design choice, several open questions were listed in Section \ref{sec:intro}. 
Having constructed and validated the technical results, we summarize the list again while pointing out the corresponding results and discussions from the earlier sections, 
including the aspects which fall outside the scope of the current paper. 

\begin{itemize} [itemsep=0.5mm]
\item[{\bf (Q1)}] The technical results in Section \ref{sec:single}--\ref{sec:multidrop} and the discussion in Section \ref{sec:disc} cumulatively tackles 
  this set of questions.
\item[{\bf (Q2)}] Fully supervised dropout does compensate for unsupervised pretraining and lack of unlabelled data. 
See Theorem \ref{thm:pretvsdrop-gen} and the corresponding remarks, the discussion in Section \ref{sec:goodpretrain}, \ref{sec:pret+drop} and \ref{sec:droprates},
Corollary \ref{thm:bestdrop}, the evaluations in Section \ref{sec:droppretexps} and the discussion from Section \ref{sec:choice-goodness-2}.
\item[{\bf (Q3)}] The design strategies from Tables \ref{tab:pret} and \ref{tab:drop}, the corresponding discussion in Section \ref{sec:design-revisit}, 
and Corollaries \ref{thm:tall-short} and \ref{thm:tall-short-drop} addresses the issue of choosing the best architecture (or network design) from a family of networks. 
\item[{\bf (Q4)}] Sections \ref{sec:da-net-conv}, \ref{sec:goodpretrain}, \ref{sec:pret+drop} and \ref{sec:only+drop} 
explore the situations where the network may not be able to learn the underlying concepts, while also pointing out the regimes where the technical results in this work tend to be loose. 
\item[{\bf (Q5)}] This aspect has not yet been answered by the analysis in this paper, and will be addressed in future work. 
\item[{\bf (Q6)}] The discussion in Sections \ref{sec:compare} and \ref{sec:comparedrop}, 
Corollaries \ref{thm:equiv-pretrain}, \ref{thm:equiv-pretrain-drop}, \ref{thm:tall-short} and \ref{thm:tall-short-drop} address the issue of comparing arbitrary networks. 
The design procedures in Section \ref{sec:design-revisit} further explore these results.
\item[{\bf (Q7)}] Theorems \ref{thm:conv1nn-mono}, \ref{thm:expgradda}, and \ref{thm:expgradmulnn} and Corollary \ref{thm:conv1nn-drop} summarizes the sample size estimates. 
The corresponding remarks and the design in Section \ref{sec:disc} provide insights into small sample deep learning. A more in depth 
study is still needed here.
\end{itemize}


\section{Conclusion}
\label{sec:conc}

Understanding the relationship and influence of network structure and learning parameters on the parameter estimation, 
and eventually, the generalization performance is vital in designing new and more specialized deep networks.
This line of work also facilitates a better understanding of deep learning architectures in general.
In this work, we studied the behavior of stochastic gradients, and their role in parameter convergence as a function of network structure and dataset. 
We showed that there is non-trivial dependence of the architecture and input data distribution on parameter estimation, and we studied this relationship. 
We provided decay bounds of gradient norms as the depth and the lengths of layers in multi-layer networks including unsupervised pretraining and fully supervised versions. 
Overall a general framework is presented that analyzes these relationships, and the resulting trends, on arbitrary deep networks.
Based on this framework, we derived some interesting observations about deep network learning, several of which have empirical evidence independent of our 
work (but little associated theoretical analysis). 
The results presented provide at least partial answers to some of the open questions related to deep learning theory.
We visualized the trends implied by our results and our experiments provide empirical justification to the derived bounds. 
Finally, we presented systematic design procedures for constructing the network architecture that is guaranteed to achieve a certain level of convergence (and generalization).
These design procedures are evaluated on a medical imaging dataset where some of the issues (different sites, data pooling problematic) are serious and yet systematic 
deployment of deep learning methods (perhaps facilitated by some strategies presented in this work) has a potential for impact.


\section{Appendix}\label{sec:app}

The proofs for all the technical results are presented here. 


\subsection{Proof of Theorem 1}\label{sec:proof1} 


This proof for the $1$-layer NN serves as a template for the results corresponding to DA and multi-layer NN.
The proof relies on two inequalities concerning the behaviour of the noisy gradients. We first derive these inequalities. 


\noindent Let $\delta = G(\eta;\W) - \nabla_{\W} f(\W)$, where by definition
\begin{equation} f(\W) = \mathbb{E}_{\eta} \mathcal{L}(\eta;\W)  \quad \quad G(\eta;\W) = \frac{1}{B} \sum_{i=1}^{B} g(\eta;\W) 
\quad \text{with} \quad g(\eta;\W) = \nabla_{\W} \mathcal{L}(\eta;\W) \end{equation}
Recall that in the single layer NN case $\eta = \{\x,\y\} \sim \mathcal{X}$. 
The Lipschitz constant of the activation function $\sigma(\cdot)$ is $\frac{1}{4}$ following the fact that $\{\sigma(\cdot)(1-\sigma(\cdot))\} \leq \frac{1}{4}$.


\noindent The first inequality that we derive computes an upper bound on the variance of noisy gradients $g(\eta;\W)$ i.e., $\mathbb{E}_{\eta} \| g(\eta;\W) - \nabla_{\W} f(\W)\|^2$. 
The second inequality quantifies the Lipschitz continuity of $\nabla_{\W} f(\W)$ using continuity and boundedness properties of $\sigma(\cdot)$. 


\noindent For the first inequality, denote $\delta$ to be the length $d_xd_y$ vector with $\delta_{ij} = g^{ij}(\eta;\W) - \nabla_{\W_{ij}} f(\W)$ where $i=1,\ldots,d_y$ and $j=1,\ldots,d_x$. Hence 
\begin{equation} \label{eq:lossdiff0}
\mathbb{E}_{\eta} \| \delta \|^2 = \mathbb{E}_{\eta} \sum_{ij} |\delta_{ij}|^2 = \sum_{ij} \mathbb{E}_{\eta} \left( g^{ij}(\eta;\W) - \nabla_{\W_{ij}} f(\W) \right)^2 
\end{equation} 
where $g^{ij}(\eta;\W)$ is noisy gradients with respect to $W_{ij}$. We upper bound this by computing the maximum variance of $g^{ij}(\eta;\W)$s. 


\noindent For the $1$-layer NN, $\mathcal{L}(\eta;\W) = \| \y - \sigma(\W\x) \|^2$. Hence 
\begin{equation} \label{eq:gradval1}
g^{ij}(\eta;\W) = \nabla^{ij}_{\W} \mathcal{L}(\eta;\W) = -2(y_i - \sigma_i(\W\x))\sigma_i(\W\x)(1-\sigma_i(\W\x))x_j
\end{equation}
Since $\x \in [0,1]^{d_x}$, $\y \in [0,1]^{d_y}$ and $\sigma(\cdot) \in [0,1]$, we have
\begin{equation} \label{eq:gradval2}
|g^{ij}(\eta;\W)| = 2 |y_i - \sigma_i(\W\x)||\sigma_i(\W\x)(1-\sigma_i(\W\x))||x_j| \leq \frac{1}{2}
\end{equation}


\noindent Hence $\mathbb{E}_{\eta} g^{ij}(\eta;\W) \leq \frac{1}{2}$.
Using the definition of variance and \eqref{eq:gradval2} we then have
\begin{equation} \begin{aligned} \label{eq:gradval3}
\Var_{\eta}(g^{ij}(\eta;\W)) &= \mathbb{E}_{\eta} (g^{ij}(\eta;\W))^2 - (\mathbb{E}_{\eta} g^{ij}(\eta;\W))^2 \\
&\leq \mathbb{E}_{\eta} g^{ij}(\eta;\W) (\frac{1}{2} - \mathbb{E}_{\eta} g^{ij}(\eta;\W)) \leq \frac{1}{16}
\end{aligned} \end{equation}
which follows from the fact that the global maxima of $t(c-t)$ is $\frac{c^2}{4}$.


\noindent Using this in \eqref{eq:lossdiff0}, we get 
\begin{equation} \label{eq:varlipbounds1}
\mathbb{E}_{\eta} \| \delta \|^2 \leq \frac{d_xd_y}{16}
\end{equation}


\noindent We now derive the second inequality required. 
Since $\nabla_{\W} f(\W) = \mathbb{E}_{\eta} \nabla_{\W}\mathcal{L}(\eta;\W)$, we have
\begin{equation} \begin{aligned} \label{eq:graddiff}
\| \nabla_{\W} f(\W) - \nabla_{\W} f(\hat{\W}) \| &\leq \mathbb{E}_{\eta} \| \nabla_{\W}\mathcal{L}(\eta;\W) - \nabla_{\hat{\W}}\mathcal{L}(\eta;\hat{\W}) \| \\
&\leq sup_{\eta} \| \nabla_{\W}\mathcal{L}(\eta;\W) - \nabla_{\hat{\W}}\mathcal{L}(\eta;\hat{\W}) \| 
\end{aligned} \end{equation}


\noindent Consider $|\nabla^{ij}_{\W}\mathcal{L}(\eta;\W) - \nabla^{ij}_{\W}\mathcal{L}(\eta;\hat{\W})|$ where $\nabla^{ij}_{\W}\mathcal{L}(\eta;\W)$ is from \eqref{eq:gradval1}.
Adding subtracting $2(y_i - \sigma_i(\W\x))\sigma_i(\hat{\W}\x)(1-\sigma_i(\hat{\W}\x))x_j$, and using triangle inequality we get
\begin{equation} \begin{aligned} \label{eq:graddiff1}
|\nabla^{ij}_{\W}\mathcal{L}(\eta;\W) - \nabla^{ij}_{\W}\mathcal{L}(\eta;\hat{\W})| &\leq 2| (y_i - \sigma_i(\W\x))\sigma_i(\W\x)(1-\sigma_i(\W\x))x_j \\
& \quad - 2(y_i - \sigma_i(\W\x))\sigma_i(\hat{\W}\x)(1-\sigma_i(\hat{\W}\x))x_j | \\
& \quad + 2| (y_i - \sigma_i(\W\x))\sigma_i(\hat{\W}\x)(1-\sigma_i(\hat{\W}\x))x_j \\
& \quad - (y_i - \sigma_i(\hat{\W}\x))\sigma_i(\hat{\W}\x)(1-\sigma_i(\hat{\W}\x))x_j |
\end{aligned} \end{equation}


\noindent Rearranging the terms within each $|\cdot|$ we have
\begin{equation} \begin{aligned} \label{eq:graddiff2}
|\nabla^{ij}_{\W}\mathcal{L}(\eta;\W) &- \nabla^{ij}_{\W}\mathcal{L}(\eta;\hat{\W})| \\
& \leq 2|x_j||y_i - \sigma_i(\W\x)||\sigma_i(\W\x)(1-\sigma_i(\W\x)) - \sigma_i(\hat{\W}\x)(1-\sigma_i(\hat{\W}\x))| \\
& \quad + 2|\sigma_i(\W\x) - \sigma_i(\hat{\W}\x)||\sigma_i(\W\x)(1-\sigma_i(\W\x))x_j|
\end{aligned} \end{equation}


\noindent Using triangle inequality again and recalling that $x_j$, $y_i$ and $\sigma_i(\cot)$ lie between $0$ and $1$, we have
\begin{equation} \begin{aligned} \label{eq:graddiff3}
|\nabla^{ij}_{\W}\mathcal{L}(\eta;\W) &- \nabla^{ij}_{\W}\mathcal{L}(\eta;\hat{\W})| \\
& \leq 2|\sigma_i(\W\x) - \sigma_i(\hat{\W}\x)| + 2|\sigma_i^2(\W\x) - \sigma_i^2(\hat{\W}\x)| \\
& \quad + 2|\sigma_i(\W\x) - \sigma_i(\hat{\W}\x)||\sigma_i(\W\x)(1-\sigma_i(\W\x))x_j| \\
& \leq 2|\sigma_i(\W\x) - \sigma_i(\hat{\W}\x)| + 2|\sigma_i^2(\W\x) - \sigma_i^2(\hat{\W}\x)| \\
& \quad + \frac{1}{2}|\sigma_i(\W\x) - \sigma_i(\hat{\W}\x)|
\end{aligned} \end{equation}
where in the last step we used the fact that $\sigma_i(\hat{\W}\x)(1-\sigma_i(\hat{\W}\x)) \leq \frac{1}{4}$. 


\noindent Splitting the second term as $|\sigma_i^2(\W\x) - \sigma_i^2(\hat{\W}\x)| = (\sigma_i(\W\x) + \sigma_i(\hat{\W}\x))|\sigma_i(\W\x) - \sigma_i(\hat{\W}\x)|$ we have
\begin{equation} \begin{aligned} \label{eq:graddiff4}
|\nabla^{ij}_{\W}\mathcal{L}(\eta;\W) &- \nabla^{ij}_{\W}\mathcal{L}(\eta;\hat{\W})| \leq \frac{13}{2}|\sigma_i(\W\x) - \sigma_i(\hat{\W}\x)| \\
&\leq \frac{13}{8}|\W_{i\cdot}\x - \hat{\W}_{i\cdot}\x| = \frac{13}{8}|\sum_{j}W_{ij}x_j - \sum_{j}\hat{W}_{ij}x_j| = \frac{13}{8}|W_{ij} - \hat{W}_{ij}|
\end{aligned} \end{equation}
where the second inequality above uses the Lipschitz constant of $\sigma(\cdot)$, and the third one uses Cauchy-Schwartz and the fact that $\W$ and $\hat{\W}$ differ only in $ij^{th}$ entry. 


\noindent Using this in \eqref{eq:graddiff} we have
\begin{equation} \begin{aligned} \label{eq:varlipbounds2}
\| \nabla_{\W} f(\W) - \nabla_{\W} f(\hat{\W}) \| &\leq sup_{\eta} \| \nabla_{\W}\mathcal{L}(\eta;\W) - \nabla_{\hat{\W}}\mathcal{L}(\eta;\hat{\W}) \| \\
&= sup_{\eta} \sqrt{\sum_{ij} | \nabla_{\W_{ij}}\mathcal{L}(\eta;\W) - \nabla_{\hat{\W}_{ij}}\mathcal{L}(\eta;\hat{\W})|} \\
&\leq \sqrt{\sum_{ij} (\frac{13}{8})^2|W_{ij} - \hat{W}_{ij}|^2} = \frac{13}{8}\|\W - \hat{\W}\|
\end{aligned} \end{equation}


\noindent A significant part of the rest of the proof adapts that of Theorem 2.1 in \cite{ghadimi2013stochastic} with several adjustments. 
We first start with $f(\W)$. Using \eqref{eq:varlipbounds2}, which corresponds to the Lipschitz constant for $f(\W)$, we have
\begin{equation} \label{eq:firstineq}
f(\W^{k+1}) \leq f(\W^k) + \langle \nabla_{\W} f(\W^k),\W^{k+1}-\W^k \rangle + \frac{13}{16} \|\W^{k+1}-\W^k\|^2   
\end{equation}
where $\langle\cdot,\cdot\rangle$ denotes inner product. $k$ denotes the iteration index $k = 1,\ldots,N$.


\noindent Substituting for the update $\W^{k+1} \leftarrow \W^k - \gamma^k G(\eta^k;\W^k)$ (where $\gamma^k$ is the stepsize) 
and using $G(\eta^k;\W^k) = \frac{1}{B} \sum_{b=1}^{B} g(\eta^{b,k};\W^k)$, we get
\begin{equation} \label{eq:firstineq1} \begin{aligned}
f(\W^{k+1}) &\leq f(\W^k) - \gamma^k \langle \nabla_{\W} f(\W^k),G(\eta^k;\W^k) \rangle + \frac{13}{16}(\gamma^k)^2 \|G(\eta^k;\W^k)\|^2 \\
&\leq f(\W^k) - \frac{\gamma^k}{B} \sum_{b=1}^{B} \langle \nabla_{\W} f(\W^k), g(\eta^{b,k};\W^k) \rangle + \frac{13}{16B^2}(\gamma^k)^2 \|\sum_{b=1}^{B} g(\eta^{b,k};\W^k)\|^2 
\end{aligned} \end{equation}
where $b=1,\ldots,B$ represents indices within the mini-batch. 
$\eta^{b,k}$ denotes the $\{\x,\y\} \in \mathcal{X}$ sampled for $b^{th}$ noisy gradient computation (within the mini-batch) at $k^{th}$ iteration. 


\noindent Let $\delta^{b,k} := g(\eta^{b,k};\W^k) - \nabla_{\W} f(\W^k)$, and so we denote $\delta^k = \frac{1}{B} \sum_{b=1}^{B} \delta^{b,k} = G(\eta^{k};\W) - \nabla_{\W} f(\W)$. 
The above inequality then reduces to
\begin{equation} \label{eq:firstineq2} \begin{aligned}
f(\W^{k+1}) &\leq f(\W^k) - \gamma^k \|\nabla_{\W} f(\W^k)\|^2 - \frac{\gamma^k}{B} \sum_{b=1}^{B} \langle \nabla_{\W} f(\W^k),\delta^{b,k} \rangle \\
& \quad + \frac{13}{16B^2}(\gamma^k)^2 \|B\nabla_{\W} f(\W) + \sum_{b=1}^{B} \delta^{b,k}\|^2 \\
f(\W^{k+1}) &\leq f(\W^k) - \gamma^k \|\nabla_{\W} f(\W^k)\|^2 - \frac{\gamma^k}{B} \sum_{b=1}^{B} \langle \nabla_{\W} f(\W^k),\delta^{b,k} \rangle \\
& \quad + \frac{13}{16B^2}(\gamma^k)^2 \biggl( B^2\|\nabla_{\W} f(\W)\|^2 + 2B\sum_{b=1}^{B} \langle \nabla_{\W} f(\W^k),\delta^{b,k} \rangle + \|\sum_{b=1}^{B} \delta^{b,k}\|^2 \biggr) \\
f(\W^{k+1}) &\leq f(\W^k) - \left( \gamma^k - \frac{13}{16}(\gamma^k)^2 \right) \|\nabla_{\W} f(\W^k)\|^2 \\
& \quad - \frac{1}{B}\biggl( \gamma^k - \frac{13}{8}(\gamma^k)^2 \biggr) \sum_{b=1}^{B} \langle \nabla_{\W} f(\W^k),\delta^{b,k} \rangle + \frac{13}{16B^2} (\gamma^k)^2 \|\sum_{b=1}^{B} \delta^{b,k}\|^2 \\
\end{aligned} \end{equation}


\noindent Adding up the above inequalities over the iterations $k = 1,\ldots,N$, we get 
\begin{equation} \begin{aligned} \label{eq:sumineq0}
\sum_{k=1}^{N} \left( \gamma^k - \frac{13}{16}(\gamma^k)^2 \right) \|\nabla_{\W} f(\W^k)\|^2 &\leq f(\W^1) - f(\W^{N+1}) \\
- \sum_{k=1}^{N} \frac{1}{B}\biggl( \gamma^k - \frac{13}{8}(\gamma^k)^2 \biggr) \sum_{b=1}^{B} &\langle \nabla_{\W} f(\W^k),\delta^{b,k} \rangle 
+ \frac{13}{16B^2} \sum_{k=1}^{N} (\gamma^k)^2 \|\sum_{b=1}^{B} \delta^{b,k}\|^2 \\
\end{aligned} \end{equation}
\begin{equation} \begin{aligned} \label{eq:sumineq1}
\sum_{k=1}^{N} \left( \gamma^k - \frac{13}{16}(\gamma^k)^2 \right) \|\nabla_{\W} f(\W^k)\|^2 &\leq f(\W^1) - f^{*} \\
- \frac{1}{B} \sum_{k=1}^{N} \sum_{b=1}^{B} \biggl( \gamma^k - \frac{13}{8}(\gamma^k)^2 \biggr) &\langle \nabla_{\W} f(\W^k),\delta^{b,k} \rangle 
+ \frac{13}{16B^2} \sum_{k=1}^{N} (\gamma^k)^2 \|\sum_{b=1}^{B} \delta^{b,k}\|^2 
\end{aligned} \end{equation}
where $\W^1$ is the initial estimate. The second inequality follows from the fact that at the optimum $\W^*$, we have $f^* \leq f(\W^{N+1})$.


\noindent Now observe that the above inequality comprises of two sets of random variables -- 
sampling over $\eta$ for constructing noisy gradients, and $N := R \sim \mathbb{P}_R(\cdot)$ which denotes the stopping iteration (which is independent of $\eta$).
We now marginalize out these variables from \eqref{eq:sumineq1}. 
Observe that, by definition, $\eta^{b,k}$ for different $b$'s and $k$'s are independent of each other. 
However, $\W^{k+1}$ depends on all $G(\eta^{b,l};{\bf W}^l)$s (being functions of $\eta^{b,k}$) for $l = 1$ to $k$.
Further, all the information needed for $k+1^{th}$ update comes from the ``state'' at $k^{th}$ iteration, implying that the updates ${\bf W}^k$ form a {\it Markov} process. 
Hence we take the expectation with respect to $p(\eta^{[B,N]},R) = p(\eta^{[B,N]})p(R)$ where $\eta^{[B,N]}$ is the random process generating $\eta^{\cdot,1},\ldots,\eta^{\cdot,N}$.


\noindent Taking expectation over $\eta^{[B,N]}$, the second to last term of \eqref{eq:sumineq1} becomes
\begin{equation} \begin{aligned} \label{eq:term1}
\mathbb{E}_{\eta^{[B,N]}} & \left[ \sum_{k=1}^{N} \sum_{b=1}^{B} \biggl( \gamma^k - \frac{13}{8}(\gamma^k)^2 \biggr) \langle \nabla_{\bf W} f(\W^k),\delta^{b,k} \rangle \right] \\
& \quad = \sum_{k=1}^{N} \sum_{b=1}^{B} \biggl( \gamma^k - \frac{13}{8}(\gamma^k)^2 \biggr) \hspace{2mm} 
\mathbb{E}_{\eta^{[b,k]}} (\langle \nabla_{\W} f(\W^k),\delta^{b,k} \rangle | \eta^{1,1},\ldots,\eta^{b,k}) = 0 
\end{aligned} \end{equation}
where the last equality follows from the definition of $\delta^{b,k} = g(\eta^{b,k};\W^k) - \nabla_{\W} f(\W^k)$ and the fact that $\mathbb{E}_{\eta} g(\eta;\W^k) = \nabla_{\W} f(\W^k)$. 


\noindent To bound the last term in \eqref{eq:sumineq1} consider the inner product $\langle \sum_{b=1}^{B-1} \delta^{b,k}, \delta^{B,k} \rangle$.
Taking the expectation with $\eta^{[B,k]}$ (for a given $k$), we have the following 
\begin{equation} \label{eq:forterm2}
\mathbb{E}_{\eta^{[B,k]}} \biggl[ \langle \sum_{b=1}^{B-1} \delta^{b,k}, \delta^{B,k} \rangle | \eta^{1,k},\ldots,\eta^{B-1,k} \biggr] 
= \langle \sum_{b=1}^{B-1} \delta^{b,k}, \mathbb{E}_{\eta^{[B,k]}} \delta^{B,k} \rangle | \eta^{1,k},\ldots,\eta^{B-1,k} = 0
\end{equation}
which follows from the fact that $\mathbb{E}_{\eta} \delta^{b,k} = \nabla_{\W} f(\W)$. Using this we have the following 
(the superscripts of $\eta^{\cdot}$ will be dropped hence forward so as to avoid reduce the clutter),
\begin{equation} \begin{aligned} \label{eq:forterm2_1}
\mathbb{E}_{\eta} \| \sum_{b=1}^{B} \delta^{b,k} \|^2 &= \mathbb{E}_{\eta} \biggl[ \| \sum_{b=1}^{B-1} \delta^{b,k} \|^2 + \mathbb{E}_{\eta} \| \delta^{B,k} \|^2 \biggr]
+ 2 \langle \sum_{b=1}^{B-1} \delta^{b,k}, \delta^{B,k} \rangle \\
&= \mathbb{E}_{\eta} \| \sum_{b=1}^{B} \delta^{b,k} \|^2 + \mathbb{E}_{\eta} \| \delta^{B,k} \|^2 = \sum_{b=1}^{B} \mathbb{E}_{\eta} \| \delta^{b,k} \|^2
\end{aligned} \end{equation} 


\noindent Further, from \eqref{eq:varlipbounds1} we have $\mathbb{E}_{\eta} \| \delta \|^2 \leq \frac{d_xd_y}{16}$. 
Using this and \eqref{eq:forterm2_1}, and taking the expectation over $\eta^{[B,N]}$ of the last term in \eqref{eq:sumineq1}, we get
\begin{equation}\label{eq:term2} 
\mathbb{E}_{\eta} \left[ \frac{13}{16B^2} \sum_{k=1}^{N} (\gamma^k)^2 \|\sum_{b=1}^{B} \delta^{b,k}\|^2 \right] 
= \frac{13}{16B^2} \sum_{k=1}^{N} (\gamma^k)^2 \mathbb{E}_{\eta} \|\sum_{b=1}^{B} \delta^{b,k}\|^2 \leq \frac{13d_xd_y}{256B} \sum_{k=1}^{N} (\gamma^k)^2 
\end{equation}


\noindent Using \eqref{eq:term1} and \eqref{eq:term2} and the inequality in \eqref{eq:sumineq1} we have
\begin{equation} \label{eq:sumineq2} 
\sum_{k=1}^{N} \left( \gamma^k - \frac{13}{16}(\gamma^k)^2 \right) \mathbb{E}_{\eta} \|\nabla_{\W} f(\W^k)\|^2 \leq f(\W^1) - f^* + \frac{e^s}{B} \sum_{k=1}^{N} (\gamma^k)^2  
\end{equation}
where the definition $e^s$ are used. Clearly for the above bound to make sense for any stepsize $\gamma^k < 1$.


\noindent Recall that we are intersted in bounding the expected gradients $\mathbb{E}_{R,\eta} \|\nabla_{\W} f(\W^k)\|^2$.
Apart from $\eta$, this expectation is over the stopping iteration $R$, where $R \sim \mathbb{P}_R(k)$ (i.e., sampled from the stopping distribution $\mathbb{P}_R(\cdot)$).
Hence we intend to bound 
\begin{equation} \label{eq:nonconstobound}
\mathbb{E}_{R,\eta} \|\nabla_{\W} f(\W^k)\|^2  := \frac{1}{\sum_{k=1}^N p^k_R} \left( p^k_R \mathbb{E}_{\eta} \|\nabla_{\W} f(\W^k)\|^2 \right)
\end{equation}
where $p^k_R$ is the probability of choosing $k^{th}$ iteration to be the stopping iteration $R$ (note that $k = 1,\ldots,N$).
Following the left hand side in \eqref{eq:sumineq2}, we see that $p^k_R$ would depend on the stepsizes $\gamma^k$.
Specifically, by choosing $p^k_R = \gamma^k - \frac{13}{16}(\gamma^k)^2$, we can compute the desired $\mathbb{E}_{R,\eta} \|\nabla_{\W} f(\W^k)\|^2$ as follows 
(clearly, whenever $\gamma^k < 1$, $\gamma^k - \frac{13}{16}(\gamma^k)^2 > 0$, and hence it makes sense to use it as $p^k_R$),
\begin{equation} \begin{aligned} \label{eq:sumineq2-gen} 
\mathbb{E}_{R,\eta} \|\nabla_{\W} f(\W^k)\|^2 &:= \frac{ \sum_{k=1}^{N} \left( \gamma^k - \frac{13}{16}(\gamma^k)^2 \right) \mathbb{E}_{\eta} \|\nabla_{\W} f(\W^k)\|^2 }
{\sum_{k=1}^N  (\gamma^k - \frac{13}{16}(\gamma^k)^2)} \\
&\leq \frac{ f(\W^1) - f^* + \frac{e^s}{B} \sum_{k=1}^{N} (\gamma^k)^2 }{\sum_{k=1}^N  (\gamma^k - \frac{13}{16}(\gamma^k)^2)}
\end{aligned} \end{equation}


\noindent Observe that for the current result, however, we are interested in the constant stepsize case where $\gamma^k = \gamma \hspace{1mm} \forall k$. 
Here the bounds, and the corresponding results, are a litlle more messier compared to the constant $\gamma$ case, nevertheless, 
as will be shown in the proof of Corollary \ref{thm:expgrad-noncon}, the overall recipe for deriving the bound is the same even in the non-constant stepsize setting. 


\noindent Using $\gamma^k = \gamma$ in \eqref{eq:sumineq2} and \eqref{eq:sumineq2-gen} we get 
\begin{equation} \begin{aligned} \label{eq:finalineq}
\mathbb{E}_{R,\eta} \|\nabla_{\W} f(\W^k)\|^2 &:= \frac{1}{N\gamma e^s_\gamma} \left( f(\W^1) - f^* + \frac{e^sN\gamma^2}{B} \right) \\
&\leq \frac{1}{e^s_\gamma} \left( \frac{D_f}{N\gamma} + \frac{e^s\gamma}{B} \right) 
\end{aligned} \end{equation}
where $D_f = f({\bf W}^1) - f^*$ is the initial deviation from optimum.
Recall the discussion in the remarks of Theorem \ref{thm:expgrad} (from Section \ref{sec:single}), 
where we argue that using constant stepsizes with no other constraint on the stopping distribution is equivalent, from the perspective of the bound, 
to choosing the stopping distribution to be uniform over all the $N$ iterations.


\noindent Observe that between the two terms within the paranthesis in \eqref{eq:finalineq}, the first term decreases as $\gamma$ increases, while the second term increases as $\gamma$ increases. 
Balancing these two, we obtain the optimal constant stepsize $\gamma_o$ (in the regime $\gamma < \frac{16}{13}$) as follows
\begin{equation} \label{eq:optstep}
\gamma_o = \sqrt{\frac{BD_f}{e^sN}} \approx \sqrt{\frac{Bf(\W^1)}{e^sN}}
\end{equation}
where without loss of generality we used the approximation that $f^{*} \sim 0$ and hence $D_f \approx f(\W^1)$. 
Observe that instead of balancing these terms, we could have explicitly worked with the entire term in the bound and differentiated w.r.t $\gamma$, which will result in a quadratic involving $\gamma$.
Note that unlike the above expression for $\gamma_o$, this latter case is much messier with no significant advantage. 



\subsection{Proof of Corollary 2}\label{sec:proof2} 


Using the result of Theorem \ref{thm:expgrad} and substituting the optimal stepsize $\gamma_o = \sqrt{\frac{Bf(\W^1)}{e^sN}}$  in \eqref{eq:finalineq}, 
with $e^s = \frac{13d_xd_y}{256}$ and $D_f$ replaced by $f(\W^1)$, we have 
\begin{equation} \label{eq:reduced1}
\mathbb{E}_{R,\eta} (\| \nabla_{\W} f(\W^R)\|^2) \leq \frac{d_xd_y\sqrt{13f(\W^1)}}{8\sqrt{B} \left( \sqrt{d_xd_yN} - \sqrt{13Bf(\W^1)} \right)} 
\end{equation}
Observe that $\gamma < \frac{16}{13}$ ensures that $d_xd_yN > 13Bf(\W^1)$ thereby making sure the denominator in the above inequality makes sense. 
We first derive the convergence rates of the above bound at this optimal choice of stepsize, 
followed by estimating the corresponding sample complexity using setup of $(\epsilon,\delta)$-solution which is based on choosing the best estimate from multiple runs of the Alg. \ref{alg:single-nn}. 


\noindent Although the right hand side above is non-trivial to interpret, the rates (vs. $N$ and $d_x$, $d_y$) can be obtained as follows. 
For a given network (i.e., for a fixed $d_xd_y$) and batch size $B$, we have 
\begin{equation} \begin{aligned} \label{eq:rate1}
\mathbb{E}_{R,\eta} &(\| \nabla_{\W} f(\W^R)\|^2) \leq \frac{P}{\sqrt{N}} \quad \forall \quad N>N_o>\frac{13Bf(\W^1)}{d_xd_y} \\
& \quad \text{where} \quad P = \frac{a\sqrt{N_o}}{\sqrt{N_o}-b} \quad a = \sqrt{\frac{13d_xd_yf(\W^1)}{64B}} \quad b = \sqrt{\frac{13Bf(\W^1)}{d_xd_y}}
\end{aligned} \end{equation}
\begin{equation} \begin{aligned} \label{eq:rate2}
\mathbb{E}_{R,\eta} &(\| \nabla_{\W} f(\W^R)\|^2) \leq Q\sqrt{d_xd_y} \quad \forall \quad d_xd_y>d_o>\frac{13Bf(\W^1)}{N} \\
& \quad \text{where} \quad Q = \frac{\sqrt{d_o}}{a(\sqrt{d_o}-b)} \quad a = 8\sqrt{\frac{BN}{13f(\W^1)}} \quad b = \sqrt{\frac{13Bf(\W^1)}{N}}
\end{aligned} \end{equation}



\subsection{Proof of Corollary 3}\label{sec:proof3} 


The proof for this result follows the same recipe as that of Theorem \ref{thm:expgrad}. We only point out the differences here.


\noindent In the non-constant stepsize case, using the stopping probability distribution such that $p^k_R = \gamma^k - \frac{13}{16}(\gamma^k)^2$ and \eqref{eq:sumineq2}, 
and computing the expected gradients both over $R$ and $\eta$, we get the following
\begin{equation} \label{eq:finalineqnonconst} 
\mathbb{E}_{R,\eta} \|\nabla_{\W} f(\W^k)\|^2  \leq \frac{f(\W^1) - f^* + \frac{e^s \sum_{i=1}^N (\gamma^k)^2}{B}}{\sum_{k=1}^N (\gamma^k - \frac{13}{16}(\gamma^k)^2)} 
\end{equation}
Whenever $\gamma^k < 1$, we have $\gamma^k(1-\frac{13\gamma^k}{16}) > \frac{3\gamma^k}{16}$, using which we have (and denoting $D_f = f(\W^1) - f^*$)
\begin{equation} \label{eq:finalineqnonconst1} 
\mathbb{E}_{R,\eta} \|\nabla_{\W} f(\W^k)\|^2 \leq \frac{D_f + \frac{e^s \sum_{i=1}^N (\gamma^k)^2}{B}}{\frac{3}{16} \sum_{k=1}^N \gamma^k}
= \frac{16}{3 \sum_{k=1}^N \gamma^k} \left( D_f + \frac{e^s \sum_{k=1}^N (\gamma^k)^2}{B} \right)
\end{equation}


\noindent Now using $\gamma^k = \frac{\gamma}{k^\alpha}$ for some $\alpha>1$, we get
\begin{equation} \begin{aligned} \label{eq:finalineq-noncon} 
\mathbb{E}_{R,\eta} \|\nabla_{\W} f(\W^k)\|^2 &\leq \frac{16}{3\gamma \sum_{k=1}^N k^{-\alpha}} \left( D_f + \frac{e^s \gamma^2 \sum_{k=1}^N k^{-2\alpha}}{B} \right) \\
&\leq \frac{16}{3\mathcal{H}_N(\alpha)} \left( \frac{D_f}{\gamma} + \frac{e^s \gamma\mathcal{H}_N(2\alpha)}{B} \right)
\end{aligned} \end{equation}
where $\mathcal{H}_n(m) = \sum_{i=1}^n \frac{1}{i^m} (m>1)$ is the generalized harmonic number.
Comparing this to \eqref{eq:finalineq} we see that the change in the bound is with respect to some constants.
Similar the computation for optimal stepsize as in \eqref{eq:optstep}, we have the following here (approximating $D_f$ with $f(\W^1)$)
\begin{equation} \label{eq:optstepnonconst}
\gamma_o = \sqrt{\frac{Bf(\W^1)}{e^s\mathcal{H}_N(2\alpha)}}
\end{equation}



\subsection{Proof of Corollary 4}\label{sec:proof4} 


The proof for this result follows the same recipe as that of Theorem \ref{thm:expgrad}. We only point out the differences here.


\noindent Recall \eqref{eq:sumineq2} from the proof of Theorem \ref{thm:expgrad} earlier. 
Using constant stepsizes $\gamma^k = \gamma$ and the definitions of $e^s$ and $e^s_\gamma$, this would reduce to
\begin{equation} \label{eq:sumineq2-general} 
\sum_{k=1}^{N} \mathbb{E}_{\eta} \|\nabla_{\W} f(\W^k)\|^2 \leq \frac{1}{\gamma e^s_\gamma} \left( f(\W^1) - f^* + \frac{e^sN}{B}\gamma^2 \right)
\end{equation}


\noindent Following a similar recipe as that of \eqref{eq:nonconstobound}-\eqref{eq:finalineq}, we have the following, 
where, without the loss of generality, we assume that the probabilites $p^k_R$ are normalized i.e., $\sum_{k=1}^N p^k_R = 1$ and $p^k_R \leq 1$, 
\begin{equation} \label{eq:nonconstobound-general}
\mathbb{E}_{R,\eta} \|\nabla_{\W} f(\W^k)\|^2  := \sum_{k=1}^N \left( p^k_R \mathbb{E}_{\eta} \|\nabla_{\W} f(\W^k)\|^2 \right)
\end{equation}
and using the condition that $p^k_R \leq p^{k+1}_R$ for all $k=2,\ldots,N$ and replacing $f(\W^1) - f^*$ with $D_f$, we have
\begin{equation} \label{eq:finalineq-mono}
\mathbb{E}_{R,\eta} \|\nabla_{\W} f(\W^k)\|^2  \leq \frac{p^N_R}{\gamma e^s_\gamma} \left( D_f + \frac{e^sN}{B}\gamma^2 \right)
\end{equation}


Depending on $p^N_R$, the right hand side above can be very lose, and hence it does not necessarily make sense to compute the optimal stepsize for any general $\mathbb{P}_R(\cdot)$.
Nevertheless, following the optimal stepsize computation above in \eqref{eq:optstep}, we have (using $D_f \approx f(\W^1)$)
\begin{equation}\label{eq:optstep-mono}
\gamma^1_o = \gamma = \sqrt{\frac{Bf(\W^1)}{e^sN}}
\end{equation}



\subsection{Proof of Theorem 6}\label{sec:proof6} 


\noindent Recall that, for a given $\epsilon > 0$ and $0 < \delta \ll 1)$, the $(\epsilon,\delta)$-solution is defined as follows:
\begin{equation} \label{eq:epsdelpf}
\W^{*} = \argmin_{t} \| \nabla_{\W} f(\W^{R_t}) \|^2 
\quad \text{s.t.} \quad Pr\left( \min_{t} \| \nabla_{\W} f(\W^{R_t}) \|^2 \leq \epsilon \right) \geq 1-\delta
\end{equation}
where $\W^{R_1},\ldots,\W^{R_T}$ correspond to estimates from the $T$ independent runs of Alg. \ref{alg:single-nn}.
Without loss of generality we assume that $N$ and $B$ are the same across all these $T$ runs.


\noindent Using some basic probability properties, 
\begin{equation} \begin{aligned} \label{eq:epsdelcomp} 
Pr\left( \min_{t} \| \nabla_{\W} f(\W^{R_t}) \|^2 \leq \epsilon \right) & = Pr\left( \exists t = 1,\ldots,T \hspace{1mm} \text{s.t.} \hspace{1mm} \| \nabla_{\W} f(\W^{R_t}) \|^2 \leq \epsilon \right) \\
& = 1 - Pr\left( \| \nabla_{\W} f(\W^{R_t}) \|^2 \geq \epsilon \hspace{1mm} \forall \hspace{1mm} t = 1,\ldots,T \right) \\
& = 1 - \prod_{t=1}^{T} Pr\left( \| \nabla_{\W} f(\W^{R_t}) \|^2 \geq \epsilon \right) 
\end{aligned} \end{equation}


\noindent Observe that we want the above probability to be larger than a given $1-\delta$. Hence we require
\begin{equation} \label{eq:epsdelcomp2} 
\prod_{t=1}^{T} Pr\left( \| \nabla_{\W} f(\W^{R_t}) \|^2 \geq \epsilon \right) \leq \delta
\end{equation}


\noindent Now recall the constraint on the stepsizes and stopping distribution i.e., $\gamma^k = \gamma$, $p^k_R \leq p^{k+1}_R \forall k$, $p^N_R = \frac{\vartheta}{N}$. 
Using this and the optimal value for $\gamma$ from \eqref{eq:optstep-mono} with $D_f \approx f(\W^1)$, and substituting it in \eqref{eq:finalineq-mono}, we get
\begin{equation} \begin{aligned} \label{eq:reduced1-mono}
\frac{p^N_R}{\gamma_o(1-\frac{13}{16}\gamma_o)} \left( D_f + \frac{1}{B} e^sN\gamma^2_o \right) \quad \text{with} \quad \gamma_o = \sqrt{\frac{Bf(\W^1)}{e^sN}} \\
\quad \quad \text{gives} \quad \frac{32p^N_Re^sN\sqrt{f(\W^1)}}{\sqrt{B} (16\sqrt{e^sN} - 13\sqrt{Bf(\W^1)})}
\end{aligned} \end{equation}
where the definition of $e^s_\gamma = 1-\frac{13\gamma}{16}$ was used.


\noindent Now using Markov inequality and optimal stepsize reduced bound from \eqref{eq:reduced1-mono}, we have the following
\begin{equation}\label{eq:epsdelcomp3} 
Pr\left( \| \nabla_{\W} f(\W^{R_t}) \|^2 \geq \epsilon \right) \leq \frac{\mathbb{E} (\nabla_{\W} f(\W^{R_t}) \|^2)}{\epsilon} 
\leq \frac{32e^s\vartheta\sqrt{f(\W^1)}}{\epsilon\sqrt{B} (16\sqrt{e^sN} - 13\sqrt{Bf(\W^1)})}
\end{equation} 


\noindent Noting that the $T$ different runs of Algorithm \ref{alg:single-nn} are independent, we can ensure \eqref{eq:epsdelcomp2} as follows
\begin{equation} \begin{aligned} \label{eq:epsdelcomp4-mono} 
\left( \frac{32e^s\sqrt{f(\W^1)}}{\epsilon\sqrt{B} (16\sqrt{e^sN} - 13\sqrt{Bf(\W^1)})} \right)^{T} &\leq \delta 
\iff 16\sqrt{e^sN} - 13\sqrt{Bf(\W^1)} \geq \frac{32e^s\vartheta\sqrt{f(\W^1)}}{\epsilon\delta^{1/T}\sqrt{B}} \\
&\iff N \geq \frac{f(\W^1)}{256e^s} \left( 13\sqrt{B} + \frac{32e^s\vartheta}{\epsilon\delta^{1/T}\sqrt{B}} \right)^2 
\end{aligned} \end{equation} 


\noindent Using $\bar{\delta} = \frac{13B\epsilon\delta^{1/T}}{32e^s}$, and observing that the denominator in \eqref{eq:reduced1-mono} is positive for any stepsize $\gamma<1$ 
(because for such stepsizes we will have $(1-\frac{13}{16}\gamma) > 0$), we finally have
\begin{equation}\label{eq:Nbound-mono} N \geq \frac{4f(\W^1)e^s}{B\delta^{2/T}\epsilon^2} (\bar{\delta} + \vartheta)^2 \end{equation}


\subsection{Proof of Corollary 7}\label{sec:proof7} 


\noindent Inducing dropout into the setting of $1$-NN from Theorem \ref{thm:expgrad} corresponds to adjusting the terms in \eqref{eq:lossdiff0}--\eqref{eq:finalineq}. 


\noindent To see this, first observe that dropout corredponds to dropping out units, and we consider the setting where this dropping is in the inputs/visible layer, 
and there is no dropout in the output layer.
Specifically, in each iteration of Algorithm \ref{alg:single-nn}, an input feature/dimension is dropped with probability $1-\zeta$. 
Hence, within each iteration, out of the $d_xd_y$ number of unknowns, only $\zeta d_xd_y$ are updated.  
If $\mathcal{J}$ denotes the $\zeta d_x$ {\it non-dropped} dimensions, then within a given iteration, the following statements will hold. 
\begin{equation} \label{eq:singledrop}
\nabla^{ij}_{W} f(\W) = 0 \quad;\quad g^{ij}(\eta;\W) = \nabla^{ij}_{\W} \mathcal{L}(\eta;\W) = 0 \quad \quad j \in \mathcal{J}^C
\end{equation}


\noindent The first and second equalities are due to the fact that, within an iteration we have a network of $\zeta d_x$ and $d_y$ outputs, 
and the transformation matrix will be of size $d_y \times \zeta d_x$.
These imply that the variance bound in \eqref{eq:varlipbounds1} would change from $\frac{d_xd_y}{16}$ to $\frac{\zeta d_xd_y}{16}$. 


\noindent Further, the inequalities from \eqref{eq:firstineq}--\eqref{eq:firstineq2} and \eqref{eq:sumineq0}--\eqref{eq:sumineq1} 
would have to be {\it corrected} for this change in the number of unknowns being updated within each iteration.
This can be done by observing changing the left hand side of \eqref{eq:sumineq2}.


\noindent Note that the $\zeta d_xd_y$ unknowns across different iterations are independent, and a given edge is dropped w.p. $1-zeta$.
Hence, whenever $N$ is reasonably large, the effective number of times a given $W_{ij}$ is updated is $\zeta N$ 
i.e., a given $W_{ij}$ appears $\zeta N$ times in the left hand side of \eqref{eq:sumineq2} (instead of $N$ as in the non-dropout setting).


\noindent Overall, this amounts to the following changes. First, the constants $e^s$ would need to be scaled down to $\zeta e^s$ (recall the definition of $e^s$ here).
Second, the left hand side in the inequality from \eqref{eq:sumineq2} would change to 
\begin{equation} \label{eq:sumineq2-drop} 
\zeta \sum_{k=1}^{N} \left( \gamma^k - \frac{13}{16}(\gamma^k)^2 \right) \mathbb{E}_{\eta} \|\nabla_{\W} f(\W^k)\|^2
\end{equation}


\noindent Incorporating these into \eqref{eq:sumineq2}--\eqref{eq:finalineq}, we have
\begin{equation} \begin{aligned} \label{eq:finalineq-drop}
\mathbb{E}_{R,\eta} \|\nabla_{\W} f(\W^k)\|^2 &:= \frac{1}{\zeta N\gamma e^s_\gamma} \left( f(\W^1) - f^* + \frac{\zeta e^sN\gamma^2}{B} \right) \\
&\leq \frac{1}{e^s_\gamma} \left( \frac{D_f}{N\gamma\zeta} + \frac{e^s\gamma}{B} \right) 
\end{aligned} \end{equation}
where $D_f = f({\bf W}^1) - f^*$ is the initial deviation from optimum. The corresponding optimal stepsize would simply be 
\begin{equation} \label{eq:optstep-drop}
\gamma_o = \sqrt{\frac{BD_f}{\zeta e^sN}} \approx \sqrt{\frac{Bf(\W^1)}{\zeta e^sN}}
\end{equation}


\noindent To compute the computational complexity, following the steps from the proof of Corollary \ref{thm:conv1nn-mono} from Section \ref{sec:proof6}, we have the following.
Using \eqref{eq:epsdelpf}--\eqref{eq:epsdelcomp2}, and \eqref{eq:finalineq-drop} and \eqref{eq:optstep-drop} from above, we have
\begin{equation} \begin{aligned} \label{eq:reduced1-drop}
\frac{1}{(1-\frac{13}{16}\gamma_o)} \left( \frac{D_f}{N\gamma_o\zeta} + \frac{e^s\gamma_o}{B} \right) \quad \text{with} \quad \gamma_o = \sqrt{\frac{Bf(\W^1)}{\zeta e^sN}} \\
\quad \quad \text{gives} \quad \frac{32e^s\sqrt{f(\W^1)}}{\sqrt{B} (16\sqrt{\zeta e^sN} - 13\sqrt{Bf(\W^1)})}
\end{aligned} \end{equation}


\noindent Using this, the definition of $\bar{\delta}$ and following the steps from \eqref{eq:epsdelcomp3}--\eqref{eq:Nbound-mono}, we get
\begin{equation}\label{eq:Nbound-drop} N \geq \frac{4f(\W^1)e^s}{\zeta B\delta^{2/T}\epsilon^2} (1 + \bar{\delta})^2 \end{equation}


\subsection{Proof of Theorem 8}\label{sec:proof8} 


This proof structure closely follows that of Theorem \ref{thm:expgrad}.
Following the setup there, we first derive the inequality concerning the behaviour of the noisy gradients (refer to \eqref{eq:varlipbounds1}). 
Using this, we then bound the expected gradients taking into account that, unlike the single-layer case, the minimization is over box constrained set $W_{ij} \in [-w_m,w_m]$ 
($\forall i$, $j$; refer to Alg. \ref{alg:da}). 
The update step in Alg. \ref{alg:da} comprises of projecting the parameter update onto this feasible set. 


\noindent The loss function for DA pretraining is $\mathcal{L}(\eta;\W) = \| \x - \sigma(\W^T\sigma(\W\tx)) \|^2$ (recall the objective $f(\W) = \mathbb{E}_{\eta} \mathcal{L}(\eta;\W)$). 
The minimization is over $W_{ij} \leq w_m$ for $i = 1,\ldots,d_h$ and $j = 1,\ldots,d_x$ with $\tilde{x}_j = x_j$ w.p $\zeta$ and $0$ otherwise. 
For some $I = 1,\ldots,d_h$ and $J = 1,\ldots,d_x$, consider the gradient of this loss with respect to $W_{IJ}$, given as follows
\begin{equation} \begin{aligned} \label{eq:da_gradval1}
g^{IJ}(\eta;\W) &= \nabla^{IJ}_{\W} \mathcal{L}(\eta;\W) \\
&= \sum_{j=1}^{d_x} -2 (x_j - \sigma_j(\W^T\sigma(\W\tx))) \sigma_j(\W^T\sigma(\W\tx)) (1-\sigma_j(\W^T\sigma(\W\tx))) \\
& \quad \quad \quad \quad[\mathbb{I}_J\sigma_I(\W\tx) + W_{Ij}\sigma_I(\W\tx)(1-\sigma_I(\W\tx))\tilde{x}_J]
\end{aligned} \end{equation}
where $\mathbb{I}_J=1$ for $j=J$ and $0$ else.


\noindent Separating the cases of $\tilde{x}_J=0$ and $\tilde{x}_J=x_J$, and using the properties of $\sigma(\cdot)$, and the fact that $\x \in [0,1]^{d_x}$ we have the following
\begin{equation} \begin{aligned} \label{eq:da_gradval2}
&|g^{IJ}(\eta;\W)|\quad | \{\tilde{x}_J=0\} \\
&\quad \quad = \quad |2 (x_J - \sigma_J(\W^T\sigma(\W\tx))) \sigma_J(\W^T\sigma(\W\tx)) (1-\sigma_J(\W^T\sigma(\W\tx))) \sigma_I(\W\tx)| \leq \frac{1}{2} \\
&|g^{IJ}(\eta;\W)|\quad | \{\tilde{x}_J=x_J\} \\
&\quad \quad = \quad |2 (x_J - \sigma_J(\W^T\sigma(\W\tx))) \sigma_J(\W^T\sigma(\W\tx)) (1-\sigma_J(\W^T\sigma(\W\tx))) \\
&\quad \quad \quad \quad \quad(\sigma_I(\W\tx) + W_{IJ}\sigma_I(\W\tx)(1-\sigma_I(\W\tx))x_J) \\
&\quad \quad \quad \quad + 2 \sum_{j\neq J} (x_j - \sigma_j(\W^T\sigma(\W\tx))) \sigma_j(\W^T\sigma(\W\tx)) (1-\sigma_j(\W^T\sigma(\W\tx))) \\
&\quad \quad \quad \quad \quad \quad \quad W_{Ij}\sigma_I(\W\tx)(1-\sigma_I(\W\tx)\tilde{x}_J| \\
&\quad \quad \leq \quad \frac{1}{2}|x_J - \sigma_J(\W^T\sigma(\W\tx))||\sigma_I(\W\tx) + W_{IJ}\sigma_I(\W\tx)(1-\sigma_I(\W\tx))x_J| \\
&\quad \quad \quad \quad + \frac{1}{2} x_J\sum_{j\neq J} |x_j - \sigma_j(\W^T\sigma(\W\tx))|\frac{1}{4}|W_{Ij}| \\
&\quad \quad \leq \quad \frac{1}{2}|x_J - \sigma_J(\W^T\sigma(\W\tx))| + |x_J - \sigma_J(\W^T\sigma(\W\tx))||W_{IJ}|\frac{1}{4}x_J \\
&\quad \quad \quad \quad + \frac{1}{2} x_J\sum_{j\neq J} |x_j - \sigma_j(\W^T\sigma(\W\tx))|\frac{1}{4}|W_{Ij}| \quad \leq \quad \frac{1}{2} + \frac{x_J}{8} \sum_{j=1}^{d_x}|W_{Ij}| 
\end{aligned} \end{equation}


\noindent Taking expectation over $\x \in \mathcal{X}$ for a given corruption of $J^{th}$ unit (i.e., independent of the input $\x$, $\tilde{x}_J$ is always $0$), we get
\begin{equation} \begin{aligned} \label{eq:da_gradval3}
\mathbb{E}_{\x} (g^{IJ}(\eta;\W)) \leq \begin{cases} \frac{1}{2} \quad \quad \quad \text{whenever $\tilde{x}_J=0$} \\ 
\frac{1}{2} + \frac{\mu_Jd_xw_m}{8}  \quad \quad \text{otherwise} \end{cases}
\end{aligned} \end{equation}
where the second case follows from the fact that $\mathbb{E}_{\x} (x_J\sum_{j=1}^{d_x}|W_{Ij}|) \leq \mathbb{E}_{\x} (x_Jd_xw_m) = \mu_Jd_xw_m$ 
which uses the max-norm boundary (or box) constraint for the parameters i.e., $W_{ij} \leq w_m \forall i,j$ (refer to the update from Alg. \ref{alg:da}). 
Here $\mu_j = \mathbb{E}_{\x} x_j$. 


\noindent Further, using the definition of variance we have 
\begin{equation} \begin{aligned} \label{eq:da_gradval3}
\Var_{\x}(g^{IJ}(\eta;\W)) &\leq \mathbb{E}_{\x} (g^{IJ}(\eta;\W))\left( sup_{\x}(g^{IJ}(\eta;\W)) - \mathbb{E}_{\x} (g^{IJ}(\eta;\W))\right) \\
&\leq \begin{cases} \frac{1}{16} \quad \quad \quad \text{whenever $\tilde{x}_J=0$} \\ 
\frac{1}{16}\left(1 + \frac{\mu_Jd_xw_m}{4} \right)^2  \quad \quad \text{otherwise} \end{cases}
\end{aligned} \end{equation}


\noindent Note that we are interested in $\Var_{\eta}(G^{ij}(\eta;\W))$ for some $i,j$. 
Here $\mu_j = \mathbb{E}_{\x}x_j$ is the average (across all inputs from $\mathcal{X}$) first moment at the $j^{th}$ input unit. 
Hence using the variance decomposition, we get
\begin{equation} \begin{aligned} \label{eq:da_gradval4}
\Var_{\eta}(g^{ij}(\eta;\W)) &= \Var_{\x,\tx}(g^{ij}(\eta;\W)) \\
&= \mathbb{E}_{\tx|\x} [\Var_{\x}(g^{ij}(\eta;\W)) | \{\tx|\x\}] + \Var_{\tx|\x} [\mathbb{E}_{\x}(g^{ij}(\eta;\W)) | \{\tx|\x\}] \\
&= \mathbb{E}_{\tilde{x}_j|x_j} [\Var_{\x}(g^{ij}(\eta;\W)) | \{\tilde{x}_j|x_j\}] + \Var_{\tilde{x}_j|x_j} [\mathbb{E}_{\x}(g^{ij}(\eta;\W)) | \{\tilde{x}_j|x_j\}] 
\end{aligned} \end{equation}
Now applying \eqref{eq:da_gradval3} and \eqref{eq:da_gradval4} and using the fact that $\tilde{x}_j=0$ w.p $1-\zeta$, we have
\begin{equation} \label{eq:da_gradval5}
\mathbb{E}_{\tilde{x}_j|x_j} [\Var_{\x}(g^{ij}(\eta;\W)) | \{\tilde{x}_j|x_j\}] \leq \frac{1-\zeta}{16} + \frac{\zeta}{16}\left(1 + \frac{\mu_jd_xw_m}{4} \right)^2 
\end{equation}
\begin{equation} \begin{aligned} \label{eq:da_gradval6}
\Var_{\tilde{x}_j|x_j} [\mathbb{E}_{\x}(g^{ij}(\eta;\W)) | \{\tilde{x}_j|x_j\}] &\leq \mathbb{E}_{\tilde{x}_j|x_j} [\mathbb{E}^2_{\x}(G^{ij}(\eta;\W))] 
- (\mathbb{E}_{\tilde{x}_j|x_j} [\mathbb{E}_{\x}(g^{ij}(\eta;\W))])^2 \\
&\leq \frac{1-\zeta}{4} + \zeta\left(\frac{1}{2} + \frac{\mu_jd_xw_m}{8}\right)^2 - \left(\frac{1-\zeta}{2} + \frac{\zeta}{2} + \frac{\zeta\mu_jd_xw_m}{8} \right)^2
\end{aligned} \end{equation} 


\noindent Combining these we finally get 
\begin{equation} \begin{aligned} \label{eq:da_gradval6}
\Var_{\eta}(g^{ij}(\eta;\W)) &\leq \frac{1}{16}\left[ 1 + \frac{\zeta\mu_jd_xw_m}{4} + \left(\frac{5\zeta}{16} - \frac{\zeta^2}{4}\right)(\zeta\mu_jd_xw_m)^2\right] 
\end{aligned} \end{equation}
Following the setup from \eqref{eq:lossdiff0}, \eqref{eq:gradval3} and \eqref{eq:varlipbounds1} from the proof of Theorem \ref{thm:expgrad} in Section \ref{sec:proof1}, 
and replicating the same here, we have
\begin{equation} \begin{aligned} \label{eq:da_varlipbounds1}
\mathbb{E}_{\eta} \| \delta \|^2 &\leq \frac{1}{16} \left[ d_xd_h + \frac{\zeta d_xw_m}{4}\sum_{ij}\mu_j + \left(\frac{5\zeta}{16} - \frac{\zeta^2}{4}\right)(\zeta d_xw_m)^2 \sum_{ij}\mu^2_j \right] \\
&= \frac{1}{16} \left[ d_xd_h + \frac{\zeta d_xd_hw_m}{4}\sum_{j}\mu_j + \left(\frac{5\zeta}{16} - \frac{\zeta^2}{4}\right)(\zeta d_xw_m)^2d_h \sum_{j}\mu^2_j \right] \\ 
&= \frac{d_xd_h}{16} \left[ 1 + \frac{\zeta d_xw_m}{4}\mu_{\x} + \left(\frac{5\zeta}{16} - \frac{\zeta^2}{4}\right)(\zeta d_xw_m)^2 \tau_{\x} \right] 
\end{aligned} \end{equation}
where $\mu_{\x} = \frac{1}{d_x}\sum_{j=1}^{d_x}\mu_j$ and $\tau_{\x} = \frac{1}{d_x}\sum_{j=1}^{d_x}\mu^2_j$ correspond to the average moment of the inputs across all $d_x$ units/dimensions.
Note that $\tau_{\x}$ is not the second moment, and instead is a function of the first moments ($\mu_j$'s) themselves. 
Recall that $\delta = g(\eta;\W) - \nabla_{\W} f(\W)$.


\noindent The rest of the proof follows the recipe from Section \ref{sec:proof1} and builds the expected bounds for Alg. \ref{alg:da} following the update 
\begin{equation}
\W_{ij}^{k+1} \leftarrow P_{\W}(\W_{ij}^k - \gamma^k G(\eta^k;\W_{ij}^k))
\end{equation}
Observe that $P_{\W}(\cdot)$ denotes the Euclidean projection on $[-w_m,w_m]$.
We will be needing the following results about the Euclidean projection for constructing the convergence bound. 
\begin{equation} \begin{aligned} \label{eq:projppts}
& \text{(a)} \quad h^{T}P_{\mathcal{\W}}(\W,h,\gamma) \geq \| P_{\mathcal{\W}}(\W,h,\gamma) \|^2 \\
& \text{(b)} \quad \| P_{\mathcal{\W}}(\W,h_1,\gamma) - P_{\mathcal{\W}}(\W,h_2,\gamma) \| \leq \| h_1 - h_2 \|
\end{aligned} \end{equation}
for some $h$, $h_1$, $h_2 \in \mathbb{R}^{d_hd_x}$. 
We do not discuss their proofs here, which can be found in \cite{rockafellar2009variational} (and Lemma 1 and Proposition 2 in \cite{ghadimi2016mini}).
\eqref{eq:projppts}(b) corresponds to the general inequality that projection does not reduce distances.
For simplicity we have the following notation for the projected versions of $g(\eta;\W)$, $G(\eta;\W)$ and $\nabla_{\W}f(\eta;\W)$.
\begin{equation} \begin{aligned} \label{eq:notations}
&P_{\mathcal{\W}}(\W,g(\eta;\W),\gamma) := \tg(\eta;\W) \\
&P_{\mathcal{\W}}(\W,G(\eta;\W),\gamma) := \tG(\eta;\W) \\
&P_{\mathcal{\W}}(\W,\nabla_{\W}f(\W),\gamma) := \nabla_{\W}\tf(\W)
\end{aligned} \end{equation} 
(i.e, by adding an additional {\it tilde}). 


\noindent We now follow the recipe from \eqref{eq:firstineq}--\eqref{eq:finalineq}, from the proof of Theorem \ref{thm:expgrad}. 
Using the Lipschitz constant $\uda$ (and denoting it by $L(d_h,w_m)$), we have the following 
\begin{equation} \begin{aligned} \label{eq:da_firstineq}
&f(\W^{k+1}) \leq f(\W^k) + \langle \nabla_{\W} f(\W^k),\W^{k+1}-\W^k \rangle + \frac{\uda}{2} \|\W^{k+1}-\W^k\|^2 \\
&\quad \text{where} \quad \uda = \left( (d_h-1) \left[ \frac{w_m}{20} + \frac{13w^2_m}{640} \right] + \frac{16}{20} + \frac{73w_m}{480} \right) 
\end{aligned} \end{equation}
By the definition of prox operation, and using some abuse of notation, we have $\W^{k+1} \leftarrow \W^{k} - \gamma^k \tG(\eta^k;\W^k)$, and recall that $\delta^k = G(\eta^k;\W^k) - \nabla_{\W} f(\W^k)$.


\noindent Using these we get
\begin{equation} \begin{aligned} \label{eq:da_firstineq1}
&f(\W^{k+1}) \leq f(\W^k) + \langle \nabla_{\W} f(\W^k),\W^{k+1}-\W^k \rangle + \frac{\uda}{2} \|\W^{k+1}-\W^k\|^2 \\
&= f(\W^k) - \gamma^k \langle \nabla_{\W} f(\W^k),\tG(\eta^k;\W^k) \rangle + \frac{\uda}{2}(\gamma^k)^2 \|\tG(\eta^k;\W^k)\|^2 \\
&= f(\W^k) - \gamma^k \langle G(\eta^k;\W^k),\tG(\eta^k;\W^k) \rangle + \frac{\uda}{2}(\gamma^k)^2 \|\tG(\eta^k;\W^k)\|^2 \\ 
&\quad \quad \quad + \gamma^k \langle \delta^k,\tG(\eta^k;\W^k) \rangle
\end{aligned} \end{equation}
and using \eqref{eq:projppts}(a) with $h = G(\eta^k;\W^k)$, we get
\begin{equation} \begin{aligned} \label{eq:da_firstineq2}
f(\W^{k+1}) &\leq f(\W^k) - \gamma^k \| \tG(\eta^k;\W^k) \|^2 + \frac{\uda}{2}(\gamma^k)^2 \|\tG(\eta^k;\W^k)\|^2 \\ 
&\quad \quad \quad + \gamma^k \langle \delta^k,\tG(\eta^k;\W^k) \rangle \\
&= f(\W^k) - \gamma^k \| \tG(\eta^k;\W^k) \|^2 + \frac{\uda}{2}(\gamma^k)^2 \|\tG(\eta^k;\W^k)\|^2 \\ 
&\quad \quad \quad + \gamma^k \langle \delta^k,\nabla_{\W}\tf(\eta^k;\W^k) \rangle + \gamma^k \langle \delta^k,\tG(\eta^k;\W^k) - \nabla_{\W}\tf(\W^k) \rangle \\
&= f(\W^k) - \left( \gamma^k - \frac{\uda}{2}(\gamma^k)^2 \right) \|\tG(\eta^k;\W^k)\|^2 \\ 
&\quad \quad \quad + \gamma^k \langle \delta^k,\nabla_{\W}\tf(\eta^k;\W^k) \rangle + \gamma^k \|\delta^k\| \|\tG(\eta^k;\W^k) - \nabla_{\W}\tf(\W^k)\|
\end{aligned} \end{equation}
where the last term in the third inequality above is obtained using Cauchy-Schwartz.
Using \eqref{eq:projppts}(b) with $h_1 = G(\eta^k;\W^k)$ and $h_2 = \nabla_{\W}f(\eta;\W)$ in the last inequality above, we have
\begin{equation} \begin{aligned} \label{eq:da_firstineq3}
f(\W^{k+1}) &\leq f(\W^k) - \left( \gamma^k - \frac{\uda}{2}(\gamma^k)^2 \right) \|\tG(\eta^k;\W^k)\|^2 \\ 
&\quad \quad \quad + \gamma^k \langle \delta^k,\nabla_{\W}\tf(\W^k) \rangle + \gamma^k \|\delta^k\|^2
\end{aligned} \end{equation}
Adding up the above inequalities over the iterations $k = 1,\ldots,N$, we get
\begin{equation} \begin{aligned} \label{eq:da_sumineq0}
\sum_{k=1}^N \left( \gamma^k - \frac{\uda}{2}(\gamma^k)^2 \right) &\|\tG(\eta^k;\W^k)\|^2 \leq f(\W^1) - f(\W^{N+1}) \\
& \quad + \sum_{k=1}^N \left[ \gamma^k \langle \delta^k,\nabla_{\W}\tf(\W^k) \rangle + \gamma^k \|\delta^k\|^2 \right] 
\end{aligned} \end{equation} 
We now take the expectation of the above inqeuality over the two sets of random variables -- the stopping iteration $R$ and the data instance $\eta$ 
(similar to the setup from \eqref{eq:term1}-\eqref{eq:sumineq2})


\noindent Note that $\delta^k = G(\eta^k;\W^k) - \nabla_{\W} f(\W^k)$ and recall the steps \eqref{eq:term1}-\eqref{eq:forterm2_1} 
(from the proof of Theorem \ref{thm:expgrad}) which were used to bound the expectations of the last two terms on the right hand side of \eqref{eq:sumineq2}.
Replicating a similar set of steps here and using \eqref{eq:da_varlipbounds1}, we have the following
\begin{equation} \label{eq:da_term1}
\mathbb{E}_{\eta} \left( \sum_{k=1}^N \gamma^k \langle \delta^k,\nabla_{\W}\tf(\W^k) \rangle \right) 
= \sum_{k=1}^N \gamma^k \mathbb{E}_{\eta} \left( \langle \delta^k,\nabla_{\W}\tf(\W^k) \rangle | \eta^1,\ldots,\eta^{k-1} \right) = 0 
\end{equation}
\begin{equation} \begin{aligned} \label{eq:da_term2}
&\mathbb{E}_{\eta} \|\delta^k\|^2 = \frac{1}{B^2} \mathbb{E}_{\eta} \|\sum_{b=1}^B \delta^{b,k}\|^2 = \sum_{b=1}^B \mathbb{E}_{\eta} \| \delta^{b,k} \|^2 \leq \frac{e^{da}}{B} \\
& \quad \text{where} \quad e^{da} := \frac{d_xd_h}{16} \left[ 1 + \frac{\zeta d_xw_m}{4}\mu_{\x} + \left(\frac{5\zeta}{16} - \frac{\zeta^2}{4}\right)(\zeta d_xw_m)^2 \tau_{\x} \right]
\end{aligned} \end{equation}
Using \eqref{eq:da_term1} and \eqref{eq:da_term2} after taking the expectation (over $\eta$) of \eqref{eq:da_sumineq0} and using the fact that $f(\W^{N+1}) \leq f^*$, we have
\begin{equation} \begin{aligned} \label{eq:da_sumineq1}
\sum_{k=1}^N \left( \gamma^k - \frac{\uda}{2}(\gamma^k)^2 \right) &\mathbb{E}_{\eta} \|\tG(\eta^k;\W^k)\|^2 \leq D_f + \frac{e^{da}}{B} \sum_{k=1}^N \gamma^k 
\end{aligned} \end{equation} 
where $D_f = f(\W^1) - f^*$.


\noindent Observe that the inequality to makes sense we need $\gamma^k < \frac{2}{\uda}$. 
With this, and following the setup described using \eqref{eq:nonconstobound} for arriving at \eqref{eq:finalineq} using constant stepsizes $\gamma^k = \gamma < \frac{2}{\uda}$ 
(refer to the proof of Theorem \ref{thm:expgrad}), we have the following
\begin{equation} \begin{aligned} \label{eq:da_finalineq}
\left( \gamma - \frac{\uda}{2}\gamma^2 \right) \sum_{k=1}^N \mathbb{E}_{\eta} \|\tG(\eta^k;\W^k)\|^2 &\leq D_f + \frac{e^{da}N\gamma}{B} \quad \text{and hence} \\
\mathbb{E}_{R,\eta} \| \tG(\eta^k;\W^k) \|^2 := \frac{1}{N} \sum_{k=1}^N \mathbb{E}_{\eta} \|\tG(\eta^k;\W^k)\|^2 &\leq \frac{D_f + \frac{e^{da} N\gamma}{B}}{N(\gamma - \frac{\uda}{2}\gamma^2)} \\
&= \left( \frac{D_f}{N\gamma} + \frac{e^{da}}{B} \right) \frac{1}{1 - \frac{\uda}{2}\gamma}
\end{aligned} \end{equation}


\noindent Recall that in all the derivations above $\eta$ essentially corresponds to sampling a data instance $\x \in \mathcal{X}$ and then corrupting it to get $\tx$.
The above inequality governs the decay of $\tG(\eta^k;\W^k)$, however, we are interested in bounding the expectation of $\nabla_{\W}\tf(\eta;\W)$. 
To that end, we have the following (the subscript of expectation is dropped to reduce the clutter),
\begin{equation} \begin{aligned} \label{eq:da_finalineq1}
\mathbb{E}_{R,\eta} \| \nabla_{\W}\tf(\W) \|^2 &= \mathbb{E}_{R,\eta} \| \nabla_{\W}\tf(\W) + \tG(\eta^k;\W^k) - \tG(\eta^k;\W^k) \|^2 \\
&\leq \mathbb{E}_{R,\eta} \| \tG(\eta^k;\W^k) \|^2 + \mathbb{E}_{R,\eta} \| \nabla_{\W}\tf(\W) - \tG(\eta^k;\W^k) \|^2 \\
&\leq \left( \frac{D_f}{N\gamma} + \frac{e^{da}}{B} \right) \frac{1}{1 - \frac{\uda}{2}\gamma} + \mathbb{E}_{R,\eta} \| \nabla_{\W}f(\W) - G(\eta^k;\W^k) \|^2 
\end{aligned} \end{equation}
where the last inequality uses \eqref{eq:da_finalineq} for the first term, and \eqref{eq:projppts}(b) with $h_1 = \nabla_{\W}f(\W)$ and $h_2 = G(\eta^k;\W^k)$ for the second term.


\noindent Recalling that $\mathbb{E}_{\eta} \| \nabla_{\W}f(\W) - G(\eta^k;\W^k) \|^2$ is simply $\mathbb{E}_{\eta} \| \delta^k \|^2$, and using \eqref{eq:da_term2} and the definition of $e^{da}_\gamma$, 
we finally get
\begin{equation} \label{eq:da_finalineq2}
\mathbb{E}_{R,\eta} \| \nabla_{\W}\tf(\W) \|^2 \leq \frac{D_f}{N\gamma e^{da}_\gamma} + \frac{e^{da}}{B} \left( 1+\frac{1}{e^{da}_\gamma} \right)
\end{equation} 


\noindent The bound in \eqref{eq:da_finalineq2} is very large whenever the stepsizes lie closer to the left and right extremes of the allowed range $0<\gamma<\frac{2}{\uda}$.
Hence, the optimal $\gamma_o$ is given by the derivative of the right hand side above. Noting that $B \ll N$ for the regimes of interest, $\gamma_o$ is given by
\begin{equation} \begin{aligned} \label{eq:da_optstep}
&\uda e^{da} N \gamma_o^2 + 2BD_f\uda \gamma_o - 2BD_f = 0 \\
&\quad \implies \gamma_o = -\frac{BD_f}{e^{da}N} + \sqrt{\left(\frac{BD_f}{e^{da}N}\right)^2 + \frac{2BD_f}{\uda e^{da}N}} \approx \sqrt{\frac{2BD_f}{\uda e^{da}N}} 
\end{aligned} \end{equation}
where $D_f \approx f(\W^1)$. 
Observe that for reasonable values of $w_m$ and $d_h$, $\uda<2$, and intandem with $B \ll N$ this result is of practical use, instead of just being for theoretical interest.
Clearly we need $\frac{2}{\uda} > \frac{BD_f}{e^{da} N}$ for the above stepsize to make sense, and in the $B \ll N$ setting with reasonable $w_m$ and $d_h$, this would be satisfied.


\subsection{Proof of Corollary 9}\label{sec:proof10} 


Recall the proofs of Theorem \ref{thm:expgrad} and Corollary \ref{thm:expgrad-mono} from Sections \ref{sec:proof1} and \ref{sec:proof4} respectively, 
and modifications required in Section \ref{sec:proof4} over the calculations from Section \ref{sec:proof1}.
The proof of this result entails a similar set of changes over the recipe and inequalities from Section \ref{sec:proof8}.
Hence, we just list down the changed set of inequalities and the resulting change in the final bound. 


\noindent Using $p^k_R \leq p^{k+1}_R \forall k$, \eqref{eq:da_finalineq} would change as follows,
\begin{equation} \begin{aligned} \label{eq:da_finalineq-mono}
\mathbb{E}_{R,\eta} \| \tG(\eta^k;\W^k) \|^2 &:= \sum_{k=1}^N p^k_R \left( \mathbb{E}_{\eta} \|\tG(\eta^k;\W^k)\|^2 \right) \leq p^N_R \sum_{k=1}^N \left( \mathbb{E}_{\eta} \|\tG(\eta^k;\W^k)\|^2 \right) \\
&\leq p^N_R \frac{D_f + \frac{e^{da} N\gamma}{B}}{\gamma - \frac{\uda}{2}\gamma^2} = \left( \frac{D_f}{\gamma} + \frac{e^{da}N}{B} \right) \frac{p^N_R}{1 - \frac{\uda}{2}\gamma}
\end{aligned} \end{equation}


\noindent The rest of the derivation would follow the structure in \eqref{eq:da_finalineq1}--\eqref{eq:da_optstep}, and hence we have the following
\begin{equation} \label{eq:da_finalineq2-mono}
\mathbb{E}_{R,\eta} \| \nabla_{\W}\tf(\W) \|^2 \leq \frac{p^N_RD_f}{\gamma e^{da}_\gamma} + \frac{e^{da}}{B} \left( 1+\frac{p^N_RN}{e^{da}_\gamma} \right)
\end{equation}


\noindent Re-computing the optimal stepsize from the above bound, following \eqref{eq:da_optstep}, we would still get (with $B \ll N$),
\begin{equation} \label{eq:da_optstep-mono}
\gamma_o = -\frac{BD_f}{e^{da}N} + \sqrt{\left(\frac{BD_f}{e^{da}N}\right)^2 + \frac{2BD_f}{\uda e^{da}N}} \approx \sqrt{\frac{2BD_f}{\uda e^{da}N}} 
\end{equation}



\subsection{Proof of Theorem 10}\label{sec:proof9-extra} 


The expected gradients bound for DA RSG is \eqref{eq:da_finalineq2}.
Now recall the large deviation estimate proof procedure for single layer RSG from Section \ref{sec:proof6}, specifically the derivation of \eqref{eq:epsdelcomp3} and \eqref{eq:epsdelcomp4-mono}.
Applying similar steps for \eqref{eq:da_finalineq2}, we get the following


\begin{equation} \label{eq:sampdaderiv1}
\frac{D_f}{N\gamma e^{da}_\gamma} + \frac{e^{da}}{B} \left( 1+\frac{1}{e^{da}_\gamma} \right) \leq \epsilon \delta^{1/T}
\end{equation}
Under the assumption that $SC \approx BN$, the two terms in the left hand side above decrease and increase respectively as $N$ increases. 
Hence, we can balance them out by forcing each to be smaller than $\frac{\epsilon \delta^{1/T}}{2}$, which then gives
\begin{equation} \label{eq:sampdaderiv2}
B \leq \frac{2e^{da}}{\epsilon \delta^{1/T}} \left( 1+\frac{1}{e^{da}_\gamma} \right) \qquad\text{and}\qquad N \geq \frac{2 f(\W^1)}{\gamma e^{da}_\gamma \epsilon \delta^{1/T}}
\end{equation}


Observe that $f(\W^1)$ is at most $d_x$ which follows from the loss function of DA in \eqref{eq:loss-da} and the fact that $\x \in [0,1]^{d_x}$. 
Using this in the second inequality above and also the fact that $SC \approx BN$, we get 
\begin{equation} \label{eq:sampdaderiv3}
N \geq \max\left\{ \frac{SC}{B}, \frac{2 f(\W^1)}{\gamma e^{da}_\gamma \epsilon \delta^{1/T}} \right\}
\end{equation}



\subsection{Proof of Theorem 11}\label{sec:proof11} 


Observe that the setup of Alg. \ref{alg:multi-nn} is to first pretrain the $L-1$ hidden layers, followed by running backpropagation through these layers using the outputs $\y$ 
(which correspond to the $L^th$ layer). 
Hence, the analysis of expected gradients for this $L$-layered network (denoted by $\W^1,\ldots,\W^L$), 
starts from Theorem \ref{thm:expgradda} where we bound the expected gradients of a single pretrained layer. 
Note the abuse of notation here: $\W^1,\ldots,\W^L$ denote the parameters of the network while $\W^{k,1},\ldots,\W^{k,L}$ 
correspond to their estimates at $k^{th}$ iteration of back propagation fine-tuning. 


\noindent Consider the term $\| \tg(\eta;\W) \| = \| P_{\W}(\W,\nabla_{\W} \mathcal{L}(\eta;\W),\gamma \|$, 
and the event $Pr(sup_{\eta} \| \tg(\eta;\W) \| \leq \alpha) = 1 - Pr(sup_{\eta} \| \tg(\eta;\W) \| \geq \alpha)$. 
We intend to make this probability as small as possible. Using Jensen and Markov inequalities, we have the following
\begin{equation} \label{eq:unsupbound0} 
sup_{\eta} \| \tg(\eta;\W) \| \geq \mathbb{E}_{\eta} \| \tg(\eta;\W) \| \geq \| \mathbb{E}_{\eta} \tg(\eta;\W) \| 
\end{equation}


\begin{equation} \begin{aligned} \label{eq:unsupbound}
\quad Pr \left(sup_{\eta} \| \tg(\eta;\W) \|^2 \geq \| \mathbb{E}_{\eta} \tg(\eta;\W) \|^2 \geq \alpha^2 \right) &\leq \frac{\mathbb{E}_{R} \| \mathbb{E} \tg(\eta;\W) \|^2}{\alpha^2} \\
&= \frac{1}{\alpha^2} \mathbb{E}_{R} \| \nabla_{\W} \tf(\eta;\W) \|^2 
\end{aligned} \end{equation}
and using \eqref{eq:convda}, we then get
\begin{equation} \label{eq:unsupbound1}
Pr(sup_{\eta} \| \tg(\eta;\W) \| \leq \alpha) \geq 1 - \frac{\mathrm{c}_{u}}{\alpha^2} \hspace{1mm} \quad \hspace{3mm} 
\mathrm{c}_{u} = \frac{1}{C_\gamma} \left( \frac{D_f}{N\gamma} + \frac{\vda}{B} \right) + \frac{\vda}{B}
\end{equation} 


\noindent Consider a $L$-layered multi-layer network with lengths $d_0,d_1,\ldots,d_L$. 
Recall that $d_0$ and $d_L$ correspond to the input ($\x$) and output ($\y$) layers respectively. 
Alg. \ref{alg:multi-nn} pretrains the $L-1$ hidden layers (corresponding to $\h^1,\ldots,\h^{L-1}$). 
If the expected gradients bounds resulting from pretraining these layers are denoted by $\mathrm{c}^1_u,\ldots,\mathrm{c}^{L-1}_u$, 
then we have the following with $\delta_{\alpha} = \frac{1}{\alpha^2}max\{\mathrm{c}^1_u,\ldots,\mathrm{c}^{L-1}_u\}$
\begin{equation} \label{eq:unsupbound2}
Pr(sup_{\eta} \| \tg_{l}(\eta;\W) \| \leq \alpha) \geq 1 - \delta_{\alpha} \quad \text{for} \quad l = 1,\ldots,L-1
\end{equation}


\noindent We then operate in this setting where hidden layers are pretrained (refer Alg. \ref{alg:multi-nn}); 
$\alpha$, $\delta_{\alpha}$ and the corresponding pretraining hyper-parameters (refer to Theorem \ref{thm:expgradda} and Corollary \ref{thm:convda}) 
are chosen such that the probability bound in \eqref{eq:unsupbound2} is satisfied.


\noindent We now setup the proof for bounding expected gradients for `tuning' the network. 
Recall the {\it projected} back propagation setting used in Alg. \ref{alg:multi-nn}, where the gradients of $l-1^{th}$ take the following from based on the updates and gradients from $l^{th}$ layer.
\begin{equation} \label{eq:mulbackprop}
g^{ij}(\eta;\W^l) = h^{l-1}_q h^{l}_p(1-h^{l}_p) \sum_{m=1}^{d_{l+1}} \tg^{m,i}_{l+1}(\eta;\W^{l+1}) W_{ij}^{l+1}
\end{equation}
where $\h^l = \sigma(\W^l\h^{l-1})$ is the $l^{th}$ hidden layer output, $i = 1,\ldots,d_l$ and $j = 1,\ldots,d_{l-1}$ ($\h^{0} = \x$). 
Using this, and following the setup of proof from Theomem \ref{thm:expgrad} we first compute the variance of the noisy gradients $g^{ij}(\eta;\W^l)$.
Since the network is pretrained, using \eqref{eq:unsupbound2} 
\begin{equation} \label{eq:mlnn_gradval1}
|g^{ij}(\eta;\W^l)| = |h^{l-1}_j h^{l}_i(1-h^{l}_i)| |\sum_{m=1}^{d_{l+1}} \tg^{m,i}_{l+1}(\eta;\W^{l+1})||W_{ij}^{l+1}| \leq \frac{\alpha d_{l+1}w^l_m}{4}
\end{equation}
and denoting $\delta^l = G_l(\eta;\W^l) - \nabla_{\W^l} f(\eta;\W^l)$, we have 
\begin{equation} \label{eq:mlnn_varlipbounds1}
\mathbb{E}_{\eta} \|\delta^l\|^2 \leq \frac{\alpha d_{l-1}d_{l}d_{l+1}w^l_m}{4}
\end{equation} 


\noindent Following the recipe Theorem \ref{thm:expgrad}, we then compute the Lipschitz constants for the hidden layer gradients $\tg_{l}(\eta;\W^l)$. 
Using \eqref{eq:mulbackprop} and rearranging few terms, we have
\begin{equation} \begin{aligned} \label{eq:mlnn_varlipbounds2}
\| g^{ij}_{l}(\eta;\W) - g^{ij}_{l}(\eta;\hW) \| &\leq |h^{l-1}_j \sigma_i(\W^l\h^{l-1})(1-\sigma_i(\W^l\h^{l-1})) \sum_{m=1}^{d_{l+1}} \tg^{m,i}_{l+1}(\eta;\W^{l+1}) W_{ij}^{l+1} \\
&\quad \quad - h^{l-1}_j \sigma_i(\hW^l\h^{l-1})(1-\sigma_i(\hW^l\h^{l-1})) \sum_{m=1}^{d_{l+1}} \tg^{m,i}_{l+1}(\eta;\W^{l+1}) W_{ij}^{l+1}| \\
&\leq |\sigma_i(\W^l\h^{l-1})(1-\sigma_i(\W^l\h^{l-1})) - \sigma_i(\hW^l\h^{l-1})(1-\sigma_i(\hW^l\h^{l-1}))| \\
&\quad \quad \sum_{m=1}^{d_{l+1}} |\tg^{m,i}_{l+1}(\eta;\W^{l+1})| |W_{ij}^{l+1}| \leq \frac{\alpha d_{l+1}w^l_m}{10} |W^l_{ij} - \hw^l_{ij}| 
\end{aligned} \end{equation} 


\noindent Using \eqref{eq:mlnn_varlipbounds1} and \eqref{eq:mlnn_varlipbounds2}, 
we now follow the recipe for the proofs of Theorem \ref{thm:expgrad} and Theorem \ref{thm:expgradda} to derive the expected gradients.
Recall the update steps from Alg. \ref{alg:multi-nn}, which essentially is back propagating the errors from the output layer $L$ to the first/input layer.
Within a single iteration of the backprop all the layers are updated i.e., $\{ \W^{k+1,l}\} \leftarrow \{ \W^{k,l}\} \hspace{1mm} \forall l$.
Further, the objective for each of the $L$ updates within this iteration is $f(\W^l) = \mathbb{E} \| \h^l - \sigma(\W^l\h^{l-1})\|^2$ ($\h^L = \y$, $\h^0 = \x$).


\noindent Using this intuition and recalling that is no pretraining for the output ($L^{th}$) layer, we start with the following inequality concerning the final layer (based on \eqref{eq:firstineq}),
\begin{equation} \label{eq:mlnn_firstineq}
f(\W^{k+1,l}) \leq f(\W^{k,l}) + \langle \nabla_{\W^l} f(\W^{k,l}),\W^{k+1,l}-\W^{k,l} \rangle + \frac{13}{16} \|\W^{k+1,l}-\W^{k,l}\|^2   
\end{equation}


\noindent Applying the same set of steps as in \eqref{eq:firstineq}--\eqref{eq:firstineq2}, and recalling that $\delta^{k,L} = G(\eta;\W^L) - \nabla_{\W^L} f(\W^L)$ we get
\begin{equation} \begin{aligned} \label{eq:mlnn_firstineq1}
f(\W^{k+1,L}) &\leq f(\W^{k,L}) - \left( \gamma_L^k - \frac{13}{16}(\gamma_L^k)^2 \right) \|\nabla_{\W^L} f(\W^{k,L})\|^2 \\
& \quad - \biggl( \gamma_L^k - \frac{13}{8}(\gamma_L^k)^2 \biggr) \langle \nabla_{\W^L} f(\W^{k,L}),\delta^{k,L} \rangle + \frac{13}{16} (\gamma_L^k)^2 \|\delta^{k,L}\|^2
\end{aligned} \end{equation}
Once the $L^{th}$ layer is updated, the remaining $L-1$ layers are updated from top to bottom (refer Alg. \ref{alg:multi-nn}). 


\noindent Based on the set of inequalities derived in \eqref{eq:da_firstineq}--\eqref{eq:da_firstineq3}, the corresponding bounds on the layer-wise objectives $f(\W^l)$ will be as follows
\begin{equation} \begin{aligned} \label{eq:mlnn_firstineq2}
f(\W^{k+1,l}) &\leq f(\W^{k,l}) - \left( \gamma_l^k - \frac{\alpha d_{l+1}w^l_m}{20}(\gamma_l^k)^2 \right) \|\tG(\eta^k;\W^{k,l})\|^2 \\ 
&\quad \quad \quad + \gamma_l^k \langle \delta^{k,l},\nabla_{\W^l}\tf(\W^{k,l}) \rangle + \gamma_l^k \|\delta^{k,l}\|^2 \quad;\quad 
l = L-1,\ldots,1
\end{aligned} \end{equation}
where the Lipschitz bound from \eqref{eq:mlnn_varlipbounds2} was used and $\delta^{k,l} = G(\eta;\W^l) - \nabla_{\W^l} f(\W^l)$. 


\noindent Observe that the final layer has no constraints imposed on the parameters $W^L_{ij}$, and so $\nabla_{\W^L}\tf(\W^{k,L}) = \nabla_{\W^L}f(\W^{k,L})$ and $\tG(\eta^k;\W^{k,l}) = G(\eta^k;\W^{k,L})$. 
Denoting $F(\W^k) := \sum_{l=1}^L f(\W^{k,l})$, and adding up \eqref{eq:mlnn_firstineq1} and \eqref{eq:mlnn_firstineq2}, we get
\begin{equation} \begin{aligned} \label{eq:mlnn_firstineq3}
&F(\W^{k+1}) \leq F(\W^k) - \left( \gamma_L^k - \frac{13}{16}(\gamma_L^k)^2 \right) \|\nabla_{\W^L} f(\W^{k,L})\|^2 \\
&- \sum_{l=1}^{L-1} \left( \gamma_l^k - \frac{\alpha d_{l+1}w^l_m}{20}(\gamma_l^k)^2 \right) \|\tG(\eta^k;\W^{k,l})\|^2 \\
& \quad \quad - \biggl( \gamma_L^k - \frac{13}{8}(\gamma_L^k)^2 \biggr) \langle \nabla_{\W^L} f(\W^{k,L}),\delta^{k,L} \rangle \\
& \quad \quad + \sum_{l=1}^{L-1} \gamma_l^k \langle \delta^{k,l},\nabla_{\W^l}\tf(\W^{k,l}) \rangle + \frac{13}{16} (\gamma_L^k)^2 \|\delta^{k,L}\|^2 + \sum_{l=1}^{L-1} \gamma_l^k \|\delta^{k,l}\|^2 
\end{aligned} \end{equation}


\noindent We restrict to the constant stesizes case, i.e., $\gamma_l^k = \gamma_l \forall k$ (for a given layer).
Now, summing the above inequality over the $N$ iterations and rearranging terms, we get
\begin{equation} \begin{aligned} \label{eq:mlnn_sumineq}
&\left( \gamma_L - \frac{13}{16}(\gamma_L)^2 \right) \sum_{k=1}^N \|\nabla_{\W^L} f(\W^{k,L})\|^2 
+ \sum_{l=1}^{L-1} \left( \gamma_l - \frac{\alpha d_{l+1}w^l_m}{20}(\gamma_l)^2 \right) \sum_{k=1}^N \|\tG(\eta^k;\W^{k,l})\|^2 \\
& \quad \leq F(\W^1) - F^* - \biggl( \gamma_L - \frac{13}{8}(\gamma_L)^2 \biggr) \sum_{k=1}^N \langle \nabla_{\W^L} f(\W^{k,L}),\delta^{k,L} \rangle \\
& \quad \quad + \sum_{l=1}^{L-1} \gamma_l \sum_{k=1}^N \langle \delta^{k,l},\nabla_{\W^l}\tf(\W^{k,l}) \rangle + \frac{13}{16} (\gamma_L)^2 \sum_{k=1}^N \|\delta^{k,L}\|^2 
+ \sum_{l=1}^{L-1} \gamma_l \sum_{k=1}^N \|\delta^{k,l}\|^2 
\end{aligned} \end{equation}
where $\W^1$ represents the initial estimate -- $\W^{1,L}$ for the final layer, and the pretrained estimates for the remaining $L-1$ layers. 


\noindent We now take the expectation of the above inequality with $\eta$, followed by the stopping iteration $R$. 
Recall the set of inequalities from \eqref{eq:term1} and \eqref{eq:da_term1}. 
Replicating a similar set of analysis, the terms involving $\langle\cdot,\cdot\rangle$ from the right hand side above will vanish. 
Further, from \eqref{eq:forterm2}--\eqref{eq:term2} and \eqref{eq:da_term2}, the last two terms in the above inequality will reduce to 
\begin{equation} \begin{aligned} \label{eq:mlnn_term2}
\mathbb{E}_{\eta} \left( \frac{13}{16} (\gamma_L)^2 \sum_{k=1}^N \|\delta^{k,L}\|^2 \right) &\leq \frac{13d_{L-1}d_L}{256B} N\gamma^2_L \\
\mathbb{E}_{\eta} \left( \sum_{l=1}^{L-1} \gamma_l \sum_{k=1}^N \|\delta^{k,l}\|^2 \right) &\leq \frac{N\alpha}{4B}\sum_{l=1}^{L-1} \gamma_l d_{l-1}d_{l}d_{l+1}w^l_m
\end{aligned} \end{equation}


\noindent The second inequality uses \eqref{eq:mlnn_varlipbounds1}. Here the size of mini-batch $B$ is the same across all layers. 
Using \eqref{eq:mlnn_term2} after applying the expectation to \eqref{eq:mlnn_firstineq}, we have the following
\begin{equation} \begin{aligned} \label{eq:mlnn_finalineq}
\left( \gamma_L - \frac{13}{16}(\gamma_L)^2 \right) &\sum_{k=1}^N \mathbb{E} \|\nabla_{\W^L} f(\W^{k,L})\|^2 
+ \sum_{l=1}^{L-1} \left( \gamma_l - \frac{\alpha d_{l+1}w^l_m}{20}(\gamma_l)^2 \right) \sum_{k=1}^N \mathbb{E} \|\tG(\eta^k;\W^{k,l})\|^2 \\
& \quad \leq D_F + \frac{N}{4B} \left( \frac{13d_{L-1}d_L\gamma^2_L}{64} + \alpha \sum_{l=1}^{L-1} \gamma_l d_{l-1}d_{l}d_{l+1}w^l_m \right) 
\end{aligned} \end{equation}


\noindent To allow for `simpler' summarization of the above inequality before taking the expectation with respect to the stopping iteration, denote
\begin{equation} \label{eq:mlnn_minstepfn}
\tilde{\gamma} = min \left[ \gamma_L - \frac{13}{16}(\gamma_L)^2, \gamma_l - \frac{\alpha d_{l+1}w^l_m}{20}(\gamma_l)^2 \right] \hspace{2mm} l = L-1,\ldots,1
\end{equation}


\noindent With this, we have
\begin{equation} \begin{aligned} \label{eq:mlnn_finalineqleft}
\tilde{\gamma} \sum_{k=1}^N &\left( \mathbb{E} \|\nabla_{\W^L} f(\W^{k,L})\|^2 + \sum_{l=1}^{L-1} \mathbb{E} \|\tG(\eta^k;\W^{k,l})\|^2 \right) \\
&\quad \leq \left( \gamma_L - \frac{13}{16}(\gamma_L)^2 \right) \sum_{k=1}^N \mathbb{E} \|\nabla_{\W^L} f(\W^{k,L})\|^2 \\
&\quad \quad \quad + \sum_{l=1}^{L-1} \left( \gamma_l - \frac{\alpha d_{l+1}w^l_m}{20}(\gamma_l)^2 \right) \sum_{k=1}^N \mathbb{E} \|\tG(\eta^k;\W^{k,l})\|^2 
\end{aligned} \end{equation}


\noindent Using the decomposition of $\mathbb{E}_{R} \| \tf(\W^{k,l}) \|^2$ based on \eqref{eq:da_finalineq1}, the above inequality gives
\begin{equation} \begin{aligned} \label{eq:mlnn_finalineq1}
\mathbb{E}_{R,\eta} &\| \nabla_{\W^1,\ldots,\W^L} \tf(\W) \|^2 := \sum_{l=1}^L \mathbb{E}_{R,\eta} \| \nabla_{\W^l} \tf(\W) \|^2 \\
&= \mathbb{E}_{R,\eta} \| \nabla_{\W^L} f(\W) \|^2 + \sum_{l=1}^{L-1} \mathbb{E}_{R,\eta} \| \nabla_{\W^l} \tf(\W) \|^2 \\
&= \mathbb{E}_{R,\eta} \| \nabla_{\W^L} f(\W) \|^2 + \sum_{l=1}^{L-1} \mathbb{E}_{R,\eta} \| \tG(\eta^k;\W^{k,l})\|^2 + \sum_{l=1}^{L-1} \| \nabla_{\W^l} \tf(\W) - \tG(\eta^k;\W^{k,l}) \|^2 \\
&= \frac{1}{N} \sum_{k=1}^N \mathbb{E}_{\eta} \| \nabla_{\W^L} f(\W^{k,L}) \|^2 + \frac{1}{N} \sum_{k=1}^N \sum_{l=1}^{L-1} \mathbb{E}_{\eta} \| \tG(\eta^k;\W^{k,l})\|^2 \\ 
&\quad \quad \quad + \sum_{l=1}^{L-1} \| \nabla_{\W^l} \tf(\W) - \tG(\eta^k;\W^{k,l}) \|^2 
\end{aligned} \end{equation}
where the last inequality uses the fact that the stopping iteration is uniformly chosen over the $k=1,\ldots,N$ iterations and \eqref{eq:mlnn_varlipbounds1}. 


\noindent Clearly, by the definition of $F(\W)$ from above, we see that $D_F = D_f$ (since $f^l(\W^{1,l}) = 0$ for $l=1,\ldots,L-1$).
Using this, \eqref{eq:mlnn_firstineq} and \eqref{eq:mlnn_finalineqleft}, we finally have 
\begin{equation} \begin{aligned} \label{eq:mlnn_finalineq2}
\mathbb{E}_{R,\eta} \| \nabla_{\W^1,\ldots,\W^L} \tf(\W) \|^2 &\leq \frac{D_f}{Ne^m_\gamma} 
+ \frac{1}{4Be^m_\gamma} \left( \frac{13d_{L-1}d_L\gamma^2_L}{64} + \alpha \sum_{l=1}^{L-1} \gamma_l d_{l-1}d_{l}d_{l+1}w^l_m \right) \\
&= \frac{1}{e^m_\gamma} \left( \frac{D_f}{N} + \frac{1}{B} (e^m_L + \alpha\sum_{l=1}^{L-1} e^m_l) \right) \\
\text{where} \quad &e^m_l = \frac{\gamma^l}{4}d_{l-1}d_{l}d_{l+1}w^l_m \quad (l=1,\ldots,L-1) \quad \text{and} \quad e^m_L = \frac{13d_{L-1}d_L\gamma^2_L}{256}
\end{aligned} \end{equation}



\subsection{Proof of Corollary 13}\label{sec:proof13} 


The proof of this result directly follows from \eqref{eq:convmulnn-gen}, which is presented below for reference. 
\begin{equation} \begin{aligned} \label{eq:convmulnn-gen-thesameone}
\mathbb{E} \| \nabla_{\W} \tf(\W^R) \|^2  \leq \frac{1}{e^m_\gamma} \left( \frac{D_f}{N} + \frac{1}{B} (e^m_L + \sum_{l=1}^{L-1} \alpha_le^m_l) \right) \\
\text{with}\quad\quad e^m_\gamma = \min \left\{ \gamma_L - \frac{13}{16}(\gamma_L)^2, \gamma_l - \frac{\alpha_ld_{l+1}w^l_m}{20}(\gamma_l)^2 \right\}
\end{aligned} \end{equation}


\noindent Under the assumption that $\gamma_l = \gamma \forall l$, and whenever $\alpha_ld_{l+1} < \frac{65}{4}$, we have 
\begin{equation} \begin{aligned} \label{eq:emgamma-reduced}
&\frac{\alpha_ld_{l+1}}{20}\gamma^2 < \frac{13}{16}\gamma^2 \quad\text{for any $\gamma$} \\
&\quad\text{and hence}\quad e^m_\gamma = \gamma - \frac{13}{16}\gamma^2 \quad\text{will be a constant} 
\end{aligned} \end{equation}


\noindent Now, for a given $B$ and $N$, and approximating $D_f \approx d_L$, 
the term that will be different across the neworks $\mathcal{D} := (\alpha_l,d_l)$ in the decay bound from \eqref{eq:convmulnn-gen-thesameone} is the following 
(using the definition of $e^m_l$s from Theorem \ref{thm:expgradmulnn}),
\begin{equation} \label{eq:rel-term}
\frac{13\gamma^2}{256}d_{L-1}d_L + \frac{\gamma}{4} \sum_{l=1}^{L-1} \alpha_ld_{l-1}d_ld_{l+1}
\end{equation}


\noindent Now all the networks that satisfy the following condition, will have the same \eqref{eq:rel-term}, 
and hence the decay bound would be the same i.e., they all are equivalent according to Definition \ref{thm:equiv}.
\begin{equation} \label{eq:equiv-pretrain-thesame}
\alpha_ld_{l-1}d_ld_{l+1} = \frac{1}{\Psi_l} \quad\text{for a given $\Psi_1,\ldots,\Psi_{L-1}$, $d_L$ and $d_{L-1}$} 
\end{equation}



\subsection{Proof of Corollary 14}\label{sec:proof14} 


The proof follwos by comparing the decay bound from \eqref{eq:convmulnn} for the two networks $\mathcal{D}_s$ (depth $L_s$) and $\mathcal{D}_t$ (depth $L_t$). 
Recall that $d_0$ and $d_L$ are given. $d_t = \frac{d_s}{1+\delta_{st}}$ with $L_t > L_s$ and the pretraining goodness is $\alpha$ for both (for all the layers) networks.
Since we are interested in the regime where the taller network $\mathcal{D}_t$ is better than the shorter one $\mathcal{D}_s$, we have the following using \eqref{eq:convmulnn},
\begin{equation} \begin{aligned} \label{eq:ds-vs-dt}
\frac{1}{e^m_\gamma} &\left( \frac{d_L}{N} + \frac{1}{B} \left( \frac{13d_sd_L\gamma^2}{256} + \frac{\alpha\gamma w}{4} (d^2_sd_L + d^2_sd_0) 
+ \frac{\alpha\gamma w}{4}(L_s-3)d^3_s \right) \right) \\
&\geq \frac{1}{e^m_\gamma} \left( \frac{d_L}{N} + \frac{1}{B} \left( \frac{13d_td_L\gamma^2}{256} + \frac{\alpha\gamma w}{4} (d^2_td_L + d^2_td_0) 
+ \frac{\alpha\gamma w}{4}(L_t-3)d^3_t \right) \right) 
\end{aligned} \end{equation} 
where $e^m_\gamma = \gamma - \frac{13}{16}\gamma^2$ and the definition of $e^m_l$s was used from Theorem \ref{thm:expgradmulnn}.


\noindent Rearranging and cancelling several terms will give
\begin{equation} \begin{aligned} \label{eq:ds-vs-dt-1}
\frac{13d_L\gamma^2}{256}(d_s-d_t) &+ \frac{\alpha\gamma w}{4} (d_L + d_0)(d^2_s - d^2_t) \\
&\geq \frac{\alpha\gamma w}{4} \left( (L_t-3)d^3_t - (L_s-3)d^3_s \right)
\end{aligned} \end{equation} 


\noindent Using $d_s = d_t(1+\delta_{st})$ in the above inequality, we get
\begin{equation} \begin{aligned} \label{eq:ds-vs-dt-2}
\frac{13d_L\gamma}{64}\delta_{st} &+ \alpha w \delta_{st}(2+\delta_{st})(d_L + d_0)d_t \\
&\geq \alpha wd^2_t \left( (L_t-3) - (L_s-3)(1+\delta_{st})^3 \right)
\end{aligned} \end{equation} 


\noindent which then gives
\begin{equation} \label{eq:ds-vs-dt-3}
L_t-3 \leq (L_s-3)(1+\delta_{st})^3 + \frac{1}{\alpha wd^2_t}\left( \frac{13d_L\gamma}{64}\delta_{st} + \alpha w \delta_{st}(2+\delta_{st})(d_L + d_0)d_t \right)
\end{equation} 


\noindent Recall that $\delta_{st}>0$, and so any $L_t$ that satisfies the following upper bound will satisfy the above inequality, 
\begin{equation} \begin{aligned} \label{eq:ds-vs-dt-4}
L_t &\leq (L_s-3)(1+3\delta_{st}) + 3 + \delta_{st} \left( \frac{13d_L\gamma}{64\alpha wd^2_t} + \frac{(1+\delta_{st})(d_L + d_0)}{d_t} \right) \\
&= (L_s-3)(1+3\delta_{st}) + 3 + \delta_{st} \left( \frac{\frac{13d_td_L\gamma^2}{256}}{\alpha\frac{\gamma w}{4}d^3_t} 
+ (1+\delta_{st}) \left( \frac{\frac{\gamma w}{4}d^2_td_L}{\frac{\gamma w}{4}d^3_t} + \frac{\frac{\gamma w}{4}d^2_td_0}{\frac{\gamma w}{4}d^3_t} \right) \right)
\end{aligned} \end{equation} 


\noindent where in the second inequaity several terms are adjusted to get the expression into a form of interest (that will be clear shortly). 
Now using the fact that $d_t = \frac{d_s}{1+\delta_{st}}$, and the definitions of $e^m_l$ from Theorem \ref{thm:expgradmulnn}, we finally have 
(recall that under the assumptions $e^m_2 = \ldots = e^m_{L-2}$ for $\mathcal{D}_s$),
\begin{equation} \begin{aligned} \label{eq:ds-vs-dt-final}
L_t &\leq L_s + 3\delta_{st}(L_s-3) + \delta_{st}(1+\delta_{st})^2 \left( \frac{\frac{13d_sd_L\gamma^2}{256}}{\alpha\frac{\gamma w}{4}d^3_s} 
+ \frac{\frac{\gamma w}{4}d^2_sd_L}{\frac{\gamma w}{4}d^3_s} + \frac{\frac{\gamma w}{4}d^2_sd_0}{\frac{\gamma w}{4}d^3_s} \right) \\ 
&= L_s + 3\delta_{st}(L_s-3) + \delta_{st}(1+\delta_{st})^2 \left( \frac{e^m_L}{\alpha e^m_2} + \frac{e^m_{L-1}}{e^m_2} + \frac{e^m_1}{e^m_2} \right) 
\end{aligned} \end{equation} 



\subsection{Proof of Theorem 15}\label{sec:proof16} 


Inducing dropout into the setting of Corollary \ref{thm:expgradmulnn} corresponds to adjusting the terms in \eqref{eq:mlnn_finalineq}--\eqref{eq:mlnn_finalineq2}. 
The arguments follow through what was presented in the proof of Corollary \ref{thm:conv1nn-drop} in Section \ref{sec:proof7}.


\noindent Those same observations hold for the multi-layer setting as well, and we follow the derivative arguments presented for $1$-NN above.
Since the setting for this results, assumes that the network has been pretrained layer-wise, the results from Corollary \ref{thm:expgradmulnn} will hold.
We now adapt them (specifically \eqref{eq:mlnn_firstineq}--\eqref{eq:mlnn_finalineq2}) to this multi-layer dropout case. 


\noindent Firstly the constants $e^m_1,\ldots,e^m_L$ would change as follows (refer to the statement of Corollary \ref{thm:expgradmulnn}),
\begin{equation} \begin{aligned} \label{eq:dropchange1}
e^m_l;& \quad \frac{\gamma^l}{4}d_{l-1}d_{l}d_{l+1}w^l_m \rightarrow \frac{\gamma^l}{4}\zeta_{l-1}\zeta_{l}\zeta_{l+1}d_{l-1}d_{l}d_{l+1}w^l_m \quad (l=1,\ldots,L-2) \\
e^m_{L-1};& \quad \frac{\gamma^l}{4}d_{L-2}d_{L-1}d_{L}w^l_m \rightarrow \frac{\gamma^l}{4}\zeta_{L-2}\zeta_{L-1}d_{L-2}d_{L-1}d_{L}w^l_m \\
e^m_L;& \quad \frac{13d_{L-1}d_L\gamma^2_L}{256} \rightarrow \frac{13\zeta_{L-1} d_{L-1}d_L\gamma^2_L}{256}
\end{aligned} \end{equation}
Secondly, the left hand side in \eqref{eq:mlnn_finalineq} changes to
\begin{equation} \begin{aligned} \label{eq:dropchange2}
& \zeta_L \left( \gamma_L - \frac{13}{16}(\gamma_L)^2 \right) \sum_{k=1}^N \mathbb{E} \|\nabla_{\W^L} f(\W^{k,L})\|^2 \\
&\quad + \sum_{l=1}^{L-1} \zeta_{l-1}\zeta_{l} \left( \gamma_l - \frac{\alpha d_{l+1}w^l_m}{20}(\gamma_l)^2 \right) \sum_{k=1}^N \mathbb{E} \|\tG(\eta^k;\W^{k,l})\|^2 
\end{aligned} \end{equation}


\noindent Note that this basically follows from the correction for effective number of times $W_{ij}$'s are getting updated in the dropout setting 
(refer to the discussion about \eqref{eq:singledrop} and \eqref{eq:sumineq2-drop} from Section \ref{sec:proof7}).
Using the fact that $\zeta_l^2 < \zeta_l$ and denoting $\underline{\zeta} = \min_{l} \zeta_l$, we get
\begin{equation} \begin{aligned} \label{eq:dropchange3}
& \underline{\zeta}^2 \left( \left( \gamma_L - \frac{13}{16}(\gamma_L)^2 \right) \sum_{k=1}^N \mathbb{E} \|\nabla_{\W^L} f(\W^{k,L})\|^2 \right) \\
&\quad + \underline{\zeta}^2 \left( \sum_{l=1}^{L-1} \left( \gamma_l - \frac{\alpha d_{l+1}w^l_m}{20}(\gamma_l)^2 \right) \sum_{k=1}^N \mathbb{E} \|\tG(\eta^k;\W^{k,l})\|^2  \right) < \\
& \zeta_L \left( \gamma_L - \frac{13}{16}(\gamma_L)^2 \right) \sum_{k=1}^N \mathbb{E} \|\nabla_{\W^L} f(\W^{k,L})\|^2 \\
&\quad + \sum_{l=1}^{L-1} \zeta_{l-1}\zeta_{l} \left( \gamma_l - \frac{\alpha d_{l+1}w^l_m}{20}(\gamma_l)^2 \right) \sum_{k=1}^N \mathbb{E} \|\tG(\eta^k;\W^{k,l})\|^2 
\end{aligned} \end{equation}


\noindent Lastly, the hyper-parameter $e^m_\gamma$ would also need to be changed, but assuming that the stepsizes $\gamma^l$s and $\alpha$ are reasonably small, 
so that, $e^m_\gamma$ would be approximately the same as in the non-dropout case.
If this is not the case, then the analysis and the inequalities derived become very messy, with a lot of clutter, and in the end, 
nothing new to infer from comapred to the case where $e^m_\gamma$ would simply depend on the $\gamma^l$s instead of $d_l$s.
Hence, we use the same $e^m_\gamma$ as was used in Theorem \ref{thm:expgradmulnn}.


\noindent With the changes in \eqref{eq:dropchange1} and \eqref{eq:dropchange3}, and using $\bar{\zeta} = \max_l \zeta_l$, we finally have (from \eqref{eq:mlnn_finalineq2})
\begin{equation} \begin{aligned} \label{eq:mlnndrop_finalineq}
\mathbb{E}_{R,\eta} &\| \nabla_{\W^1,\ldots,\W^L} \tf(\W) \|^2 \leq \frac{D_F}{N\underline{\zeta}^2e^m_\gamma} 
+ \frac{1}{4B\underline{\zeta}^2e^m_\gamma} \left( \frac{13\zeta_{L-1} d_{L-1}d_L\gamma^2_L}{64} \right) \\
&\quad + \frac{1}{4B\underline{\zeta}^2e^m_\gamma} \left( \alpha\gamma_l \zeta_{L-2}\zeta_{L-1} d_{L-2}d_{L-1}d_{L}w^{L-1}_m 
+ \alpha \sum_{l=1}^{L-2} \gamma_l \zeta_{l-1}\zeta_l\zeta_{l+1} d_{l-1}d_{l}d_{l+1}w^l_m \right) \\
&\leq \frac{D_F}{N\underline{\zeta}^2e^m_\gamma} 
+ \frac{1}{4B\underline{\zeta}^2e^m_\gamma} \left( \frac{13\bar{\zeta} d_{L-1}d_L\gamma^2_L}{64} \right) \\
&\quad + \frac{1}{4B\underline{\zeta}^2e^m_\gamma} \left( \alpha\gamma_l \bar{\zeta}^2 d_{L-2}d_{L-1}d_{L}w^{L-1}_m 
+ \alpha \sum_{l=1}^{L-2} \gamma_l \bar{\zeta}^3 d_{l-1}d_{l}d_{l+1}w^l_m \right) \\
\end{aligned} \end{equation} 


\noindent Using the definitions of $e^m_l \forall l$ (refer to the statement of Theorem \ref{thm:expgradmulnn}) reduces to
\begin{equation} \label{eq:mlnndrop_finalineq1}
\mathbb{E}_{R,\eta} \| \nabla_{\W} \tf(\W) \|^2 \leq \frac{1}{e^m_\gamma} \left( \frac{D_f}{N\underline{\zeta}^2} 
+ \frac{1}{B} \left( \frac{\bar{\zeta}e^m_L}{\underline{\zeta}^2} + \frac{\alpha\bar{\zeta}^2e^m_{L-1}}{\underline{\zeta}^2} 
+ \frac{\alpha\bar{\zeta}^3}{\underline{\zeta}^2}\sum_{l=1}^{L-2} e^m_l \right) \right) 
\end{equation}
which further reduces to the following, using that fact that $\bar{\zeta}^3 < \bar{\zeta}^2$ and recalling the definition of $\zeta_{\mathcal{B}}$ from the Theorem statement, 
\begin{equation} \label{eq:mlnndrop_finalineq2}
\mathbb{E}_{R,\eta} \| \nabla_{\W} \tf(\W) \|^2 \leq \frac{1}{e^m_\gamma} \left( \frac{D_f}{N\underline{\zeta}^2} 
+ \frac{1}{B} \left( \frac{\zeta_{\mathcal{B}}e^m_L}{\underline{\zeta}} + \alpha\zeta^2_{\mathcal{B}}e^m_{L-1} + \alpha\zeta^2_{\mathcal{B}} \sum_{l=1}^{L-2} e^m_l \right) \right) 
\end{equation}



\subsection{Proof of Corollary 16}\label{sec:proof17} 


The proof corresponds to some adjustments in \eqref{eq:dropchange1}--\eqref{eq:mlnndrop_finalineq2} from Section \ref{sec:proof13}.
We list them here, followed by the changed bound.


\noindent The constants $e^m_1,\ldots,e^m_L$ would change as follows,
\begin{equation} \begin{aligned} \label{eq:dropchange1-cons}
e^m_l;& \quad \frac{\gamma^l}{4}d_{l-1}d_{l}d_{l+1}w^l_m \rightarrow \frac{\gamma^l}{4}\zeta^3d_{l-1}d_{l}d_{l+1}w^l_m \quad (l=1,\ldots,L-2) \\
e^m_{L-1};& \quad \frac{\gamma^l}{4}d_{L-2}d_{L-1}d_{L}w^l_m \rightarrow \frac{\gamma^l}{4}\zeta^2d_{L-2}d_{L-1}d_{L}w^l_m \\
e^m_L;& \quad \frac{13d_{L-1}d_L\gamma^2_L}{256} \rightarrow \frac{13\zeta d_{L-1}d_L\gamma^2_L}{256}
\end{aligned} \end{equation}
The left hand side in \eqref{eq:mlnn_finalineq} changes to
\begin{equation} \begin{aligned} \label{eq:dropchange2-cons}
& \zeta \left( \gamma_L - \frac{13}{16}(\gamma_L)^2 \right) \sum_{k=1}^N \mathbb{E} \|\nabla_{\W^L} f(\W^{k,L})\|^2 \\
&\quad + \zeta^2 \sum_{l=1}^{L-1} \left( \gamma_l - \frac{\alpha d_{l+1}w^l_m}{20}(\gamma_l)^2 \right) \sum_{k=1}^N \mathbb{E} \|\tG(\eta^k;\W^{k,l})\|^2 
\end{aligned} \end{equation}


\noindent Using these and the fact that $\zeta^2 < \zeta$, we have (from \eqref{eq:mlnn_finalineq2})
\begin{equation} \begin{aligned} \label{eq:mlnndrop_finalineq-cons}
\mathbb{E}_{R,\eta} &\| \nabla_{\W^1,\ldots,\W^L} \tf(\W) \|^2 \leq \frac{D_F}{N\zeta^2e^m_\gamma} 
+ \frac{1}{4B\zeta^2e^m_\gamma} \left( \frac{13\zeta d_{L-1}d_L\gamma^2_L}{64} \right) \\
&\quad + \frac{1}{4B\zeta^2e^m_\gamma} \left( \alpha\gamma_l \zeta^2 d_{L-2}d_{L-1}d_{L}w^{L-1}_m 
+ \alpha\zeta^3 \sum_{l=1}^{L-2} \gamma_l d_{l-1}d_{l}d_{l+1}w^l_m \right) \\
\end{aligned} \end{equation} 
which then reduces to 
\begin{equation} \label{eq:mlnndrop_finalineq2-cons}
\mathbb{E}_{R,\eta} \| \nabla_{\W} \tf(\W) \|^2 \leq \frac{1}{e^m_\gamma} \left( \frac{D_f}{N\zeta^2} 
+ \frac{1}{B} \left( \frac{e^m_L}{\zeta} + \alpha e^m_{L-1} + \alpha\zeta \sum_{l=1}^{L-2} e^m_l \right) \right) 
\end{equation}



\subsection{Proof of Theorem 17}\label{sec:proof21-extra} 


The proof for this result is similar to that of Theorem \ref{thm:convda} from Section \ref{sec:proof9-extra}.
The only difference is instead of using the expected gradients for DA, we will now be using the bound for multilayer RSG from \eqref{eq:mlnndrop_finalineq2-cons}, and everything else follows the setup from Section \ref{sec:proof9-extra}.



\subsection{Proof of Corollary 18}\label{sec:proof18} 


The proof follws by working with the terms in the decay bound from \eqref{eq:mlnnpretdrop} in Theorem \ref{thm:pretvsdrop}.
Using the definitions of $e^m_l$s from Theorem \ref{thm:expgradmulnn}, and the assumption that $\gamma^l = \gamma$, $w^l_m = w > 0 \forall l$ and $D_f \approx d_L$, \eqref{eq:mlnnpretdrop} reduces to
\begin{equation} \begin{aligned} \label{eq:mlnnpretdrop-changed}
\mathbb{E} &\| \nabla_{\W} \tf(\W^R) \|^2  \lesssim \frac{d_L}{e^m_\gamma N\zeta^2} \\
& + \frac{1}{e^m_\gamma B} \left( \frac{13\gamma^2}{256\zeta}d_{L-1}d_L + \zeta \sum_{l=1}^{L-2} \frac{\alpha\gamma w}{4} d_{l-1}d_ld_{l+1} + \frac{\alpha\gamma w}{4} d_{L-2}d_{L-1}d_{L} \right) 
\end{aligned} \end{equation}


\noindent Using the fact that $\zeta \leq 1$, the bound in \eqref{eq:mlnnpretdrop-changed} can be further reduced to
\begin{equation} \label{eq:mlnnpretdrop-changed-more}
\mathbb{E} \| \nabla_{\W} \tf(\W^R) \|^2  \lesssim \frac{1}{e^m_\gamma\zeta^2} \left( \frac{d_L}{N} + \frac{13d^2\gamma^2}{256B} \right) 
+ \frac{\zeta\alpha\gamma w}{4e^m_\gamma B} \left( d_0d^2 + (L-3)d^3 + d^2d_L \right) 
\end{equation}


\noindent Balancing the terms involving $\zeta$, we then have the following (using the fact that $\zeta \in (0,1]$)
\begin{equation} \begin{aligned} \label{eq:mlnnpretdrop-opt}
\frac{1}{e^m_\gamma\zeta^2} \left( \frac{d_L}{N} + \frac{13d^2\gamma^2}{256B} \right) &\approx \frac{\zeta\alpha\gamma w}{4e^m_\gamma B} \left( d_0d^2 + (L-3)d^3 + d^2d_L \right) \\
\text{and hence the optimal $\zeta$} &:= \text{median}\left\{0,\sqrt[3]{\frac{C_1}{C_2}},1\right\} \\
\text{where}\quad C_1 = &\left( \frac{d_L}{N} + \frac{13d^2\gamma^2}{256B} \right) \quad C_2 = \frac{\alpha\gamma w}{4B} \left( d_0d^2 + (L-3)d^3 + d^2d_L \right) 
\end{aligned} \end{equation}



\subsection{Proof of Corollary 19}\label{sec:proof19} 


The proof for this result follws very closely to that of Corollary \ref{thm:equiv-pretrain} from Section \ref{sec:proof13}.
We will be usig \eqref{eq:mlnnpretdrop} instead of \eqref{eq:convmulnn}, and \eqref{eq:emgamma-reduced} would still hold.


\noindent Unlike in Corollary \ref{thm:equiv-pretrain}, here we do not fix up $d_L$ among the family of netowrks considered.
Hence, following the recipe from Section \ref{sec:proof13}, similar to \eqref{eq:rel-term}, the term that would be different across the two networks is 
\begin{equation} \label{eq:rel-term-drop}
\frac{d_L}{N\zeta^2} + \frac{1}{B}\left( \frac{13\gamma^2}{256\zeta}d_{L-1}d_L + \alpha_{L-1}d_{L-2}d_{L-1}d_L + \frac{\gamma}{4} \sum_{l=1}^{L-2} \alpha_ld_{l-1}d_ld_{l+1} \right)
\end{equation}


\noindent Now all the networks that satisfy the following condition, will have the same \eqref{eq:rel-term-drop}, 
and hence the decay bound would be the same i.e., they all are equivalent according to Definition \ref{thm:equiv}.
\begin{equation} \begin{aligned} \label{eq:equiv-pretrain-drop-thesame}
&\zeta\alpha_l d_{l-1}d_ld_{l+1} = \frac{1}{\Psi_l} \quad\quad \alpha_{L-1}d_{L-2}d_{L-1}d_{L} = \frac{1}{\Psi_{L-1}} \\
&\frac{d_{L-1}d_L}{\zeta} = \frac{1}{\Psi_L} \quad\quad \frac{d_L}{\zeta^2} = \frac{1}{\Psi_f} \quad\quad\text{for a given $\Psi_1,\ldots,\Psi_L$ and $\Psi_{f}$}
\end{aligned} \end{equation}



\subsection{Proof of Corollary 20}\label{sec:proof20} 


The proof follows exactly the same set of steps of comparing the bound for $\mathcal{D}_S$ and $\mathcal{D}_t$, 
as in \eqref{eq:ds-vs-dt}--\eqref{eq:ds-vs-dt-final} from the proof for Corollary \ref{thm:tall-short}.
The only change is the presence of the dropout rate $\zeta$, which then implies that, 
instead of the bound from \eqref{eq:convmulnn} for constructing \eqref{eq:ds-vs-dt} we use \eqref{eq:mlnnpretdrop}. 
This then gives the following, similar to \eqref{eq:ds-vs-dt},
\begin{equation} \begin{aligned} \label{eq:ds-vs-dt-drop}
\frac{1}{e^m_\gamma} &\left( \frac{d_L}{N\zeta^2} + \frac{1}{B} \left( \frac{13d_sd_L\gamma^2}{256\zeta} + \frac{\alpha\gamma w}{4} (d^2_sd_L + \zeta d^2_sd_0) 
+ \frac{\alpha\gamma\zeta w}{4}(L_s-3)d^3_s \right) \right) \\
&\geq \frac{1}{e^m_\gamma} \left( \frac{d_L}{N\zeta^2} + \frac{1}{B} \left( \frac{13d_td_L\gamma^2}{256\zeta} + \frac{\alpha\gamma w}{4} (d^2_td_L + \zeta d^2_td_0) 
+ \frac{\alpha\gamma\zeta w}{4}(L_t-3)d^3_t \right) \right) \\
&\quad\quad\text{where}\quad e^m_\gamma = \gamma - \frac{13}{16}\gamma^2
\end{aligned} \end{equation} 


\noindent Following the same recipe from \eqref{eq:ds-vs-dt-1}--\eqref{eq:ds-vs-dt-final}, we will arrive at 
\begin{equation} \label{eq:ds-vs-dt-drop-3}
L_t-3 \leq (L_s-3)(1+\delta_{st})^3 + \frac{1}{\alpha\zeta wd^2_t}\left( \frac{13d_L\gamma}{64\zeta}\delta_{st} + \alpha w \delta_{st}(2+\delta_{st})(d_L + \zeta d_0)d_t \right)
\end{equation} 


\noindent Since $\delta_{st}>0$, any $L_t$ that satistifes the following set of inequalities will follow the above one 
(where we used the definitions of $e^m_l$s and the fact that $d_t = \frac{d_s}{1+\delta_{st}}$). 
\begin{equation} \begin{aligned} \label{eq:ds-vs-dt-drop-final}
L_t &\leq L_s + 3\delta_{st}(L_s-3) + \delta_{st}(1+\delta_{st})^2 \left( \frac{\frac{13d_sd_L\gamma^2}{256}}{\alpha\zeta^2\frac{\gamma w}{4}d^3_s} 
+ \frac{\frac{\gamma w}{4}d^2_sd_L}{\zeta\frac{\gamma w}{4}d^3_s} + \frac{\frac{\gamma w}{4}d^2_sd_0}{\frac{\gamma w}{4}d^3_s} \right) \\ 
&= L_s + 3\delta_{st}(L_s-3) + \delta_{st}(1+\delta_{st})^2 \left( \frac{e^m_L}{\alpha\zeta^2 e^m_2} + \frac{e^m_{L-1}}{\zeta e^m_2} + \frac{e^m_1}{e^m_2} \right) 
\end{aligned} \end{equation} 



\bibliographystyle{icml2016}
\bibliography{dlinterplay-arxiv}

\begin{thebibliography}{91}
\providecommand{\natexlab}[1]{#1}
\providecommand{\url}[1]{\texttt{#1}}
\expandafter\ifx\csname urlstyle\endcsname\relax
  \providecommand{\doi}[1]{doi: #1}\else
  \providecommand{\doi}{doi: \begingroup \urlstyle{rm}\Url}\fi

\bibitem[Abdel-Hamid et~al.(2013)Abdel-Hamid, Deng, and Yu]{abdel2013exploring}
Ossama Abdel-Hamid, Li~Deng, and Dong Yu.
\newblock Exploring convolutional neural network structures and optimization
  techniques for speech recognition.
\newblock In \emph{INTERSPEECH}, pages 3366--3370, 2013.

\bibitem[Ahmad et~al.(1990)Ahmad, Tesauro, and He]{ahmad1990asymptotic}
Subutai Ahmad, Gerald Tesauro, and Yu~He.
\newblock Asymptotic convergence of backpropagation: numerical experiments.
\newblock In \emph{Advances in neural information processing systems}, pages
  606--613, 1990.

\bibitem[Andrychowicz et~al.(2016)Andrychowicz, Denil, Gomez, Hoffman, Pfau,
  Schaul, and de~Freitas]{andrychowicz2016learning}
Marcin Andrychowicz, Misha Denil, Sergio Gomez, Matthew~W Hoffman, David Pfau,
  Tom Schaul, and Nando de~Freitas.
\newblock Learning to learn by gradient descent by gradient descent.
\newblock \emph{arXiv preprint arXiv:1606.04474}, 2016.

\bibitem[Arlot et~al.(2010)Arlot, Celisse, et~al.]{arlot2010survey}
Sylvain Arlot, Alain Celisse, et~al.
\newblock A survey of cross-validation procedures for model selection.
\newblock \emph{Statistics surveys}, 4:\penalty0 40--79, 2010.

\bibitem[Arora et~al.(2014)Arora, Bhaskara, Ge, and Ma]{arora2014provable}
Sanjeev Arora, Aditya Bhaskara, Rong Ge, and Tengyu Ma.
\newblock Provable bounds for learning some deep representations.
\newblock In \emph{ICML}, pages 584--592, 2014.

\bibitem[Arora et~al.(2015)Arora, Liang, and Ma]{arora2015deep}
Sanjeev Arora, Yingyu Liang, and Tengyu Ma.
\newblock Why are deep nets reversible: A simple theory, with implications for
  training.
\newblock \emph{arXiv preprint arXiv:1511.05653}, 2015.

\bibitem[Ashburner et~al.(2008)Ashburner, Barnes, Chen, Daunizeau, Flandin,
  Friston, Gitelman, Kiebel, Kilner, Litvak, et~al.]{ashburner2008spm8}
John Ashburner, Gareth Barnes, C~Chen, Jean Daunizeau, Guillaume Flandin, Karl
  Friston, D~Gitelman, S~Kiebel, J~Kilner, V~Litvak, et~al.
\newblock Spm8 manual.
\newblock \emph{Functional Imaging Laboratory, Institute of Neurology},
  page~41, 2008.

\bibitem[Bach(2014)]{bach2014breaking}
Francis Bach.
\newblock Breaking the curse of dimensionality with convex neural networks.
\newblock \emph{arXiv preprint arXiv:1412.8690}, 2014.

\bibitem[Baldi and Sadowski(2014)]{baldi2014dropout}
Pierre Baldi and Peter Sadowski.
\newblock The dropout learning algorithm.
\newblock \emph{Artificial intelligence}, 210:\penalty0 78--122, 2014.

\bibitem[Baldi et~al.(2014)Baldi, Sadowski, and Whiteson]{baldi2014searching}
Pierre Baldi, Peter Sadowski, and Daniel Whiteson.
\newblock Searching for exotic particles in high-energy physics with deep
  learning.
\newblock \emph{Nature communications}, 5, 2014.

\bibitem[Becker and Le~Cun(1988)]{becker1988improving}
Sue Becker and Yann Le~Cun.
\newblock Improving the convergence of back-propagation learning with second
  order methods.
\newblock In \emph{Proceedings of the 1988 connectionist models summer school},
  pages 29--37. San Matteo, CA: Morgan Kaufmann, 1988.

\bibitem[Bellido and Fiesler(1993)]{bellido1993backpropagation}
I~Bellido and Emile Fiesler.
\newblock Do backpropagation trained neural networks have normal weight
  distributions?
\newblock In \emph{ICANN’93}, pages 772--775. Springer, 1993.

\bibitem[Bengio(2009)]{bengio2009learning}
Y.~Bengio.
\newblock Learning deep architectures for {AI}.
\newblock \emph{Foundations and trends in Machine Learning}, 2\penalty0
  (1):\penalty0 1--127, 2009.

\bibitem[Bengio et~al.(2007)Bengio, Lamblin, Popovici, Larochelle,
  et~al.]{bengio2007greedy}
Y.~Bengio, P.~Lamblin, D.~Popovici, H.~Larochelle, et~al.
\newblock Greedy layer-wise training of deep networks.
\newblock \emph{Advances in Neural information processing systems},
  19:\penalty0 153, 2007.

\bibitem[Bengio(2012)]{bengio2012practical}
Yoshua Bengio.
\newblock Practical recommendations for gradient-based training of deep
  architectures.
\newblock In \emph{Neural Networks: Tricks of the Trade}, pages 437--478.
  Springer, 2012.

\bibitem[Bengio et~al.(2013)Bengio, Courville, and
  Vincent]{bengio2013representation}
Yoshua Bengio, Aaron Courville, and Pascal Vincent.
\newblock Representation learning: A review and new perspectives.
\newblock \emph{Pattern Analysis and Machine Intelligence, IEEE Transactions
  on}, 35\penalty0 (8):\penalty0 1798--1828, 2013.

\bibitem[Bishop(1995)]{bishop1995training}
Chris~M Bishop.
\newblock Training with noise is equivalent to tikhonov regularization.
\newblock \emph{Neural computation}, 7\penalty0 (1):\penalty0 108--116, 1995.

\bibitem[Blundell et~al.(2015)Blundell, Cornebise, Kavukcuoglu, and
  Wierstra]{blundell2015weight}
Charles Blundell, Julien Cornebise, Koray Kavukcuoglu, and Daan Wierstra.
\newblock Weight uncertainty in neural networks.
\newblock \emph{arXiv preprint arXiv:1505.05424}, 2015.

\bibitem[Bottou(1991)]{bottou1991stochastic}
L{\'e}on Bottou.
\newblock Stochastic gradient learning in neural networks.
\newblock \emph{Proceedings of Neuro-N{\i}mes}, 91\penalty0 (8), 1991.

\bibitem[Bottou(2010)]{bottou2010large}
L{\'e}on Bottou.
\newblock Large-scale machine learning with stochastic gradient descent.
\newblock In \emph{Proceedings of COMPSTAT'2010}, pages 177--186. Springer,
  2010.

\bibitem[Castillo et~al.(2006)Castillo, Guijarro-Berdi{\~n}as, Fontenla-Romero,
  and Alonso-Betanzos]{castillo2006very}
Enrique Castillo, Bertha Guijarro-Berdi{\~n}as, Oscar Fontenla-Romero, and
  Amparo Alonso-Betanzos.
\newblock A very fast learning method for neural networks based on sensitivity
  analysis.
\newblock \emph{The Journal of Machine Learning Research}, 7:\penalty0
  1159--1182, 2006.

\bibitem[Ciregan et~al.(2012)Ciregan, Meier, and Schmidhuber]{ciregan2012multi}
Dan Ciregan, Ueli Meier, and J{\"u}rgen Schmidhuber.
\newblock Multi-column deep neural networks for image classification.
\newblock In \emph{Computer Vision and Pattern Recognition (CVPR), 2012 IEEE
  Conference on}, pages 3642--3649. IEEE, 2012.

\bibitem[Corouge et~al.(2004)Corouge, Gouttard, and Gerig]{corouge2004towards}
Isabelle Corouge, Sylvain Gouttard, and Guido Gerig.
\newblock Towards a shape model of white matter fiber bundles using diffusion
  tensor mri.
\newblock In \emph{Biomedical Imaging: Nano to Macro, 2004. IEEE International
  Symposium on}, pages 344--347. IEEE, 2004.

\bibitem[Dahl et~al.(2013)Dahl, Sainath, and Hinton]{dahl2013improving}
George~E Dahl, Tara~N Sainath, and Geoffrey~E Hinton.
\newblock Improving deep neural networks for {LVCSR} using rectified linear
  units and dropout.
\newblock In \emph{Acoustics, Speech and Signal Processing (ICASSP), 2013 IEEE
  International Conference on}, pages 8609--8613. IEEE, 2013.

\bibitem[Dauphin et~al.(2014)Dauphin, Pascanu, Gulcehre, Cho, Ganguli, and
  Bengio]{dauphin2014identifying}
Yann~N Dauphin, Razvan Pascanu, Caglar Gulcehre, Kyunghyun Cho, Surya Ganguli,
  and Yoshua Bengio.
\newblock Identifying and attacking the saddle point problem in
  high-dimensional non-convex optimization.
\newblock In \emph{Advances in Neural Information Processing Systems}, pages
  2933--2941, 2014.

\bibitem[Erhan et~al.(2009)Erhan, Manzagol, Bengio, Bengio, and
  Vincent]{erhan2009difficulty}
D.~Erhan, P.~Manzagol, Y.~Bengio, S.~Bengio, and P.~Vincent.
\newblock The difficulty of training deep architectures and the effect of
  unsupervised pre-training.
\newblock In \emph{Proceedings of the International Conference on Artificial
  Intelligence and Statistics}, pages 153--160, 2009.

\bibitem[Erhan et~al.(2010)Erhan, Bengio, Courville, Manzagol, Vincent, and
  Bengio]{erhan2010does}
D.~Erhan, Y.~Bengio, A.~Courville, P.~Manzagol, P.~Vincent, and S.~Bengio.
\newblock Why does unsupervised pre-training help deep learning?
\newblock \emph{The Journal of Machine Learning Research}, 11:\penalty0
  625--660, 2010.

\bibitem[Fei-Fei et~al.(2007)Fei-Fei, Fergus, and Perona]{fei2007learning}
Li~Fei-Fei, Rob Fergus, and Pietro Perona.
\newblock Learning generative visual models from few training examples: An
  incremental bayesian approach tested on 101 object categories.
\newblock \emph{Computer Vision and Image Understanding}, 106\penalty0
  (1):\penalty0 59--70, 2007.

\bibitem[Frisoni et~al.(2010)Frisoni, Fox, Jack, Scheltens, and
  Thompson]{frisoni2010clinical}
Giovanni~B Frisoni, Nick~C Fox, Clifford~R Jack, Philip Scheltens, and Paul~M
  Thompson.
\newblock The clinical use of structural mri in alzheimer disease.
\newblock \emph{Nature Reviews Neurology}, 6\penalty0 (2):\penalty0 67--77,
  2010.

\bibitem[Gerber et~al.(2009)Gerber, Tasdizen, Joshi, and
  Whitaker]{gerber2009manifold}
Samuel Gerber, Tolga Tasdizen, Sarang Joshi, and Ross Whitaker.
\newblock On the manifold structure of the space of brain images.
\newblock In \emph{International Conference on Medical Image Computing and
  Computer-Assisted Intervention}, pages 305--312. Springer, 2009.

\bibitem[Ghadimi and Lan(2013)]{ghadimi2013stochastic}
S.~Ghadimi and G.~Lan.
\newblock Stochastic first-and zeroth-order methods for nonconvex stochastic
  programming.
\newblock \emph{SIAM Journal on Optimization}, 23\penalty0 (4):\penalty0
  2341--2368, 2013.

\bibitem[Ghadimi et~al.(2016)Ghadimi, Lan, and Zhang]{ghadimi2016mini}
Saeed Ghadimi, Guanghui Lan, and Hongchao Zhang.
\newblock Mini-batch stochastic approximation methods for nonconvex stochastic
  composite optimization.
\newblock \emph{Mathematical Programming}, 155\penalty0 (1-2):\penalty0
  267--305, 2016.

\bibitem[Goh et~al.(2013)Goh, Thome, Cord, and Lim]{goh2013top}
Hanlin Goh, Nicolas Thome, Matthieu Cord, and Joo-Hwee Lim.
\newblock Top-down regularization of deep belief networks.
\newblock In \emph{Advances in Neural Information Processing Systems}, pages
  1878--1886, 2013.

\bibitem[Goodfellow et~al.(2013)Goodfellow, Warde-Farley, Mirza, Courville, and
  Bengio]{goodfellow2013maxout}
Ian~J Goodfellow, David Warde-Farley, Mehdi Mirza, Aaron~C Courville, and
  Yoshua Bengio.
\newblock Maxout networks.
\newblock \emph{ICML (3)}, 28:\penalty0 1319--1327, 2013.

\bibitem[Haidich(2011)]{haidich2011meta}
AB~Haidich.
\newblock Meta-analysis in medical research.
\newblock \emph{Hippokratia}, 14\penalty0 (1):\penalty0 29--37, 2011.

\bibitem[Hardt et~al.(2015)Hardt, Recht, and Singer]{hardt2015train}
Moritz Hardt, Benjamin Recht, and Yoram Singer.
\newblock Train faster, generalize better: Stability of stochastic gradient
  descent.
\newblock \emph{arXiv preprint arXiv:1509.01240}, 2015.

\bibitem[Hinrichs et~al.(2011)Hinrichs, Singh, Xu, Johnson, Initiative,
  et~al.]{hinrichs2011predictive}
Chris Hinrichs, Vikas Singh, Guofan Xu, Sterling~C Johnson, Alzheimers
  Disease~Neuroimaging Initiative, et~al.
\newblock Predictive markers for ad in a multi-modality framework: an analysis
  of mci progression in the adni population.
\newblock \emph{Neuroimage}, 55\penalty0 (2):\penalty0 574--589, 2011.

\bibitem[Hinton and Salakhutdinov(2006)]{hinton2006reducing}
G.~Hinton and R.~Salakhutdinov.
\newblock Reducing the dimensionality of data with neural networks.
\newblock \emph{Science}, 313\penalty0 (5786):\penalty0 504--507, 2006.

\bibitem[Hinton(2010)]{hinton2010practical}
Geoffrey Hinton.
\newblock A practical guide to training restricted boltzmann machines.
\newblock \emph{Momentum}, 9\penalty0 (1):\penalty0 926, 2010.

\bibitem[Hinton et~al.(2012)Hinton, Deng, Yu, Dahl, Mohamed, Jaitly, Senior,
  Vanhoucke, Nguyen, Sainath, et~al.]{hinton2012deep}
Geoffrey Hinton, Li~Deng, Dong Yu, George~E Dahl, Abdel-rahman Mohamed, Navdeep
  Jaitly, Andrew Senior, Vincent Vanhoucke, Patrick Nguyen, Tara~N Sainath,
  et~al.
\newblock Deep neural networks for acoustic modeling in speech recognition: The
  shared views of four research groups.
\newblock \emph{Signal Processing Magazine, IEEE}, 29\penalty0 (6):\penalty0
  82--97, 2012.

\bibitem[Ioffe and Szegedy(2015)]{ioffe2015batch}
Sergey Ioffe and Christian Szegedy.
\newblock Batch normalization: Accelerating deep network training by reducing
  internal covariate shift.
\newblock \emph{arXiv preprint arXiv:1502.03167}, 2015.

\bibitem[Ithapu et~al.(2015)Ithapu, Singh, Okonkwo, Chappell, Dowling, Johnson,
  Initiative, et~al.]{ithapu2015imaging}
Vamsi~K Ithapu, Vikas Singh, Ozioma~C Okonkwo, Richard~J Chappell, N~Maritza
  Dowling, Sterling~C Johnson, Alzheimer's Disease~Neuroimaging Initiative,
  et~al.
\newblock Imaging-based enrichment criteria using deep learning algorithms for
  efficient clinical trials in mild cognitive impairment.
\newblock \emph{Alzheimer's \& Dementia}, 11\penalty0 (12):\penalty0
  1489--1499, 2015.

\bibitem[Jack et~al.(2008)Jack, Bernstein, Fox, Thompson, Alexander, Harvey,
  Borowski, Britson, L~Whitwell, Ward, et~al.]{jack2008alzheimer}
Clifford~R Jack, Matt~A Bernstein, Nick~C Fox, Paul Thompson, Gene Alexander,
  Danielle Harvey, Bret Borowski, Paula~J Britson, Jennifer L~Whitwell,
  Chadwick Ward, et~al.
\newblock The alzheimer's disease neuroimaging initiative (adni): Mri methods.
\newblock \emph{Journal of Magnetic Resonance Imaging}, 27\penalty0
  (4):\penalty0 685--691, 2008.

\bibitem[Janzamin et~al.(2015)Janzamin, Sedghi, and
  Anandkumar]{janzamin2015beating}
Majid Janzamin, Hanie Sedghi, and Anima Anandkumar.
\newblock Beating the perils of non-convexity: Guaranteed training of neural
  networks using tensor methods.
\newblock \emph{CoRR abs/1506.08473}, 2015.

\bibitem[Kavukcuoglu et~al.(2010)Kavukcuoglu, Sermanet, Boureau, Gregor,
  Mathieu, and LeCun]{kavukcuoglu2010learning}
K.~Kavukcuoglu, P.~Sermanet, Y.~Boureau, K.~Gregor, M.~Mathieu, and Y.~LeCun.
\newblock Learning convolutional feature hierarchies for visual recognition.
\newblock In \emph{Advances in Neural Information Processing Systems},
  volume~1, page~5, 2010.

\bibitem[Killiany et~al.(2000)Killiany, Gomez-Isla, Moss, Kikinis, Sandor,
  Jolesz, Tanzi, Jones, Hyman, and Albert]{killiany2000use}
Ronald~J Killiany, Teresa Gomez-Isla, Mark Moss, Ron Kikinis, Tamas Sandor,
  Ferenc Jolesz, Rudolph Tanzi, Kenneth Jones, Bradley~T Hyman, and Marilyn~S
  Albert.
\newblock Use of structural magnetic resonance imaging to predict who will get
  alzheimer's disease.
\newblock \emph{Annals of neurology}, 47\penalty0 (4):\penalty0 430--439, 2000.

\bibitem[Klunk et~al.(2015)Klunk, Koeppe, Price, Benzinger, Devous, Jagust,
  Johnson, Mathis, Minhas, Pontecorvo, et~al.]{klunk2015centiloid}
William~E Klunk, Robert~A Koeppe, Julie~C Price, Tammie~L Benzinger, Michael~D
  Devous, William~J Jagust, Keith~A Johnson, Chester~A Mathis, Davneet Minhas,
  Michael~J Pontecorvo, et~al.
\newblock The centiloid project: standardizing quantitative amyloid plaque
  estimation by pet.
\newblock \emph{Alzheimer's \& Dementia}, 11\penalty0 (1):\penalty0 1--15,
  2015.

\bibitem[Krizhevsky et~al.(2012)Krizhevsky, Sutskever, and
  Hinton]{krizhevsky2012imagenet}
A.~Krizhevsky, I.~Sutskever, and G.~Hinton.
\newblock Imagenet classification with deep convolutional neural networks.
\newblock In \emph{Advances in Neural Information Processing Systems},
  volume~1, page~4, 2012.

\bibitem[Krizhevsky and Hinton(2009)]{krizhevsky2009learning}
Alex Krizhevsky and Geoffrey Hinton.
\newblock Learning multiple layers of features from tiny images.
\newblock 2009.

\bibitem[Lazar et~al.(2003)Lazar, Weinstein, Tsuruda, Hasan, Arfanakis,
  Meyerand, Badie, Rowley, Haughton, Field, et~al.]{lazar2003white}
Mariana Lazar, David~M Weinstein, Jay~S Tsuruda, Khader~M Hasan, Konstantinos
  Arfanakis, M~Elizabeth Meyerand, Benham Badie, Howard~A Rowley, Victor
  Haughton, Aaron Field, et~al.
\newblock White matter tractography using diffusion tensor deflection.
\newblock \emph{Human brain mapping}, 18\penalty0 (4):\penalty0 306--321, 2003.

\bibitem[LeCun et~al.(1998)LeCun, Bottou, Bengio, and
  Haffner]{lecun1998gradient}
Yann LeCun, L{\'e}on Bottou, Yoshua Bengio, and Patrick Haffner.
\newblock Gradient-based learning applied to document recognition.
\newblock \emph{Proceedings of the IEEE}, 86\penalty0 (11):\penalty0
  2278--2324, 1998.

\bibitem[LeCun et~al.(2015)LeCun, Bengio, and Hinton]{hintonnature}
Yann LeCun, Yoshua Bengio, and Geoffrey Hinton.
\newblock Deep learning.
\newblock \emph{Nature}, 521:\penalty0 436--44, 2015.

\bibitem[LeCun et~al.(2012)LeCun, Bottou, Orr, and
  M{\"u}ller]{lecun2012efficient}
Yann~A LeCun, L{\'e}on Bottou, Genevieve~B Orr, and Klaus-Robert M{\"u}ller.
\newblock Efficient backprop.
\newblock In \emph{Neural networks: Tricks of the trade}, pages 9--48.
  Springer, 2012.

\bibitem[Lee et~al.(2009)Lee, Grosse, Ranganath, and Ng]{lee2009convolutional}
H.~Lee, R.~Grosse, R.~Ranganath, and A.~Ng.
\newblock Convolutional deep belief networks for scalable unsupervised learning
  of hierarchical representations.
\newblock In \emph{Proceedings of the 26th International Conference on Machine
  Learning (ICML-09)}, pages 609--616, 2009.

\bibitem[Lenz et~al.(2015)Lenz, Lee, and Saxena]{lenz2015deep}
Ian Lenz, Honglak Lee, and Ashutosh Saxena.
\newblock Deep learning for detecting robotic grasps.
\newblock \emph{The International Journal of Robotics Research}, 34\penalty0
  (4-5):\penalty0 705--724, 2015.

\bibitem[Li et~al.(2016)Li, Jamieson, DeSalvo, Rostamizadeh, and
  Talwalkar]{li2016efficient}
Lisha Li, Kevin Jamieson, Giulia DeSalvo, Afshin Rostamizadeh, and Ameet
  Talwalkar.
\newblock Efficient hyperparameter optimization and infinitely many armed
  bandits.
\newblock \emph{arXiv preprint arXiv:1603.06560}, 2016.

\bibitem[Livni et~al.(2014)Livni, Shalev-Shwartz, and
  Shamir]{livni2014computational}
R.~Livni, S.~Shalev-Shwartz, and O.~Shamir.
\newblock On the computational efficiency of training neural networks.
\newblock In \emph{Advances in Neural Information Processing Systems}, pages
  855--863, 2014.

\bibitem[Long et~al.(2015)Long, Cao, Wang, and Jordan]{long2015learning}
Mingsheng Long, Yue Cao, Jianmin Wang, and Michael~I Jordan.
\newblock Learning transferable features with deep adaptation networks.
\newblock In \emph{ICML}, pages 97--105, 2015.

\bibitem[Lusci et~al.(2013)Lusci, Pollastri, and Baldi]{lusci2013deep}
Alessandro Lusci, Gianluca Pollastri, and Pierre Baldi.
\newblock Deep architectures and deep learning in chemoinformatics: the
  prediction of aqueous solubility for drug-like molecules.
\newblock \emph{Journal of chemical information and modeling}, 53\penalty0
  (7):\penalty0 1563--1575, 2013.

\bibitem[Magoulas et~al.(1999)Magoulas, Vrahatis, and
  Androulakis]{magoulas1999improving}
George~D. Magoulas, Michael~N. Vrahatis, and George~S Androulakis.
\newblock Improving the convergence of the backpropagation algorithm using
  learning rate adaptation methods.
\newblock \emph{Neural Computation}, 11\penalty0 (7):\penalty0 1769--1796,
  1999.

\bibitem[Matsuoka(1992)]{matsuoka1992noise}
Kiyotoshi Matsuoka.
\newblock Noise injection into inputs in back-propagation learning.
\newblock \emph{Systems, Man and Cybernetics, IEEE Transactions on},
  22\penalty0 (3):\penalty0 436--440, 1992.

\bibitem[Mnih et~al.(2013)Mnih, Kavukcuoglu, Silver, Graves, Antonoglou,
  Wierstra, and Riedmiller]{mnih2013playing}
Volodymyr Mnih, Koray Kavukcuoglu, David Silver, Alex Graves, Ioannis
  Antonoglou, Daan Wierstra, and Martin Riedmiller.
\newblock Playing atari with deep reinforcement learning.
\newblock \emph{arXiv preprint arXiv:1312.5602}, 2013.

\bibitem[Montufar et~al.(2014)Montufar, Pascanu, Cho, and
  Bengio]{montufar2014number}
Guido~F Montufar, Razvan Pascanu, Kyunghyun Cho, and Yoshua Bengio.
\newblock On the number of linear regions of deep neural networks.
\newblock In \emph{Advances in neural information processing systems}, pages
  2924--2932, 2014.

\bibitem[Mueller et~al.(2005)Mueller, Weiner, Thal, Petersen, Jack,
  et~al.]{mueller2005ways}
Susanne~G Mueller, Michael~W Weiner, Leon~J Thal, Ronald~C Petersen, Clifford~R
  Jack, et~al.
\newblock Ways toward an early diagnosis in alzheimer’s disease: the
  alzheimer’s disease neuroimaging initiative (adni).
\newblock \emph{Alzheimer's \& Dementia}, 1\penalty0 (1):\penalty0 55--66,
  2005.

\bibitem[Murty and Kabadi(1987)]{murty1987some}
Katta~G Murty and Santosh~N Kabadi.
\newblock Some np-complete problems in quadratic and nonlinear programming.
\newblock \emph{Mathematical programming}, 39\penalty0 (2):\penalty0 117--129,
  1987.

\bibitem[Nagi et~al.(2011)Nagi, Ducatelle, Di~Caro, Ciresan, Meier, Giusti,
  Nagi, Schmidhuber, and Gambardella]{nagi2011max}
Jawad Nagi, Frederick Ducatelle, Gianni~A Di~Caro, Dan Ciresan, Ueli Meier,
  Alessandro Giusti, Farrukh Nagi, J{\"u}rgen Schmidhuber, and Luca~Maria
  Gambardella.
\newblock Max-pooling convolutional neural networks for vision-based hand
  gesture recognition.
\newblock In \emph{Signal and Image Processing Applications (ICSIPA), 2011 IEEE
  International Conference on}, pages 342--347. IEEE, 2011.

\bibitem[Nair and Hinton(2010)]{nair2010rectified}
Vinod Nair and Geoffrey~E Hinton.
\newblock Rectified linear units improve restricted boltzmann machines.
\newblock In \emph{Proceedings of the 27th International Conference on Machine
  Learning (ICML-10)}, pages 807--814, 2010.

\bibitem[Ngiam et~al.(2011)Ngiam, Coates, Lahiri, Prochnow, Le, and
  Ng]{ngiam2011optimization}
J.~Ngiam, A.~Coates, A.~Lahiri, B.~Prochnow, Q.~Le, and A.~Ng.
\newblock On optimization methods for deep learning.
\newblock In \emph{Proceedings of the 28th International Conference on Machine
  Learning (ICML-11)}, pages 265--272, 2011.

\bibitem[Patel et~al.(2015)Patel, Nguyen, and Baraniuk]{patel2015probabilistic}
Ankit~B Patel, Tan Nguyen, and Richard~G Baraniuk.
\newblock A probabilistic theory of deep learning.
\newblock \emph{arXiv preprint arXiv:1504.00641}, 2015.

\bibitem[Pesaran and Weeks(2001)]{pesaran2001non}
M~Hashem Pesaran and Melvyn Weeks.
\newblock Non-nested hypothesis testing: an overview.
\newblock \emph{A Companion to Theoretical Econometrics}, pages 279--309, 2001.

\bibitem[Plis et~al.(2014)Plis, Hjelm, Salakhutdinov, Allen,
  et~al.]{plis2014deep}
Sergey~M Plis, Devon~R Hjelm, Ruslan Salakhutdinov, Elena~A Allen, et~al.
\newblock Deep learning for neuroimaging: a validation study.
\newblock \emph{Frontiers in neuroscience}, 8, 2014.

\bibitem[Rifai et~al.(2011)Rifai, Glorot, Bengio, and Vincent]{rifai2011adding}
Salah Rifai, Xavier Glorot, Yoshua Bengio, and Pascal Vincent.
\newblock Adding noise to the input of a model trained with a regularized
  objective.
\newblock \emph{arXiv preprint arXiv:1104.3250}, 2011.

\bibitem[Rockafellar and Wets(2009)]{rockafellar2009variational}
R~Tyrrell Rockafellar and Roger J-B Wets.
\newblock \emph{Variational analysis}, volume 317.
\newblock Springer Science \& Business Media, 2009.

\bibitem[Romero et~al.(2014)Romero, Ballas, Kahou, Chassang, Gatta, and
  Bengio]{romero2014fitnets}
Adriana Romero, Nicolas Ballas, Samira~Ebrahimi Kahou, Antoine Chassang, Carlo
  Gatta, and Yoshua Bengio.
\newblock Fitnets: Hints for thin deep nets.
\newblock \emph{arXiv preprint arXiv:1412.6550}, 2014.

\bibitem[Roth and Black(2005)]{roth2005fields}
Stefan Roth and Michael~J Black.
\newblock Fields of experts: A framework for learning image priors.
\newblock In \emph{2005 IEEE Computer Society Conference on Computer Vision and
  Pattern Recognition (CVPR'05)}, volume~2, pages 860--867. IEEE, 2005.

\bibitem[Sachidanandam et~al.(2001)Sachidanandam, Weissman, Schmidt, Kakol,
  Stein, Marth, Sherry, Mullikin, Mortimore, Willey,
  et~al.]{sachidanandam2001map}
Ravi Sachidanandam, David Weissman, Steven~C Schmidt, Jerzy~M Kakol, Lincoln~D
  Stein, Gabor Marth, Steve Sherry, James~C Mullikin, Beverley~J Mortimore,
  David~L Willey, et~al.
\newblock A map of human genome sequence variation containing 1.42 million
  single nucleotide polymorphisms.
\newblock \emph{Nature}, 409\penalty0 (6822):\penalty0 928--933, 2001.

\bibitem[Saxe et~al.(2011)Saxe, Koh, Chen, Bhand, Suresh, and
  Ng]{saxe2011random}
A.~Saxe, P.~Koh, Z.~Chen, M.~Bhand, B.~Suresh, and A.~Ng.
\newblock On random weights and unsupervised feature learning.
\newblock In \emph{Proceedings of the 28th International Conference on Machine
  Learning (ICML-11)}, pages 1089--1096, 2011.

\bibitem[Shao and Zheng(2011)]{shao2011convergence}
Hongmei Shao and Gaofeng Zheng.
\newblock Convergence analysis of a back-propagation algorithm with adaptive
  momentum.
\newblock \emph{Neurocomputing}, 74\penalty0 (5):\penalty0 749--752, 2011.

\bibitem[Simonyan and Zisserman(2014)]{simonyan2014very}
Karen Simonyan and Andrew Zisserman.
\newblock Very deep convolutional networks for large-scale image recognition.
\newblock \emph{arXiv preprint arXiv:1409.1556}, 2014.

\bibitem[Snoek et~al.(2012)Snoek, Larochelle, and Adams]{snoek2012practical}
Jasper Snoek, Hugo Larochelle, and Ryan~P Adams.
\newblock Practical bayesian optimization of machine learning algorithms.
\newblock In \emph{Advances in neural information processing systems}, pages
  2951--2959, 2012.

\bibitem[Sperling et~al.(2014)Sperling, Rentz, Johnson, Karlawish, Donohue,
  Salmon, and Aisen]{sperling2014a4}
Reisa~A Sperling, Dorene~M Rentz, Keith~A Johnson, Jason Karlawish, Michael
  Donohue, David~P Salmon, and Paul Aisen.
\newblock The a4 study: stopping ad before symptoms begin?
\newblock \emph{Science translational medicine}, 6\penalty0 (228):\penalty0
  228fs13--228fs13, 2014.

\bibitem[Srivastava et~al.(2014)Srivastava, Hinton, Krizhevsky, Sutskever, and
  Salakhutdinov]{srivastava2014dropout}
Nitish Srivastava, Geoffrey Hinton, Alex Krizhevsky, Ilya Sutskever, and Ruslan
  Salakhutdinov.
\newblock Dropout: A simple way to prevent neural networks from overfitting.
\newblock \emph{The Journal of Machine Learning Research}, 15\penalty0
  (1):\penalty0 1929--1958, 2014.

\bibitem[Sutskever et~al.(2014)Sutskever, Vinyals, and
  Le]{sutskever2014sequence}
Ilya Sutskever, Oriol Vinyals, and Quoc~VV Le.
\newblock Sequence to sequence learning with neural networks.
\newblock In \emph{Advances in neural information processing systems}, pages
  3104--3112, 2014.

\bibitem[Szegedy et~al.(2014)Szegedy, Liu, Jia, Sermanet, Reed, Anguelov,
  Erhan, Vanhoucke, and Rabinovich]{szegedy2014going}
Christian Szegedy, Wei Liu, Yangqing Jia, Pierre Sermanet, Scott Reed, Dragomir
  Anguelov, Dumitru Erhan, Vincent Vanhoucke, and Andrew Rabinovich.
\newblock Going deeper with convolutions.
\newblock \emph{arXiv preprint arXiv:1409.4842}, 2014.

\bibitem[Vincent et~al.(2010)Vincent, Larochelle, Lajoie, Bengio, and
  Manzagol]{vincent2010stacked}
P.~Vincent, H.~Larochelle, I.~Lajoie, Y.~Bengio, and P.~Manzagol.
\newblock Stacked denoising autoencoders: {L}earning useful representations in
  a deep network with a local denoising criterion.
\newblock \emph{The Journal of Machine Learning Research}, 9999:\penalty0
  3371--3408, 2010.

\bibitem[Vogl et~al.(1988)Vogl, Mangis, Rigler, Zink, and
  Alkon]{vogl1988accelerating}
Thomas~P Vogl, JK~Mangis, AK~Rigler, WT~Zink, and DL~Alkon.
\newblock Accelerating the convergence of the back-propagation method.
\newblock \emph{Biological cybernetics}, 59\penalty0 (4-5):\penalty0 257--263,
  1988.

\bibitem[Wager et~al.(2013)Wager, Wang, and Liang]{wager2013dropout}
Stefan Wager, Sida Wang, and Percy~S Liang.
\newblock Dropout training as adaptive regularization.
\newblock In \emph{Advances in Neural Information Processing Systems}, pages
  351--359, 2013.

\bibitem[Wan et~al.(2013)Wan, Zeiler, Zhang, Cun, and
  Fergus]{wan2013regularization}
Li~Wan, Matthew Zeiler, Sixin Zhang, Yann~L Cun, and Rob Fergus.
\newblock Regularization of neural networks using dropconnect.
\newblock In \emph{Proceedings of the 30th International Conference on Machine
  Learning (ICML-13)}, pages 1058--1066, 2013.

\bibitem[Wang and Raj(2015)]{wang2015survey}
Haohan Wang and Bhiksha Raj.
\newblock A survey: Time travel in deep learning space: An introduction to deep
  learning models and how deep learning models evolved from the initial ideas.
\newblock \emph{arXiv preprint arXiv:1510.04781}, 2015.

\bibitem[Wei et~al.(2016)Wei, Wang, Rui, and Chen]{wei2016network}
Tao Wei, Changhu Wang, Rong Rui, and Chang~Wen Chen.
\newblock Network morphism.
\newblock \emph{arXiv preprint arXiv:1603.01670}, 2016.

\bibitem[Yosinski et~al.(2014)Yosinski, Clune, Bengio, and
  Lipson]{yosinski2014transferable}
Jason Yosinski, Jeff Clune, Yoshua Bengio, and Hod Lipson.
\newblock How transferable are features in deep neural networks?
\newblock In \emph{Advances in neural information processing systems}, pages
  3320--3328, 2014.

\end{thebibliography}

\end{document}